\documentclass[runningheads]{llncs}

\PassOptionsToPackage{table}{xcolor}

\usepackage{eccv}

 \usepackage{subcaption}
 \usepackage{wrapfig}
\usepackage{eccvabbrv}
\usepackage{threeparttable}
\usepackage{graphicx}
\usepackage{booktabs}
\usepackage{multirow}
\usepackage{makecell}
\usepackage{pifont} 
\usepackage{dsfont}
\usepackage{iftex}
\ifPDFTeX
\usepackage[accsupp]{axessibility}  
\fi
\newcommand{\gain}[1]{\textcolor{RoyalBlue}{\scriptsize\, \textbf{(#1$ \uparrow$)}}}

\usepackage{booktabs}
\usepackage{multirow}
\usepackage{xcolor}   
\usepackage{colortbl}

\usepackage{hyperref}
\hypersetup{
        final,
        colorlinks,
        linkcolor={Maroon},
        citecolor={MidnightBlue},
        urlcolor={teal!85!black}
        }
\newcommand{\suppref}[1]{Supp.~\ref{#1}}
\newif\ifarxivversion
\arxivversionfalse
\newcommand{\projectvideolink}{%
  \href{https://rathgrith.github.io/PeCA/\#results-video}{video}%
}

\usepackage{orcidlink}
\usepackage{pifont}
\usepackage{xcolor}
\usepackage[most]{tcolorbox}
\usepackage{epigraph}
\newcommand{\cmark}{\textcolor{ForestGreen}{\ding{51}}} 
\newcommand{\xmark}{\textcolor{BrickRed}{\ding{55}}} 
\definecolor{PromptTitle}{RGB}{92,190,204}
\definecolor{PromptBody}{RGB}{246,246,255}
\newtcolorbox{promptblock}[1]{
  enhanced,
  width=0.94\linewidth,
  colframe=PromptTitle,
  colback=PromptBody,
  colbacktitle=PromptTitle,
  coltitle=black,
  title={#1},
  fonttitle=\bfseries,
  boxrule=0.8pt,
  arc=2mm,
  left=2mm,
  right=2mm,
  top=1.2mm,
  bottom=1.2mm,
  toptitle=1mm,
  bottomtitle=1mm,
  boxsep=0pt
}
\DeclareRobustCommand{\zebraswatch}{%
  \texorpdfstring{%
    \tikz[baseline=-0.45ex,x=1ex,y=1ex]{%
      \path[fill=red] (0,0) rectangle (1.15,1.15);
      \clip (0,0) rectangle (1.15,1.15);
      \draw[white,line width=0.20ex] (-0.25,0) -- (0.90,1.15);
      \draw[white,line width=0.20ex] (0.25,0) -- (1.40,1.15);
      \draw[black,line width=0.04ex] (0,0) rectangle (1.15,1.15);
    }%
  }{red-white zebra}%
}

\arxivversiontrue
\renewcommand{\suppref}[1]{App.~\ref{#1}}

\begin{document}

\title{\textsc{PeCA}: Palette Context Assisted Inference for Test-Time Paint-Bucket Colourisation\\ on Animation Videos} 

\titlerunning{\textsc{PeCA}: Palette Context Assisted Inference}

\author{Dongheng Lin\orcidlink{0009-0004-5834-9077} \and Jianbo Jiao\orcidlink{0000-0003-0833-5115}}

\authorrunning{D.~Lin and J.~Jiao}

\institute{The \href{https://mix.jianbojiao.com/}{\textcolor{teal!85!black}{MIx Group}}, University of Birmingham, Birmingham, United Kingdom \\\email{dxl594@student.bham.ac.uk, j.jiao@bham.ac.uk}\\
{Project page: \textcolor{teal!85!black}{\url{https://rathgrith.github.io/PeCA/}}}
}

\maketitle

\begin{abstract}
In animation production, paint-bucket colourisation for hand-drawn animation is a labour-intensive procedure that assigns each enclosed region in line sketches a colour from reference design sheets. Recent automatic paint-bucket colourisation pipelines mirror this workflow via region correspondence, but correspondences can be brittle when regions are ambiguous fragments without proper context. In this paper, we propose Palette Context Assisted (\textsc{PeCA}), a new training-free, plug-and-play framework for animation video colourisation that aims to close this gap at test-time via reasoning over spatial and temporal contexts. Extensive experiments on existing benchmarks and a newly introduced long-video test case show consistent performance boosts. 
\end{abstract}

\epigraph{\makebox[0.9\textwidth][c]{\hspace{3mm}``A colour shines in its surroundings.''}}{\textit{Ludwig Wittgenstein}}

\section{Introduction}
\label{sec:intro}

Hand-drawn animation colourisation is not an unconstrained image synthesis problem. In production, region colours must strictly follow a discrete celluloid palette defined by design sheets, and colours must be correctly assigned despite deformation, occlusion, and changes in the layout of line-enclosed regions \cite{nakanishi2013modeling}. Consequently, transforming line sketches into correctly coloured animation remains a major bottleneck \cite{guajardo2024generative}. A particularly labour-intensive stage is the paint-bucket colourisation, where artists meticulously assign colours to a massive number of enclosed regions \cite{nakanishi2013modeling}. This step is repetitive yet unforgiving: even minor colourisation mistakes or boundary leaks can break production-level quality and lead to costly correction. Although automatic colourisation has improved markedly with the advancement of generative computer vision techniques \cite{tang2025generative, guajardo2024generative}, fully reliable automated paint-bucket colourisation remains a challenge. 

Existing automated colourisation approaches can be broadly split into pixel-based generative models and segment-based pipelines. Pixel-generative methods \cite{shi2022reference,cao2024animediffusion,zhuang2024colorflow,huang2024lvcd,liu2025manganinja,xing2024tooncrafter,sadihin2026timecolor,zhang2025animecolor}, including recent DiT-based generative models\cite{zhuang2025cobra,zhang2025followyourcolormultiinstancesketchcolorization,li2025tooncomposer, Yang_2025_ICCV}, can produce seemingly pleasing renders, but often violate production constraints: colours bleed across ink boundaries and discrete colour restriction is weak or absent \cite{tang2025generative}. Segment-based methods \cite{cao2021line, casey2021animation,maejima2019graph,ramassamy2018pre,dai2024learning,dai2024paint,nagata2025dacon} instead mirror the paint-bucket workflow by treating colourisation as assigning a colour to enclosed regions in the target frame from those in reference frames (\ie coloured key-frames or design sheets) by regional correspondences. This formulation preserves the exact colour and region constraints by construction. The remaining difficulty lies in finding robust correspondences between references and target frame regions that may appear to be completely different.

\begin{figure}[t]
    \centering
    \includegraphics[width=\linewidth]{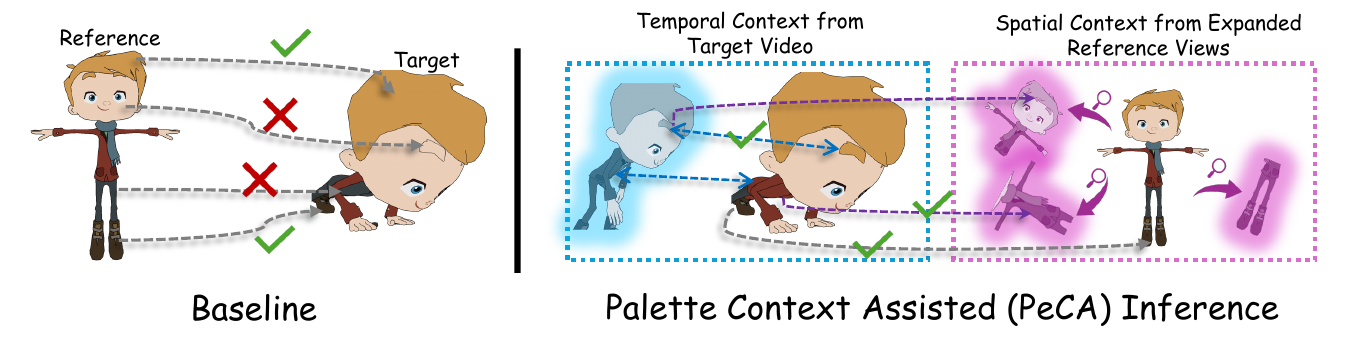}
    \caption{\textbf{Conceptual overview of \textsc{PeCA}.}
    Compared to a baseline that colours regions using isolated correspondences, \textsc{PeCA} leverages \textcolor{RoyalBlue}{temporal} and \textcolor{Plum}{spatial} context to pick supports for a more reliable colourisation.}
    \label{fig:teaser}
\end{figure}

Most segment-based pipelines solve paint-bucket colourisation by region correspondences, which parse regional similarities to colour label estimations either by top matches or a weighted combination of them \cite{maejima2019graph, dai2024learning, casey2021animation, nagata2025dacon, Feng_2025_ICCV}. Although some methods introduce structural or temporal constraints, the per-region assignment is still largely driven by local correspondence scores. In practice, failures are often not due to a complete absence of correct matches, but arise when correspondences are ambiguous or noisy due to view/pose variations in animation videos \cite{10.1023/B:VISI.0000029664.99615.94,nagata2025dacon}. For example, thin fragments under occlusion can be visually ambiguous on their own, and multiple reference regions may look similarly plausible. This ambiguity persists and becomes a bottleneck. As a result, despite being trained on correspondence-based colourisation, such a direct segment matching \& colour propagation pipeline still struggles to deal with spurious correspondences under reference-target appearance gaps \cite{10.1023/B:VISI.0000029664.99615.94,nagata2025dacon}.

A more robust view is that region identity in animation is rarely resolved from a single isolated match. Humans rely on context when understanding colours in art \cite{wittgenstein1977remarks}: both from similar views in reference images, and the temporal continuity in target videos themselves. This suggests a complementary direction to training a stronger backbone: when a direct reference-to-target match is uncertain, spatially similar reference views and temporally neighbouring frames can provide indirect context support that refines ambiguous colour propagation.

To this end, we propose a Palette Context Assisted (\textsc{PeCA}) inference framework that improves segment matching colourisation with test-time context, as shown in \cref{fig:teaser}. \textsc{PeCA} is training-free at inference; it fits well to either trained colourisation models or frozen foundation backbones. It first constructs a target-conditioned support reference bank, so that reference views are expanded to have better coverage of the current target shot, reducing the visual gap with spatially-close context (\cref{sec:are}). We then resolve noisy top correspondences with a soft voting scheme (\cref{sec:pa}) to reach a consensus that blocks spurious correspondences. Finally, we refine per-region assignments across time, leveraging the continuity between adjacent frames while avoiding unreliable transfers via a gated mechanism (\cref{sec:ct}). Together, \textsc{PeCA} turns noisy, isolated segment colourisation into context-aware colour assignments without task-specific training. We summarise our contributions as follows:

\begin{itemize}
    \item We propose a plug-and-play Palette Context Assisted (\textsc{PeCA}) inference framework for paint-bucket colourisation, leveraging context at test-time.
    \item \textsc{PeCA} improves the default inference that struggles with region ambiguity, by aggregating contexts from both spatially-supportive reference views and temporal continuity in animation videos.
    \item Extensive experimental analysis on existing benchmarks and a newly introduced long-shot benchmark show consistent gains on both task-trained and frozen foundation backbones, with particularly larger improvements in training-free settings.
\end{itemize}

\section{Related Works}
\label{sec:related_works}

\subsection{Automated Paint-bucket Colourisation in Animation}

Production paint-bucket colourisation assigns each segmented region a colour from a fixed palette, thus fundamentally relies on region correspondence \cite{nakanishi2013modeling}. Early segment-based methods modelled this problem as geometric or graph-based matching between regions \cite{maejima2019graph,ramassamy2018pre}. These approaches typically assume moderate motion and rely on handcrafted similarity measures or motion cues, which fail to handle large appearance gaps and longer videos in production \cite{Feng_2025_ICCV}.
BasicPBC \cite{dai2024learning} departs from this simple region matching formulation by explicitly modelling topology with inclusion and subset matching to handle split/merge events, and designed propagation strategies tailored to these cases. Feng \etal \cite{Feng_2025_ICCV} further extended this direction with a unified pipeline that augments adjacent-frame matching with temporal-structural constraints and additional refinement modules. Compared with earlier methods, they incorporate more structured matching rules and model-specific post-processing to refine colour propagation. But still, these works primarily operate under a temporal-local assumption, where the target frame and reference frames are in the same video.

Key-frame colourisation task relaxes this assumption and considers arbitrary reference--target pairs, such as design sheets and distant shots. BasicPBC-Ref \cite{dai2024paint} adapts the segment-matching framework to this setting by incorporating stronger semantic features to bridge pose and layout gaps. DACoN \cite{nagata2025dacon} shows that region matching with powerful foundation model descriptors can already serve as a strong baseline for key-frame colourisation, and further improves performance through additional training and model components. Despite such progress, these prior works still report failure modes on extreme views/poses that are visually different from reference shots, and the gains from introducing more reference shots become more marginal as the number of references goes up \cite{nagata2025dacon}. This saturation revealed a universal bottleneck identified in many tasks, when models fail to \textit{find direct correspondences under huge visual gaps} \cite{karaev2024cotracker}.

As a response to this bottleneck, our \textsc{PeCA} builds on the simplest segment-matching formulation. Rather than introducing a stronger model for direct correspondence, we focus on improving robustness from a model-agnostic perspective. By keeping the underlying model unchanged, our approach remains compatible with both trained segment matching models and frozen foundation backbones, and isolates the effect of \textsc{PeCA} reasoning from model-specific design choices.

\subsection{Visual Correspondence by Foundation Models at Test-time}
Large pretrained models provide transferable representations that enable training-free or low-supervision correspondence and region reasoning \cite{radford2021learning,oquab2023dinov2,kirillov2023segment,ravi2024sam}. A common practice is to pool dense features within regions to build local descriptors, then perform similarity-based retrieval for region matching and propagation \cite{schuurmans2018efficient,shlapentokh2024region,kim2023semantic}. Beyond such representation reuse, another line of work~\cite{zhang2021tip,wang2021tent,shu2022tpt,qiao2025v,zhaommicl, zhao2023improving, RTTLC} improves the performance of various downstream tasks (including generic video colourisation) via test-time adaptation, updating model parameters or refinement at inference steps.

In production-oriented animation paint-bucket colourisation, models must generalise to binarised line sketches with sparse appearance cues \cite{dai2024learning, nakanishi2013modeling}. As a result, even strong pretrained region descriptors \cite{radford2021learning,oquab2023dinov2,ravi2024sam} can be brittle when correspondence is solved purely by similarity retrieval without task-specific training \cite{nagata2025dacon}. We thus take a test-time perspective and aim to make inference more reliable with minimal assumptions on the backbone. Instead of treating each region match as an isolated decision, we organise region-level evidence into a palette-space belief and refine it using spatial support from references and temporal support from the target sequence.

\section{Methodology}
\label{sec:methodology}

\subsection{Problem Formulation}
\label{sec:formulation}
We study production-oriented paint-bucket colourisation for hand-drawn animation, where each enclosed region should be assigned a colour from a discrete palette in reference regions. Given a target video clip of $T$ sketch frames $\{I_t\}_{t=1}^{T}$, each frame is partitioned into closed regions $\mathcal{S}_t=\{s_{t,i}\}_{i=1}^{N_t}$. A reference set $\mathcal{R}$ is provided as $\mathcal{R}=\{(I^{(r)}, \mathcal{S}^{(r)}, Y^{(r)})\}_{r=1}^{R}$,
where $Y^{(r)}$ assigns a ground truth colour label (from a finite palette $\mathcal{C}=\{c_1,\dots,c_{|\mathcal{C}|}\}$) to each reference region in $\mathcal{S}^{(r)}$ (segmented from line-sketches $I^{(r)}$ by flood fill~\cite{10.1145/800249.807456}). Our goal is to assign each target region $s_{t,i}$ a colour label $\hat y_{t,i}\in\mathcal{C}$ correctly throughout the video.

Following segment-matching pipelines \cite{nagata2025dacon, schuurmans2018efficient}, we compute a descriptor for each region by average pooling dense features inside its mask in \cref{eq:feature_pooling}, and define cosine similarity 
$S_{(t,i),(r,j)}=\langle \bar{\mathbf{f}}_{t,i},\bar{\mathbf{f}}^{(r)}_j\rangle$. A retrieval-style baseline in previous SOTA \cite{nagata2025dacon} copies the colour label from the top match as in \cref{eq:hardmatch}:
\begin{equation}
\label{eq:feature_pooling}
\resizebox{.9\textwidth}{!}{$
\mathbf{f}_{t,i}=\textrm{AvgPool}\big(\{\phi(I_t)[p]\mid p\in s_{t,i}\}\big),\qquad
\mathbf{f}^{(r)}_{j}=\textrm{AvgPool}\big(\{\phi(I^{(r)})[p]\mid p\in s^{(r)}_{j}\}\big),
$}
\end{equation}
\begin{equation}
\label{eq:hardmatch}
(r^\ast,j^\ast)=\arg\max_{r,j} S_{(t,i),(r,j)},\qquad \hat y_{t,i}=y^{(r^\ast)}_{j^\ast}.
\end{equation}

As discussed in \cref{sec:intro} and \cref{sec:related_works}, direct correspondence probability $S_{(t,i),(r,j)}$ can be ambiguous for colour propagation. This motivates us to propose the Palette Context Assisted (\textsc{PeCA}) test-time reasoning framework that exploits spatial and temporal context of correspondences and colours during inference.

\begin{figure}[t]
    \centering
    \includegraphics[width=\linewidth]{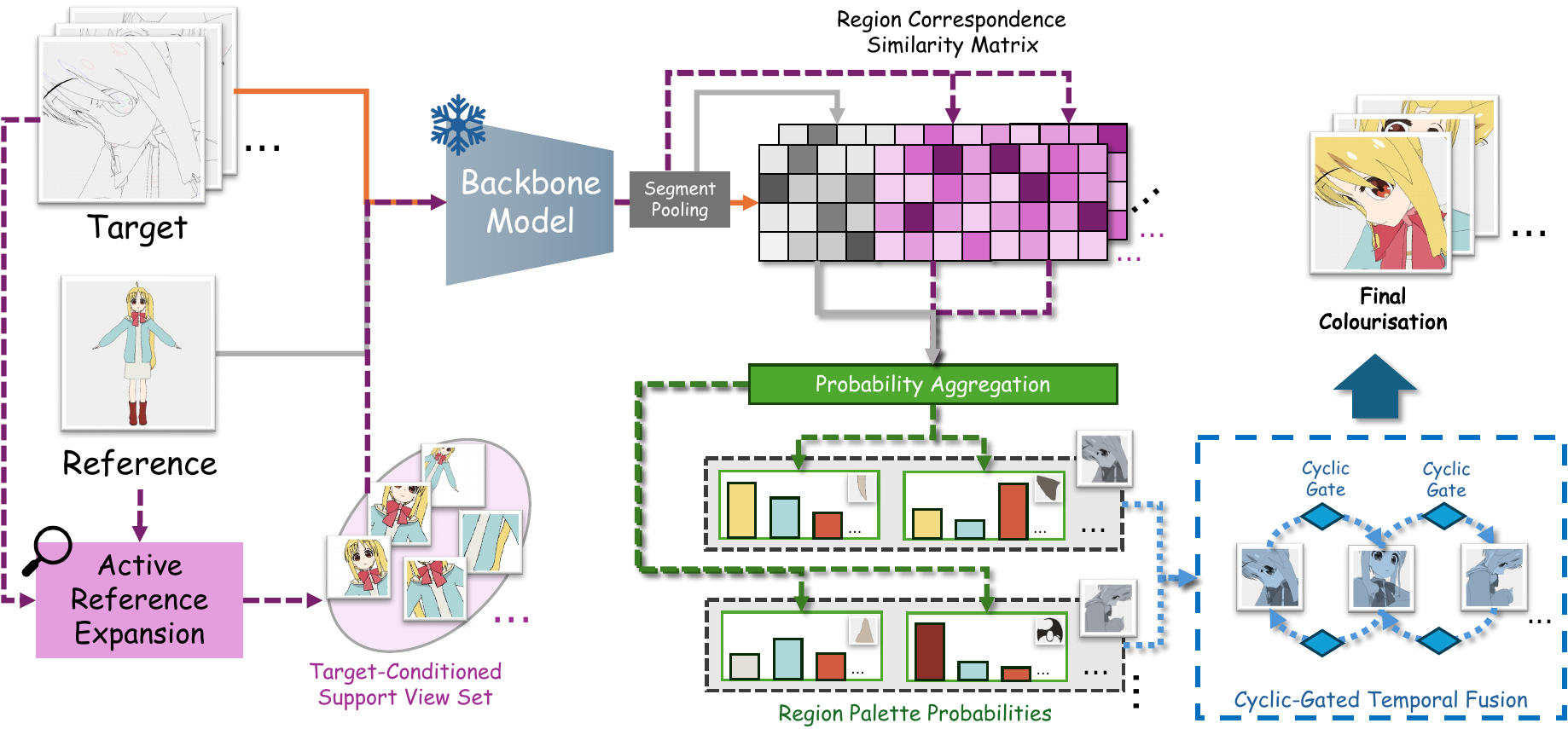}
    \caption{
    \textbf{The proposed \textsc{PeCA} framework overview.}
    \textcolor{Plum}{\textbf{Active Reference Expansion}} (\cref{sec:are}) builds a target-conditioned reference support set.
    \textcolor{ForestGreen}{\textbf{Probability Aggregation}} (\cref{sec:pa}) aggregates noisy matches into per-region palette colour probabilities via soft voting.
    \textcolor{RoyalBlue}{\textbf{Cyclic-gated Temporal Fusion}} (\cref{sec:ct}) fuses colour probabilities across adjacent frames through cycle-consistent temporal links, altogether improving colourisation with \textcolor{Plum}{\textbf{spatial}}, \textcolor{ForestGreen}{\textbf{probabilistic}}, and \textcolor{RoyalBlue}{\textbf{temporal}} context.
    }
    \label{fig:overview}
\end{figure}
\subsection{Palette Context Assisted (\textsc{PeCA}) Framework Overview}
We propose \textsc{PeCA}, a training-free and plug-and-play inference framework that improves region matching-based colourisation by constructing and exploiting context at test time (\cref{fig:overview}). \textsc{PeCA} first builds a target-conditioned support bank by expanding the given reference shots to a limited support set that maximises spatial coverage to the target video (\cref{sec:are}). Then, from the multi-source correspondences, a soft top-$k$ voting converts multiple plausible matches into per-region palette-colour consensus probabilities (\cref{sec:pa}). Finally, we refine these probabilities along reliable correspondence links between adjacent frames serving as temporal context (\cref{sec:ct}).

\subsection{Spatially-Supportive Reference Views Selection}

\label{sec:are}

Prior works \cite{dai2024paint, nagata2025dacon} have shown that increasing the number and diversity of reference shots brings consistent gains for segment-matching-based colourisation, largely because it improves the chance that a target region finds a reference region from a similar layout. In real productions, however, we are often given limited reference RGB images due to the labour-intensive nature of colourisation. In this regard, a natural next step is to expand this limited reference bank with test-time augmentation \cite{shanmugam2021better}, generating new views to reference shots that possibly provide easier ``shortcut'' matchings for colour propagation. 

\begin{wrapfigure}[18]{r}{0.43\linewidth} 
\centering 
\includegraphics[width=\linewidth]{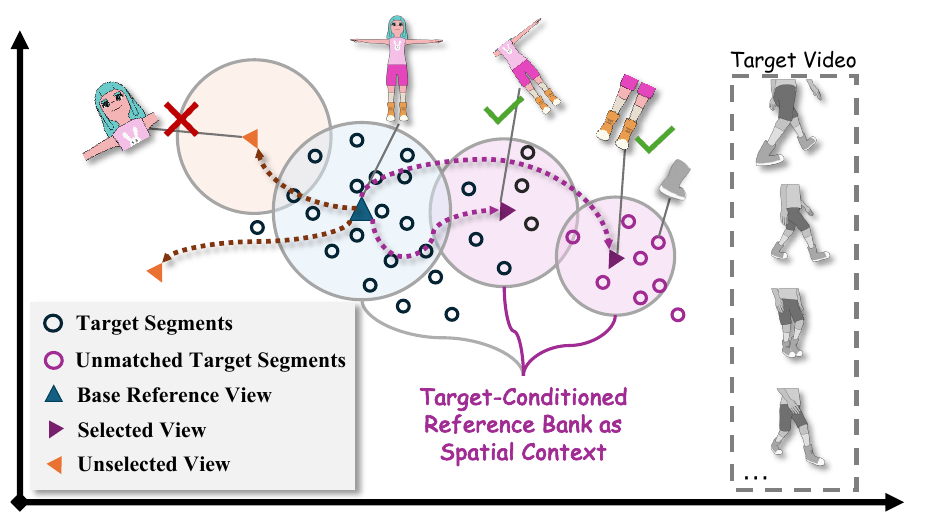} 
\caption{\textbf{Active Reference Expansion.} Regions are encoded as features (\textbf{$\circ$}). There are target segments that hardly match the reference \textcolor{Plum}{(\textbf{$\circ$)}}. Selecting \textcolor{Plum}{Target-conditioned views ($\blacktriangleright$)}, can maximise coverage to targets, while \textcolor{Orange}{unselected views $\blacktriangleleft$} provide limited support to target video.} 
\label{fig:are_vis}  
\end{wrapfigure}

However, naive test-time augmentation is unlikely to scale efficiently at inference time under this setting \cite{shorten2019surveytru}. In basic inference pipelines \cite{casey2021animation,nagata2025dacon}, region assignment is resolved from a similarity matrix over all reference-region candidates. Naively augmenting views, which increases the candidate set indiscriminately, may provide useful evidence. But the additional views may also increase exposure to high-similarity distractors \cite{radovanovic2010hubs}. This makes top matches less stable for ambiguous regions, resulting in a limited performance gain (see \cref{tab:ablation_merged}). This also aligns with a trend in previous works where further adding more references (>5) shows marginal gains \cite{nagata2025dacon}.

We therefore designed active reference expansion (\cref{fig:are_vis}) to make additional reference views both {budgeted} and {target-aware}, forming a \textcolor{Plum}{spatial context} that best covers the target regions. Starting from the original reference views $\mathcal{V}_0$, we first generate an augmented candidate pool $\mathcal{V}_{\mathrm{aug}}$ by applying geometric transformations (flips, rotations, affine transforms; details in \suppref{sec:are_supp}). Rather than keeping all candidates, we select only $B$ views that best {support} the target video in the feature space, so the bank is strengthened without indiscriminately enlarging the region candidate set. Specifically, for each candidate view $v\in\mathcal{V}_{\mathrm{aug}}$ with region descriptors $\{\mathbf{f}^{(v)}_j\}_{j=1}^{N_v}$, we measure its support to a target region $(t,i)$ by the similarity of its best-matching region:
\begin{equation}
\text{score}_v(t,i) \;=\; \max_{1\le j\le N_v}\; \left\langle \bar{\mathbf{f}}_{t,i},\, \bar{\mathbf{f}}^{(v)}_j \right\rangle.
\end{equation}

We expect the expansion to keep a subset $V\subseteq\mathcal{V}_{\mathrm{aug}}$, such that the resulting support for $(t,i)$ is $\max_{v\in V}\text{score}_v(t,i)$, \ie the best support offered by selected views. We therefore choose $B$ views by maximising a facility-location objective over target regions in uniformly sampled frames $\mathcal{T}_s \subset \{1,\dots,T\}$:
\begin{equation}
\label{eq:facility}
F(V)\;=\;\sum_{t\in\mathcal{T}_s}\sum_{i=1}^{N_t}\max_{v\in V} \text{score}_v(t,i),
\qquad
\max_{V\subseteq\mathcal{V}_{\mathrm{aug}}} F(V)\ \text{s.t.}\ |V| = B.
\end{equation}

In practice, we form an augmented candidate pool of size $|\mathcal{V}_{\mathrm{aug}}|=mB$ and select $B$ views from it. This selection is performed once per target video, with overhead depending only on $mB$ (see \suppref{sec:are_supp} and \suppref{sec:runtime} for details). Since $F(V)$ is monotone submodular, a greedy algorithm provides a standard approximation guarantee \cite{krause2014submodular}. The resulting support bank $\mathcal{V}_0 \cup V$ improves target-shot coverage under a fixed budget by actively prioritising views that best support the current video. Importantly, this selection step controls the exposure to distractor regions, yielding a more relevant candidate set for subsequent matching. On the other hand, with more reference shots, naturally, for each target region, it introduces additional correspondence candidates from different views. This motivates our next step, which votes multiple high-confidence correspondences to palette colour probabilities (\cref{sec:pa}).

\subsection{Correspondence Candidate Voting for Colour Palette}
\label{sec:pa}

When more than one reference shot is available, a target region is naturally supported by multiple high-confidence candidates from different sources. To better utilise such \textcolor{ForestGreen}{probabilistic context}, instead of committing to top-1 retrieved region \cite{nagata2025dacon} or full linear combinations \cite{casey2021animation}, we wish to resolve the correspondences to colour estimations by \textit{soft-voting} over the top-$k$ correspondence hypotheses. This produces per-region colour consensus probability over the finite palette from the relevant correspondence probabilistic context.

Specifically, for each target region $(t,i)$, let $\mathcal{N}_k(t,i)$ denote the top-$k$ candidate reference regions under similarity $S_{(t,i),(r,j)}$. We convert similarities into normalised weights with temperature $\tau \in (0,1]$:
\begin{equation}
\label{eq:pa_soft}
p_{t,i}(r,j)=
\frac{\exp\big(S_{(t,i),(r,j)}/\tau\big)}
{\sum\limits_{(r',j')\in \mathcal{N}_k(t,i)} \exp\big(S_{(t,i),(r',j')}/\tau\big)},
\qquad (r,j)\in \mathcal{N}_k(t,i).
\end{equation}

We then vote these weighted matches into the palette colour space:

\begin{equation}
\label{eq:pa}
P_{t,i}(c)=\sum_{(r,j)\in\mathcal{N}_k(t,i)} p_{t,i}(r,j)\cdot \mathds{1}\!\left[y^{(r)}_j=c\right],
\qquad c\in\mathcal{C}.
\end{equation}

Here, $P_{t,i}\in\Delta^{|\mathcal{C}|}$ summarises the correspondence evidence as colour-level probabilities, where $\Delta^{|\mathcal{C}|}$ denotes the probability simplex over $|\mathcal{C}|$ colour entries. The temperature $\tau$ controls the sharpness of this soft vote. When $k{=}1$ and $\tau\!\rightarrow\!0$, it reduces to hard copying as in baseline, while moderately larger $k$ or $\tau$ pool evidence across matches and reduce sensitivity to spurious correspondence.

Serving as an interface from region correspondence to per-region colourisation results, the aggregation has two practical benefits for spatial/temporal context: First, restricting aggregation to top-$k$ candidates avoids the dilution effect of naive global mixing when the candidate pool gets larger after active expansion in \cref{sec:are}, leading to robust colourisations from soft voting by multiple candidates. Second, colour probabilities live in a fixed label simplex shared by all frames, making them directly comparable and therefore suitable for supporting each other as temporal contexts in \cref{sec:ct}, unlike matching probabilities that may change across regions with the same colour identity but in different frame pairs.

\subsection{Temporal Cycle-Consistency for Colour Refinement}
\label{sec:ct}

\begin{wrapfigure}[17]{r}{0.45\linewidth} 
\centering 
\includegraphics[width=\linewidth]{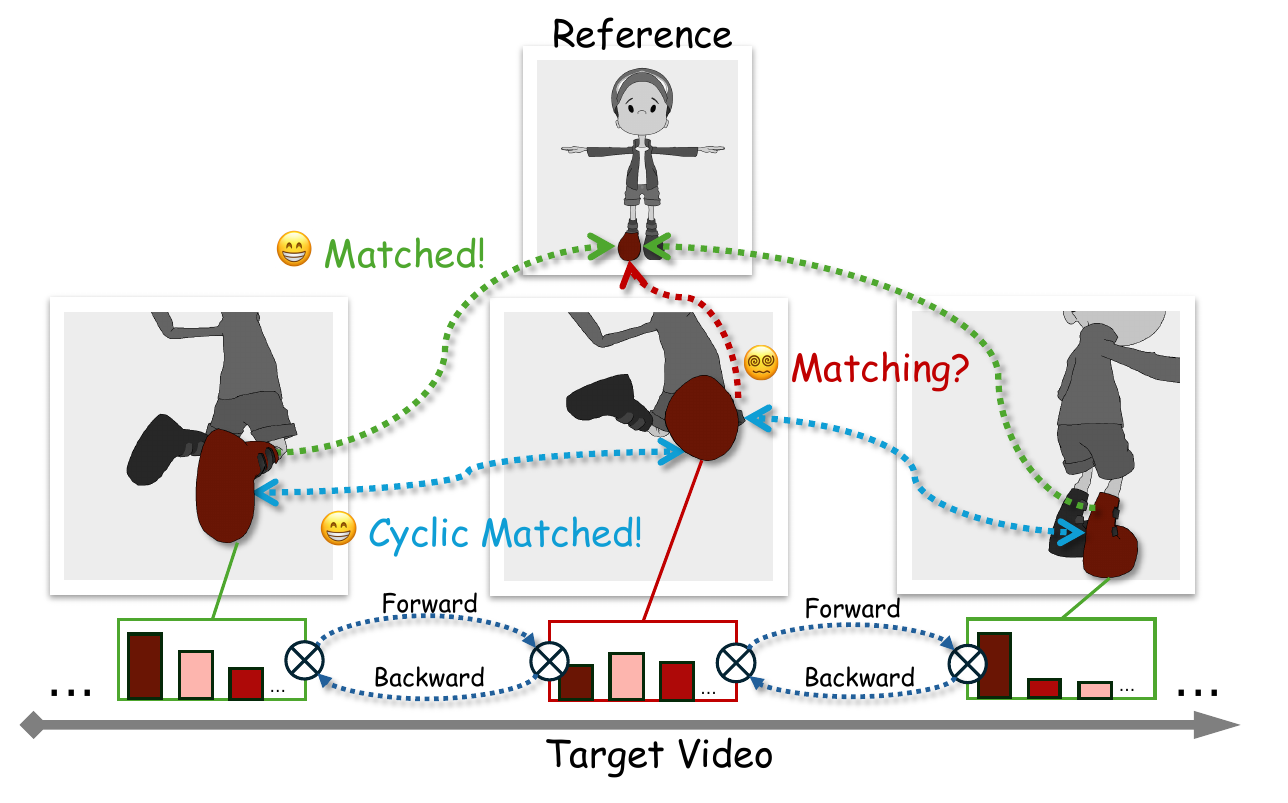} 
\caption{\textbf{Cyclic-Gated Temporal Fusion} utilises temporal context by fusing per-region colour probabilities only along cycle-consistent matches between adjacent frames, refining colours while avoiding unreliable temporal fusions.}
\label{fig:ct_vis} 
\end{wrapfigure}

Now each frame has per-region palette-colour probabilities constructed from a strengthened reference support. In addition, animation videos also provide a temporal cue, in which adjacent frames within the same video look more similar due to temporal continuity of videos \cite{tang2025generative, wang2019learning, carreira2017quo}. Motivated by this, if a direct match between reference and target failed, an easier transitive colour-propagation shortcut can be parsed from the adjacent frames' correspondences (a \textcolor{RoyalBlue}{temporal context}) to refine the colourisation.

However, region matching is not always reliable under changes between adjacent frames. If we indiscriminately fuse information across time, a single spurious match can be propagated to other frames and amplified. We therefore refine colour probabilities only along {cycle-consistent} temporal links, so that temporal context is used when it is reliable and ignored otherwise, as shown in \cref{fig:ct_vis}. 

Specifically, between adjacent frames, we compute adjacent-frame region similarity
$A_t[i,j]=\langle \bar{\mathbf{f}}_{t,i},\bar{\mathbf{f}}_{t-1,j}\rangle$.
We define the forward nearest-neighbour match from frame $t$ to $t{-}1$ as $\pi_t(i) \;=\; \arg\max_{j} A_t[i,j],~~~i\in\{1,\dots,N_t\},$ and symmetrically, the backward match from frame $t{-}1$ to $t$ as $\rho_t(j) \;=\; \arg\max_{i} A_t[i,j],~~~j\in\{1,\dots,N_{t-1}\}$.
We treat a temporal link as {cyclic stable} only if it is bidirectional:
\begin{equation}
\label{eq:cycle}
(t,i)\ \text{is stable if}\ \rho_t\!\big(\pi_t(i)\big)=i.
\end{equation}

This cycle-check conservatively filters unreliable correspondences, which is crucial as the matching between adjacent frames can still be noisy (see \cref{tab:ablation_internal_extra_kf}). For cyclic stable links, we fuse colour probabilities with a product update:
\begin{equation}
\label{eq:ct}
\tilde P_{t,i}(c)\propto P_{t,i}(c)\cdot P_{t-1,\pi_t(i)}(c),\qquad
P_{t,i}\leftarrow \textrm{Normalise}(\tilde P_{t,i}),\quad c\in\mathcal{C}.
\end{equation}

Intuitively, Eq.~\eqref{eq:ct} reinforces colour propagation for hard cases by leveraging palette probabilities carried by regions in other frames of the same video, while a cycle-consistency gate prevents error propagation. We perform one forward sweep ($t=2\!\rightarrow\!T$) and one backward sweep ($t=T-1\!\rightarrow\!1$). In the backward sweep, we apply the same matching, cycle check, and fusion with indices swapped. In practice, the two passes are complementary: each direction conditions on different neighbouring frames that may already have more reliable estimates, so bidirectional sweeping improves robustness and yields more stable colourisation.

\section{Experiments}
\label{sec:experiments}

\subsection{Experiment Setup}
\label{sec:exp_setup}
We evaluate on the existing PaintBucket benchmarks \cite{dai2024learning, dai2024paint} sourced from CG-Rendered and hand-drawn videos. We further test our method against previous works on a newly constructed long-shot dataset, with 10--20$\times$ the length of target video, compared to previous datasets. Further experiments of segment matching on generic video dataset and ablations are provided in \suppref{sec:vipseg_extension}.

{\textbf{PaintBucket-Character (PBC-3D)}} dataset contains 22 characters with 9--16 reference design-sheets per character; following prior work~\cite{dai2024paint,nagata2025dacon}, we report results on the official test split of 3,000 frames.

{\textbf{PaintBucket-Real (PBC-Real)}} is a hand-drawn test set collected from professional animation, with 200 frames in total (20 short clips). Since it is not provided with design-sheet references, we use the first coloured frame from the clips as references, following previous works \cite{nagata2025dacon, dai2024learning, Feng_2025_ICCV}. 

{\textbf{Anita-Pirate.}} We annotated a new long-shot benchmark featuring a hand-drawn 206-frame sequence, with raw data from the Anita dataset~\cite{Anita2024}, which is more challenging than PBC-Real due to its 10--20$\times$ longer time horizon with roughly 140 segments per frame. It comes with frames with complex region layouts over a long time horizon. This new challenging test set is available on our project page, with annotation protocol and licensing details in \suppref{sec:pirate-detail}.

\textbf{Evaluation protocols:} We consider several production-relevant settings. Firstly, we evaluate standard {design-sheet key-frame colourisation}, following previous works \cite{dai2024paint, nagata2025dacon}, in \cref{tab:pbc_kf_merged}. This setting provides a few coloured design sheets without assuming any temporal relation to the target video, and is therefore the most general reference protocol \cite{dai2024paint,nagata2025dacon}.  We also evaluate video colourisation under the same-video key-frame colourisation, in which references are sampled from the target video itself. Note that this is the only reference type available for PBC-Real and Anita-Pirate without design sheets. We report two protocols from prior works. 1) First-frame reference (\cref{tab:consecutive_one_shot}): only the first frame is given as an RGB reference \cite{nagata2025dacon}. 2) Two-sided reference (in-between): only the first and last frames are given as RGB references \cite{Feng_2025_ICCV}  (\cref{tab:inbetweening_continuous}). Following previous works \cite{dai2024learning, nagata2025dacon, Feng_2025_ICCV, dai2024paint}, we report both segment- and pixel-level metrics: \texttt{Acc}, \texttt{Acc-Thresh} (segments$>$10 pixels), \texttt{Pix-Acc}, \texttt{Pix-F-Acc} (for foreground), and \texttt{Pix-B-MIoU} (for background). All metrics are reported in percentages, and larger values mean better performance. See \suppref{sec:metrics} for further details.

\textbf{Implementation details:} We compare against representative pixel-generative methods \cite{zhuang2024colorflow, liu2025manganinja, meng2025anidoc} and segment-based pipelines \cite{nagata2025dacon, dai2024paint, dai2024learning} under official protocols~\cite{dai2024paint,nagata2025dacon}. We build a \emph{Base} inference for SOTA \cite{nagata2025dacon} and frozen foundation models that perform colourisation using region correspondence, following prior works. We then apply \textsc{PeCA} as a plug-and-play inference framework to various backbone models, either with or without colourisation task training, and fixed hyperparameters (top-$k{=}64$, $\tau{=}0.05$ and \#views $B=31$, $m{=}4$.) across all experiments. Further implementation details, computational costs, and hyperparameter settings are provided in \suppref{sec:imple}.

\subsection{Main Experimental Results}
\label{sec:main_results}

\begin{figure*}[h!]
    \centering
    \setlength{\tabcolsep}{1pt}
    \renewcommand{\arraystretch}{1.0}

    \begin{subfigure}[t]{0.9\textwidth}
        \centering
        \resizebox{\textwidth}{!}{%
        \begin{tabular}{c c c c c c}
            \tiny \textbf{Reference} &
            \tiny \textbf{Target} &
            \tiny \textbf{BasicPBC-Ref~\cite{dai2024paint}} &
            \tiny \textbf{DACoN~1.1~\cite{nagata2025dacon}} &
            \tiny \textbf{Ours} &
            \tiny \textbf{Groundtruth} \\[-2pt]

            \includegraphics[width=0.16\textwidth]{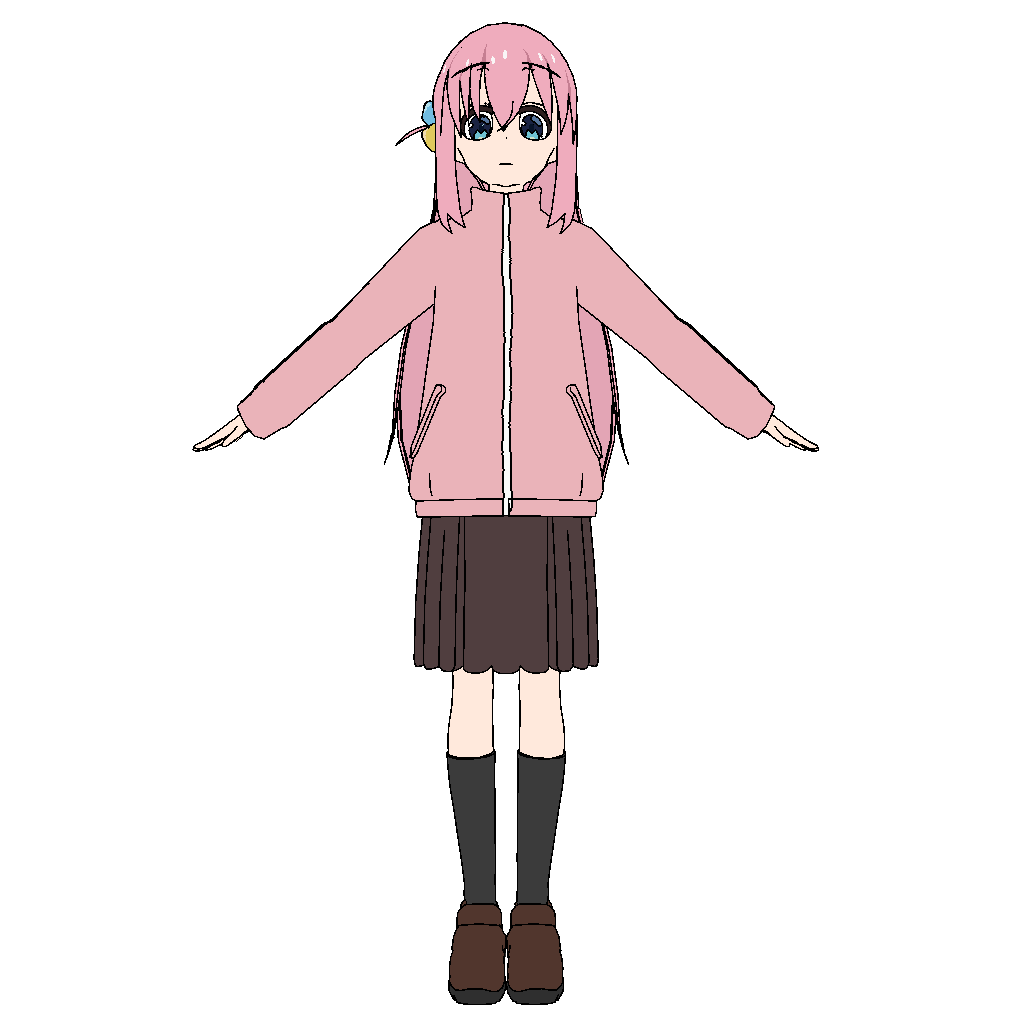} &
            \includegraphics[width=0.16\textwidth]{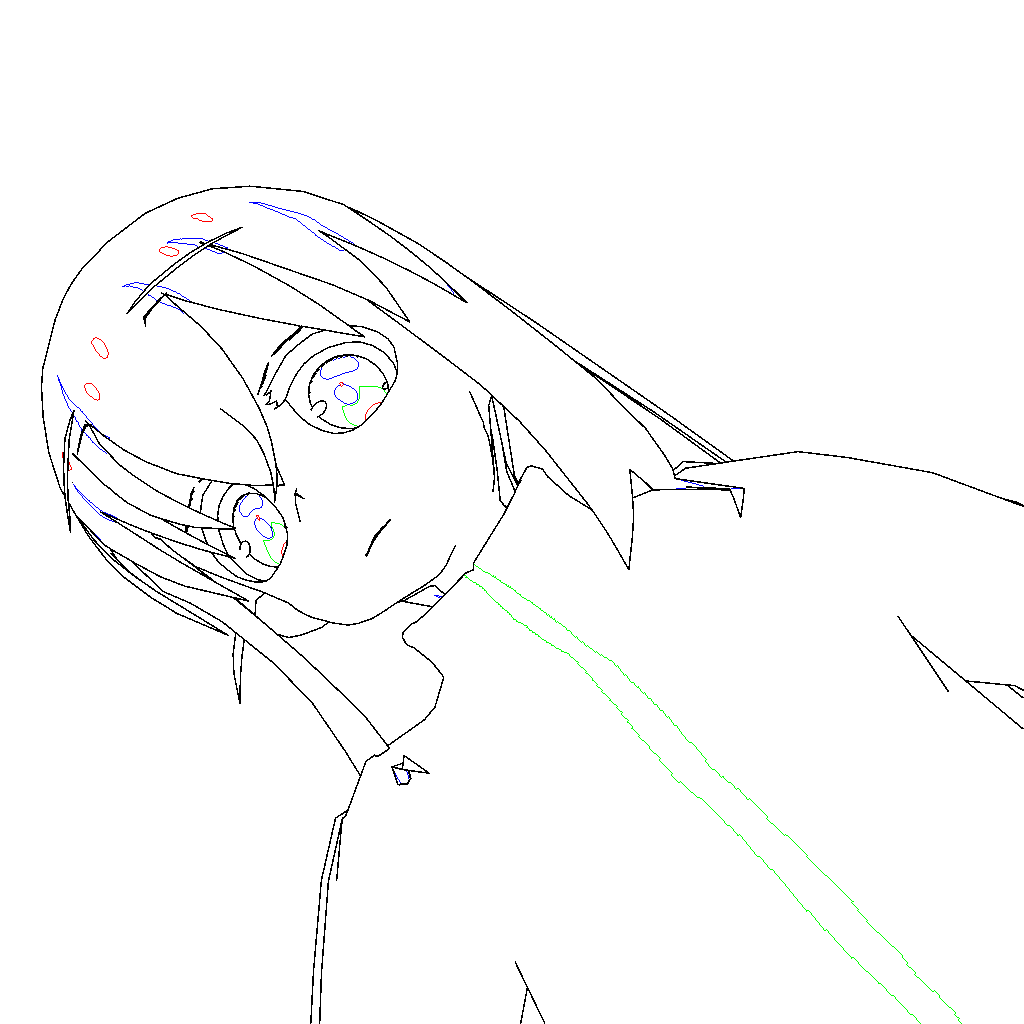} &
            \includegraphics[width=0.16\textwidth]{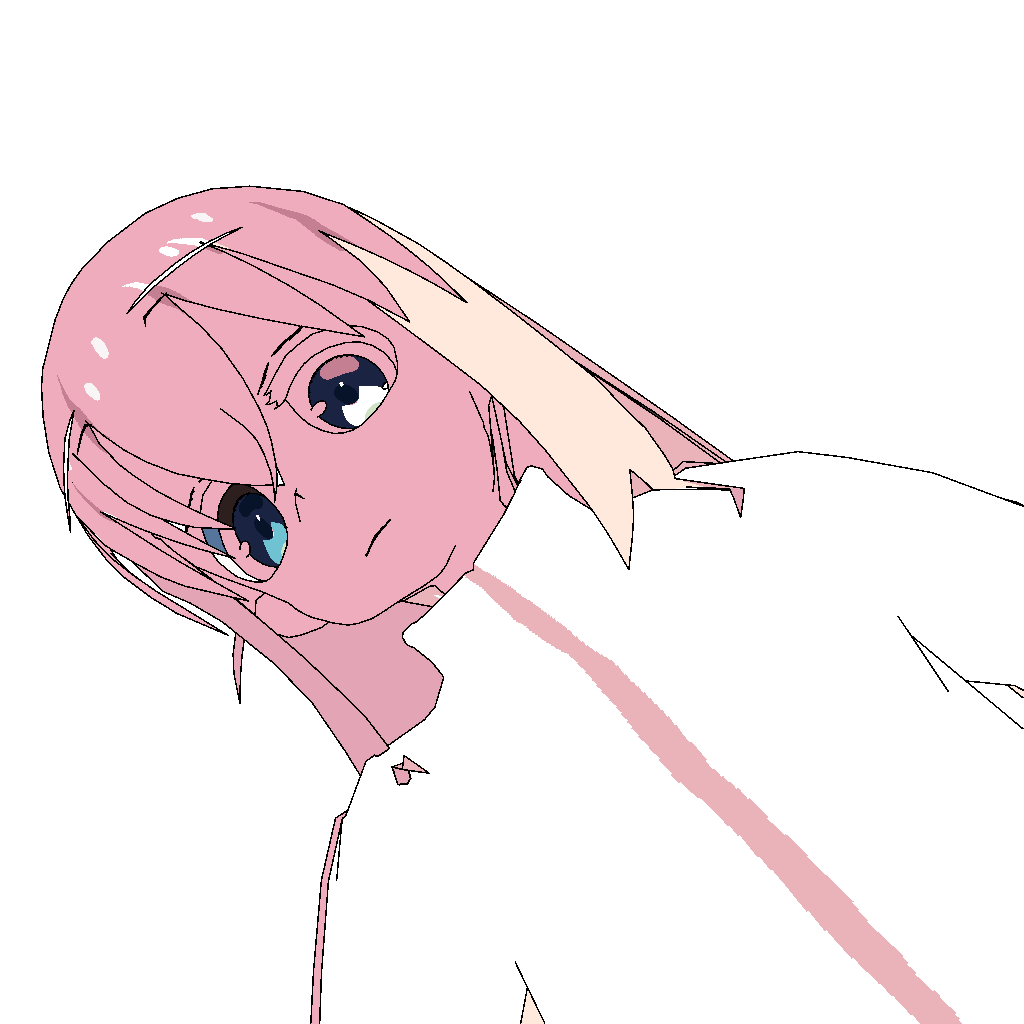} &
            \includegraphics[width=0.16\textwidth]{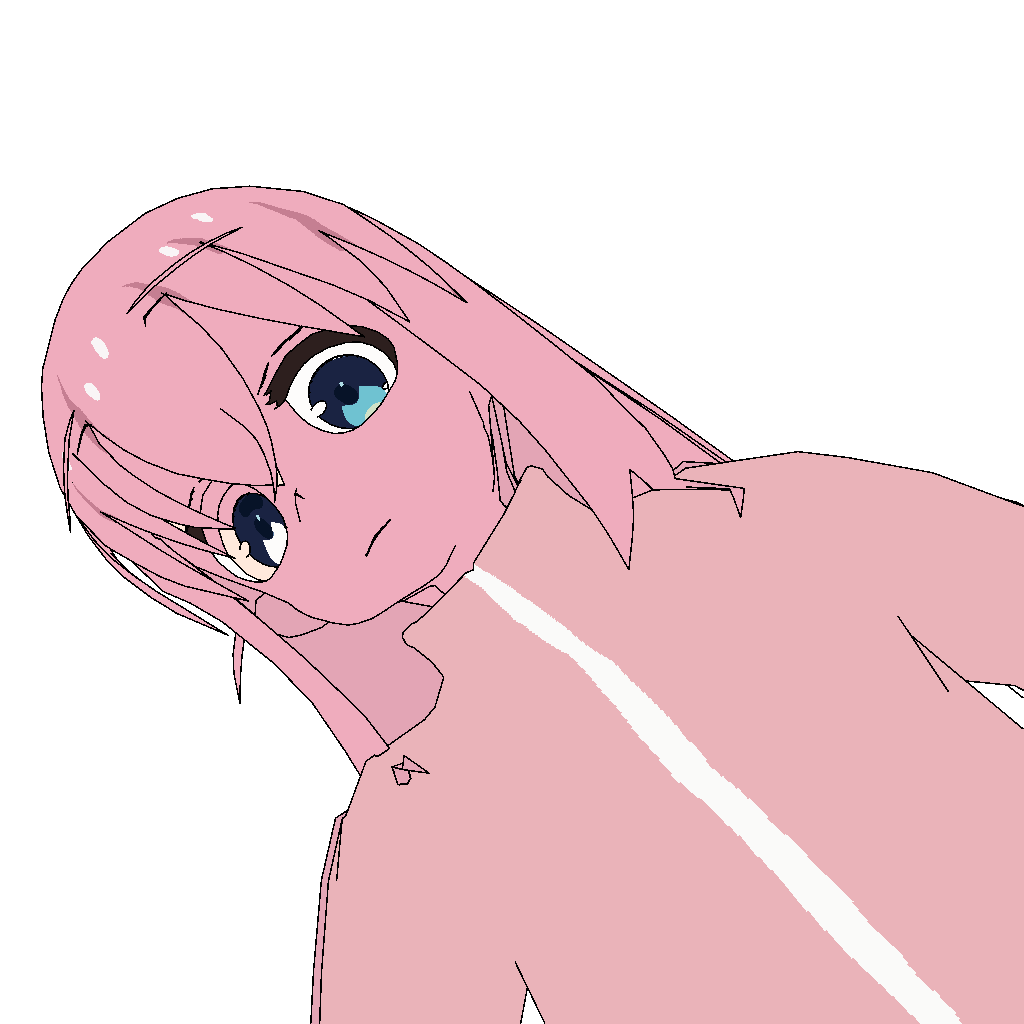} &
            \includegraphics[width=0.16\textwidth]{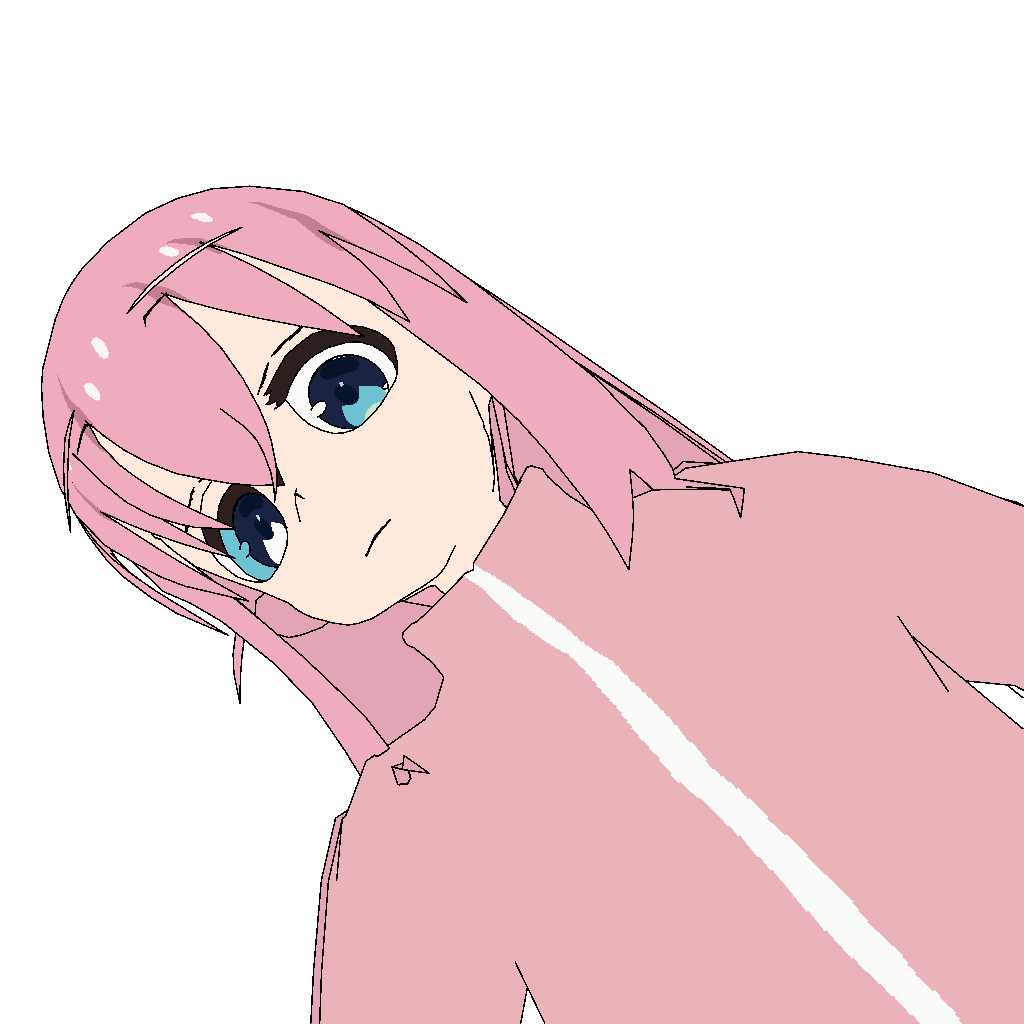} &
            \includegraphics[width=0.16\textwidth]{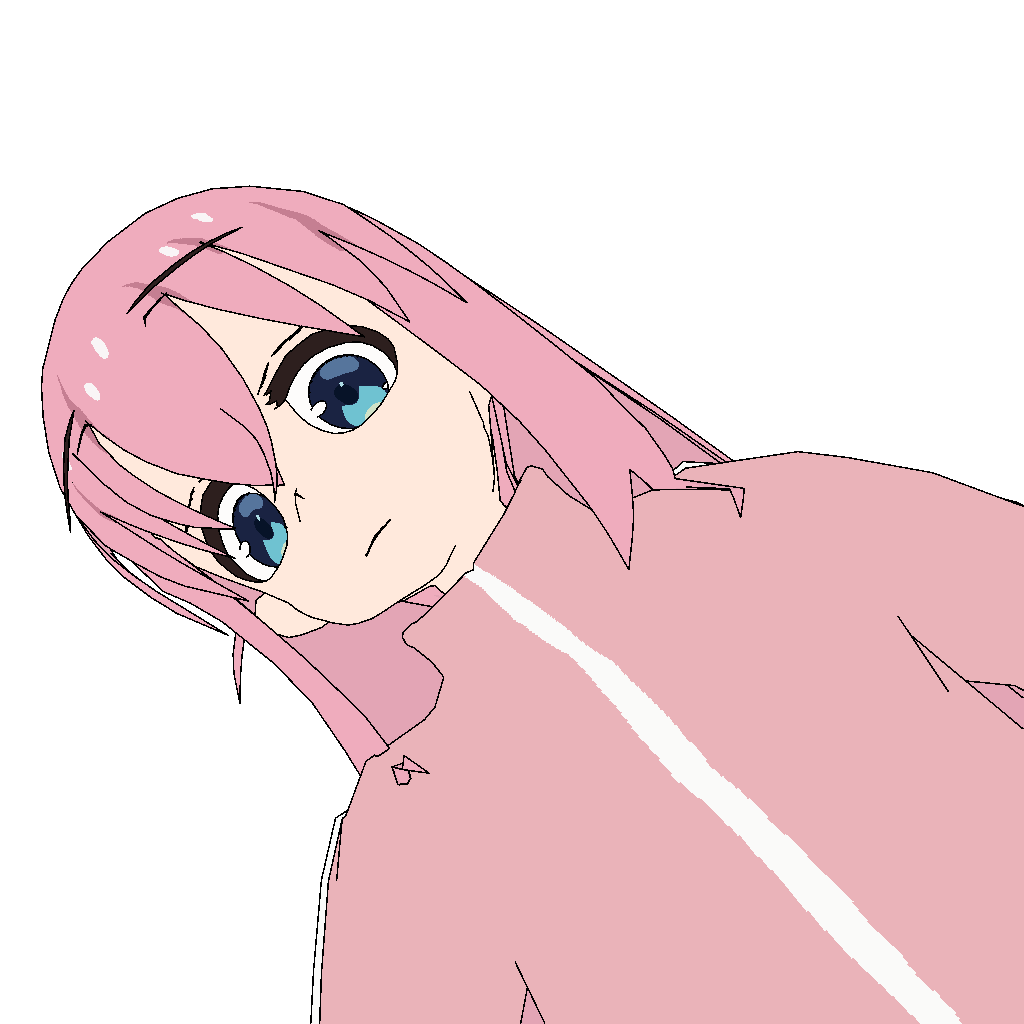} \\

            \includegraphics[width=0.16\textwidth]{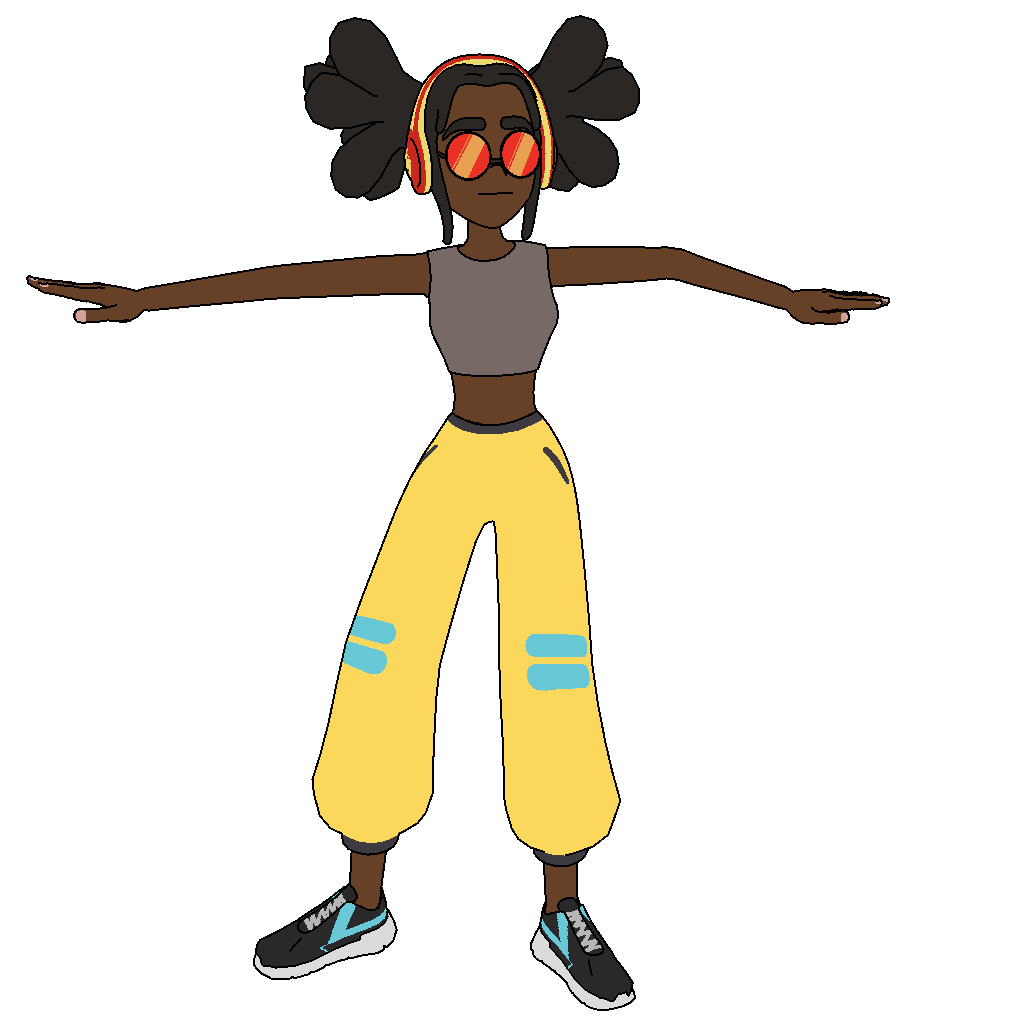} &
            \includegraphics[width=0.16\textwidth]{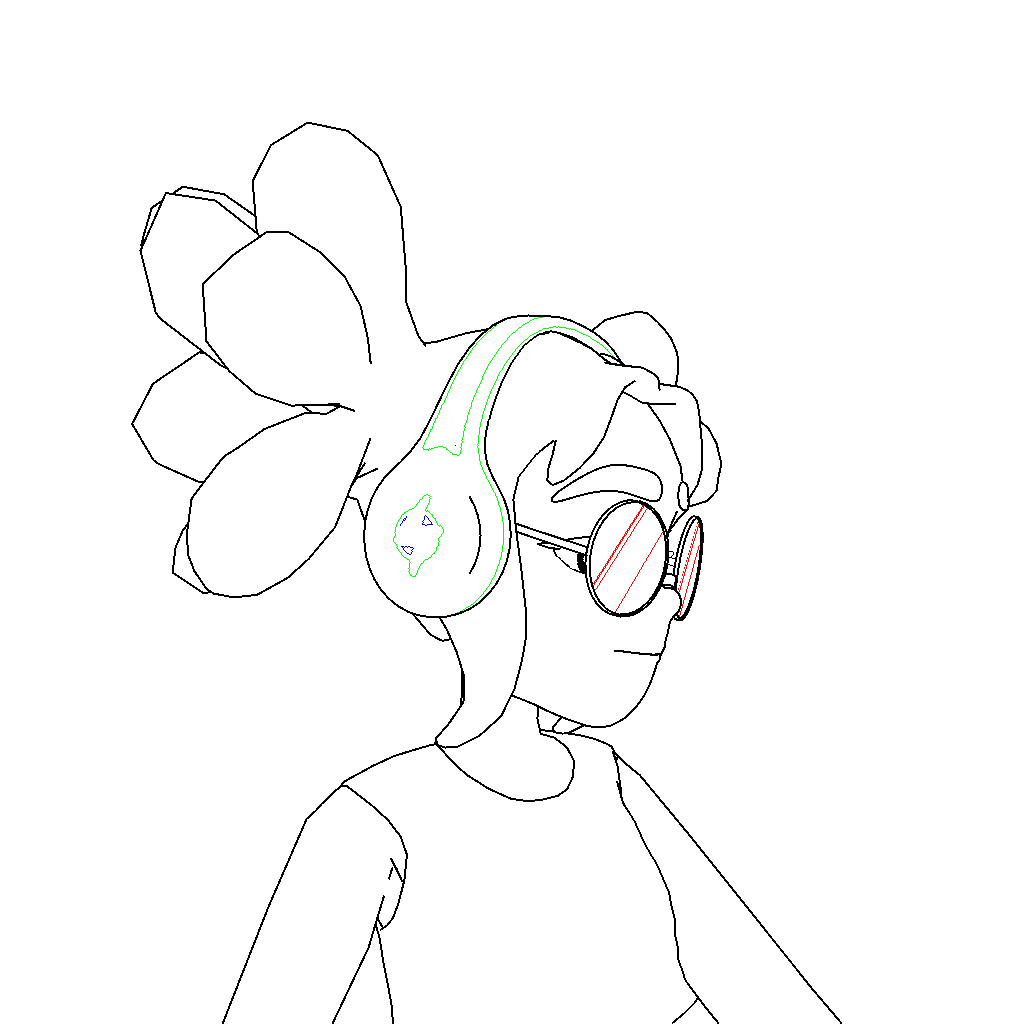} &
            \includegraphics[width=0.16\textwidth]{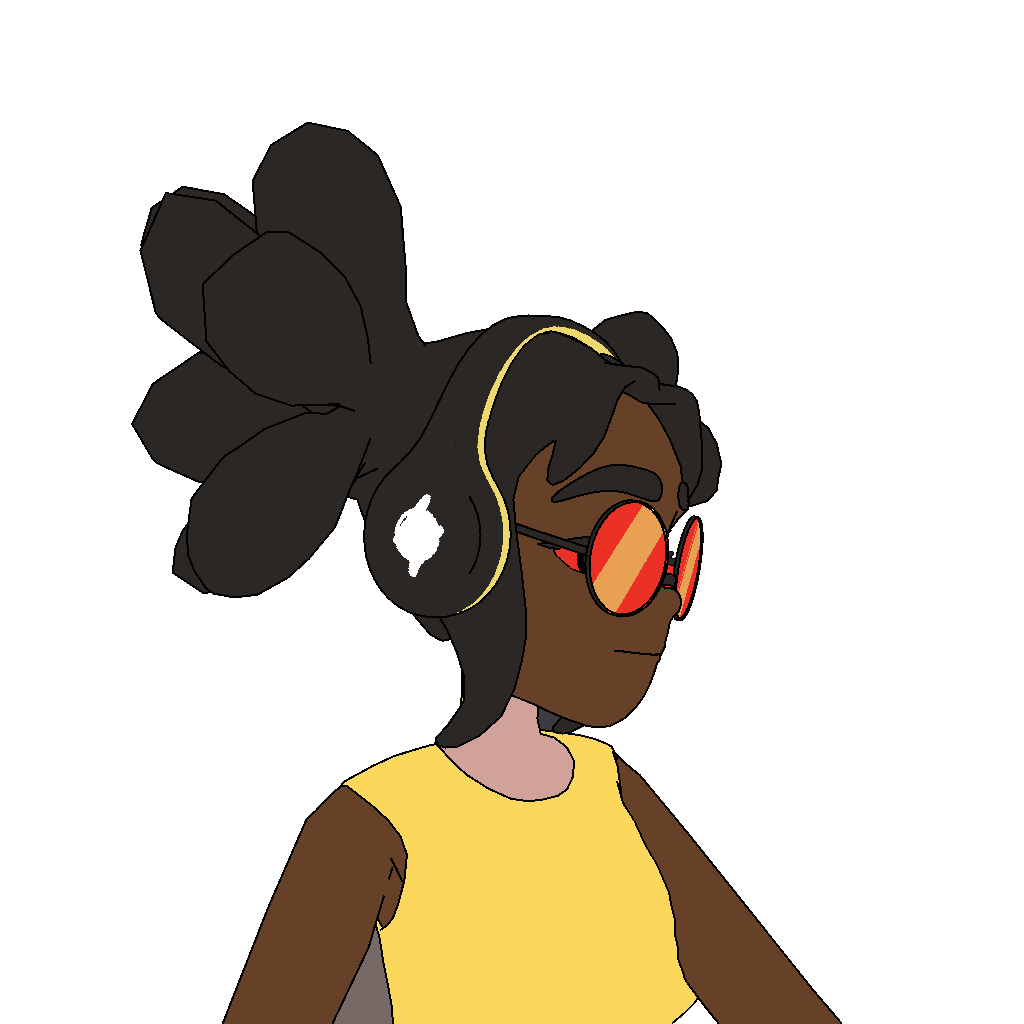} &
            \includegraphics[width=0.16\textwidth]{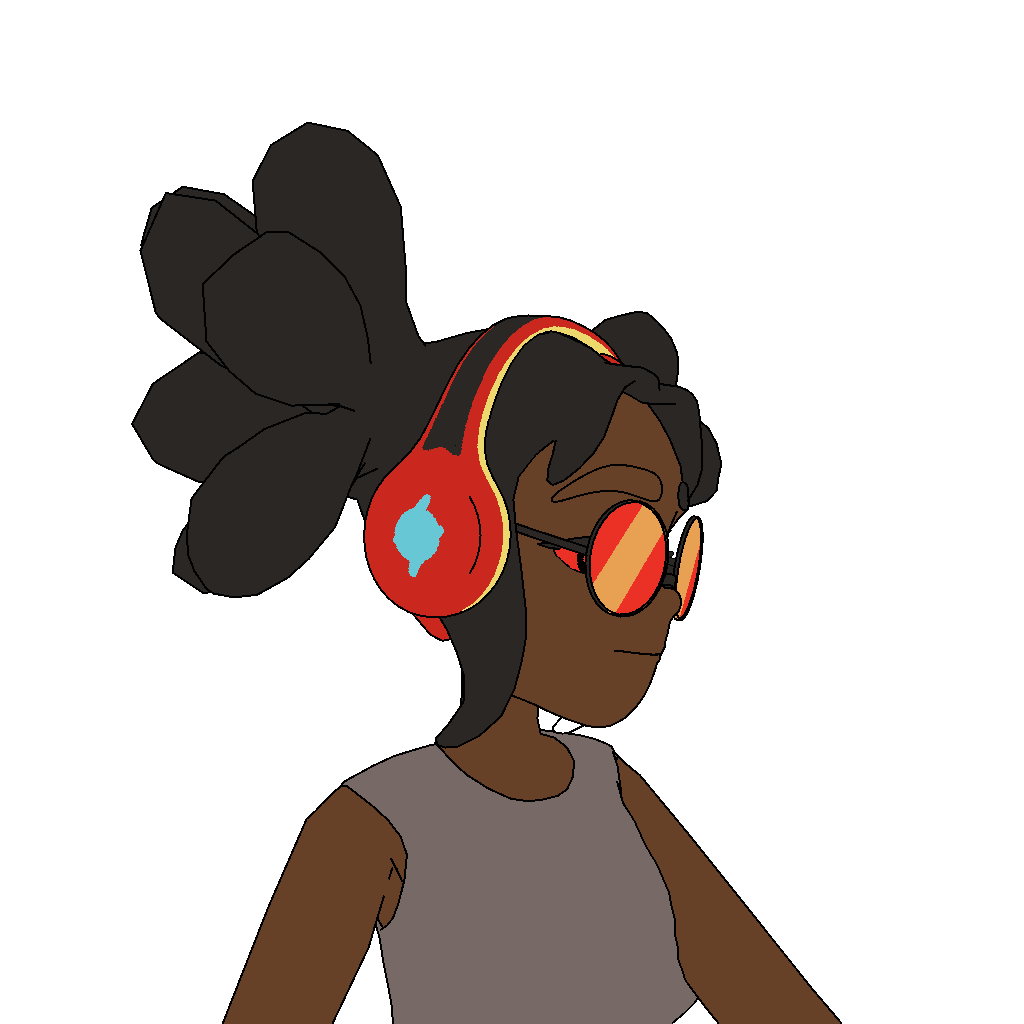} &
            \includegraphics[width=0.16\textwidth]{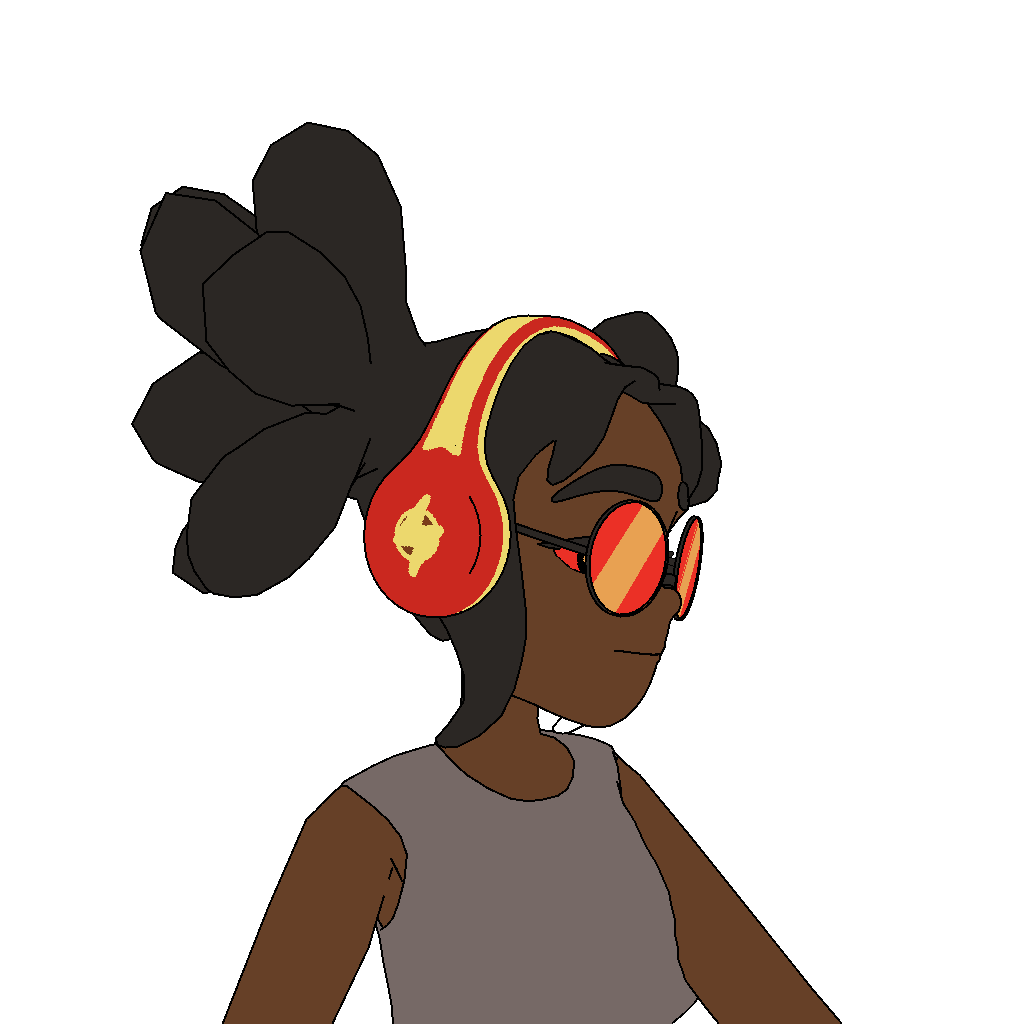} &
            \includegraphics[width=0.16\textwidth]{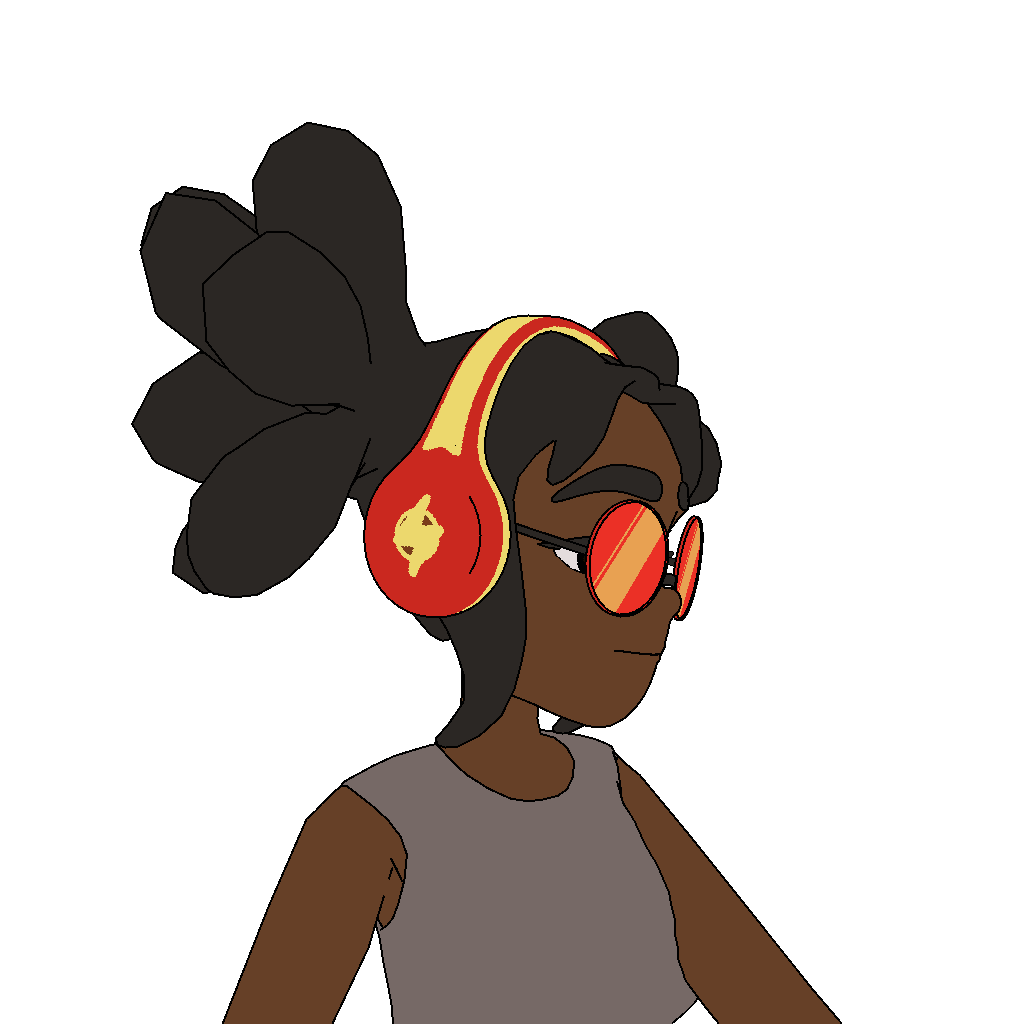} \\
        \end{tabular}%
        }

        \caption{Key-frame (Design-sheet) Reference Colourisation Results}
        \label{fig:qual_keyframe}
    \end{subfigure}
    \begin{subfigure}[t]{\textwidth}
        \centering
        \resizebox{0.9\textwidth}{!}{%
        \begin{tabular}{c c c c c c}
            \tiny \textbf{Reference} &
            \tiny \textbf{Target} &
            \tiny \textbf{BasicPBC~\cite{dai2024learning}} &
            \tiny \textbf{DACoN~1.1~\cite{nagata2025dacon}} &
            \tiny \textbf{Ours} &
            \tiny \textbf{Groundtruth} \\[-2pt]

            \includegraphics[width=0.16\textwidth]{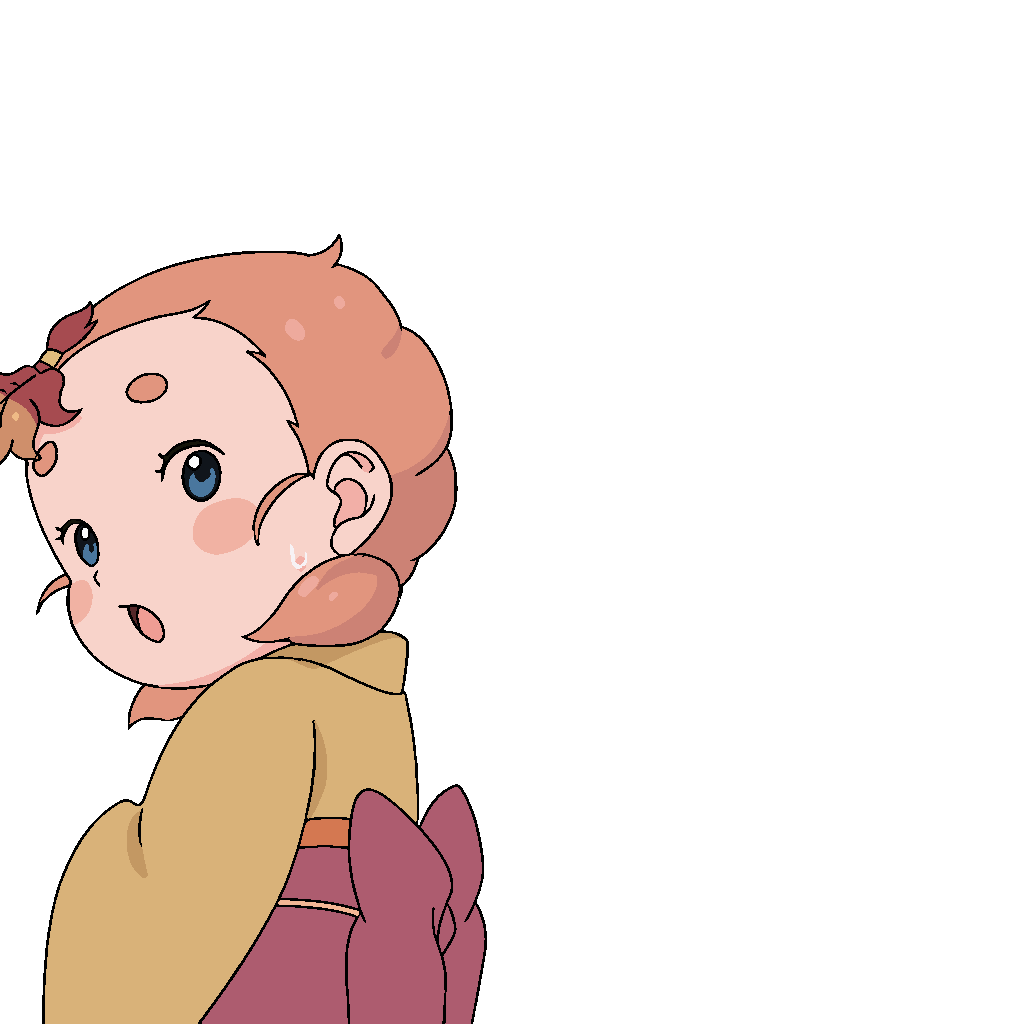} &
            \includegraphics[width=0.16\textwidth]{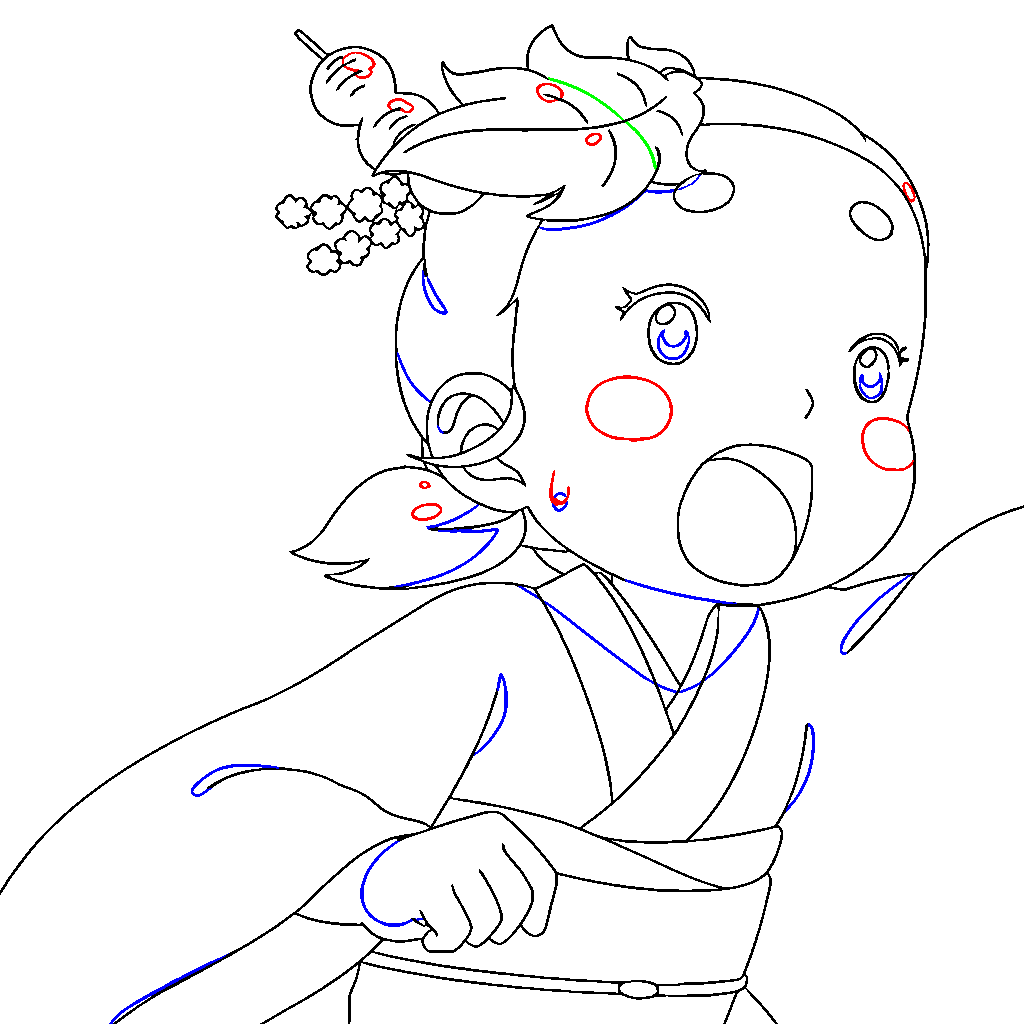} &
            \includegraphics[width=0.16\textwidth]{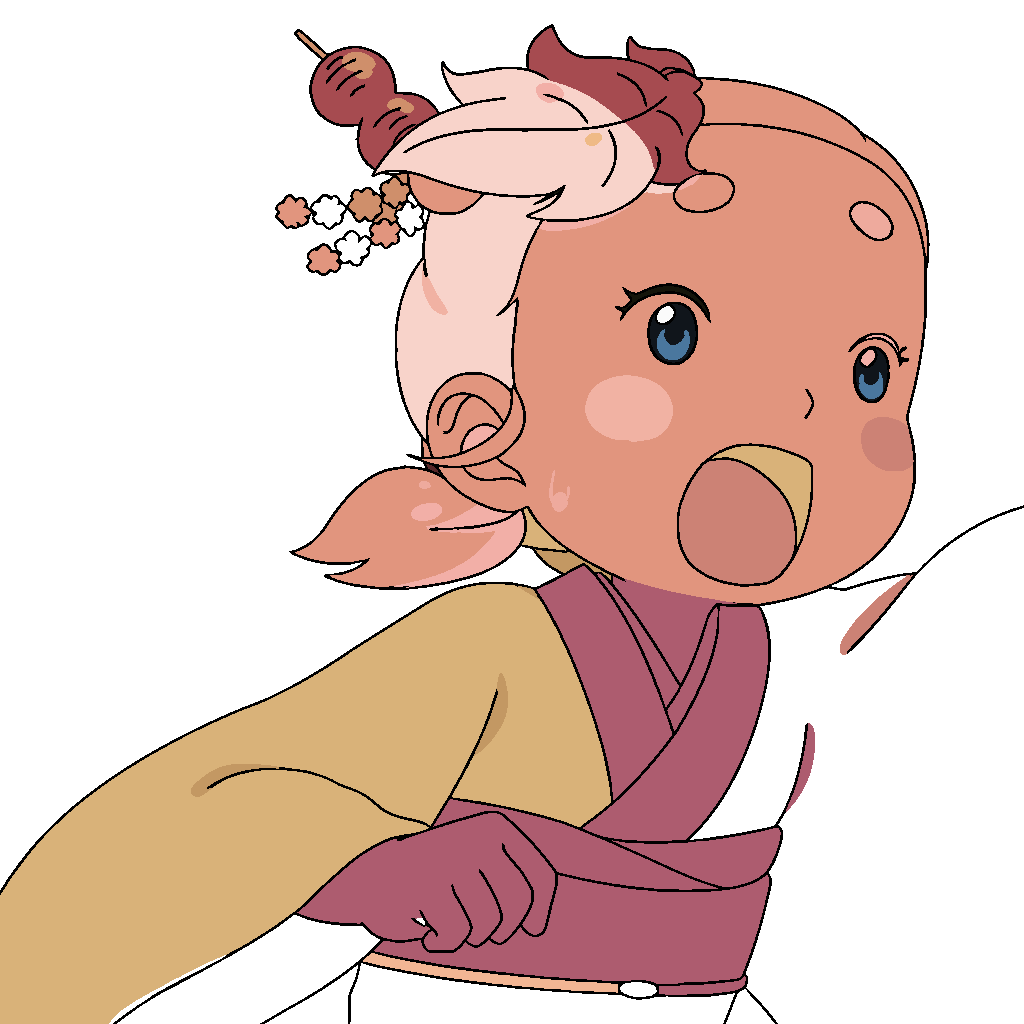} &
            \includegraphics[width=0.16\textwidth]{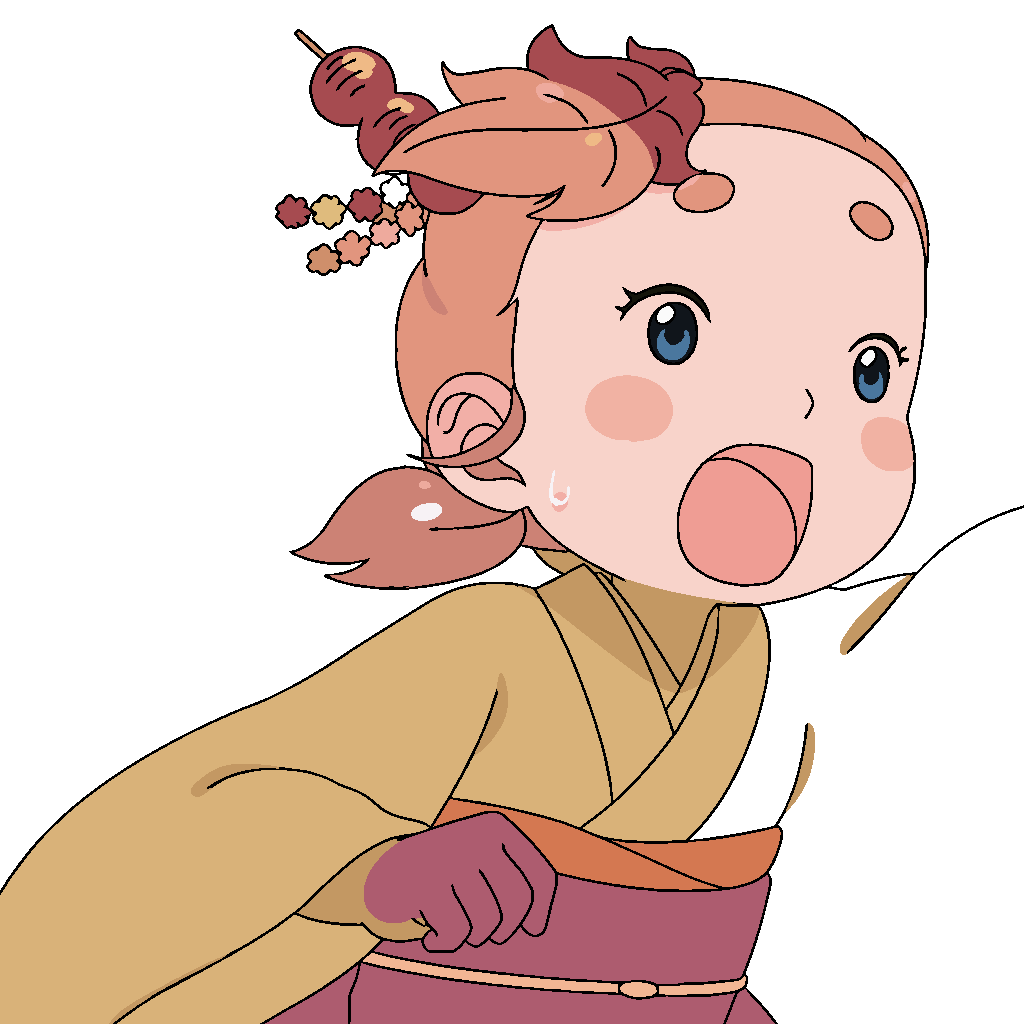} &
            \includegraphics[width=0.16\textwidth]{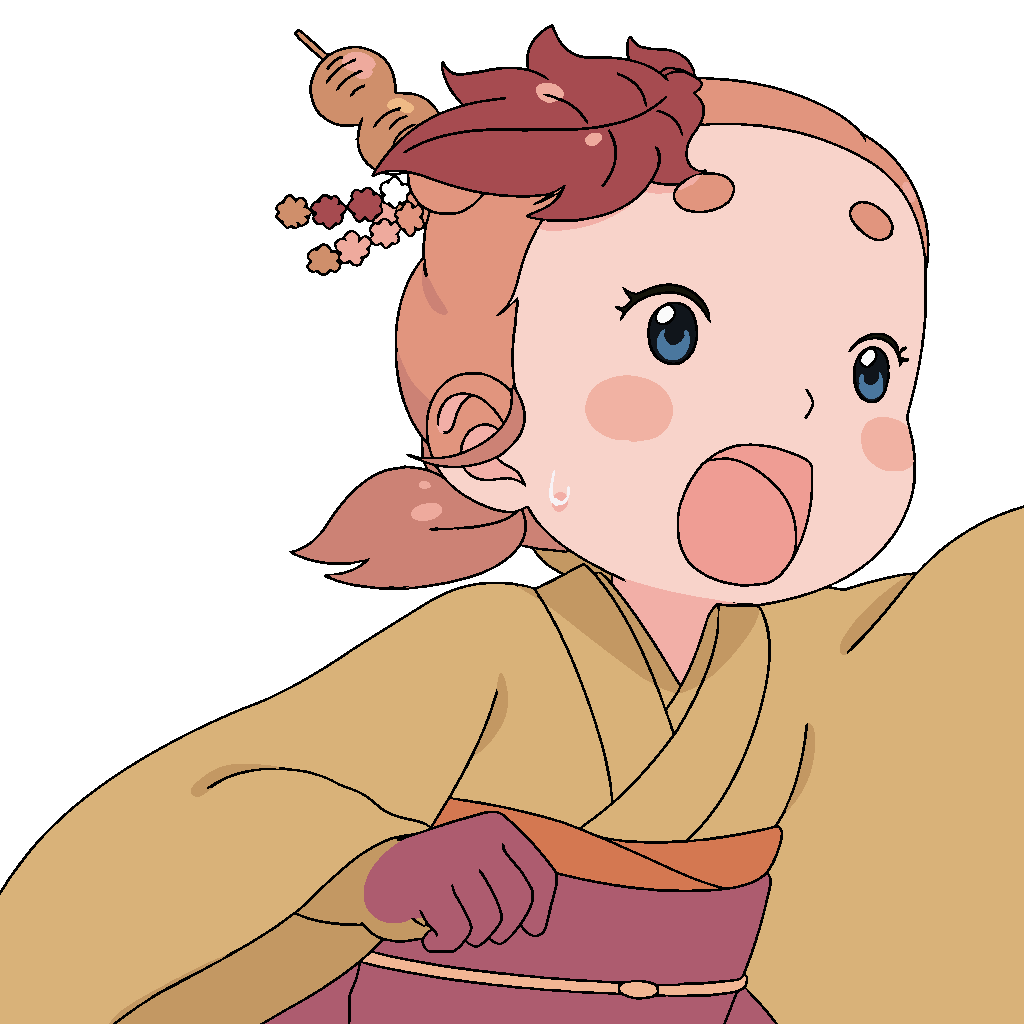} &
            \includegraphics[width=0.16\textwidth]{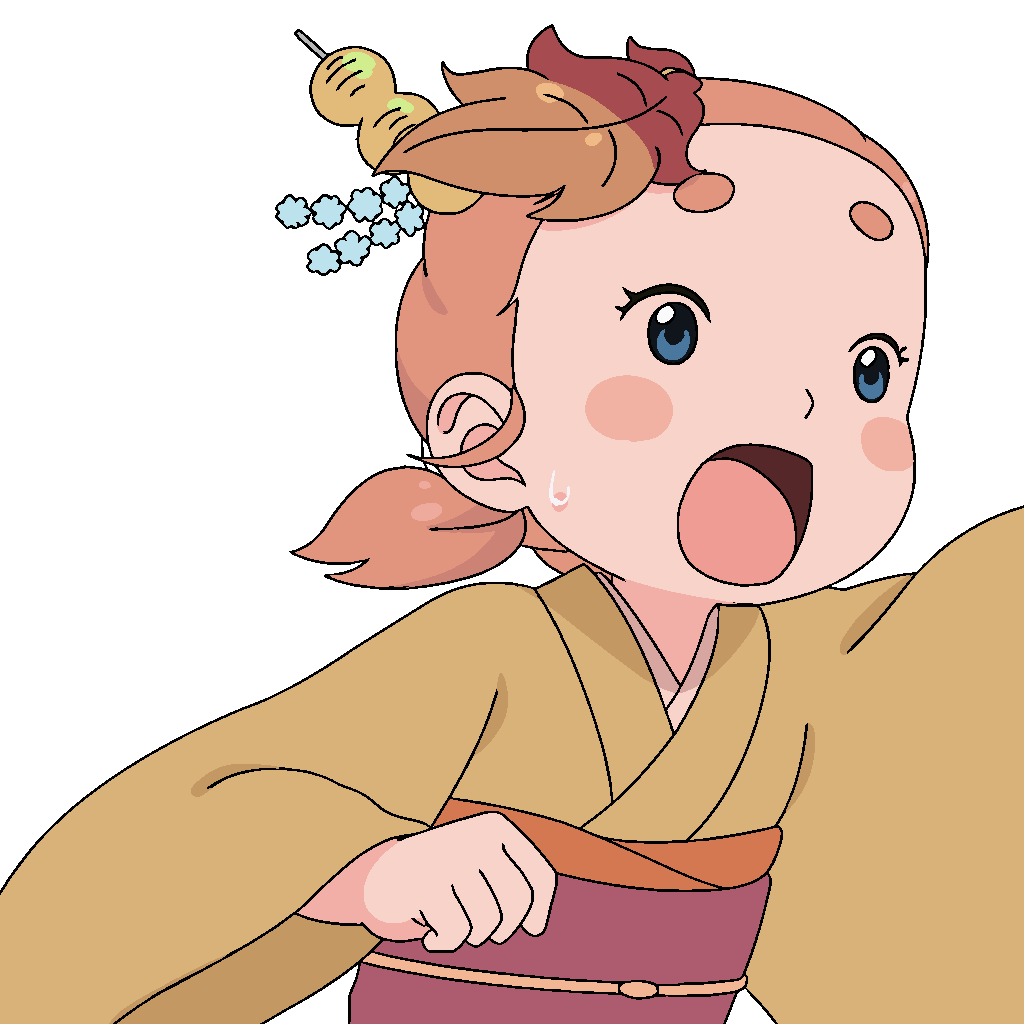} \\
            \includegraphics[width=0.16\textwidth]{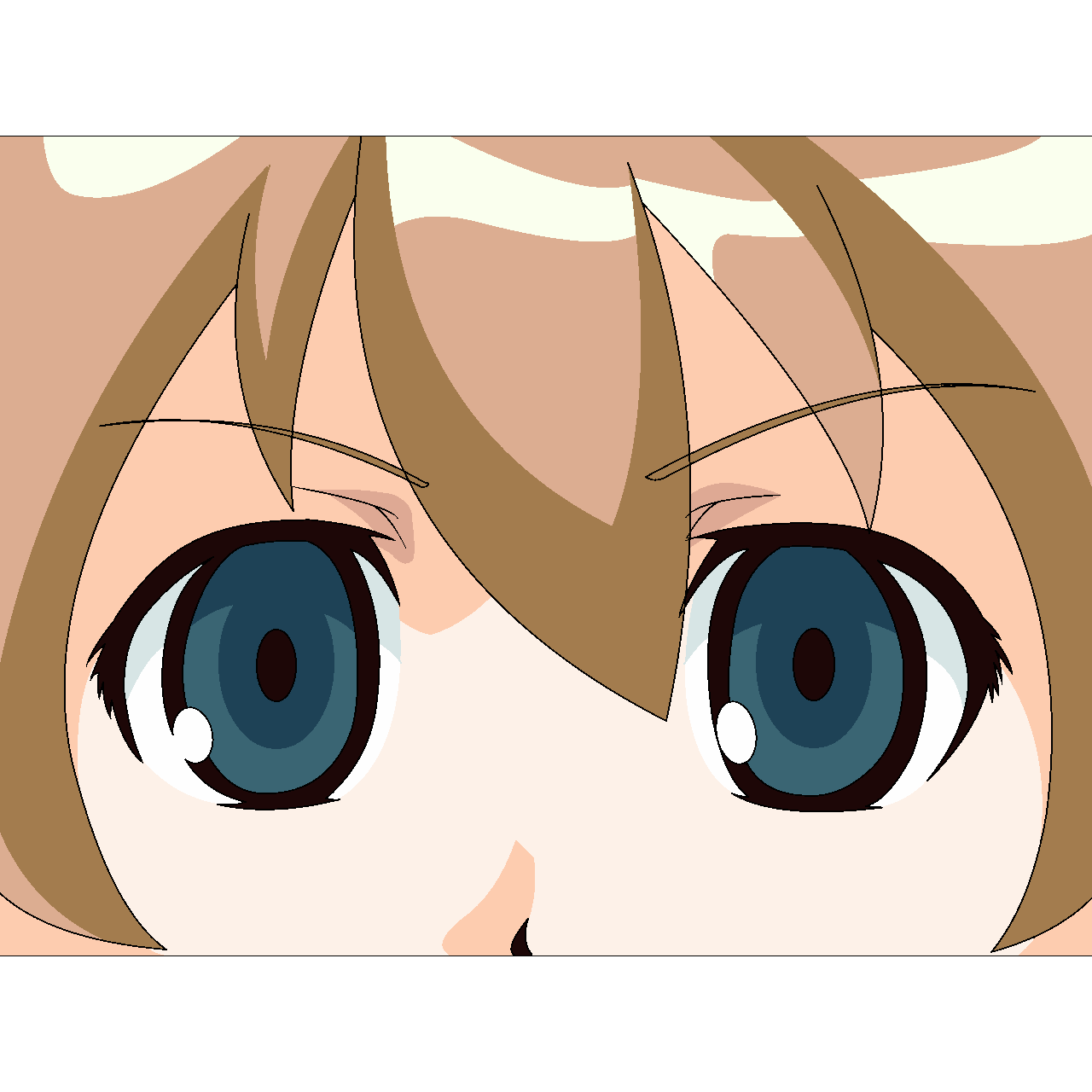} &
            \includegraphics[width=0.16\textwidth]{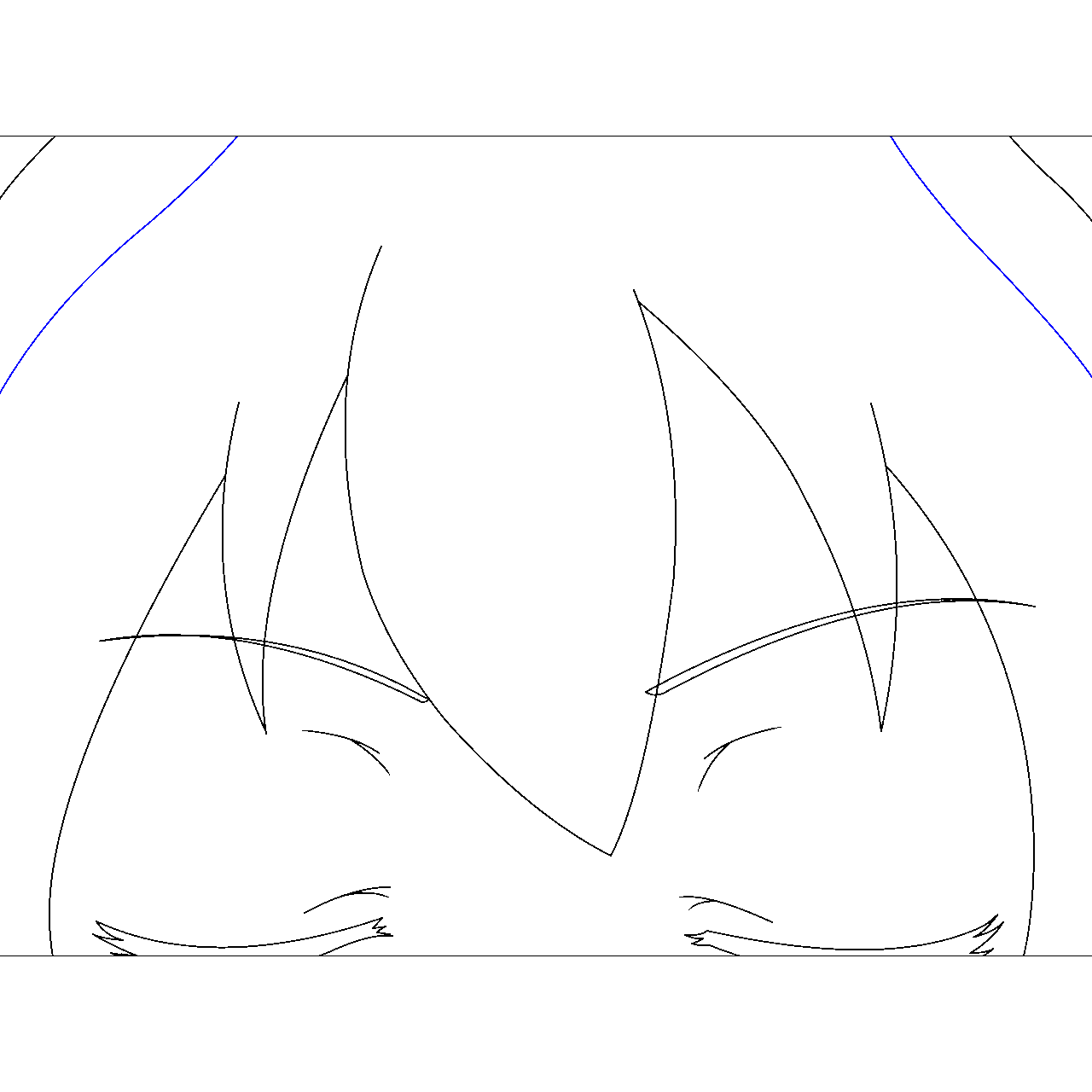} &
            \includegraphics[width=0.16\textwidth]{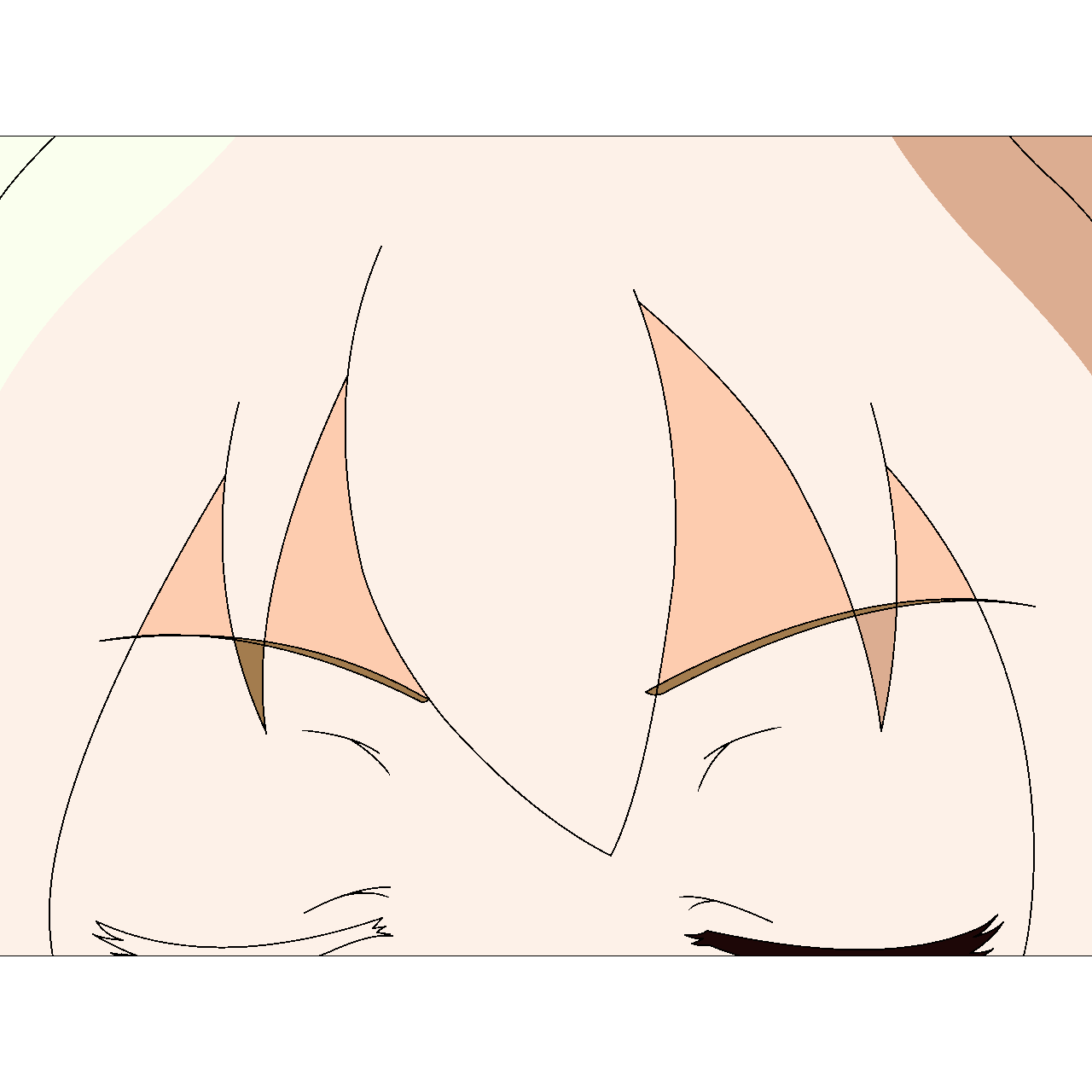} &
            \includegraphics[width=0.16\textwidth]{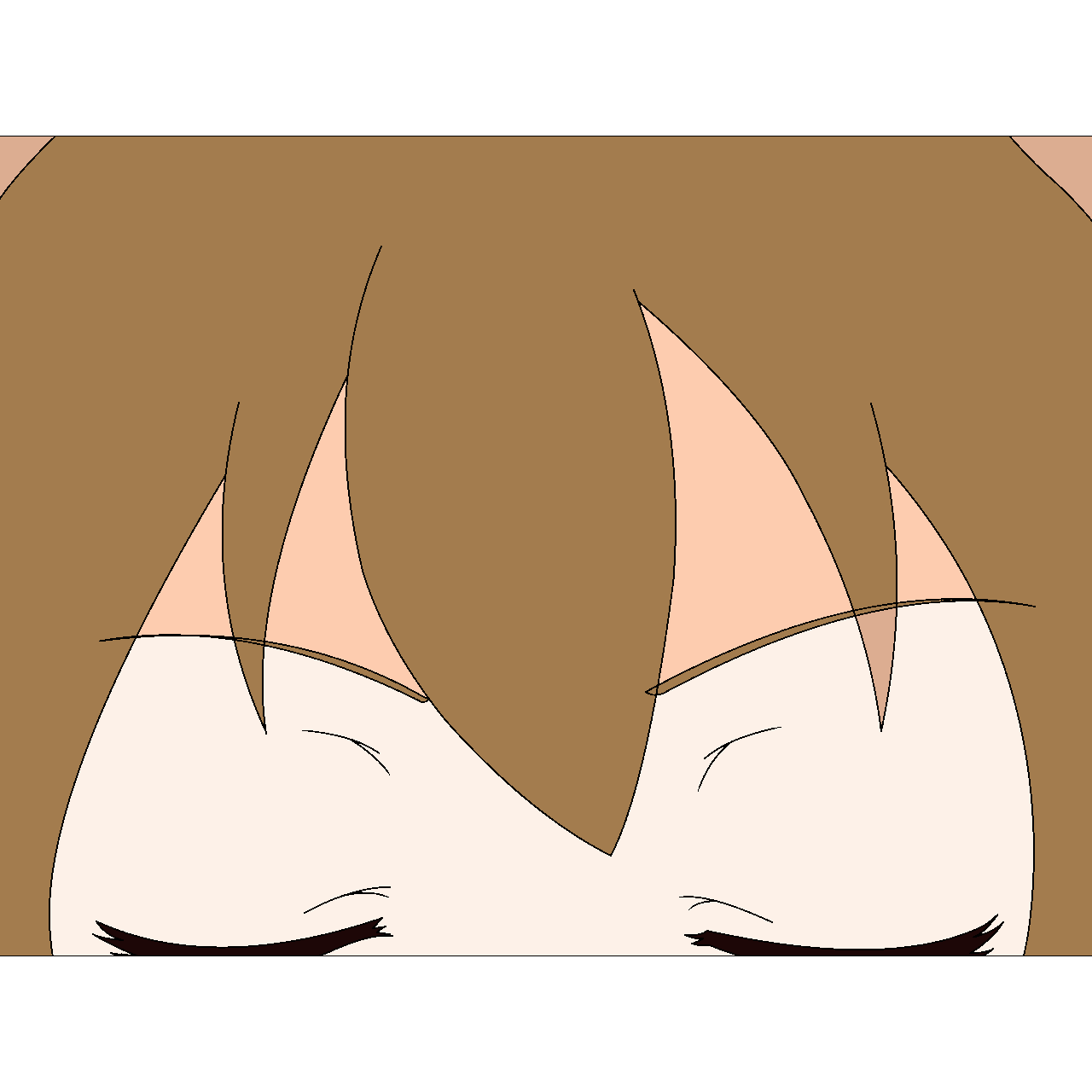} &
            \includegraphics[width=0.16\textwidth]{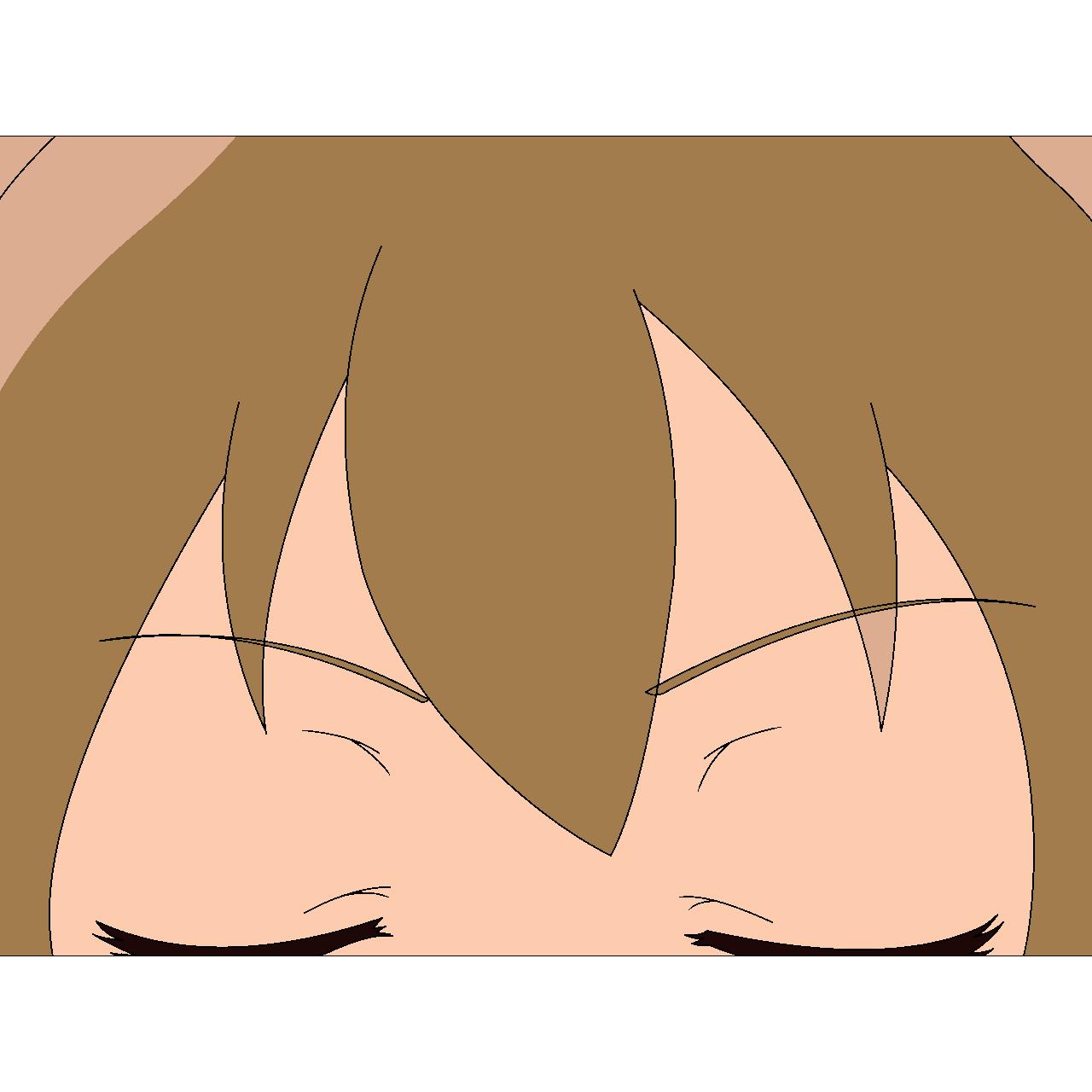} &
            \includegraphics[width=0.16\textwidth]{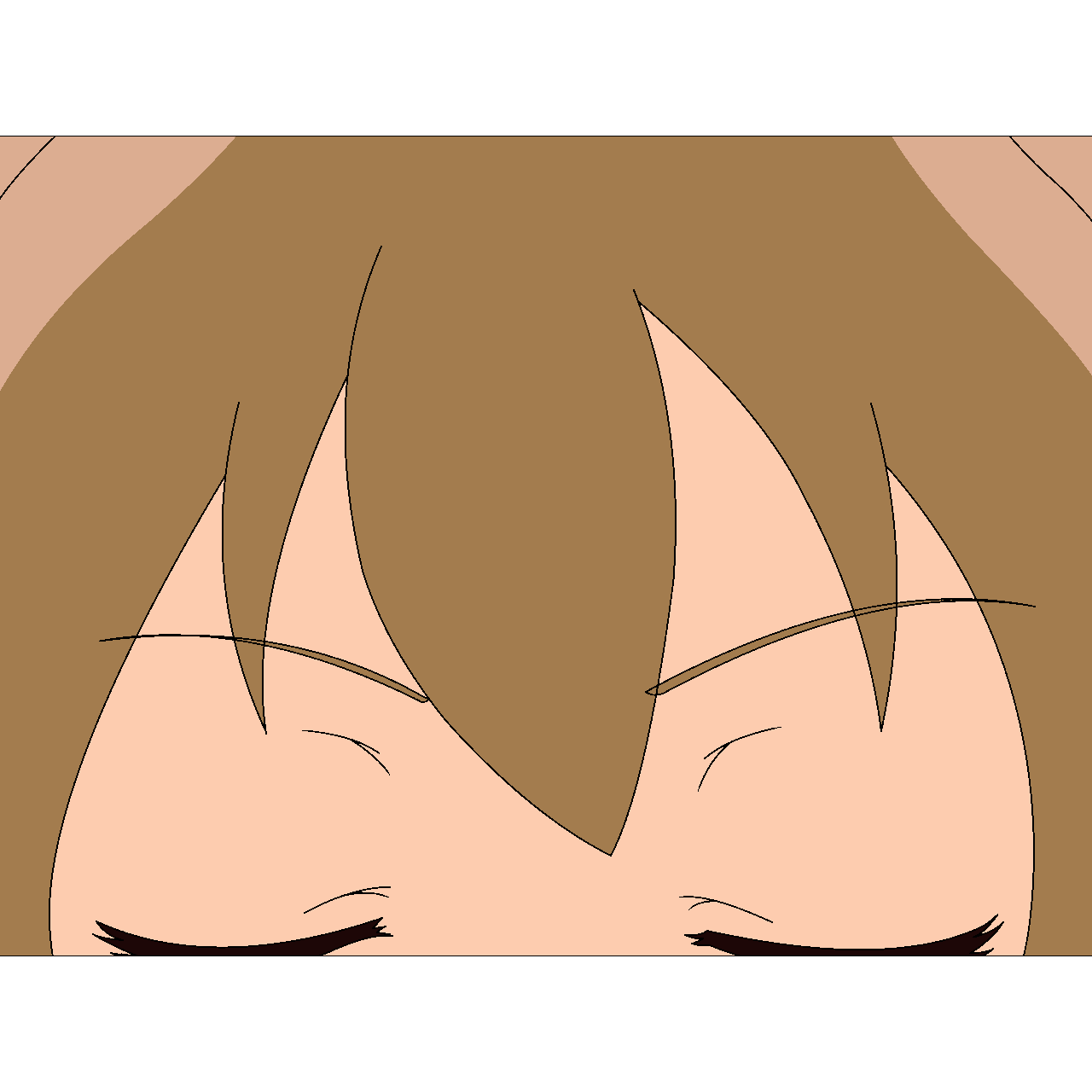} \\
        \end{tabular}%
        }
        \caption{First-frame Reference Colourisation Results}
        \label{fig:qual_continuous}
    \end{subfigure}
    \caption{\textbf{Qualitative comparison under two production formulations.}
    We show one-shot reference and target line sketch, followed by results from
    \textbf{BasicPBC(-Ref)}~\cite{dai2024learning,dai2024paint}, \textbf{DACoN~1.1}~\cite{nagata2025dacon}, our \textbf{\textsc{PeCA}} on DACoN~1.1, and colour ground-truth (right).}
    \label{fig:qual_merged}
\end{figure*}

\paragraph{\textbf{Results on design-sheet key-frame colourisation.}}
\label{sec:keyframe_results} \cref{tab:pbc_kf_merged} reports one-shot key-frame results with design-sheet references on PBC-3D. On task-trained models, the Palette Context Assisted (\textsc{PeCA}) framework further improves the current SOTA on all metrics consistently. Notably, \textsc{PeCA} yields substantially larger gains on training-free backbones, indicating that \textsc{PeCA}'s test-time context reasoning unlocks region matching animation colourisation even for foundation models without colourisation training. Such gain also persists when more key-frame references are given, as shown in \cref{tab:pbc_kf_multishot}, which reports 5-shot and max-shot results on PBC-3D. This further shows that more reference shots do not dilute \textsc{PeCA}'s contribution to the task. \cref{fig:qual_keyframe} shows representative qualitative results, showing our method can help the model overcome a huge appearance gap between references and target frames when previous base inference failed to.

\begin{table}[t]
\centering
\caption{\textbf{One-shot key-frame (design-sheet) colourisation on PBC-3D.} We follow prior work and use a single design-sheet reference image to colourise the target video. We compared the \textbf{\textsc{PeCA}} framework on both the trained DACoN 1.1 Model and other frozen foundation models with the \textbf{base} setting that matches the segmented region features for prediction. \textbf{Training-free} indicates whether the backbone is used without further training (\cmark) or requires training (\xmark) on colourisation tasks.}
\label{tab:pbc_kf_merged}
\resizebox{\textwidth}{!}{%
\begin{tabular}{l|c|ccccc}
\toprule
\textbf{Method / Backbone} & \textbf{Training-free} &
\textbf{Acc (\%)} & \textbf{Acc-Thresh (\%)} & \textbf{Pix-Acc (\%)} & \textbf{Pix-F-Acc (\%)} & \textbf{Pix-B-MIoU (\%)} \\
\midrule
ColorFlow~\cite{zhuang2024colorflow}         & \xmark &  9.72 & 10.81 & 50.64 &  9.16 & 57.17 \\
MangaNinja~\cite{liu2025manganinja}         & \xmark & 14.86 & 16.73 &  7.11 & 28.52 &  0.00 \\
AniDoc~\cite{meng2025anidoc}                & \xmark & 19.80 & 22.68 & 77.38 & 46.46 & 87.32 \\
Cobra~\cite{zhuang2025cobra}   & \xmark & 15.06 & 17.26 & 69.20 & 19.72 & 82.69 \\
MagicColor~\cite{zhang2025followyourcolormultiinstancesketchcolorization} & \xmark & 21.48 & 24.81 & 16.34 & 44.04 & 7.63 \\
BasicPBC-Ref~\cite{dai2024paint}            & \xmark & 52.55 & 56.73 & 90.53 & 72.33 & 94.56 \\
DACoN~\cite{nagata2025dacon}                & \xmark & 67.87 & 72.58 & 96.99 & 91.00 & 99.08 \\
DACoN~1.1~\cite{nagata2025dacon}            & \xmark & 68.01 & 72.87 & 96.97 & 91.03 & 99.11 \\
\rowcolor{gray!12}
\textbf{DACoN~1.1 + \textsc{PeCA}}                   & \xmark &
\textbf{72.04}\gain{4.03} &
\textbf{77.08}\gain{4.21} &
\textbf{97.90}\gain{0.93} &
\textbf{94.04}\gain{3.01} &
\textbf{99.42}\gain{0.31} \\
\midrule
SAM2.1-Large (Base)~\cite{ravi2024sam}       & \cmark & 34.54 & 38.95 & 86.76 & 54.12 & 88.37 \\
\rowcolor{gray!12}
\textbf{SAM2.1-Large + \textsc{PeCA}}                & \cmark &
\textbf{46.65}\gain{12.11} &
\textbf{49.92}\gain{10.97} &
\textbf{88.70}\gain{1.94} &
\textbf{66.96}\gain{12.84} &
\textbf{96.70}\gain{8.33} \\
\midrule
DINOv3 ConvNeXT-L (Base)~\cite{simeoni2025dinov3}    & \cmark & 34.90 & 36.35 & 71.32 & 49.79 & 75.93 \\
\rowcolor{gray!12}
\textbf{DINOv3 ConvNeXT-L + \textsc{PeCA}}           & \cmark &
\textbf{45.88}\gain{10.98} &
\textbf{46.97}\gain{10.62} &
\textbf{80.13}\gain{8.81} &
\textbf{60.15}\gain{10.36} &
\textbf{85.38}\gain{9.45} \\
\midrule
SigLIPv2 ViT-B/16 (Base)~\cite{tschannen2025siglip}   & \cmark & 48.64 & 51.68 & 89.24 & 70.05 & 91.03 \\
\rowcolor{gray!12}
\textbf{SigLIPv2 ViT-B/16 + \textsc{PeCA}}          & \cmark &
\textbf{55.34}\gain{6.70} &
\textbf{58.88}\gain{7.20} &
\textbf{92.48}\gain{3.24} &
\textbf{80.37}\gain{10.32} &
\textbf{93.88}\gain{2.85} \\
\midrule
DINOv2 ViT-L/14 (Base)~\cite{oquab2023dinov2}          & \cmark & 57.49 & 61.86 & 95.35 & 87.24 & 97.45 \\
\rowcolor{gray!12}
\textbf{DINOv2 ViT-L/14 + \textsc{PeCA}}            & \cmark &
\textbf{61.38}\gain{3.89} &
\textbf{65.58}\gain{3.72} &
\textbf{96.25}\gain{0.90} &
\textbf{89.31}\gain{2.07} &
\textbf{98.62}\gain{1.17} \\
\bottomrule
\end{tabular}%
}
\end{table}

\begin{table}[h!]
\centering
\caption{\textbf{Key-frame (design-sheet) colourisation on PBC-3D with more references.} \textsc{PeCA} show consistent gains under multi-reference protocols from \cite{dai2024paint, nagata2025dacon}.}
\label{tab:pbc_kf_multishot}

\resizebox{\textwidth}{!}{%
\begin{tabular}{c|l|c|ccccc}
\toprule
\textbf{\# of Refs} & \textbf{Method / Backbone} & \textbf{Training-free} &
\textbf{Acc} & \textbf{Acc-Thresh} & \textbf{Pix-Acc} & \textbf{Pix-F-Acc} & \textbf{Pix-B-MIoU} \\
\midrule

& ColorFlow~\cite{zhuang2024colorflow}                & \xmark & 12.64 & 14.37 & 54.51 & 15.26 & 61.22 \\
& BasicPBC-Ref~\cite{dai2024paint}                    & \xmark & --    & 64.59 & 96.12 & 83.17 & 98.67 \\
& DACoN~\cite{nagata2025dacon}                        & \xmark & 73.25 & 77.44 & 97.74 & 93.70 & 99.13 \\
& DACoN~1.1~\cite{nagata2025dacon}                    & \xmark & 73.91 & 78.23 & 97.84 & 94.28 & 98.92 \\
\rowcolor{gray!12}
\cellcolor{white} &
\textbf{DACoN~1.1 + \textsc{PeCA}}                    & \xmark &
\textbf{77.73}\gain{3.82} &
\textbf{82.39}\gain{4.16} &
\textbf{98.87}\gain{1.03} &
\textbf{97.02}\gain{2.74} &
\textbf{99.45}\gain{0.53} \\
\cmidrule(lr){2-8}
& SAM2.1-Large (Base)~\cite{ravi2024sam}              & \cmark & 43.80 & 46.59 & 87.66 & 62.25 & 96.75 \\
\rowcolor{gray!12}
\cellcolor{white} &
\textbf{SAM2.1-Large + \textsc{PeCA}}                 & \cmark &
\textbf{57.23}\gain{13.43} &
\textbf{60.96}\gain{14.37} &
\textbf{91.50}\gain{3.84} &
\textbf{76.52}\gain{14.27} &
\textbf{97.18}\gain{0.43} \\
\cmidrule(lr){2-8}
& DINOv2 ViT-L/14 (Base)~\cite{oquab2023dinov2}       & \cmark & 62.65 & 66.42 & 96.77 & 91.54 & 97.96 \\
\rowcolor{gray!12}
\multirow{-9}{*}{\cellcolor{white}\shortstack{\textbf{5-shot}\\\textbf{refs}}}%
& \textbf{DINOv2 ViT-L/14  + \textsc{PeCA}}           & \cmark &
\textbf{66.46}\gain{3.81} &
\textbf{70.01}\gain{3.59} &
\textbf{97.73}\gain{0.96} &
\textbf{93.57}\gain{2.03} &
\textbf{98.83}\gain{0.87} \\

\midrule

& DACoN~\cite{nagata2025dacon}                        & \xmark & 74.31 & 78.48 & 98.04 & 94.27 & 99.10 \\
& DACoN~1.1~\cite{nagata2025dacon}                    & \xmark & 75.05 & 79.23 & 98.19 & 94.79 & 99.16 \\
\rowcolor{gray!12}
\cellcolor{white} &
\textbf{DACoN~1.1 + \textsc{PeCA}}                    & \xmark &
\textbf{79.03}\gain{3.98} &
\textbf{83.43}\gain{4.20} &
\textbf{99.01}\gain{0.82} &
\textbf{97.21}\gain{2.42} &
\textbf{99.55}\gain{0.39} \\
\cmidrule(lr){2-8}
& SAM2.1-Large (Base)~\cite{ravi2024sam}              & \cmark & 46.40 & 49.30 & 87.98 & 63.27 & 96.59 \\
\rowcolor{gray!12}
\cellcolor{white} &
\textbf{SAM2.1-Large + \textsc{PeCA}}                 & \cmark &
\textbf{56.88}\gain{10.48} &
\textbf{60.50}\gain{11.20} &
\textbf{91.94}\gain{3.96} &
\textbf{77.49}\gain{14.22} &
\textbf{97.29}\gain{0.70} \\
\cmidrule(lr){2-8}
& DINOv2 ViT-L/14 (Base)~\cite{oquab2023dinov2}       & \cmark & 63.84 & 67.67 & 97.07 & 91.70 & 98.28 \\
\rowcolor{gray!12}
\multirow{-7}{*}{\cellcolor{white}\shortstack{\textbf{max-shot}\\\textbf{refs}}}%
& \textbf{DINOv2 ViT-L/14 + \textsc{PeCA}}            & \cmark &
\textbf{67.28}\gain{3.44} &
\textbf{70.82}\gain{3.15} &
\textbf{97.71}\gain{0.64} &
\textbf{93.63}\gain{1.93} &
\textbf{98.59}\gain{0.31} \\

\bottomrule
\end{tabular}%
}
\end{table}

\begin{table}[t]
\centering
\caption{\textbf{First-frame Colourisation on PBC-3D and PBC-Real.}
We include both colourisation-trained methods (\xmark) and results using frozen backbones (\cmark). Additional analyses of more baselines \cite{blogNanoBanana, li2025tooncomposer, zhang2025animecolor} are provided in \suppref{sec:supp_modern_pixel_baselines}.}

\label{tab:consecutive_one_shot}
\resizebox{\textwidth}{!}{%
\begin{tabular}{l|c|ccccc|ccccc}
\toprule
\textbf{Method / Backbone} & \textbf{Training-free} &
\multicolumn{5}{c|}{\textbf{PBC-3D}} & \multicolumn{5}{c}{\textbf{PBC-Real}} \\
\cmidrule(lr){3-7} \cmidrule(lr){8-12}
& &
\textbf{Acc} & \textbf{Acc-Thresh} & \textbf{Pix-Acc} & \textbf{Pix-F-Acc} & \textbf{Pix-B-MIoU} &
\textbf{Acc} & \textbf{Acc-Thresh} & \textbf{Pix-Acc} & \textbf{Pix-F-Acc} & \textbf{Pix-B-MIoU} \\
\midrule
BasicPBC~\cite{dai2024learning}                     & \xmark & 56.28 & 60.14 & 93.00 & 77.25 & 97.19 & 59.31 & 62.00 & 91.84 & 72.50 & 98.39 \\
BasicPBC ~\cite{dai2024learning} (Online$^*$)           & \xmark & 53.18 & 58.28 & 93.57 & 79.92 & 96.19 & 57.28 & 60.47 & 92.74 & 74.92 & 98.35 \\
DACoN~\cite{nagata2025dacon}                        & \xmark & 69.91 & 73.59 & 97.30 & - & - & 65.85 & 69.15 & 93.50 & - & - \\
DACoN~1.1~\cite{nagata2025dacon}                    & \xmark & 70.34 & 74.04 & 97.30 & 91.13 & 99.17 & 65.82 & 69.11 & 94.18 & 80.68 & 98.76 \\
\rowcolor{gray!12}
\textbf{DACoN~1.1 + \textsc{PeCA}}                           & \xmark &
\textbf{74.41} & \textbf{78.08} & \textbf{98.11} & \textbf{94.06} & \textbf{99.50} &
\textbf{67.64} & \textbf{71.29} & \textbf{94.70} & \textbf{82.11} & \textbf{99.48} \\
\midrule
StableDiffusion 2.1 (Base)~\cite{Rombach_2022_CVPR}              & \cmark & 32.93 & 34.52 & 87.38 & 58.70 & 94.40 & 46.45 & 48.84 & 89.91 & 64.13 & 97.96 \\
\rowcolor{gray!12}
\textbf{StableDiffusion 2.1 + \textsc{PeCA}}                 & \cmark & \textbf{40.50} & \textbf{42.01} & \textbf{90.87} & \textbf{71.01} & \textbf{96.51} & \textbf{48.11} & \textbf{49.70} & \textbf{90.89} & \textbf{67.45} & \textbf{98.18} \\
\midrule
SAM2.1-Large (Base)~\cite{ravi2024sam}              & \cmark & 49.10 & 52.46 & 91.64 & 72.40 & 97.38 & 55.63 & 58.31 & 90.32 & 69.21 & 98.73 \\
\rowcolor{gray!12}
\textbf{SAM2.1-Large + \textsc{PeCA}}                 & \cmark & \textbf{58.98} & \textbf{62.89} & \textbf{93.65} & \textbf{79.72} & \textbf{98.11} & \textbf{60.41} & \textbf{63.44} & \textbf{93.25} & \textbf{75.99} & \textbf{99.00} \\
\bottomrule
\multicolumn{12}{l}{\footnotesize $^*$ Online setting: the first frame uses the ground-truth reference, and each subsequent frame is colourised using the previous frame's prediction as the reference.}
\end{tabular}%
}

\end{table}

\begin{figure}[t]
    \centering
    \includegraphics[width=0.9\linewidth]{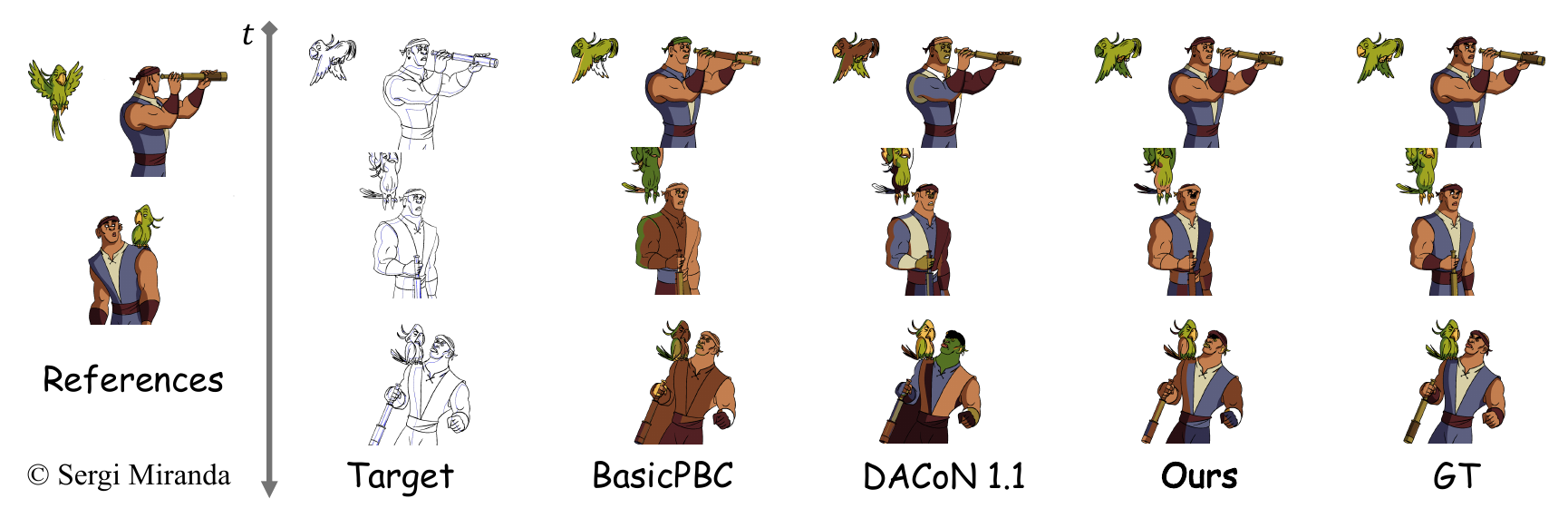}
    \caption{\textbf{Qualitative results of in-between colourisation on Anita-Pirate.} It compares how our proposed \textsc{PeCA} performs against existing methods when dealing with a 10$\times$ longer target frame sequence. }
    \label{fig:Inbetweening}
\end{figure}

\paragraph{\textbf{Results on same-video key-frame colourisation.}}
\label{sec:consecutive_results} \cref{tab:consecutive_one_shot} reports colourisation results on PBC-3D and PBC-Real using only the first frame as a reference to colourise the rest of the video. Including \textsc{PeCA} inference provides consistent improvements on the trained DACoN~1.1 pipeline on both domains, and again yields larger gains on training-free backbones. For in-between colourisation, \cref{tab:inbetweening_continuous} shows that our \textsc{PeCA} consistently improves multiple backbones on both PBC-3D (20-frame clips) and the new long-shot Anita-Pirate dataset. In the latter, only the first and last reference frames are provided with colours, while all the remaining 204 frames have to be coloured, given such a limited reference. All these suggest that test-time context reasoning with \textsc{PeCA} can substantially strengthen existing models' performance with different model types and supervisions. Such improvements persist when we naturally have better feature coverage by the temporal proximity of reference-target frames. Corresponding qualitative comparisons are provided in \cref{fig:qual_continuous} and \cref{fig:Inbetweening}. See more comparisons in \suppref{sec:more_qual}.

\begin{table}[t]
\centering
\caption{\textbf{In-between colourisation results.} We report results on short video clips on both the existing PBC-3D \cite{dai2024learning} testset (20 frames) and a new challenging testset, Anita-Pirate, with $10\times$ longer frame sequence containing multiple complex subjects.}
\label{tab:inbetweening_continuous}
\resizebox{\textwidth}{!}{%
\begin{tabular}{l|c|ccccc|ccccc}
\toprule
\textbf{Method/Backbone} & \textbf{Training-free} &
\multicolumn{5}{c|}{\textbf{PBC-3D}} & \multicolumn{5}{c}{\textbf{Anita-Pirate}} \\
\cmidrule(lr){3-7} \cmidrule(lr){8-12}
& &
\textbf{Acc} & \textbf{Acc-Thresh} & \textbf{Pix-Acc} & \textbf{Pix-F-Acc} & \textbf{Pix-B-MIoU} &
\textbf{Acc} & \textbf{Acc-Thresh} & \textbf{Pix-Acc} & \textbf{Pix-F-Acc} & \textbf{Pix-B-MIoU} \\
\midrule
BasicPBC~\cite{dai2024learning}            & \xmark & 63.38 & 67.77 & 94.84 & 84.20 & 97.54 & 28.54 & 28.97 & 88.52 & 39.77 & 96.63 \\
BasicPBC~\cite{dai2024learning} (Online$^*$)           & \xmark & 53.97 & 59.13 & 93.74 & 80.62 & 96.32 & 7.71 & 7.94 & 32.97 & 17.00 & 35.93 \\
DACoN~1.1~\cite{nagata2025dacon}           & \xmark & 78.02 & 82.11 & 98.48 & 95.51 & 99.47 & 38.16 & 39.36 & 94.29 & 61.65 & 99.16 \\
\rowcolor{gray!12}
\textbf{DACoN~1.1 + \textsc{PeCA}}                  & \xmark &
\textbf{80.80} & \textbf{84.82} & \textbf{99.00} & \textbf{97.18} & \textbf{99.58} &
\textbf{41.24} & \textbf{42.16} & \textbf{94.29} & \textbf{62.78} & \textbf{99.43} \\
\midrule
DINOv2 ViT-L/14 (Base)~\cite{oquab2023dinov2}        & \cmark & 66.25 & 70.17 & 97.73 & 93.36 & 98.89 & 28.55 & 29.30 & 93.06 & 53.88 & 99.40 \\
\rowcolor{gray!12}
\textbf{DINOv2 ViT-L/14~\cite{oquab2023dinov2}  + \textsc{PeCA}}                 & \cmark &
\textbf{69.29} & \textbf{72.60} & \textbf{98.23} & \textbf{94.49} & \textbf{99.33} &
\textbf{31.01} & \textbf{31.81} & \textbf{93.39} & \textbf{57.18} & \textbf{99.49} \\
\bottomrule
\multicolumn{12}{l}{\footnotesize $^*$ Online setting: the first frame uses the ground-truth reference, and each subsequent frame is colourised using the previous frame's prediction as the reference.}
\end{tabular}%
}
\end{table}

\begin{table}[t]
\centering
\caption{\textbf{Ablation study on colourisation under different reference and model types.}
{DACoN~1.1} is pretrained on colourisation; {DINOv2} (ViT-L/14) is used frozen in a zero-shot manner.
ARE: Active Reference Expansion in \cref{sec:are}, PA: Probability Aggregation in \cref{sec:pa}, CT: Cyclic-gated Temporal-fusion in \cref{sec:ct}.}
\label{tab:ablation_merged}

\resizebox{\linewidth}{!}{%
\begin{tabular}{l|c c c|ccccc|ccccc}
\toprule
\multirow{2}{*}{\textbf{Backbone}} &
\multirow{2}{*}{\textbf{ARE}} &
\multirow{2}{*}{\textbf{PA}} &
\multirow{2}{*}{\textbf{CT}} &
\multicolumn{5}{c|}{\textbf{Design-sheet (one-shot)}} &
\multicolumn{5}{c}{\textbf{First-frame (one-shot)}} \\
\cmidrule(lr){5-9}\cmidrule(lr){10-14}
& & & &
\textbf{Acc} & \textbf{Acc-Thresh} & \textbf{Pix-Acc} & \textbf{Pix-F-Acc} & \textbf{Pix-B-MIoU} &
\textbf{Acc} & \textbf{Acc-Thresh} & \textbf{Pix-Acc} & \textbf{Pix-F-Acc} & \textbf{Pix-B-MIoU} \\
\midrule

& \xmark & \xmark & \xmark & 68.01 & 72.87 & 96.97 & 91.03 & 99.11 & 70.34 & 74.04 & 97.30 & 91.13 & 99.17 \\
& \cmark & \xmark & \xmark & 69.04 & 74.02 & 97.22 & 91.67 & 99.24 & 70.65 & 74.54 & 97.29 & 91.34 & 99.25 \\
& \cmark & \cmark & \xmark & 70.61 & 75.40 & 97.43 & 92.52 & 99.28 & 72.50 & 76.17 & 97.64 & 92.09 & 99.38 \\
& \xmark & \cmark & \cmark & 70.30 & 75.27 & 97.21 & 92.47 & 99.38 & 72.78 & 76.56 & 97.83 & 92.87 & 99.35 \\
& \cmark & \xmark & \cmark & 69.02 & 74.07 & 97.19 & 91.68 & 99.26 & 70.87 & 74.80 & 97.37 & 91.44 & 99.31 \\
\rowcolor{gray!12}
\multirow{-6}{*}{\cellcolor{white}\textbf{DACoN~1.1}}%
& \cmark & \cmark & \cmark &
\textbf{72.04} & \textbf{77.08} & \textbf{97.90} & \textbf{94.04} & \textbf{99.42} &
\textbf{74.41} & \textbf{78.08} & \textbf{98.11} & \textbf{94.06} & \textbf{99.50} \\
\midrule

& \xmark & \xmark & \xmark & 57.49 & 61.86 & 95.35 & 87.24 & 97.45 & 59.50 & 62.99 & 96.25 & 87.95 & 98.47 \\
& \cmark & \xmark & \xmark & 58.74 & 63.43 & 95.96 & 88.14 & 98.57 & 60.92 & 64.53 & 96.52 & 88.87 & 98.64 \\
& \cmark & \cmark & \xmark & 60.64 & 64.66 & 95.94 & 88.32 & 98.60 & 62.51 & 65.61 & 96.80 & 89.00 & 99.06 \\
& \xmark & \cmark & \cmark & 58.68 & 62.35 & 94.56 & 84.52 & 97.99 & 61.00 & 63.77 & 96.54 & 88.76 & 99.07 \\
& \cmark & \xmark & \cmark & 58.51 & 63.29 & 95.89 & 88.08 & 98.52 & 61.16 & 64.85 & 96.70 & 89.10 & 98.97 \\
\rowcolor{gray!12}
\multirow{-6}{*}{\cellcolor{white}\textbf{DINOv2 ViT-L/14}}%
& \cmark & \cmark & \cmark &
\textbf{61.38} & \textbf{65.58} & \textbf{96.25} & \textbf{89.31} & \textbf{98.62} &
\textbf{63.28} & \textbf{66.47} & \textbf{97.13} & \textbf{90.45} & \textbf{99.22} \\
\bottomrule
\end{tabular}%
}
\end{table}

\subsection{Ablations and Analysis}
\label{sec:ablation}

We analyse the three steps in the Palette Context Assisted (\textsc{PeCA}) inference framework on both task-trained and frozen foundation backbone under different key-frame reference types. Quantitative results are shown in \cref{tab:ablation_merged}.

\paragraph{\textbf{Three steps are jointly contributing.}} Across both reference types and both backbones, we observe a consistent pattern. Active Reference Expansion (ARE) only gives a minor performance improvement. That means expanding the reference bank, even if in a target-aware way, can still lead to confusion, as it may dilute the exact match with more region candidates. Therefore, further adding Probability Aggregation (PA) yields substantial gains by soft-voting across multiple plausible matches, which reduces sensitivity to spurious matches. Cyclic-gated Temporal-fusion (CT) then provides additional improvements by exploiting temporal context inside the video, refining per-region colour probabilities. Combining all steps gives the strongest results, while removing any of them results in a considerable performance drop, indicating these context cues are complementary and reciprocal as expected, instead of isolated heuristics.

\paragraph{\textbf{Step-wise contributions differ by reference types.}} The relative contributions of ARE and CT differ between the two reference types. Under design-sheet references, errors are often driven by spatial mismatch because design sheets can be far from the target shot in pose and region layout; correspondingly, ARE tends to contribute more by improving target-conditioned reference support. Under first-frame references, the reference comes from the same video, and the appearance gap is smaller, so temporal continuity becomes a stronger cue; CT therefore tends to provide more noticeable gains. The balance also depends on the backbone: frozen features benefit more from improved support and aggregation, whereas the task-trained model benefits more from temporal refinement, consistent with its stronger correspondence quality.

\begin{table}[t]

\centering
\caption{\textbf{Additional one-shot key-frame (design-sheet) colourisation ablations on \textsc{PeCA}'s internal designs.}
ARE Selection uses \xmark=\textit{random}, \cmark=\textit{greedy} (\cref{eq:facility}); CT Cycle Gate uses \xmark=\textit{off}, \cmark=\textit{on} (\cref{eq:cycle}).}
\label{tab:ablation_internal_extra_kf}

\resizebox{\linewidth}{!}{%
\begin{tabular}{l|c|c|ccccc}
\toprule
\textbf{Backbone} & \textbf{Selection} & \textbf{Cycle Gate} &
\textbf{Acc} & \textbf{Acc-Thresh} & \textbf{Pix-Acc} & \textbf{Pix-F-Acc} & \textbf{Pix-B-MIoU} \\
\midrule

& \xmark & \cmark & 47.90 & 51.41 & 87.60 & 63.84 & 96.74 \\
& \cmark & \xmark & 45.92 & 49.11 & 88.19 & 65.49 & 97.18 \\
\rowcolor{gray!12}
\multirow{-3}{*}{\cellcolor{white}SAM2.1-Large~\cite{ravi2024sam}}%
& \cmark & \cmark & \textbf{50.69} & \textbf{54.59} & \textbf{89.10} & \textbf{69.05} & \textbf{97.39} \\
\midrule

& \xmark & \cmark & 46.45 & 48.70 & 88.61 & 68.66 & 90.44 \\
& \cmark & \xmark & 40.30 & 41.83 & 88.64 & 67.45 & 91.37 \\
\rowcolor{gray!12}
\multirow{-3}{*}{\cellcolor{white}CLIP ViT-L/14~\cite{radford2021learning}}%
& \cmark & \cmark & \textbf{48.58} & \textbf{50.94} & \textbf{90.51} & \textbf{73.77} & \textbf{92.19} \\
\bottomrule
\end{tabular}%
}
\end{table}

\paragraph{\textbf{Ablation on view selection and cyclic gating.}}
We additionally ablate two module-internal designs in \cref{tab:ablation_internal_extra_kf}. For ARE, greedy facility-location selection (\cref{eq:facility}) consistently outperforms keeping all the random views under the same budget, supporting that the gain comes from coverage-aware selection rather than merely adding more augmented views. For CT, disabling the cyclic gate (\cref{eq:cycle}) leads to a substantial performance drop, indicating that gating is necessary to prevent unreliable temporal matches from propagating errors over time.

\begin{figure}[t]
    \centering
    \begin{subfigure}[t]{0.22\linewidth}
        \centering
        \includegraphics[width=\linewidth]{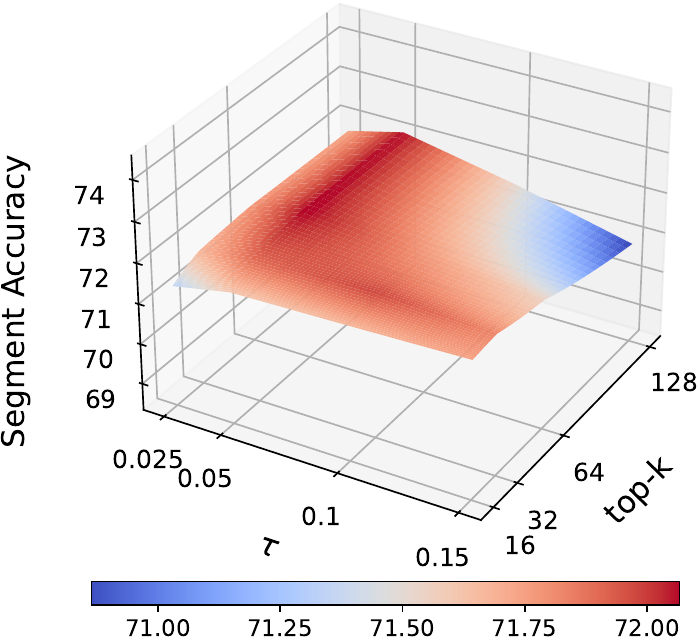}
        \caption{Segment-Acc \\(top-$k$ vs. $\tau$).}
        \label{fig:ablate_tau_topk_seg}
    \end{subfigure}\hfill
    \begin{subfigure}[t]{0.22\linewidth}
        \centering
        \includegraphics[width=\linewidth]{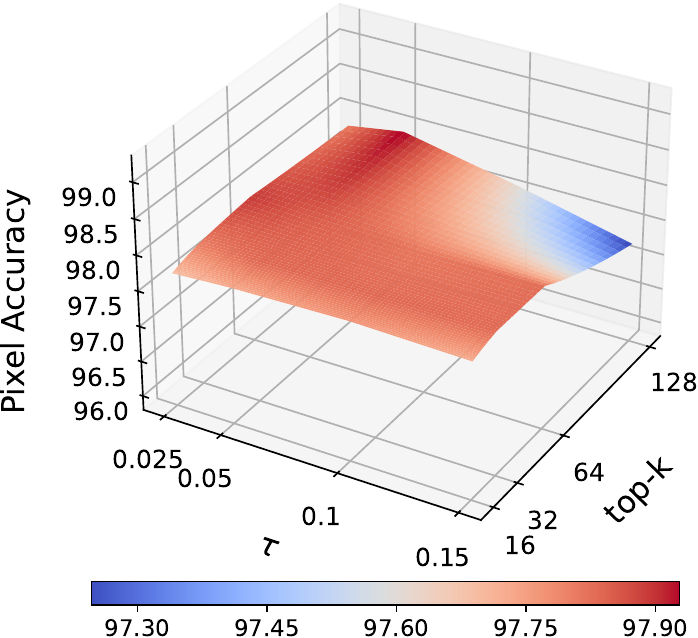}
        \caption{Pixel-Acc \\ (top-$k$ vs. $\tau$).}
        \label{fig:ablate_tau_topk_pix}
    \end{subfigure}\hfill
    \begin{subfigure}[t]{0.22\linewidth}
        \centering
        \includegraphics[width=\linewidth]{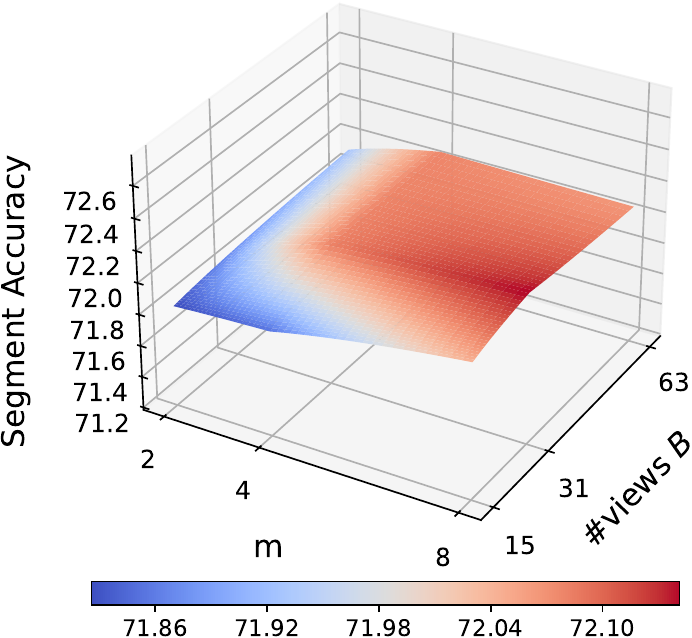}
        \caption{Segment-Acc \\(\#views $B$ vs. $m$).}
        \label{fig:ablate_views_m_seg}
    \end{subfigure}\hfill
    \begin{subfigure}[t]{0.22\linewidth}
        \centering
        \includegraphics[width=\linewidth]{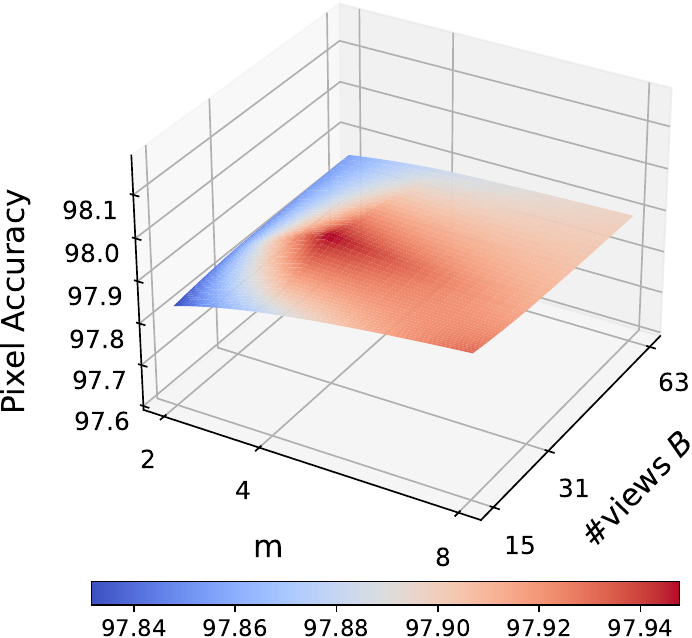}
        \caption{Pixel-Acc \\(\#views $B$ vs. $m$).}
        \label{fig:ablate_views_m_pix}
    \end{subfigure}

    \caption{\textbf{Hyperparameter sensitivity for \textsc{PeCA} inference under key-frame (one-shot design-sheet referenced) colourisation.}
    We visualise the sensitivity of (top-$k$, $\tau$) in PA (\cref{sec:pa}) and (\#views $B$ vs. $m$) in ARE \cref{sec:are}, reported under Segment Accuracy and Pixel Accuracy.}
    \label{fig:ablation_surfaces}
\end{figure}

\paragraph{\textbf{Hyperparameter sensitivity.}}
\cref{fig:ablate_tau_topk_seg} and \cref{fig:ablate_tau_topk_pix} visualise the sensitivity of PA (top-$k$, $\tau$) and ARE (\#views $B$, exploration factor $m$) under key-frame colourisation. Across a wide range of top-$k$ and $\tau$, all metrics vary mildly, indicating stable behaviour. The observed degradations match expected trends. As $\tau\!\rightarrow\!0$ and top-$k\!\rightarrow\!1$, probability aggregation degenerates to simple nearest-neighbour \cite{nagata2025dacon}. Increasing top-$k$ and $\tau$ makes aggregation less selective and approaches a global mixing of candidates, similar in spirit to linear combination schemes used in prior work \cite{casey2021animation}, which can overly flatten the colour distribution and let low-quality matches dilute the prediction. Thus, we select the default setting ($\tau{=}0.05$, top-$k{=}64$) that retains multiple plausible matches while keeping the palette colour probability mass concentrated on high-confidence candidates.

Meanwhile, \cref{fig:ablate_views_m_seg} and \cref{fig:ablate_views_m_pix} show that in ARE, the number of selected views $B$ trades off coverage against cost, while $m$ controls the size of the explored candidate pool ($mB$) for greedy selection. Performance saturates quickly as $B$ increases, and increasing $m$ yields only marginal gains beyond moderate search breadth. We therefore adopt $B{=}31$ and $m{=}4$ as a cost-effective operating point. We provide further numerical analysis and discussions in \suppref{sec:pa_additional_analysis}.

\section{Conclusion}
To sum up, we introduced a training-free and plug-and-play {Palette Context Assisted (\textsc{PeCA})} inference framework for production-level paint-bucket colourisation, where each region must follow a correct colour in a predefined palette. \textsc{PeCA} improves region matching by constructing and validating test-time context: it strengthens spatial context with a target-aware reference expansion, unifying noisy matching evidence in colour space, and uses adjacent frames as temporal context to support prediction. In particular, such context can provide additional support for colour propagation paths when direct matches are ambiguous. Experiments on existing benchmarks and a new long-video test case showed consistent gains on both task-trained models and frozen backbones.

\emph{Limitations.}
Similar to prior paint-bucket colourisation methods that rely on region segmentation and a well-defined palette, \textsc{PeCA} cannot fully resolve failures caused by line leakage or by colours/regions missing from the available references. This limitation is inherent to palette-constrained paint-bucket colourisation, which requires strict, discrete colour assignment from reference palette to target sketch regions. This construction prevents the system from hallucinating colours that are not supported by reference evidence. We provide further discussions and failure case analysis of the task formulation and current methods in \suppref{sec:limitations}.

Beyond these limitations, a promising direction is to move from a fixed-reference setting to an interactive or open-world reference-growth setting. Instead of relying only on the provided key frames or design sheets, future systems may actively acquire additional colour evidence through artist correction feedback, generated reference views, or external character-sheet retrieval. Such extensions would help cover missing poses, parts, and colours, while preserving the strict palette constraints required by paint-bucket colourisation. We view this as a natural next step for extending the concept of test-time context reasoning introduced by \textsc{PeCA} toward practical production toolkits.

\section*{Acknowledgement}
This project is partially supported by an Amazon Research Award, the EPSRC Doctoral Landscape Award (DLA) Collaborative Studentships with Industry and Allsee Technologies Ltd. The computations in this research were partially performed using the Baskerville Tier 2 HPC service. Baskerville was funded by the EPSRC and UKRI through the World Class Labs scheme (EP\textbackslash T022221\textbackslash1) and the Digital Research Infrastructure programme (EP\textbackslash W032244\textbackslash1) and is operated by Advanced Research Computing at the University of Birmingham.

\clearpage
\appendix
\section*{Appendix}
\renewcommand{\theHsection}{app.\Alph{section}}
\renewcommand{\theHsubsection}{app.\Alph{section}.\arabic{subsection}}
\renewcommand{\theHsubsubsection}{app.\Alph{section}.\arabic{subsection}.\arabic{subsubsection}}
\renewcommand{\theHfigure}{app.\Alph{section}.\arabic{figure}}
\renewcommand{\theHtable}{app.\Alph{section}.\arabic{table}}
\renewcommand{\theHequation}{app.\arabic{equation}}
\setcounter{section}{0}
\renewcommand{\thesection}{\Alph{section}}
\setcounter{figure}{0}
\setcounter{table}{0}
\setcounter{equation}{0}
\renewcommand{\theequation}{\roman{equation}}

\ifarxivversion
\renewcommand{\thefigure}{\thesection\arabic{figure}}
\renewcommand{\thetable}{\thesection\arabic{table}}
\makeatletter
\@addtoreset{figure}{section}
\@addtoreset{table}{section}
\makeatother
This appendix provides additional details on dataset, implementation, computation cost, diagnostic analyses, extended experiments, qualitative results, and limitations. For ease of navigation, the sections are organised as follows:
\fi

\begin{itemize}
    \item \textbf{Sec.~\ref{sec:pirate-detail}: Details on Annotation and Statistics of the New Test Data.}
    This section expands the details on the newly introduced Anita-Pirate data, with the complete construction pipeline, licensing information, and dataset statistics/comparisons that justify its role as a long-video stress test.

    \item \textbf{Sec.~\ref{sec:imple}: Additional Implementation Details.}
    This section provides further details for reproduction: metric definitions, backbone configurations, detailed view expansion steps and end-to-end runtime comparisons.
    \item \textbf{Sec.~\ref{sec:fengetal}: Further Comparisons to Previous Works.}
    This section reports an additional evaluation under the same shorter-clip in-between protocol related to Feng~\etal~\cite{Feng_2025_ICCV} and additional details on pixel-generative baselines.

    \item \textbf{Sec.~\ref{sec:pa_additional_analysis}: Additional Analysis on Reference Expansion and Probability Aggregation.}
    This section supports the motivation in \cref{sec:are,sec:pa} by analysing how the ``quality'' of colour probabilities changes as the number of reference views and correspondence aggregation differs.

    \item \textbf{Sec.~\ref{sec:supp_longclip_stability}: Additional Temporal Stability Analysis.}
    This section further proves that our method achieved better video-level temporal consistency. We report an additional temporal stability metric together with curves and qualitative results, revealing how performance evolves over time.

    \item \textbf{Sec.~\ref{sec:vipseg_extension}: Extension to Natural Video Region Label Propagation.}
    This section supports the generality of \textsc{PeCA} beyond paint-bucket colourisation task by conceptually extending it to a new task of semantic label propagation over regions from a generic video segmentation dataset.

    \item \textbf{Sec.~\ref{sec:more_qual}: More Qualitative Results.}
    This section provides additional visual comparisons under different reference settings and backbones, complementing the representative examples shown in \cref{fig:qual_merged}.

    \item \textbf{Sec.~\ref{sec:limitations}: Limitations.}
    This section expands the limitations discussion by detailing failure modes related to imperfect line segmentation and incomplete reference coverage, together with future directions.
\end{itemize}

Additional qualitative results are available in a demo \projectvideolink{} we provide.

\section{Details on Annotation and Statistics of Anita-Pirate}
\label{sec:pirate-detail}

For the same-video colourisation experiments in \cref{sec:consecutive_results}, we introduce Anita-Pirate, a stress-test case study with a much longer video, to validate the long-horizon stability of our method. To construct this stress test, we select the longest continuous sequence from Anita~\cite{Anita2024}, which provides hand-drawn line sketches paired with intermediate colourisation targets. The source animation video is licensed under a CC-BY licence. Our goal is to convert the raw data into production-ready line sketches paired with region-level colour annotations for standard paint-bucket evaluation. This requires addressing several mismatches between the raw data format and production-style annotation.

\begin{figure*}[h!]
    \centering
    \includegraphics[width=\textwidth]{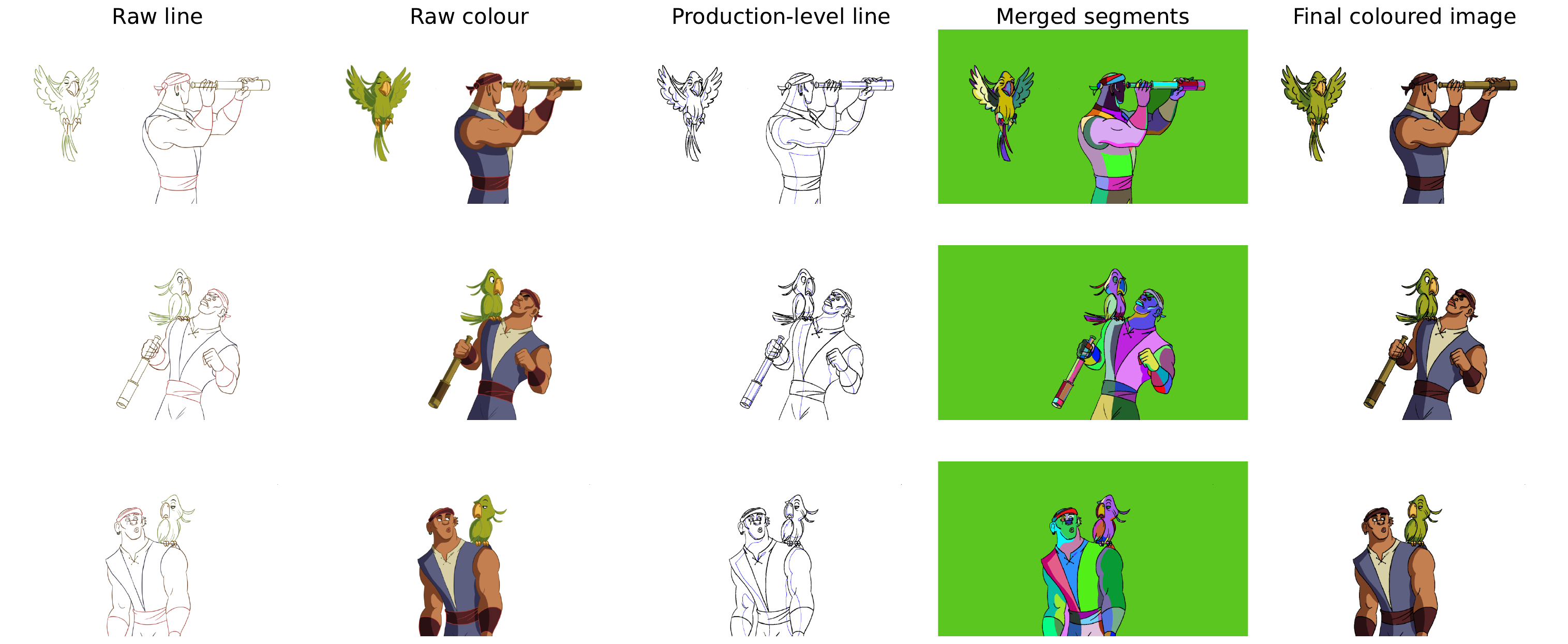}
    \caption{\textbf{Visualisation of Anita-Pirate construction pipeline:}
    Raw line, Raw colour, Production-level line, Merged segments, and Final coloured image, each corresponds to an intermediate stage of annotation.}
    \label{fig:anita_pipeline_random3}
\end{figure*}

To begin with, we first verified that all frames have properly enclosed regions by applying flood-fill segmentation \cite{10.1145/800249.807456} that reveals enclosed regions, and manually fixed the detected leakages. After that, another issue in the original data is that shadow and highlight boundaries are completely absent from the raw line sketches (first column in \cref{fig:anita_pipeline_random3}). We therefore infer such boundaries from colour discontinuities by running flood-fill segmentation on coloured images (second column in \cref{fig:anita_pipeline_random3}) and merging the boundaries into the line map. In the resulting line sketches, original artist lines are preserved in black, while newly introduced shadow/highlight separators are encoded in pure blue following industrial convention, yielding near production-level line sketches in the third column of \cref{fig:anita_pipeline_random3}.

After line merging, we performed flood-fill segmentation again on the constructed production-level lines and assigned the majority colour to each region as illustrated in the last two columns of \cref{fig:anita_pipeline_random3}. It can be seen that the resulting frames recover the appearances and colours from the raw coloured frames successfully. We then export both the region index map and the segment-to-RGBA mapping, which are directly compatible with PBC datasets \cite{dai2024paint}.

\begin{figure*}[t]
    \centering
    \begin{subfigure}[t]{0.49\textwidth}
        \centering
        \includegraphics[width=\linewidth]{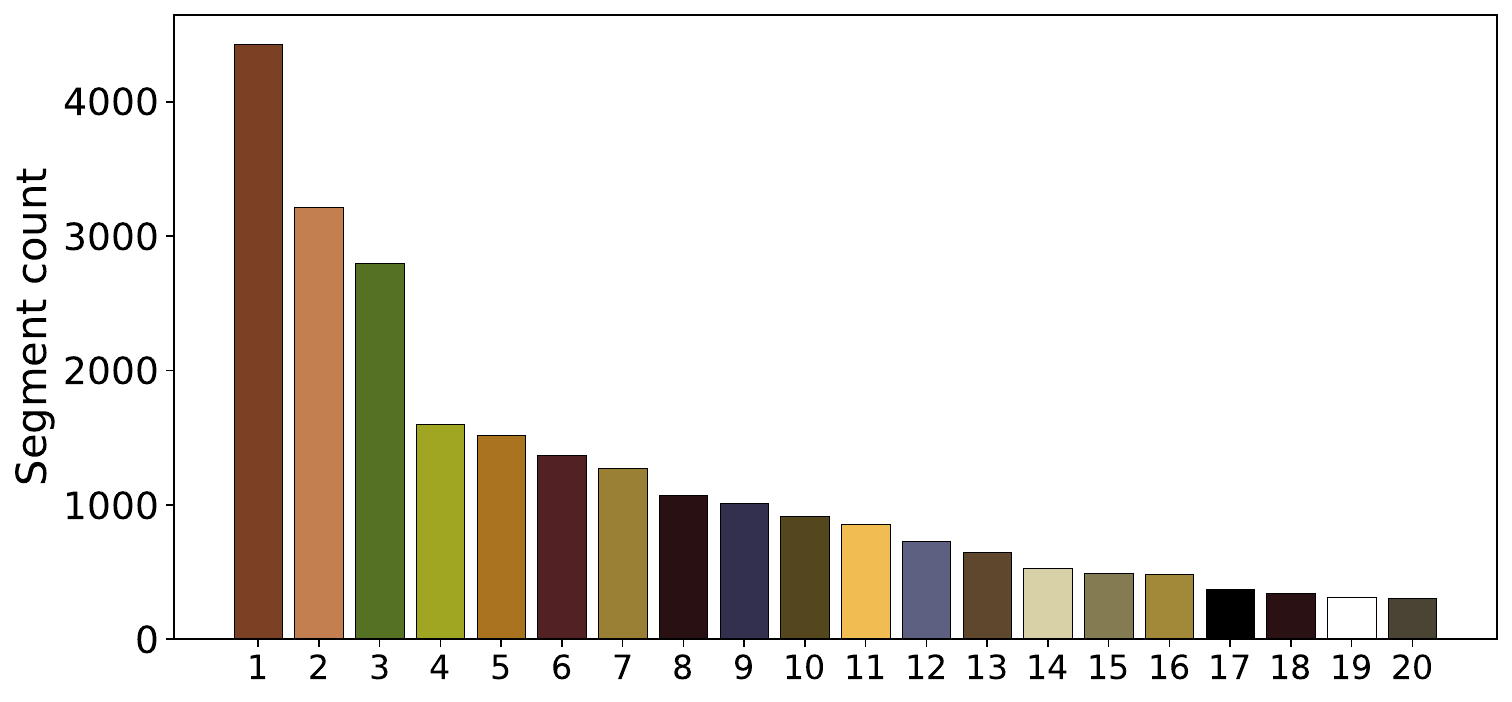}
        \caption{Top-20 colours by segment count.}
        \label{fig:anita_palette_seg}
    \end{subfigure}
    \hfill
    \begin{subfigure}[t]{0.49\textwidth}
        \centering
        \includegraphics[width=\linewidth]{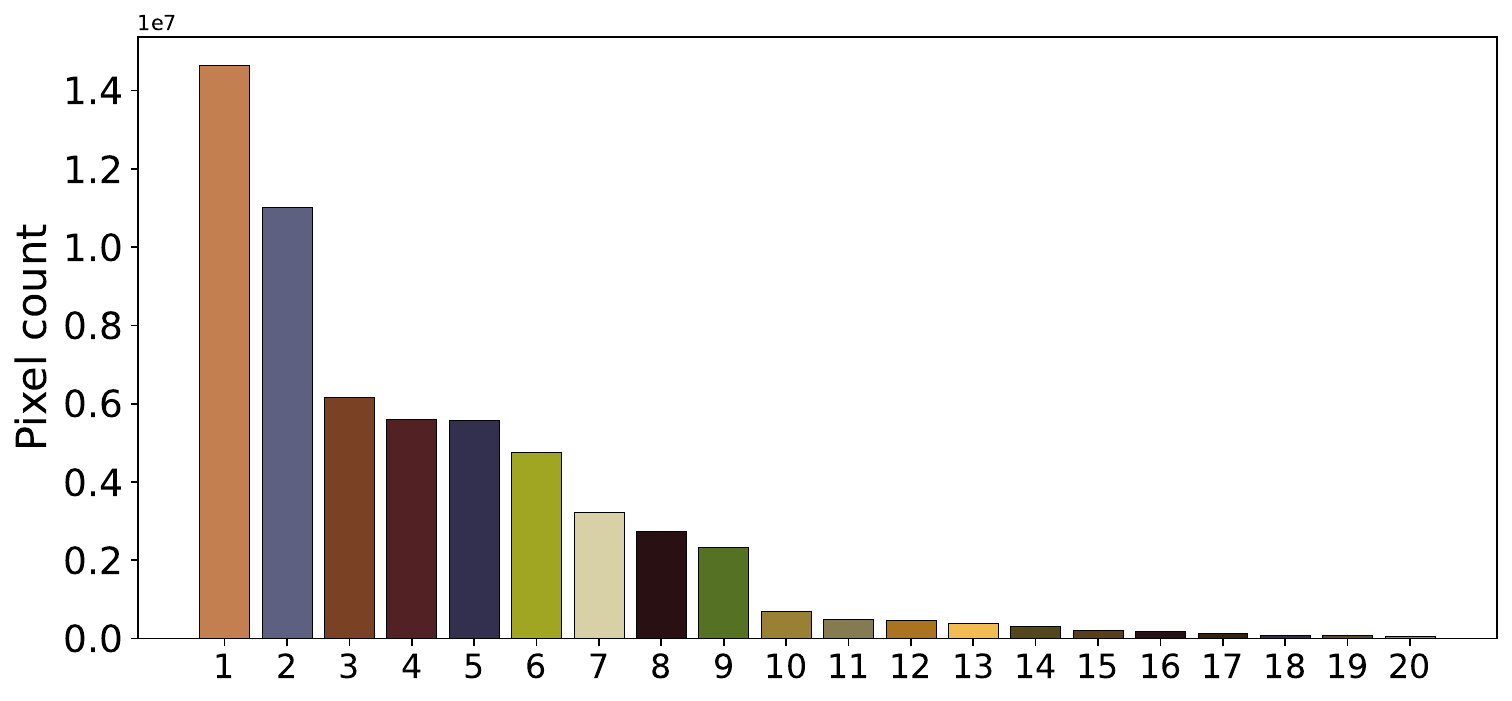}
        \caption{Top-20 colours by pixel count.}
        \label{fig:anita_palette_pix}
    \end{subfigure}
    \caption{Non-transparent colour-distribution visualisation for Anita-Pirate.}
    \label{fig:anita_palette_dist}
\end{figure*}

The final Anita-Pirate test set contains 28,773 annotated segments in total. Key dataset-level comparisons to PBC-3D and PBC-Real \cite{dai2024learning} are summarised in \cref{tab:dataset_stat_compare_transposed}. Beyond these per-frame statistics, we further visualise the palette distribution of Anita-Pirate in \cref{fig:anita_palette_dist}. These statistics indicate that the newly introduced test case Anita-Pirate features a longer sequence, denser region layouts, and a richer colour palette.

\begin{table}[t]
\centering
\caption{Statistical comparison of Anita-Pirate against existing test sets.}
\label{tab:dataset_stat_compare_transposed}
\begin{tabular}{l|ccc}
\toprule
\textbf{Dataset} 
& \textbf{PBC-3D} \cite{dai2024learning}
& \textbf{PBC-Real} \cite{dai2024learning}
& \textbf{Anita-Pirate (Ours)} \\
\midrule
Avg. Video Length (Frame) 
& 20 
& 10
& \textbf{206} \\

Frame Resolution 
& \(1024\times1024\) 
& mixed$^\dagger$ 
& \textbf{\(1920\times1080\)} \\

Avg. Regions per Frame 
& 68.26 
& 89.19 
& \textbf{139.67} \\

Unique RGBA Colours
& 196 
& 271 
& \textbf{366} \\
\bottomrule
\multicolumn{4}{l}{\tiny{$^\dagger$PBC-Real contains mixed resolutions: \(512\times512\), \(1024\times1024\), \(1280\times1280\), and \(1600\times1600\).}}
\end{tabular}
\end{table}

\section{Additional Implementation Details}
\label{sec:imple}
\subsection{Evaluation Metrics}
\label{sec:metrics}
In our experiments (\cref{sec:experiments}), we evaluate at both segment level and pixel level using exact discrete RGBA equality after palette colour decoding. Let $N$ be the number of valid target segments in a frame, $a_i$ be the pixel area of segment $i$, $y_i$ and $\hat y_i$ be ground-truth and predicted RGBA labels, and $\alpha(\cdot)$ denote the alpha channel. We list the complete metric calculations below:

\begin{equation}
\text{Acc}=\frac{1}{N}\sum_{i=1}^{N}\mathds{1}[\hat y_i=y_i].
\end{equation}

\begin{equation}
\text{Acc-Thresh}=\frac{1}{|I_{>10}|}\sum_{i\in I_{>10}}\mathds{1}[\hat y_i=y_i],\quad
I_{>10}=\{i\mid a_i>10\}.
\end{equation}

\begin{equation}
\text{Pix-Acc}=\frac{\sum_i a_i\,\mathds{1}[\hat y_i=y_i]}{\sum_i a_i}.
\end{equation}

\begin{equation}
\text{Pix-F-Acc}=\frac{\sum_i a_i\,\mathds{1}[\alpha(y_i)>0]\,\mathds{1}[\hat y_i=y_i]}
{\sum_i a_i\,\mathds{1}[\alpha(y_i)>0]},
\end{equation}
where the background is defined by transparency ($\alpha=0$). Let $b_i=\mathds{1}[\alpha(y_i)=0]$ and $\hat b_i=\mathds{1}[\alpha(\hat y_i)=0]$. The background IoU at pixel level is:
\begin{equation}
\text{Pix-B-MIoU}=
\frac{\sum_i a_i\,\mathds{1}[b_i=1\land \hat b_i=1]}
{\sum_i a_i\,\mathds{1}[b_i=1\lor \hat b_i=1]}.
\end{equation}

All final metrics are averaged over all per-frame metrics evaluated. Here, \texttt{Acc} measures segment-wise exact RGBA accuracy (all segments equally weighted), while \texttt{Acc-Thresh} excludes tiny segments to reduce noise. \texttt{Pix-Acc} is equivalent to pixel-value accuracy across all pixels, \texttt{Pix-F-Acc} restricts \texttt{Pix-Acc} to foreground (non-transparent) regions, and \texttt{Pix-B-MIoU} measures the IoU of predicted vs.\ ground-truth background (transparent) pixels.

\subsection{Details on Foundation Model Usages}
\label{sec:models}
In \cref{sec:experiments}, we evaluate several vision backbone models. For these foundation models or colourisation-pretrained models, segmented region descriptors are obtained by region pooling over dense feature maps within region masks obtained by flood-fill \cite{10.1145/800249.807456}, and matched by cosine similarity in $L_2$-normalised feature space. For colourisation-trained methods/backbones, we leverage their official checkpoints provided \cite{dai2024learning, dai2024paint, nagata2025dacon}. In addition to these previous works, we list the details of other frozen foundation model sources in \cref{tab:backbone_ckpt_details}. (Note that for Stable Diffusion \cite{Rombach_2022_CVPR}, we follow diffusion-feature extraction practice from existing works \cite{nagata2025dacon, tang2023emergent} with prompt ``a photo of an anime character.'') As shown, our experiments cover a broad spectrum of models with different capacities, supervisions, usages and resolutions, demonstrating the model-agnostic generality of \textsc{PeCA}.

Performance differences between backbones should therefore be interpreted as differences in descriptor quality and pretraining task bias. Self-supervised features tend to provide stronger region identity, while visual-language and generative diffusion model features are less directly optimised for local region correspondences. \textsc{PeCA} uses the same region-pooling and matching interface for all backbones, so its gains are measured relative to each backbone's base inference.

\begin{table}[h!]
\centering
\caption{Foundation backbones and configurations used in our inference pipeline.}
\label{tab:backbone_ckpt_details}
\resizebox{\linewidth}{!}{%
\begin{tabular}{l|l|c|l|c}
\toprule
\textbf{Backbone} & \textbf{Checkpoint / Model ID} & \textbf{Input Size} & \textbf{Feature Used} & \textbf{Source} \\
\midrule
DINOv2 ViT-L/14 & \texttt{facebookresearch/dinov2:dinov2\_vitl14} & $518{\times}518$ & final patch-token feature map & \cite{oquab2023dinov2} \\
CLIP ViT-L/14 & \texttt{ViT-L/14@336px} & $336{\times}336$ & visual encoder patch features & \cite{radford2021learning} \\
DINOv3 ConvNeXT-L & \texttt{timm/convnext\_large.dinov3\_lvd1689m} & $512{\times}512$ & \texttt{forward\_features} map (timm) & \cite{simeoni2025dinov3} \\
SigLIPv2 ViT-B/16 & \texttt{timm/vit\_base\_patch16\_siglip\_512.v2\_webli} & $512{\times}512$ & \texttt{forward\_features} map (timm) & \cite{tschannen2025siglip} \\
SAM2.1-Large & \texttt{facebook/sam2.1-hiera-large} & $512{\times}512$ & image predictor visual features & \cite{ravi2024sam} \\
Stable Diffusion 2.1 & \texttt{sd2-community/stable-diffusion-2-1} & $768{\times}768$ & U-Net first upsampling block, $t = 261/1000$ & \cite{Rombach_2022_CVPR} \\
\bottomrule
\end{tabular}%
}
\end{table}

\subsection{More Details on Active Reference Expansion (ARE).}
\label{sec:are_supp}
In \cref{sec:are}, we mentioned that ARE candidates are generated by applying joint palette-preserving geometric transforms to the reference triplet (line image, segment map, colour image). Specifically, the transform order follows:
\begin{equation}
T = T_{\text{affine}} \circ T_{90^\circ} \circ T_{\text{vflip}} \circ T_{\text{hflip}} .
\end{equation}

Each transform is independently determined by probability $p$. For affine, we use rotation + uniform scale + translation (no shear). Given an original pixel in homogeneous coordinates $\mathbf{p}=[x,y,1]^\top$, the transformed point is
\begin{equation}
\mathbf{p}'=\mathbf{A}\mathbf{p},\quad
\mathbf{A}=
\begin{bmatrix}
s\cos\theta & -s\sin\theta & \Delta x\\
s\sin\theta & \;\;s\cos\theta & \Delta y\\
0 & 0 & 1
\end{bmatrix},
\end{equation}
where $\theta\sim\mathcal U[-30^\circ,30^\circ]$, $s\sim\mathcal U[0.5,2.0]$, and $(\Delta x,\Delta y)$ is sampled translation. Following standard image affine warping, the transform is applied around the image centre $(c_x,c_y)$:
\begin{equation}
\mathbf{A}= \mathbf{T}(\Delta x,\Delta y)\,\mathbf{T}(c_x,c_y)\,\mathbf{R}(\theta)\,\mathbf{S}(s)\,\mathbf{T}(-c_x,-c_y).
\end{equation}

\begin{table}[t]
\centering
\caption{Geometric transformation parameters used in the Active Reference Expansion step (\cref{sec:are}).}
\label{tab:are_aug_params}
\resizebox{0.9\linewidth}{!}{%
\begin{tabular}{l|c|l}
\toprule
\textbf{Transform} & \textbf{Apply Prob.} & \textbf{Parameters} \\
\midrule
Horizontal flip ($T_{\text{hflip}}$) & $0.5$ & NA \\
Vertical flip ($T_{\text{vflip}}$) & $0.1$ & NA \\
$90^\circ$ rotation ($T_{90^\circ}$) & $0.2$ & $k\sim\mathrm{Unif}\{1,2,3\}$, rotate by $90^\circ\times k$ \\
\midrule
Affine ($T_{\text{affine}}$) & $1.0$ &
\makecell[l]{
angle $\theta\sim\mathrm{Unif}[-30^\circ,30^\circ]$; \\
translation $(\Delta x,\Delta y)$ up to $\pm 50\%$ of image width/height; \\
scale $s\sim\mathrm{Unif}[0.5,2.0]$; shear $=0$
} \\
\bottomrule
\end{tabular}%
}
\end{table}

We fix these transformation parameters (\cref{tab:are_aug_params}) for all experiments. When multiple reference images are provided, we split the budget \(B\) evenly across references: we generate \(mB/|\mathcal{R}|\) candidates and select \(B/|\mathcal{R}|\) views per reference. Lastly, the selection is computed against \(|\mathcal{T}_s|=20\) target frames uniformly subsampled from the target video to bound the selection cost.

\subsection{Computation Cost}
\label{sec:runtime}

\begin{figure}[t]
\centering 
\includegraphics[width=0.92\linewidth]{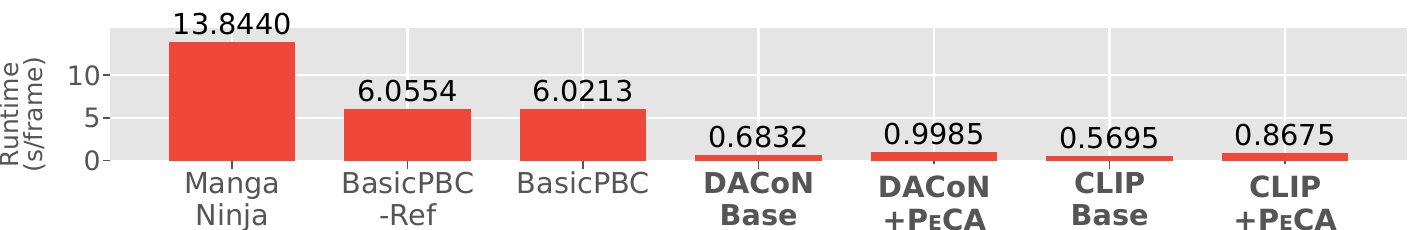}
\caption{\textbf{Overall latency per frame on Anita-Pirate.}
All measurements are obtained on an NVIDIA A100 GPU with batch size 1 and FP32 inference.}
\label{fig:cost_pirate}  
\end{figure}

We report overall runtime on Anita-Pirate, which exhibits the highest per-frame segment complexity among benchmarks (\cref{sec:pirate-detail}) using different backbone models with \textsc{PeCA}. All measurements are obtained on a single NVIDIA A100 GPU with batch size 1 and FP32 inference. We report the total runtime of each method under the same evaluation setting. As shown in \cref{fig:cost_pirate}, although \textsc{PeCA} introduces additional test-time computation, the overall pipeline remains substantially faster than earlier diffusion-based or inclusion-matching pipelines \cite{liu2025manganinja,dai2024paint,dai2024learning}.

Overall, these results suggest that the full method remains efficient. Note that all these runtimes are practical in production settings, where manual colouring typically requires minutes per tens of frames~\cite{maejima2021anime}.

\section{Further Comparisons to Previous Works}
\label{sec:fengetal}

\subsection{Comparison to a Unified Framework \cite{Feng_2025_ICCV}}
We did not report a direct quantitative comparison to Feng~\etal~\cite{Feng_2025_ICCV}, since their code, checkpoints, and new evaluation data are not publicly available at the time of submission. Moreover, their in-between evaluation protocol focuses on short clips (lengths of 3, 5, and 10 frames), rather than a complete video, which is not directly comparable to our in-between setting, which considers the rest of the whole video as the target. Nevertheless, we managed to test our method under shorter frame sequences (10 frames) as Feng~\etal~\cite{Feng_2025_ICCV} did on the PBC-3D dataset, with results shown in \cref{tab:feng_protocol_pbc3d_s10_final}. The simplest \textit{Base} training-free feature-matching baseline already achieves competitive performance with previous works. Adding \textsc{PeCA} on top of this baseline further improves the results.

\begin{table*}[h!]
\centering
\caption{\textbf{Short-sequence inbetweening colourisation on PBC-3D with shorter clips of 10 frames, following Feng~\etal~\cite{Feng_2025_ICCV}.} Training-free indicates whether the backbone model is trained on colourisation tasks. (Note that RAFT \cite{teed2020raft} used optical flow-based matching.)}
\label{tab:feng_protocol_pbc3d_s10_final}
\setlength{\tabcolsep}{5pt}
\renewcommand{\arraystretch}{1.05}
\resizebox{0.9\linewidth}{!}{
\begin{tabular}{l | c ccccc}
\toprule
\textbf{Method} & \textbf{Training-free} &
\textbf{Acc} $\uparrow$ &
\textbf{Acc-Thresh} $\uparrow$ &
\textbf{Pix-Acc} $\uparrow$ &
\textbf{Pix-F-Acc} $\uparrow$ &
\textbf{Pix-B-MIoU} $\uparrow$ \\
\midrule
ToonCrafter~\cite{xing2024tooncrafter} & \xmark & 9.69  & 13.11 & 22.48 & 12.92 & 25.64 \\
MangaNinja~\cite{liu2025manganinja}   & \xmark & 14.21 & 14.44 & 53.47 & 24.73 & 57.84 \\
LVCD~\cite{huang2024lvcd}             & \xmark & 26.59 & 28.66 & 58.38 & 42.94 & 60.19 \\
BasicPBC~\cite{dai2024learning}       & \xmark & 53.26 & 56.66 & 90.88 & 71.92 & 96.56 \\
Feng~\etal~\cite{Feng_2025_ICCV}      & \xmark & 68.67 & 72.63 & 95.42 & 87.09 & 97.80 \\
\midrule
RAFT~\cite{teed2020raft}              & \cmark & 32.06 & 36.07 & 60.74 & 52.88 & 88.80 \\
\midrule
SAM2.1                          & \cmark & 67.92 & 72.13 & 96.40 & 89.33 & 98.49 \\
\rowcolor{gray!12}\textbf{SAM2.1 + \textsc{PeCA}} & \cmark & \textbf{73.18} & \textbf{77.17} & \textbf{97.16} & \textbf{92.48} & \textbf{98.73} \\
\bottomrule
\end{tabular}}
\end{table*}

\subsection{Details on More Recent Pixel-Generative Baselines}
\label{sec:supp_modern_pixel_baselines}

\begin{figure}[h!]
    \centering
    {
    \setlength{\tabcolsep}{1pt}
    \resizebox{\linewidth}{!}{%
    \begin{tabular}{cccccc}
        \multicolumn{6}{c}{\textit{PBC-3D: key-frame reference}} \\
        \scriptsize Reference & \scriptsize Target & \scriptsize Cobra \cite{zhuang2025cobra} & \scriptsize MagicColor \cite{zhang2025followyourcolormultiinstancesketchcolorization} & \scriptsize \textsc{\textbf{PeCA (Ours)}} & \scriptsize GT \\
        \includegraphics[width=0.152\linewidth]{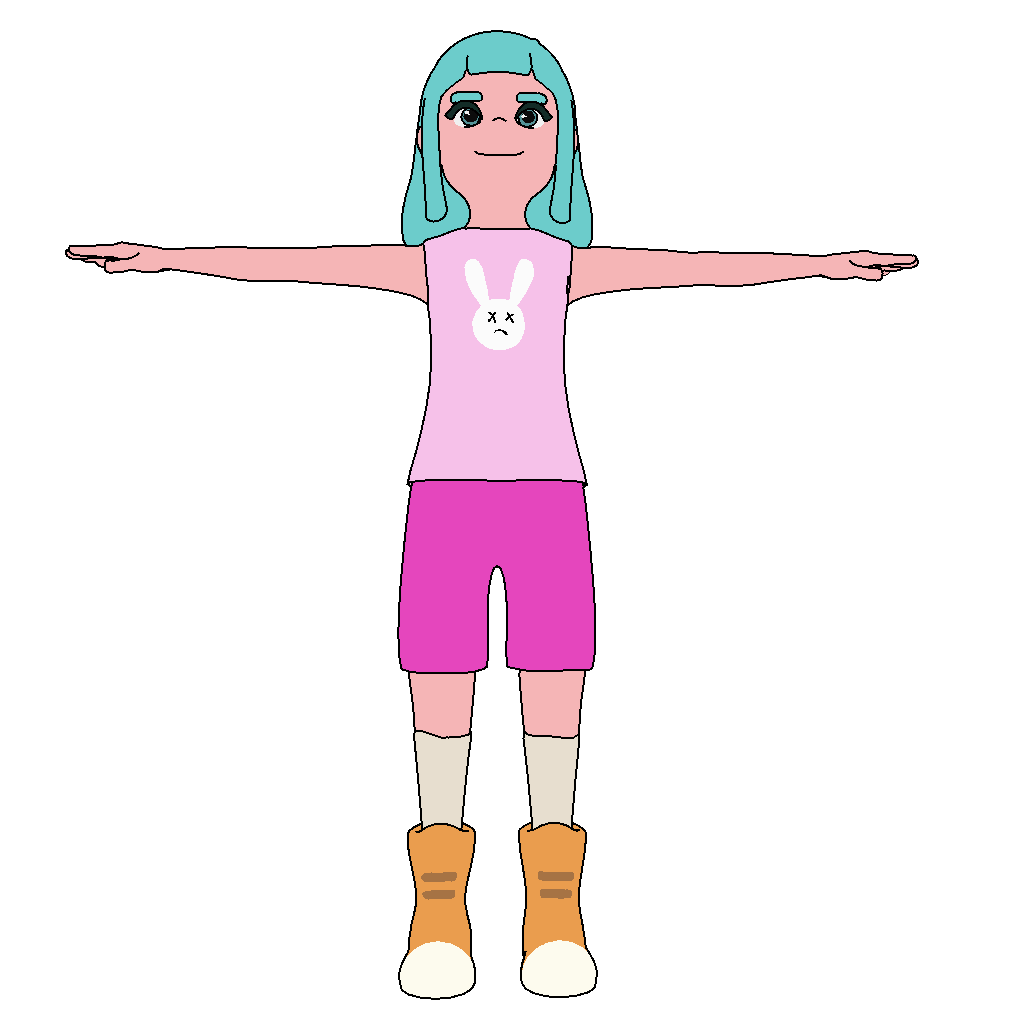} &
        \includegraphics[width=0.152\linewidth]{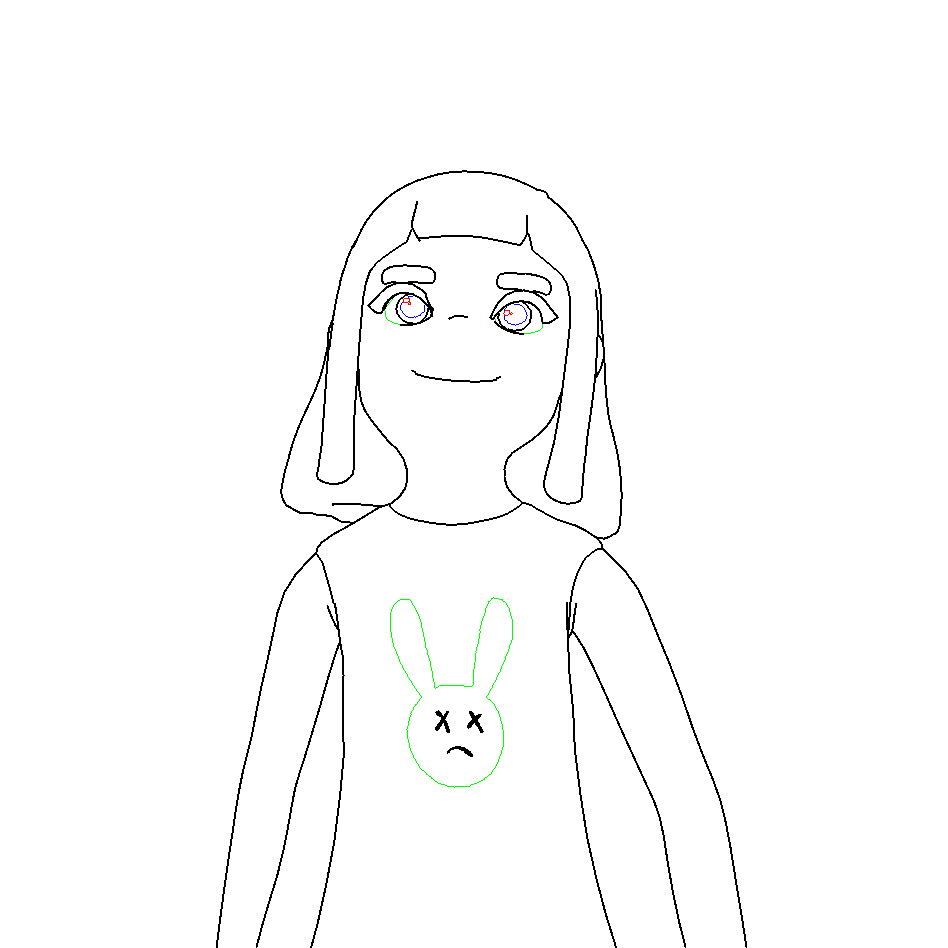} &
        \includegraphics[width=0.152\linewidth]{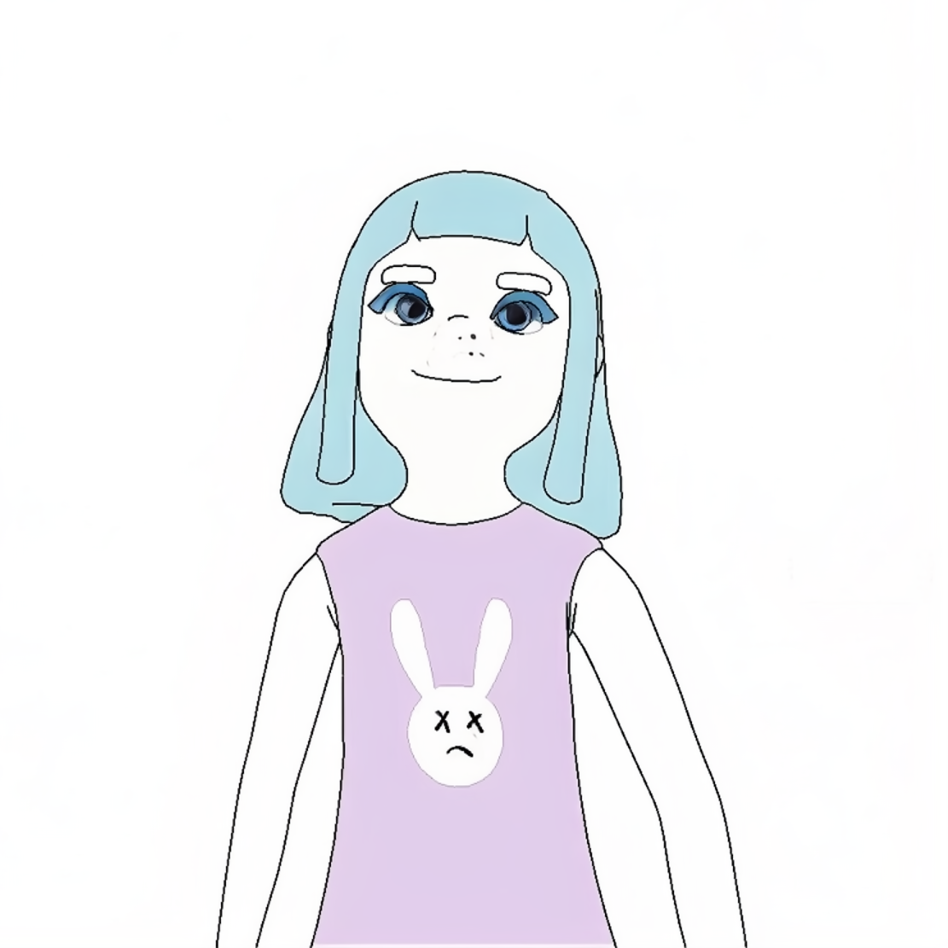} &
        \includegraphics[width=0.152\linewidth]{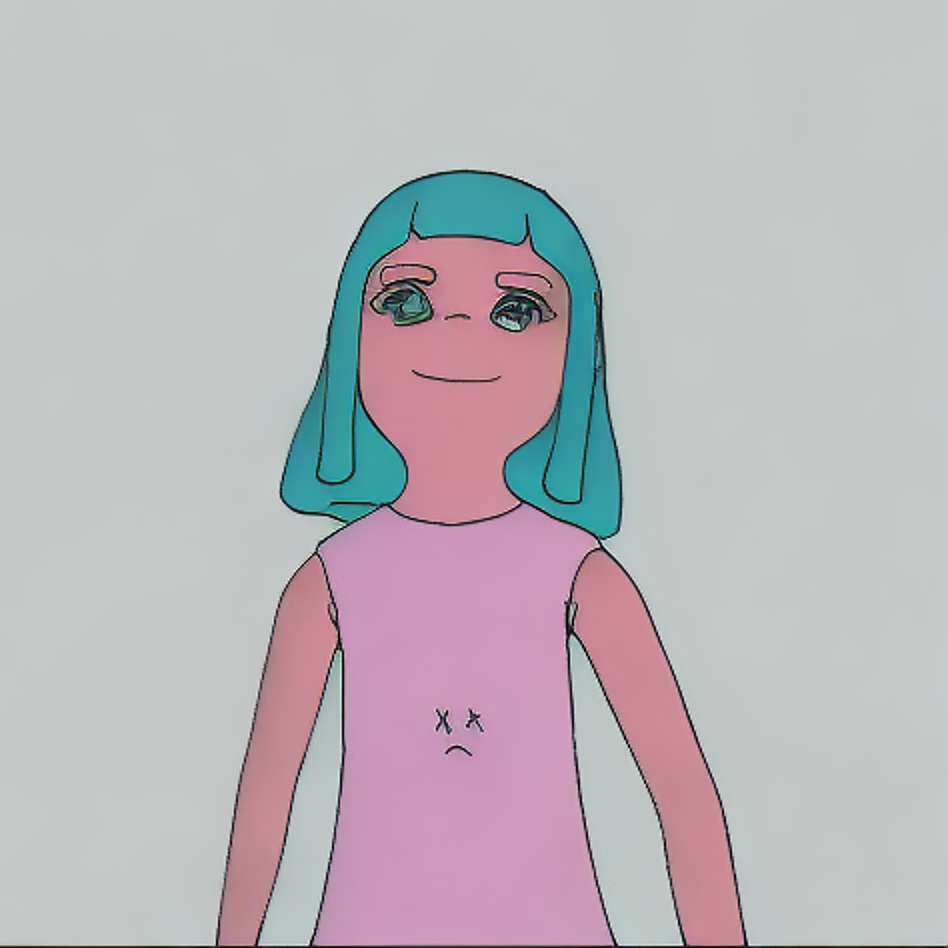} &
        \includegraphics[width=0.152\linewidth]{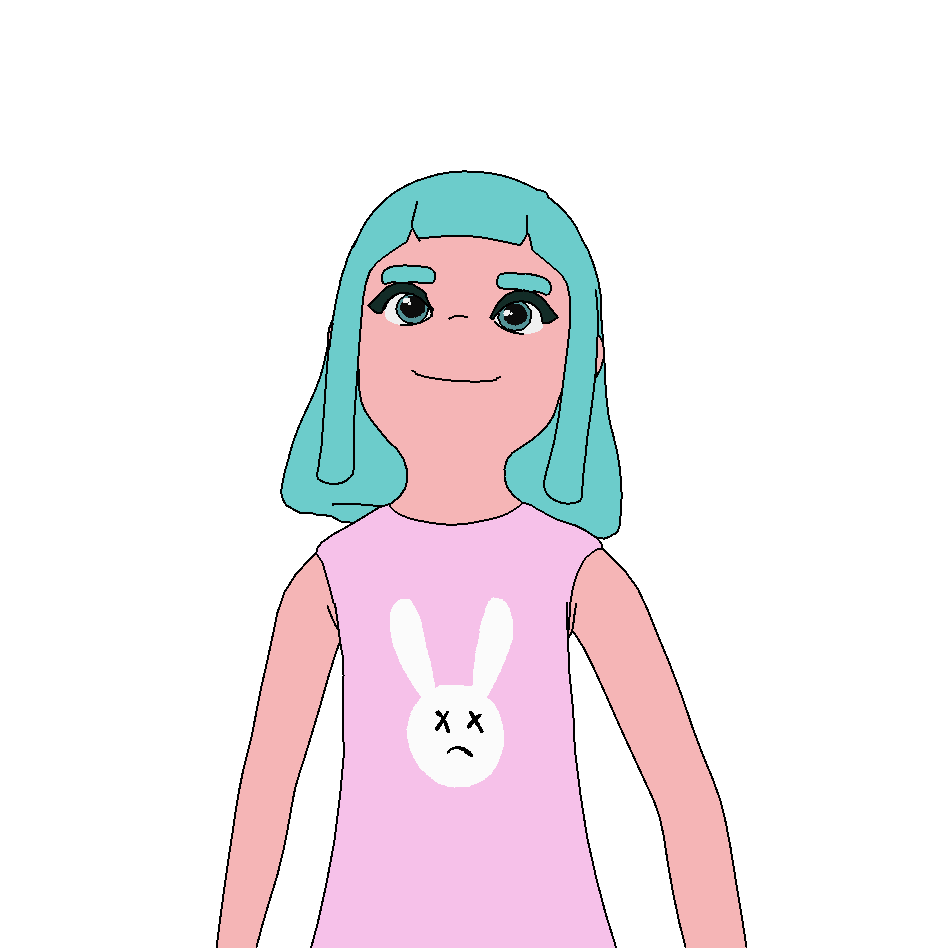} &
        \includegraphics[width=0.152\linewidth]{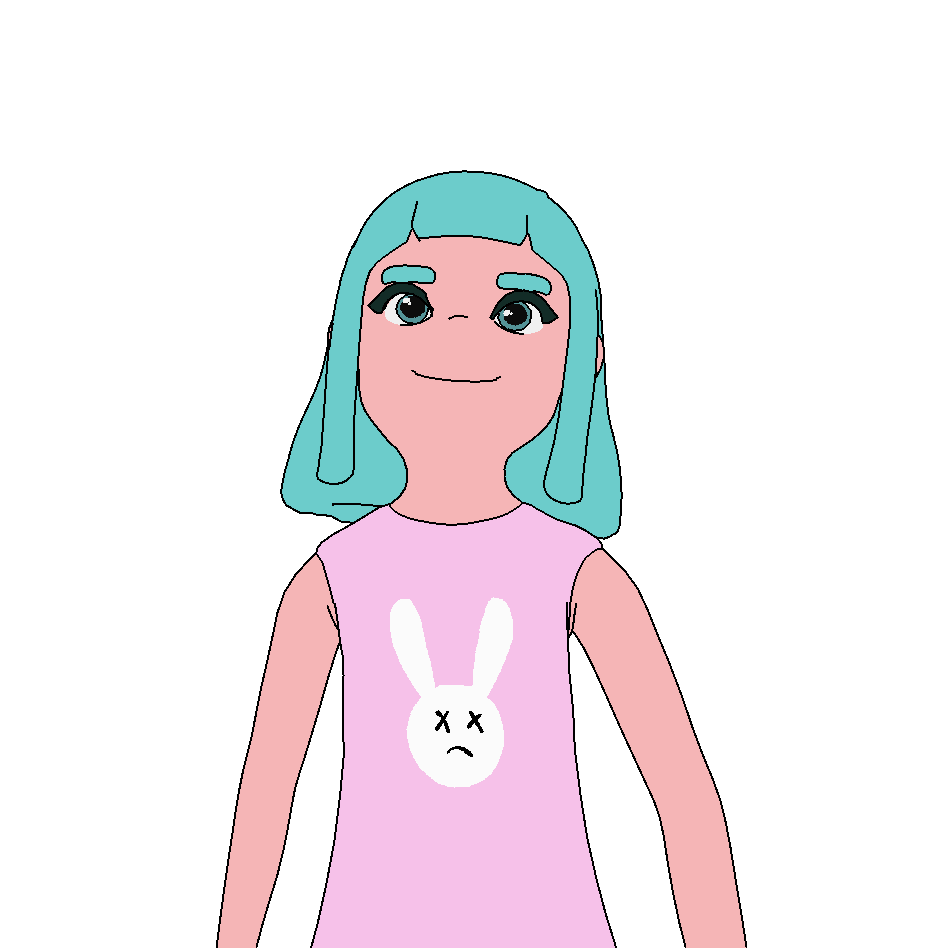} \\[0.2em]
        \multicolumn{6}{c}{\textit{PBC-Real: first-frame reference}} \\
        \scriptsize Reference & \scriptsize Target & \scriptsize ToonComposer \cite{li2025tooncomposer} & \scriptsize Nano Banana 2 \cite{blogNanoBanana} & \scriptsize \textsc{\textbf{PeCA (Ours)}} & \scriptsize GT \\
        \includegraphics[width=0.152\linewidth]{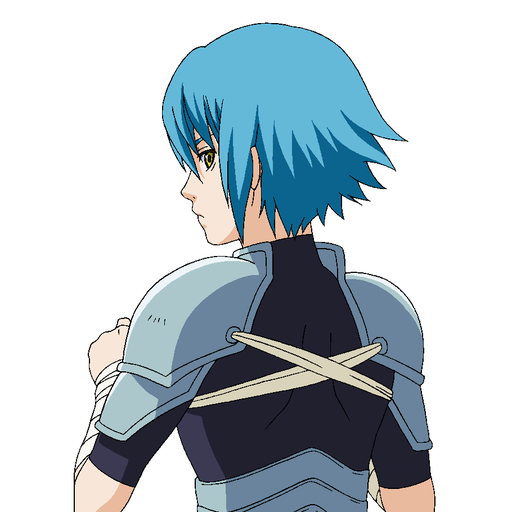} &
        \includegraphics[width=0.152\linewidth]{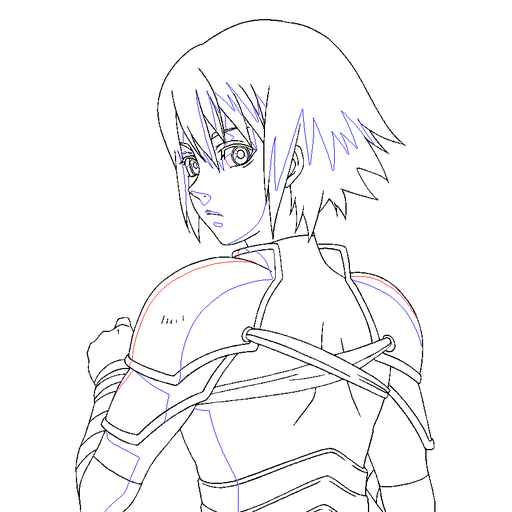} &
        \includegraphics[width=0.152\linewidth]{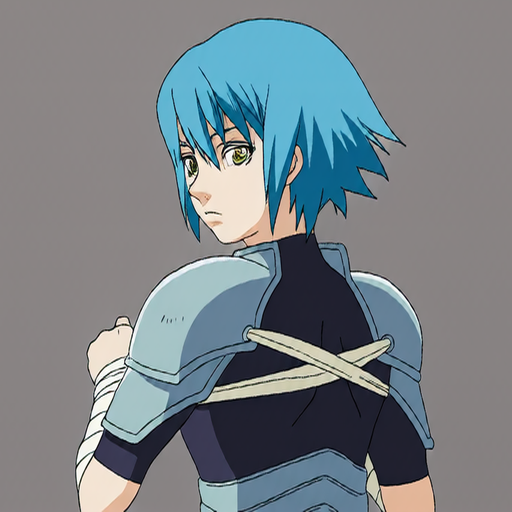} &
        \includegraphics[width=0.152\linewidth]{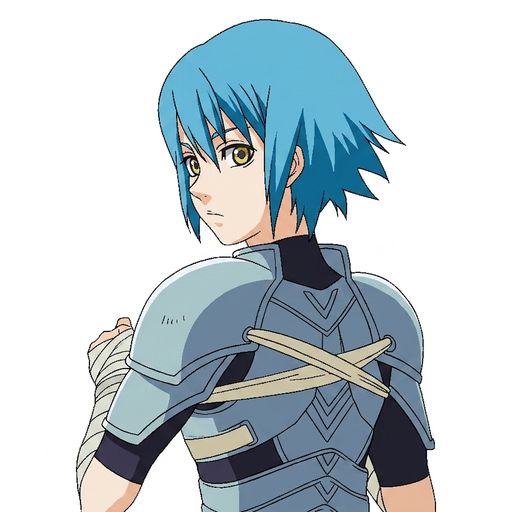} &
        \includegraphics[width=0.152\linewidth]{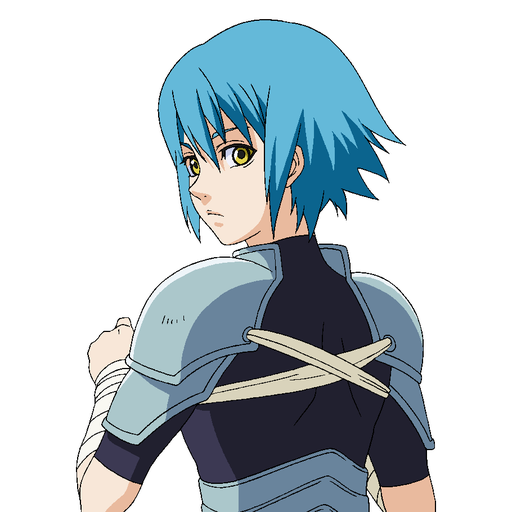} &
        \includegraphics[width=0.152\linewidth]{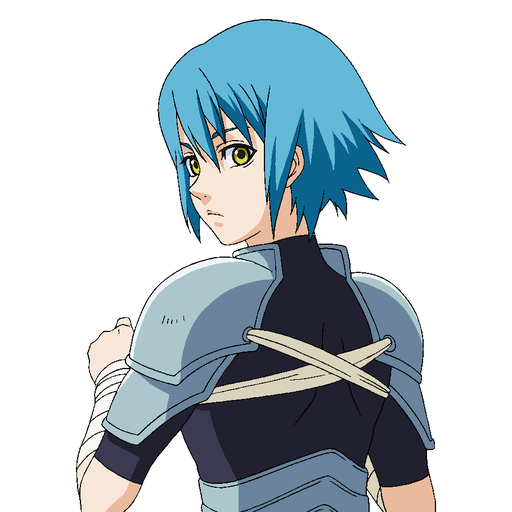} \\
    \end{tabular}%
    }}
    \caption{\textbf{Raw qualitative comparison for modern pixel-generative baselines.} We show the references, target line sketches, raw RGB outputs and GT under the PBC-3D key-frame reference protocol and the PBC-Real first-frame reference protocol.}
    \label{fig:supp_modern_pixel_qual}
\end{figure}

Here we provide further details and additional comparisons against more recent pixel-generative baselines. Since these methods output RGB images (as shown in \cref{fig:supp_modern_pixel_qual}) rather than region-to-palette assignments, we follow the DACoN post-processing protocol~\cite{nagata2025dacon}. \Ie, We resize each generated image to the target resolution, replace each pixel with the nearest colour from the reference palette, and unify each line-enclosed segment to its most frequent projected colour. The resulting palette-preserving images are then evaluated by the same metrics to previous works \cite{nagata2025dacon, dai2024paint}.

Unless otherwise stated, we use the official/default inference settings for each baseline. For Nano Banana 2, a closed-source model, we use the Google Cloud API \texttt{(gemini-3.1-flash-image-preview) \cite{blogNanoBanana}} with the default generation api call and the following prompt:

\begin{center}
\begin{promptblock}{\scriptsize Prompt for Nano Banana 2 \texttt{(gemini-3.1-flash-image-preview) \cite{blogNanoBanana}}}
{\small\ttfamily\color{black}\raggedright
 \scriptsize Colorize the target line art using the colored reference character image. The first image is the colored reference. The second image is the target line art. Preserve the target line art, pose, composition, and plain background. Use only colors visible in the reference image. Return only the final colored image.\par}
\end{promptblock}
\end{center}

\begin{table}[h!]
\centering
\caption{\textbf{Modern pixel-generative baseline comparisons.} Left: PBC-3D under the one-shot design-sheet key-frame reference protocol; right: PBC-Real under the first-frame reference protocol. RGB generation baselines are evaluated after DACoN-style post-processing before computing paint-bucket metrics.}
\label{tab:supp_modern_pixel_baselines}
\label{tab:supp_modern_pbc3d}
\label{tab:supp_modern_pbcreal}
{
\begin{minipage}[t]{0.495\linewidth}
\centering
\resizebox{\linewidth}{!}{%
\begin{tabular}{l|ccccc}
\toprule
\multicolumn{6}{c}{\textit{PBC-3D: key-frame reference}} \\
\midrule
\textbf{Method} & \textbf{Acc} & \textbf{Acc-Thresh} & \textbf{Pix-Acc} & \textbf{Pix-F-Acc} & \textbf{Pix-B-MIoU} \\
\midrule
ColorFlow~\cite{zhuang2024colorflow} & 9.72 & 10.81 & 50.64 & 9.16 & 57.17 \\
AniDoc~\cite{meng2025anidoc} & 19.80 & 22.68 & 77.38 & 46.46 & 87.32 \\
Cobra~\cite{zhuang2025cobra} & 15.06 & 17.26 & 69.20 & 19.72 & 82.69 \\
MagicColor~\cite{zhang2025followyourcolormultiinstancesketchcolorization} & 21.48 & 24.81 & 16.34 & 44.04 & 7.63 \\
\rowcolor{gray!12}
\textbf{DACoN~1.1 + \textsc{PeCA}} & \textbf{72.04} & \textbf{77.08} & \textbf{97.90} & \textbf{94.04} & \textbf{99.42} \\
\bottomrule
\end{tabular}%
}
\end{minipage}\hfill
\begin{minipage}[t]{0.495\linewidth}
\centering

\resizebox{\linewidth}{!}{%
\begin{tabular}{l|ccccc}
\toprule
\multicolumn{6}{c}{\textit{PBC-Real: first-frame reference}} \\
\midrule
\textbf{Method} & \textbf{Acc} & \textbf{Acc-Thresh} & \textbf{Pix-Acc} & \textbf{Pix-F-Acc} & \textbf{Pix-B-MIoU} \\
\midrule
AnimeColor~\cite{zhang2025animecolor} & 37.39 & 40.22 & 85.25 & 59.24 & 90.19 \\
ToonComposer~\cite{li2025tooncomposer} & 29.03 & 31.28 & 32.43 & 48.02 & 22.62 \\
Nano Banana 2~\cite{blogNanoBanana} & 47.78 & 52.17 & 90.39 & 71.63 & 98.46 \\
DACoN~1.1~\cite{nagata2025dacon} & 65.82 & 69.11 & 94.18 & 80.68 & 98.76 \\
\rowcolor{gray!12}
\textbf{DACoN~1.1 + \textsc{PeCA}} & \textbf{67.64} & \textbf{71.29} & \textbf{94.70} & \textbf{82.11} & \textbf{99.48} \\
\bottomrule
\end{tabular}%
}
\end{minipage}}
\end{table}

 The evaluation results in \cref{tab:supp_modern_pixel_baselines} and \cref{fig:supp_modern_pixel_qual} show that pixel generative methods, even when using much more powerful model structures \cite{Rombach_2022_CVPR, Peebles2022DiT} failed to provide precise colourisation over the simple line sketch inputs. All of them appear to be altering the original line structure with some artefacts. Specifically, a stronger closed-source Nano Banana 2 model \cite{blogNanoBanana} hallucinates significant patterns that never existed in input line sketches. Apart from these, some methods \cite{zhang2025followyourcolormultiinstancesketchcolorization, li2025tooncomposer} show background bias, which failed to predict emptiness for background, this may be inherited from their pretraining dataset Sakuga-42M \cite{sakuga42m2024}, which always has rich backgrounds from completed animation.

\section{Additional Analysis on Reference Expansion and Probability Aggregation}
\label{sec:pa_additional_analysis}

As discussed in \cref{sec:are,sec:pa}, increasing the number of reference views has two opposing effects. On the one hand, more views improve target coverage and increase the chance of retrieving the correct correspondence, which is the main motivation behind Active Reference Expansion (ARE). On the other hand, a larger reference pool also introduces more distractor matches, so the benefit of additional views can only be realised when the aggregation rule is sufficiently selective. This is exactly the role of our Probability Aggregation (PA). In other words, ARE enlarges the pool of potentially useful colour evidence, while PA is needed to convert that larger evidence pool into actual gains without suffering from dilution.

To make this connection more explicit, we analyse how the quality of the induced colour distribution changes as the number of reference views increases. We use DINOv2 \cite{oquab2023dinov2} as the frozen backbone, keep all other settings fixed, and vary the total number of random reference views included as \(R\in\{1,31,63,128\}\).

We compare three inference-time colour propagation rules from correspondence probabilities: top-1 hard copy as in \cref{eq:hardmatch}, full linear combination over all source matches \cite{casey2021animation} (can be viewed as a special case of \cref{eq:pa} with $k = +\infty, \tau=1$), and the PA soft-voting introduced in \cref{sec:pa}. Note that PA soft-voting uses the default hyperparameters in \cref{eq:pa}.

To characterise the quality of the colourisation probability \(P_{t,i}\), we measure it from two complementary aspects. First, \textit{Uncertainty} is measured by \texttt{Entropy}, which quantifies how concentrated the predicted colour distribution is (lower is better). Second, \textit{Discriminability} is measured by \texttt{GT Margin}, which quantifies how strongly the ground-truth colour is separated from the strongest competing colour (higher is better). We compute both metrics on non-transparent segments only (\(\alpha(y^{gt}_{t,i})>0\)). Let \(N_{\mathrm{nt}}\) denote the total number of non-transparent segments in the evaluation frames:
\begin{equation}
\mathrm{Entropy}
=
\frac{1}{N_{\mathrm{nt}}}
\sum_{\alpha(y^{gt}_{t,i})>0}
\left[
-\sum_{c\in\mathcal{C}} P_{t,i}(c)\log P_{t,i}(c)
\right],
\end{equation}
\begin{equation}
\mathrm{GT\ Margin}
=
\frac{1}{N_{\mathrm{nt}}}
\sum_{\alpha(y^{gt}_{t,i})>0}
\left[
P_{t,i}(y^{gt}_{t,i})-\max_{c\neq y^{gt}_{t,i}}P_{t,i}(c)
\right].
\end{equation}

\begin{figure}[tbp]
    \centering
    \includegraphics[width=0.95\linewidth]{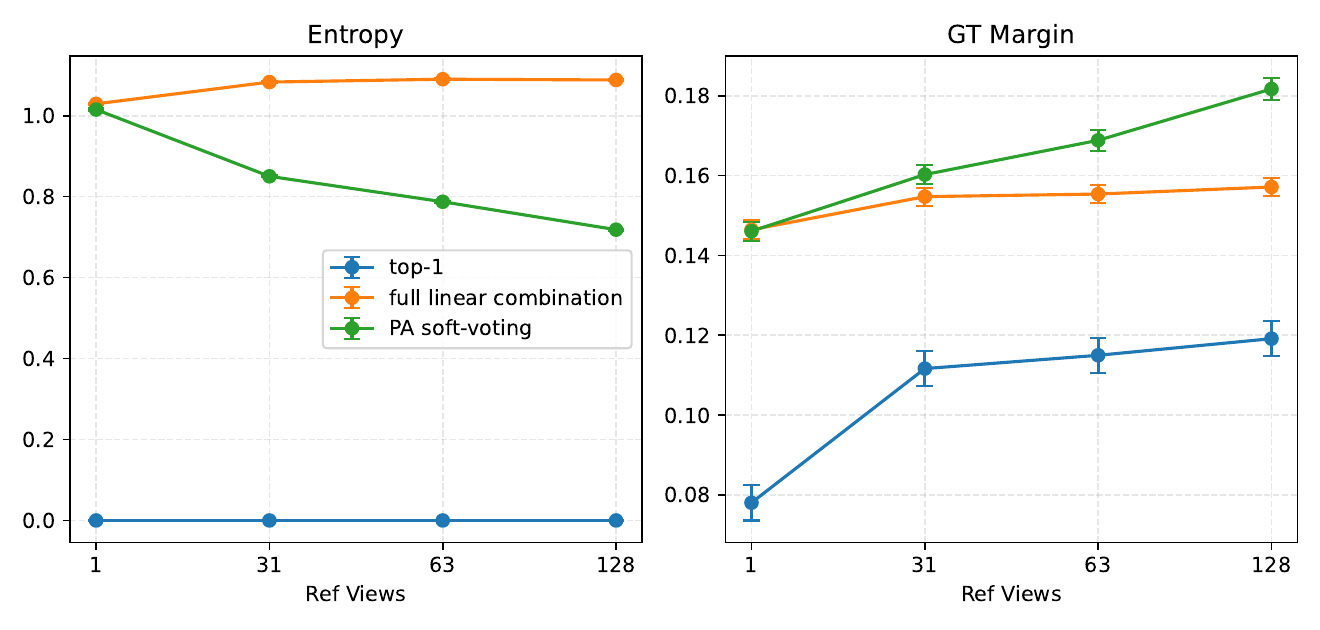}
    \caption{\textbf{Colour probability quality vs. number of reference views \(R\).}
    \texttt{Entropy} ($\downarrow$) measures uncertainty of \(P_{t,i}\), and \texttt{GT Margin} ($\uparrow$) measures the probability gap between the ground-truth colour and the strongest competitor. Top-1 hard copy yields near-zero \texttt{Entropy} by construction (one-hot), but has the lowest \texttt{GT Margin}, indicating less robust decisions under ambiguous matches. As \(R\) increases, PA soft-voting improves both confidence (lower \texttt{Entropy}) and discriminability (higher \texttt{GT Margin}), while full linear combination saturates due to evidence dilution.}
    \label{fig:pa_entropy_margin_refviews}
\end{figure}

We conduct the above experiments under the one-shot key-frame colourisation setting on PBC-3D \cite{dai2024learning}. From \cref{fig:pa_entropy_margin_refviews}, we observe three consistent trends beyond the performance gains already reported in the ablations (\cref{tab:ablation_merged}). First, top-1 prediction keeps near-zero \texttt{Entropy} across all \(R\), which is expected from its one-hot prediction, but it also yields the lowest \texttt{GT Margin}, indicating limited robustness when the best match is ambiguous. Second, a full linear combination shows the opposite trend: as \(R\) grows, \texttt{Entropy} further increases and then saturates at a relatively high level, while \texttt{GT Margin} improves only mildly. This indicates that simply adding more views is not sufficient; without the selective soft-voting, the additional evidence is increasingly diluted by distractor correspondences. This observation is also consistent with the hyperparameter analysis in \cref{fig:ablate_tau_topk_seg,fig:ablate_tau_topk_pix}: enlarging the aggregation range by increasing top-\(k\) or \(\tau\), thereby moving the behaviour closer to full combination, leads to clear performance drop, while reducing top-\(k\), which pushes the model towards top-1 direct matching, also weakens the benefit of expanding reference set.

In contrast, PA soft-voting exhibits the desired behaviour for leveraging larger reference pools: as \(R\) increases, \texttt{Entropy} decreases while \texttt{GT Margin} increases steadily, and the gap to both top-1 and full linear combination becomes more pronounced at larger \(R\). Taken together, these results support both the motivation that expanding reference views indeed provides more useful matching evidence, but such possibly noisy evidence requires a controlled aggregation mechanism to be effectively converted into performance gains rather than through indiscriminate mixing.

\begin{figure}[h!]
    \centering
    {
    \scriptsize
    \setlength{\tabcolsep}{0.35pt}
    \begin{tabular}{@{}cc|cccc|c@{}}
    \toprule
        Ref. & Target & w/o PA & \makecell{$k{=}16$\\$\tau{=}0.025$} & \makecell{\boldmath$k{=}64$\\\boldmath$\tau{=}0.05$} & \makecell{$k{=}128$\\$\tau{=}0.1$} & GT \\
        \midrule
        \includegraphics[width=0.132\linewidth]{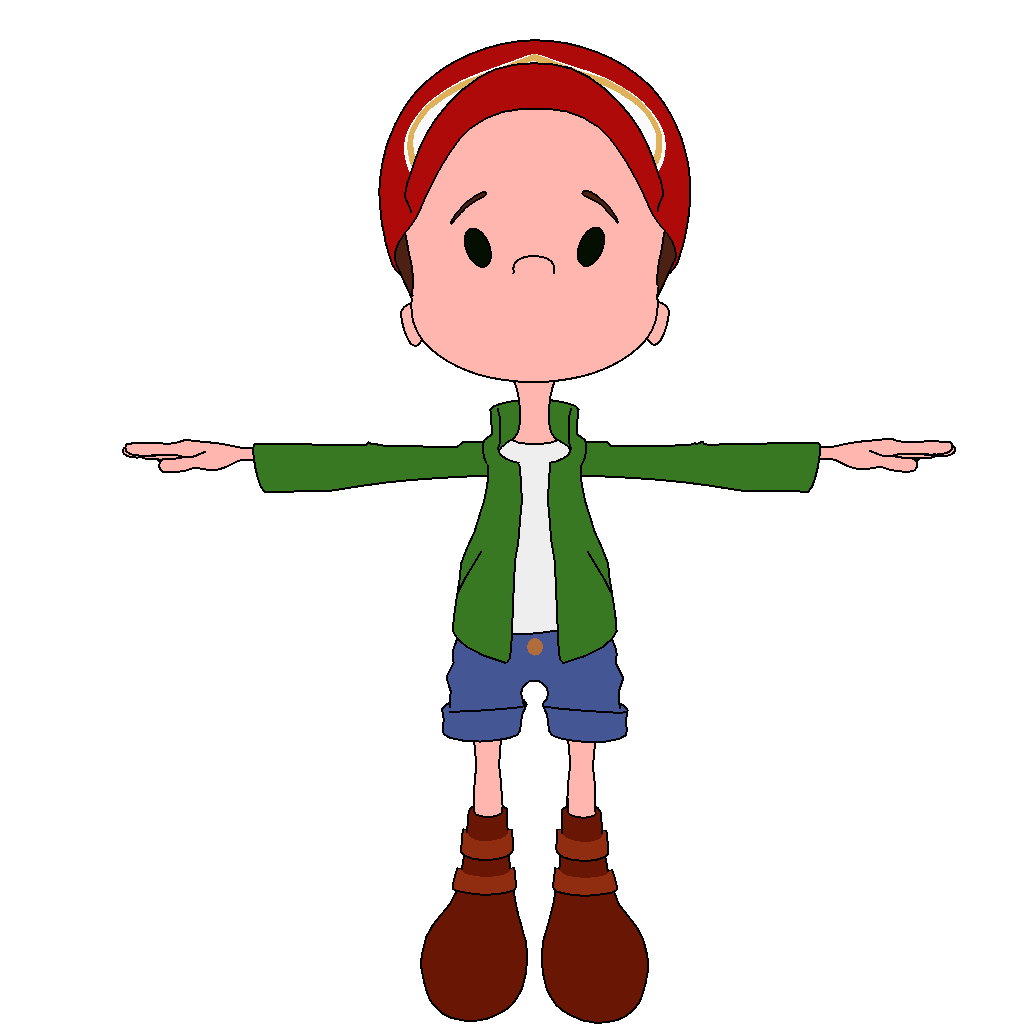} &
        \includegraphics[width=0.132\linewidth]{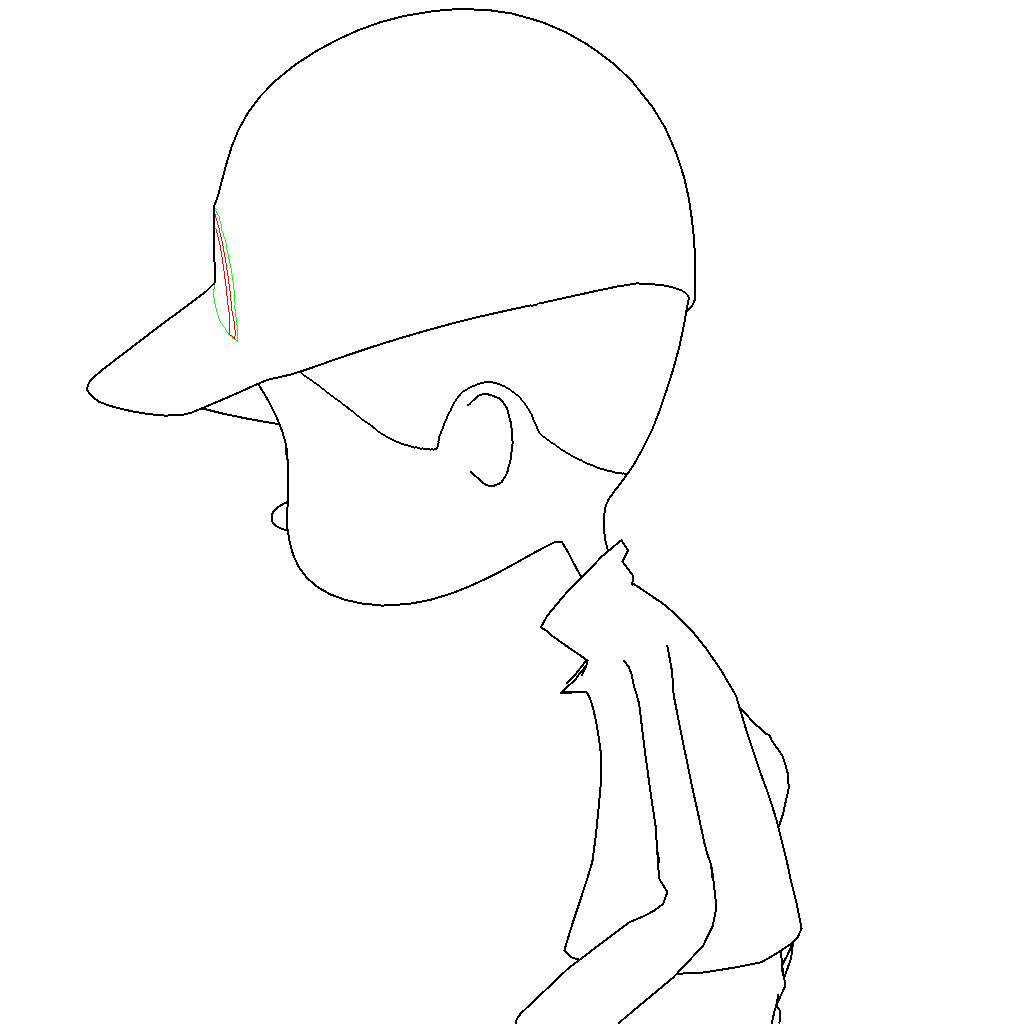} &
        \includegraphics[width=0.132\linewidth]{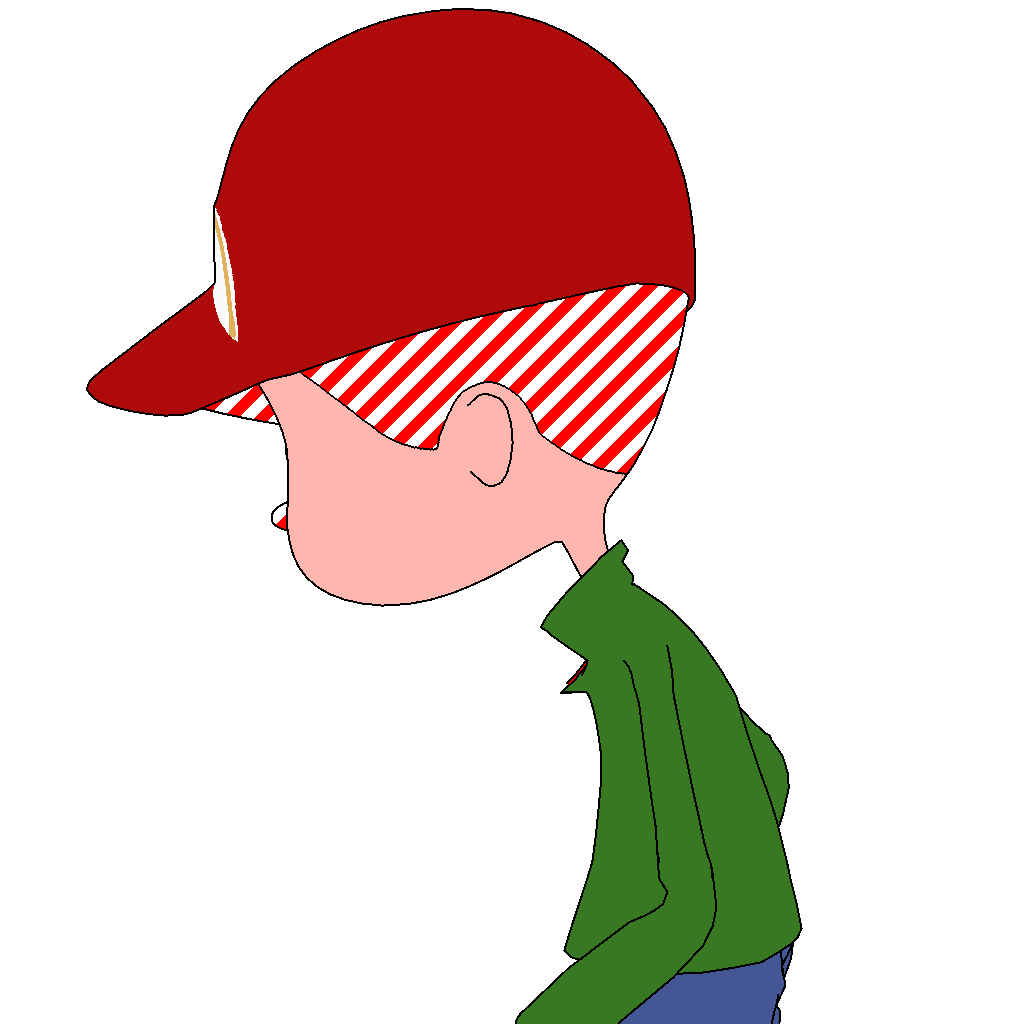} &
        \includegraphics[width=0.132\linewidth]{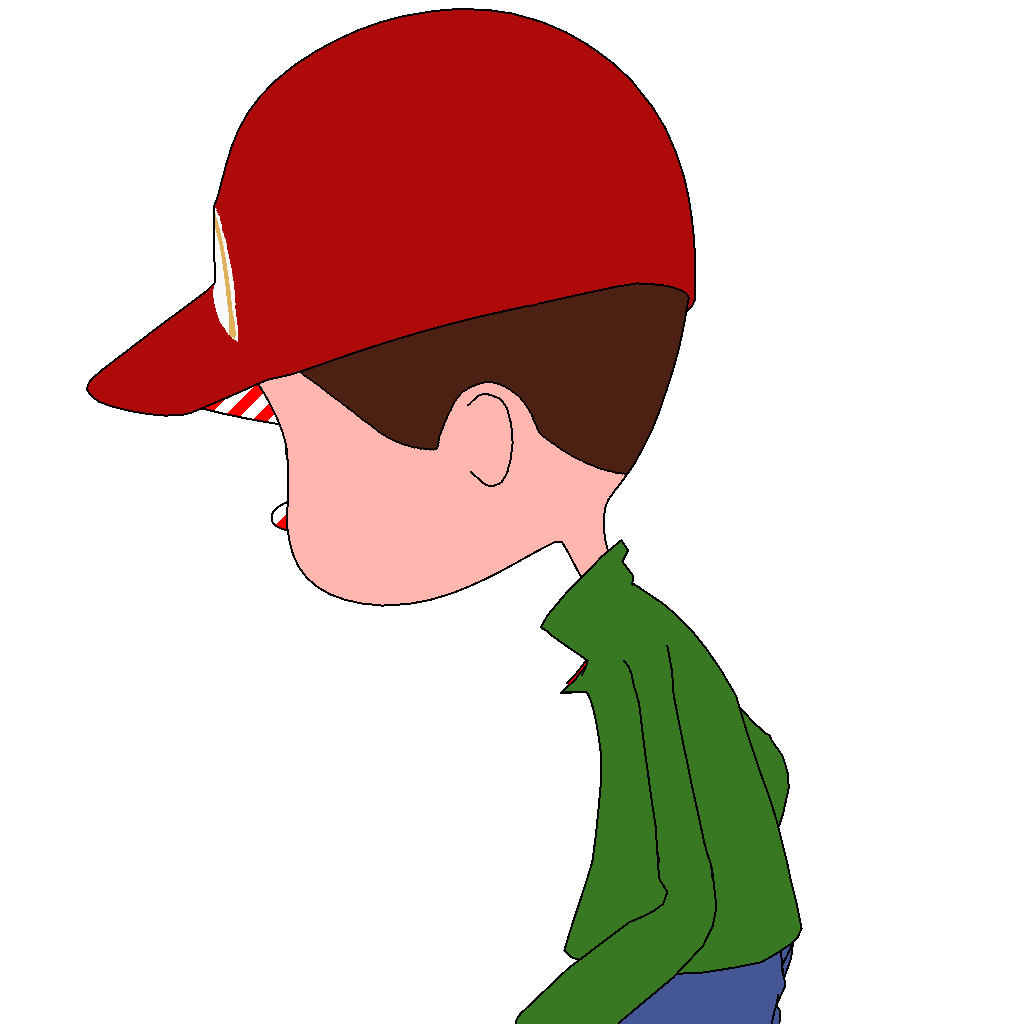} &
        \includegraphics[width=0.132\linewidth]{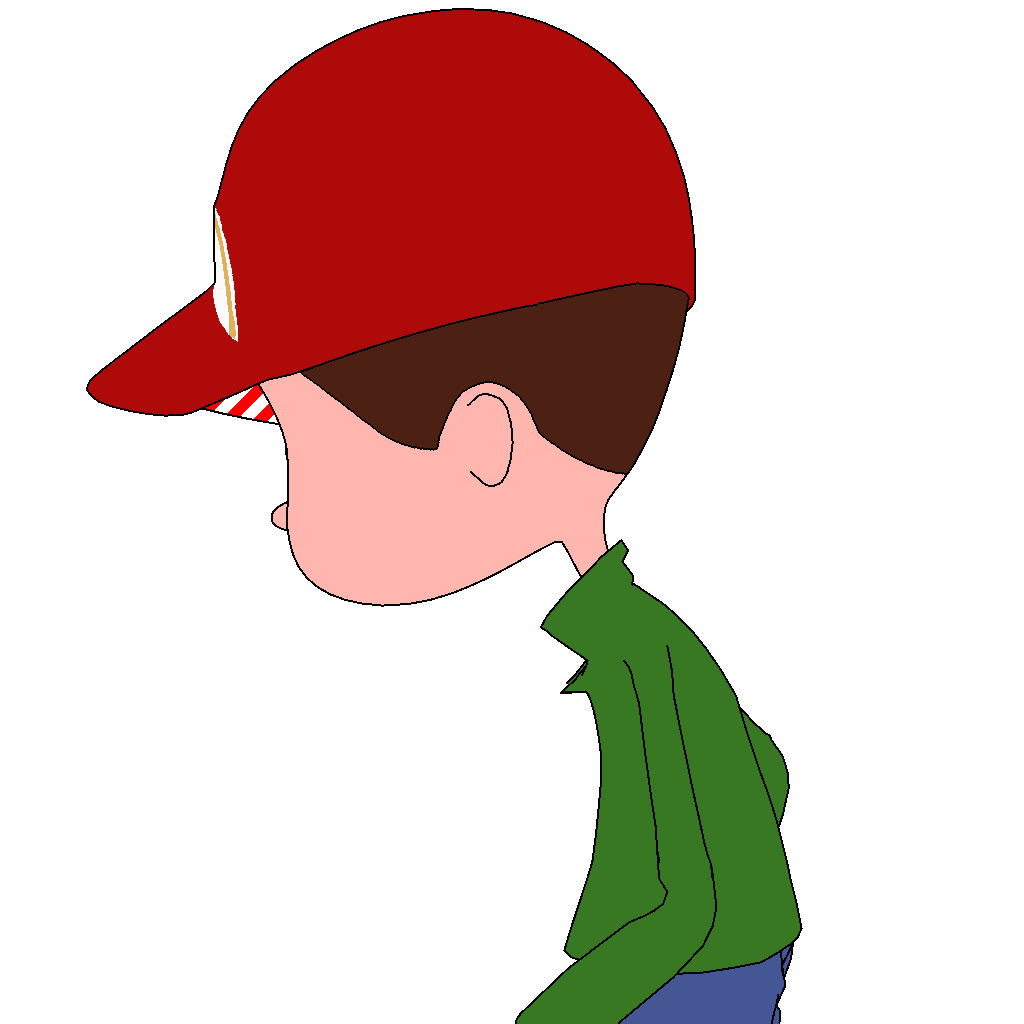} &
        \includegraphics[width=0.132\linewidth]{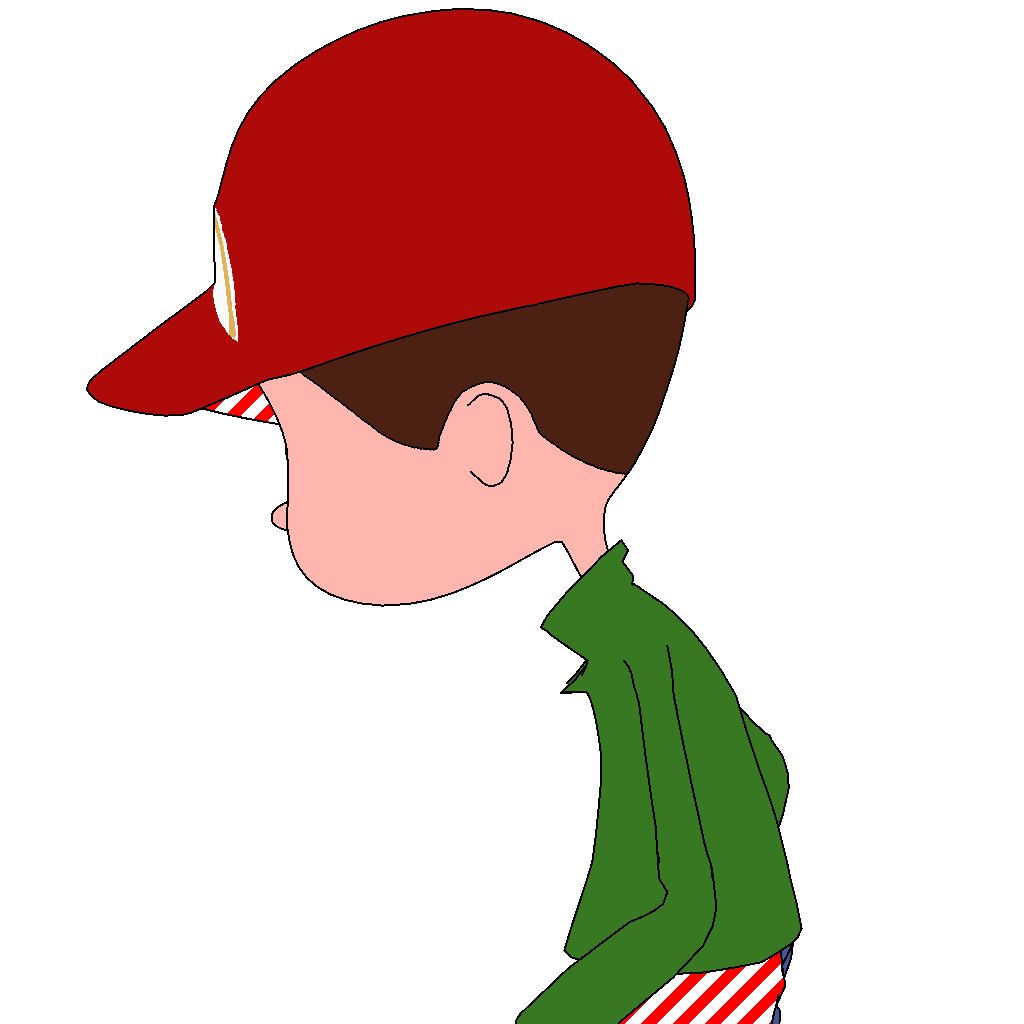} &
        \includegraphics[width=0.132\linewidth]{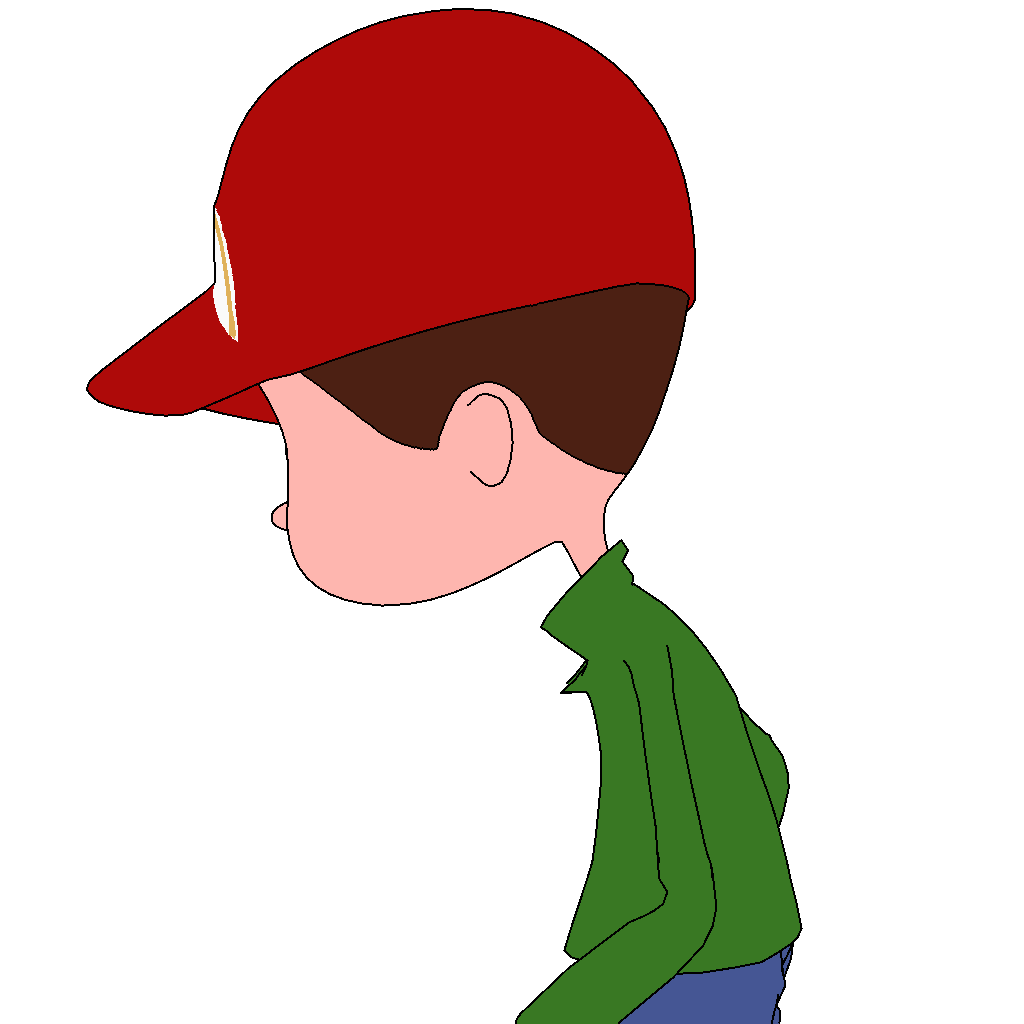} \\
        \includegraphics[width=0.132\linewidth]{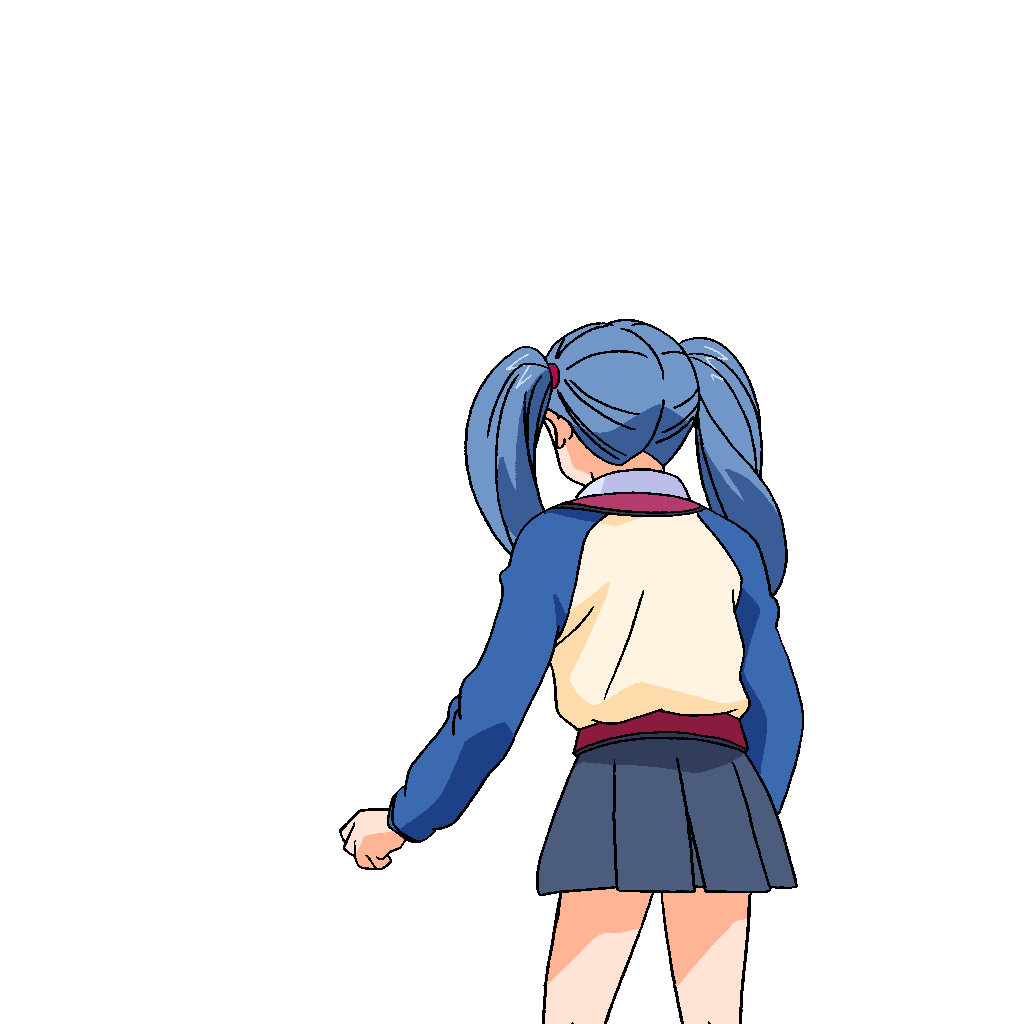} &
        \includegraphics[width=0.132\linewidth]{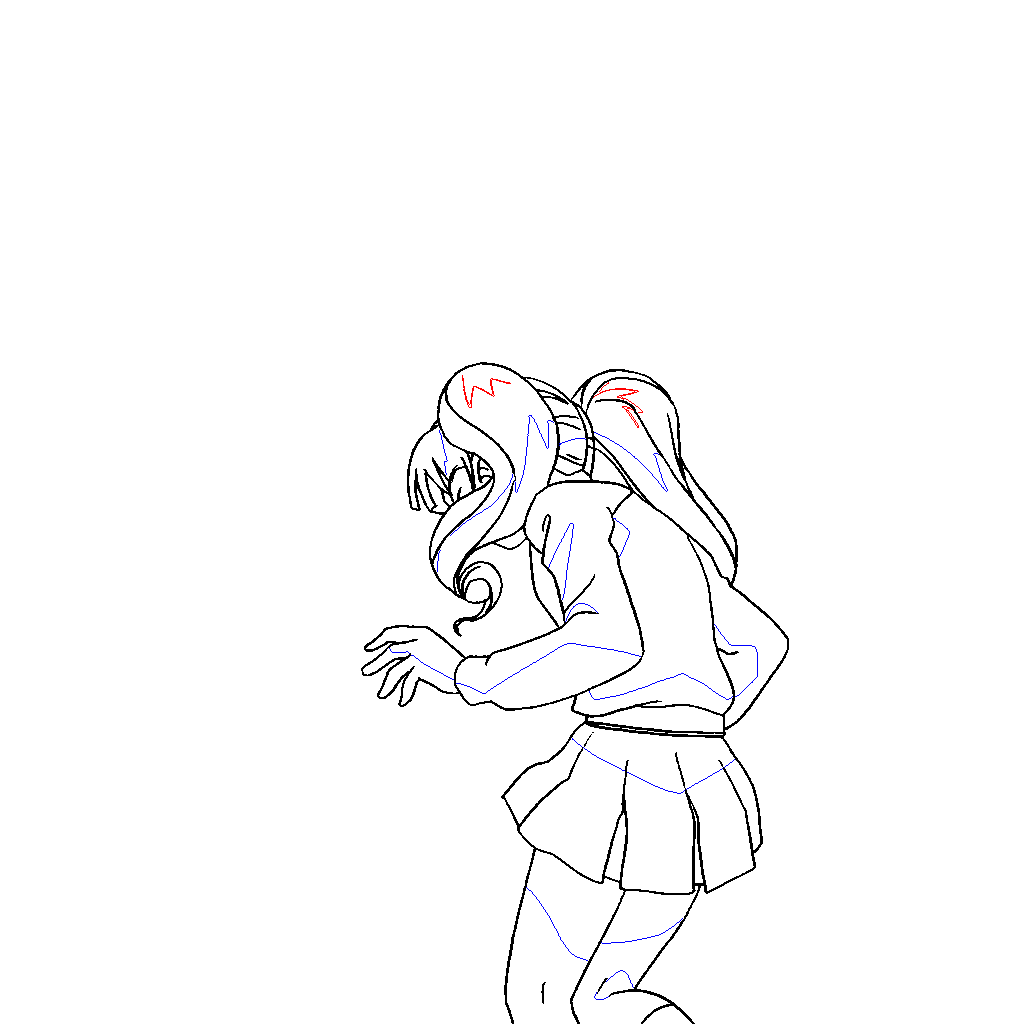} &
        \includegraphics[width=0.132\linewidth]{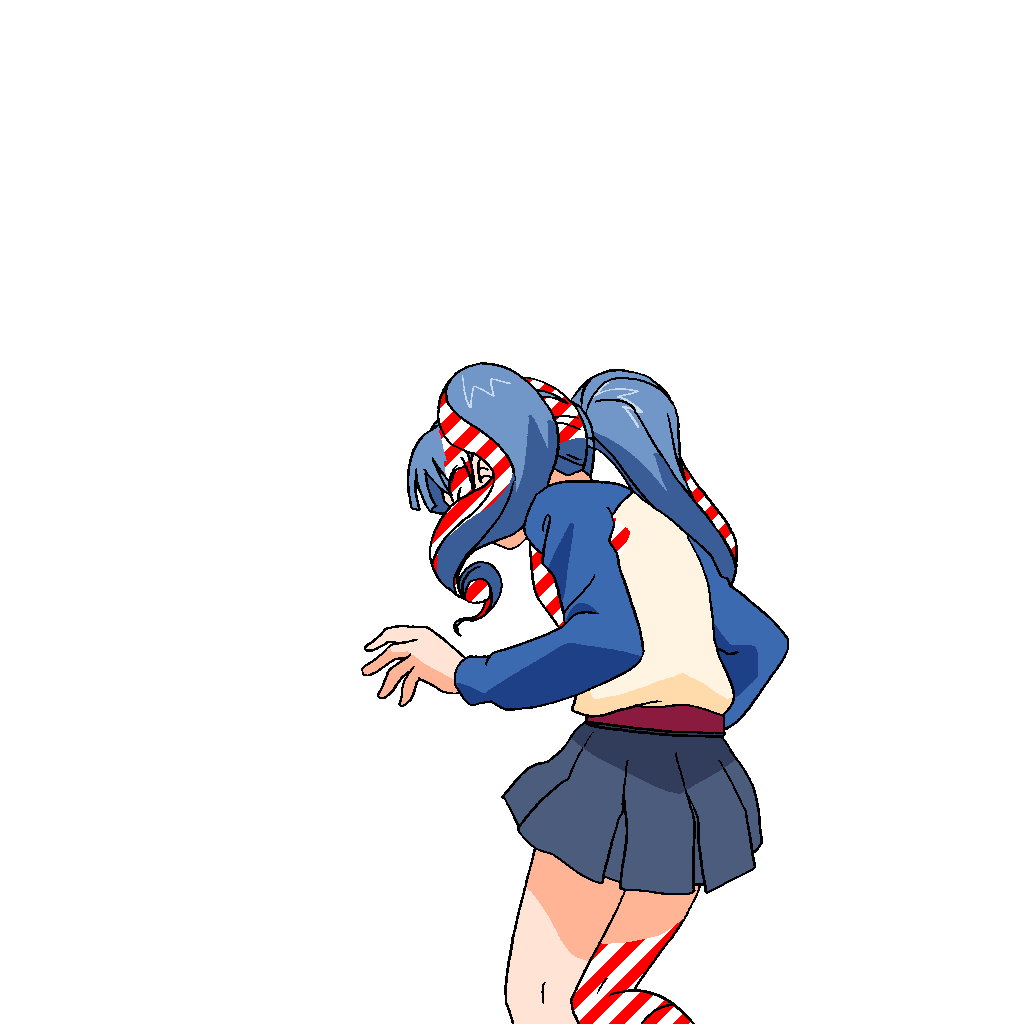} &
        \includegraphics[width=0.132\linewidth]{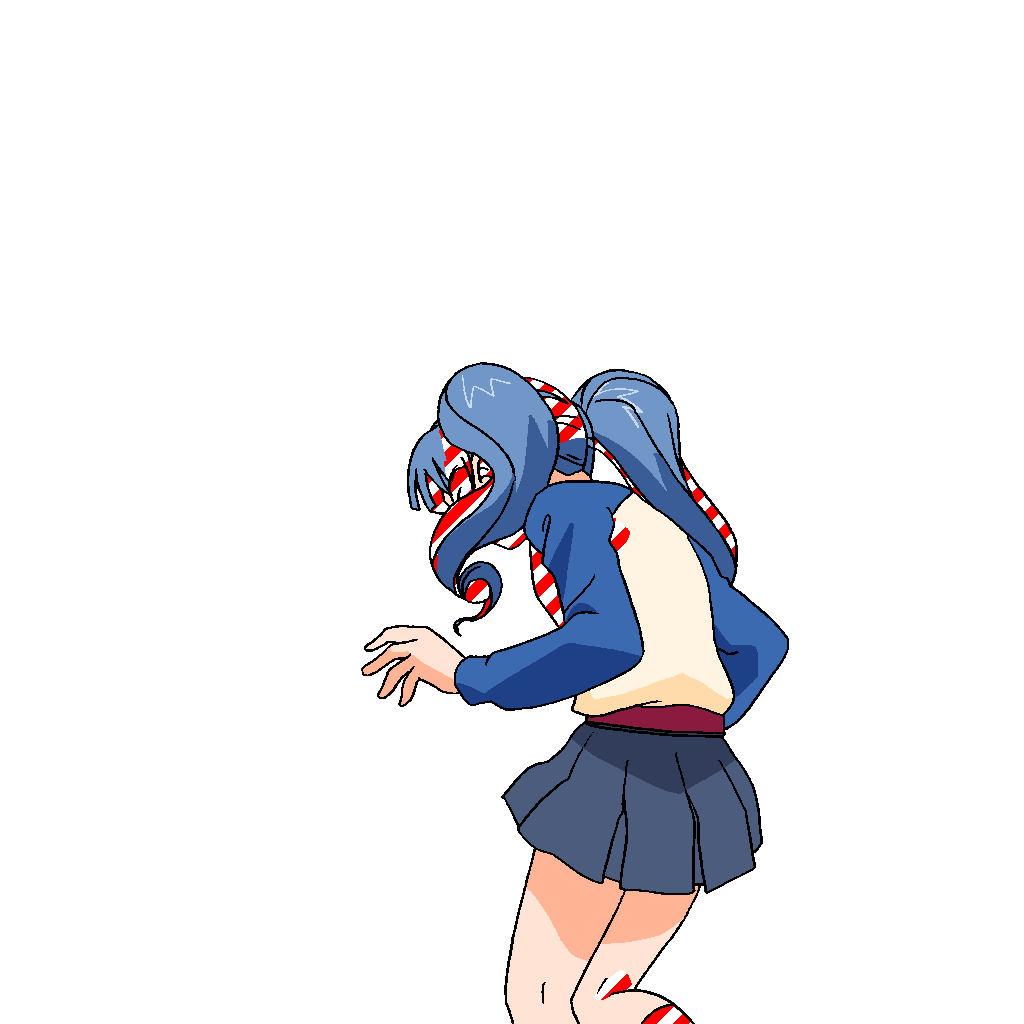} &
        \includegraphics[width=0.132\linewidth]{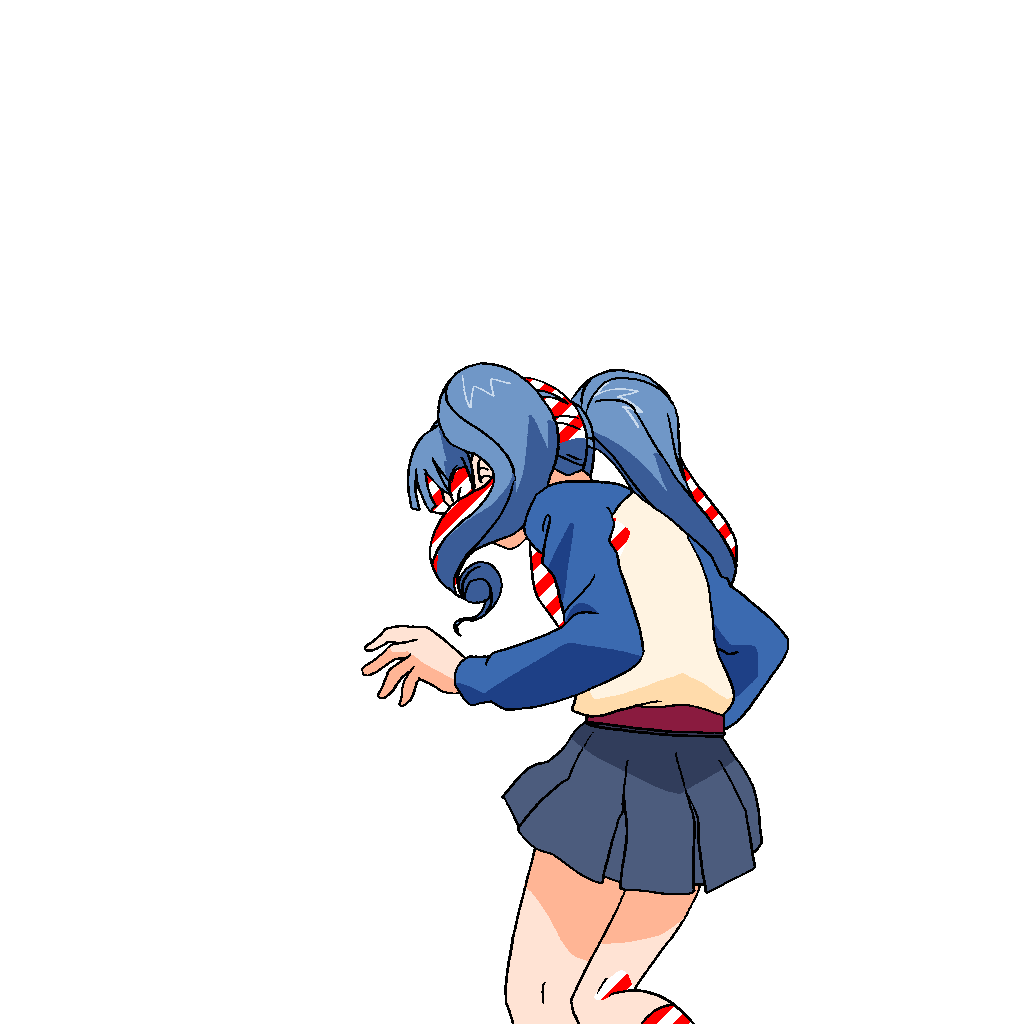} &
        \includegraphics[width=0.132\linewidth]{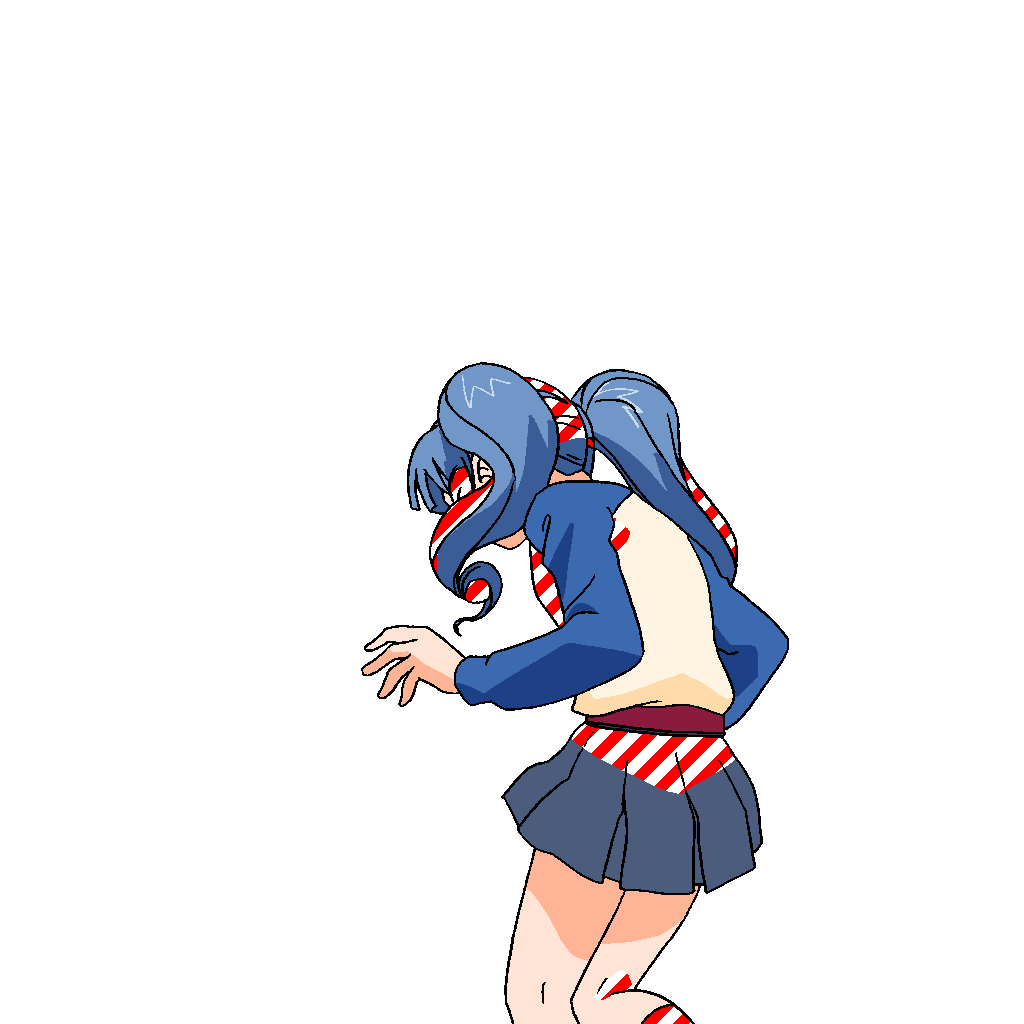} &
        \includegraphics[width=0.132\linewidth]{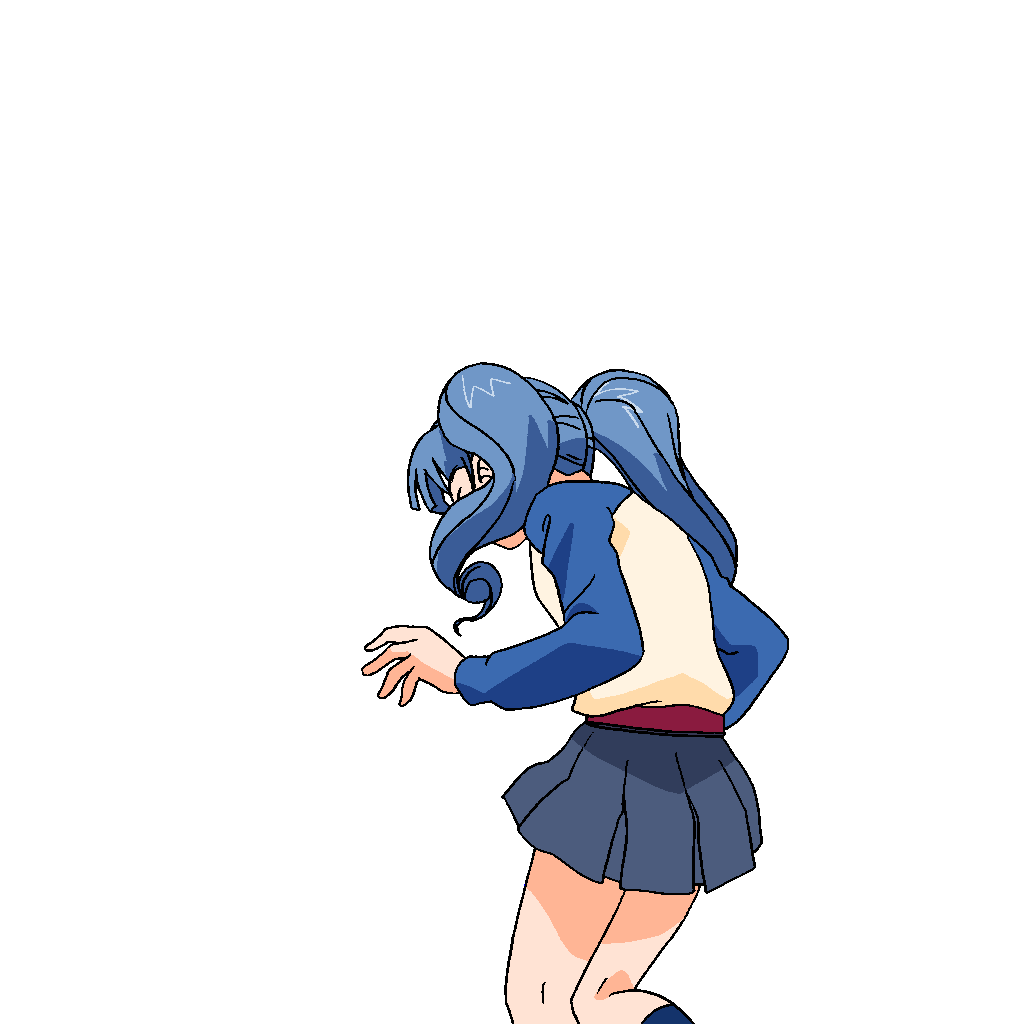} \\[0.25em]
        \midrule
        Ref. & Target & w/o ARE & \makecell{$B{=}15$\\$m{=}4$} & \makecell{\boldmath$B{=}31$\\\boldmath$m{=}4$} & \makecell{$B{=}63$\\$m{=}8$} & GT \\
        \midrule
        \includegraphics[width=0.132\linewidth]{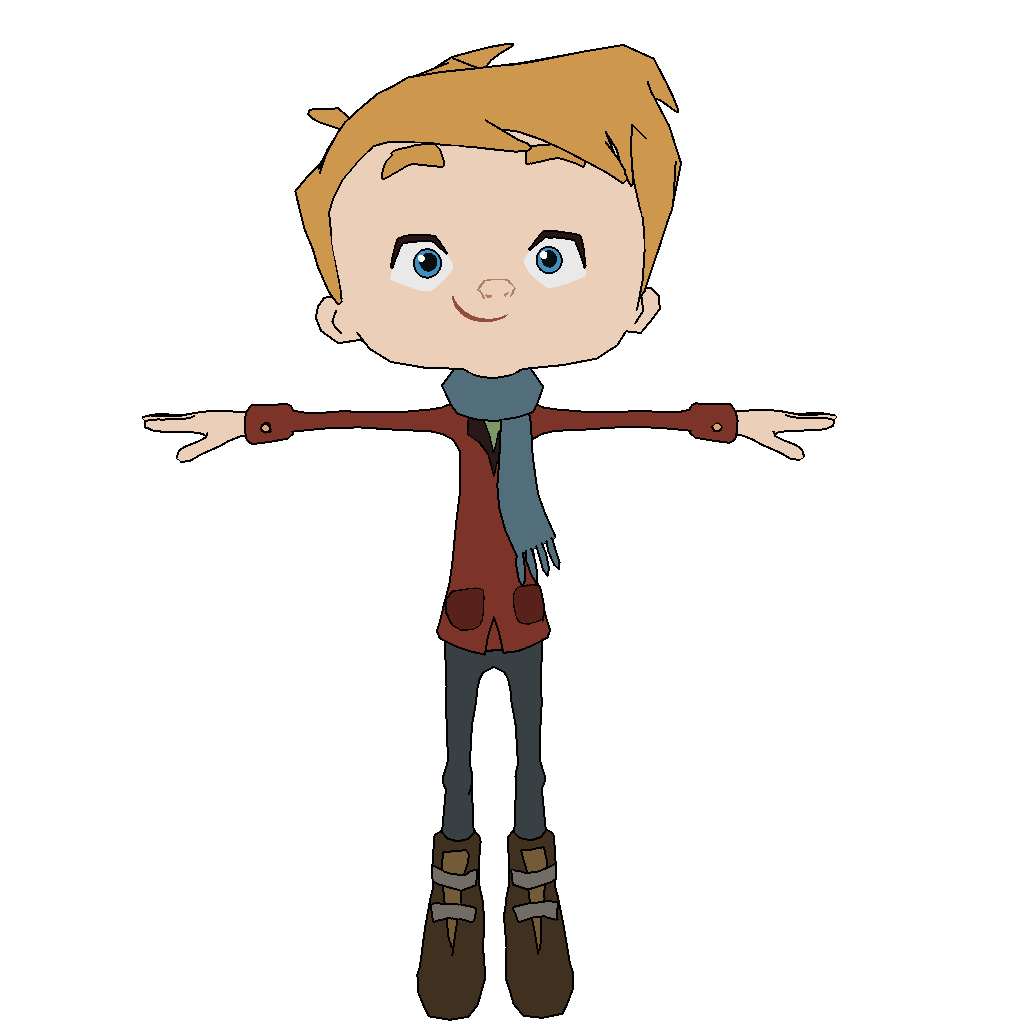} &
        \includegraphics[width=0.132\linewidth]{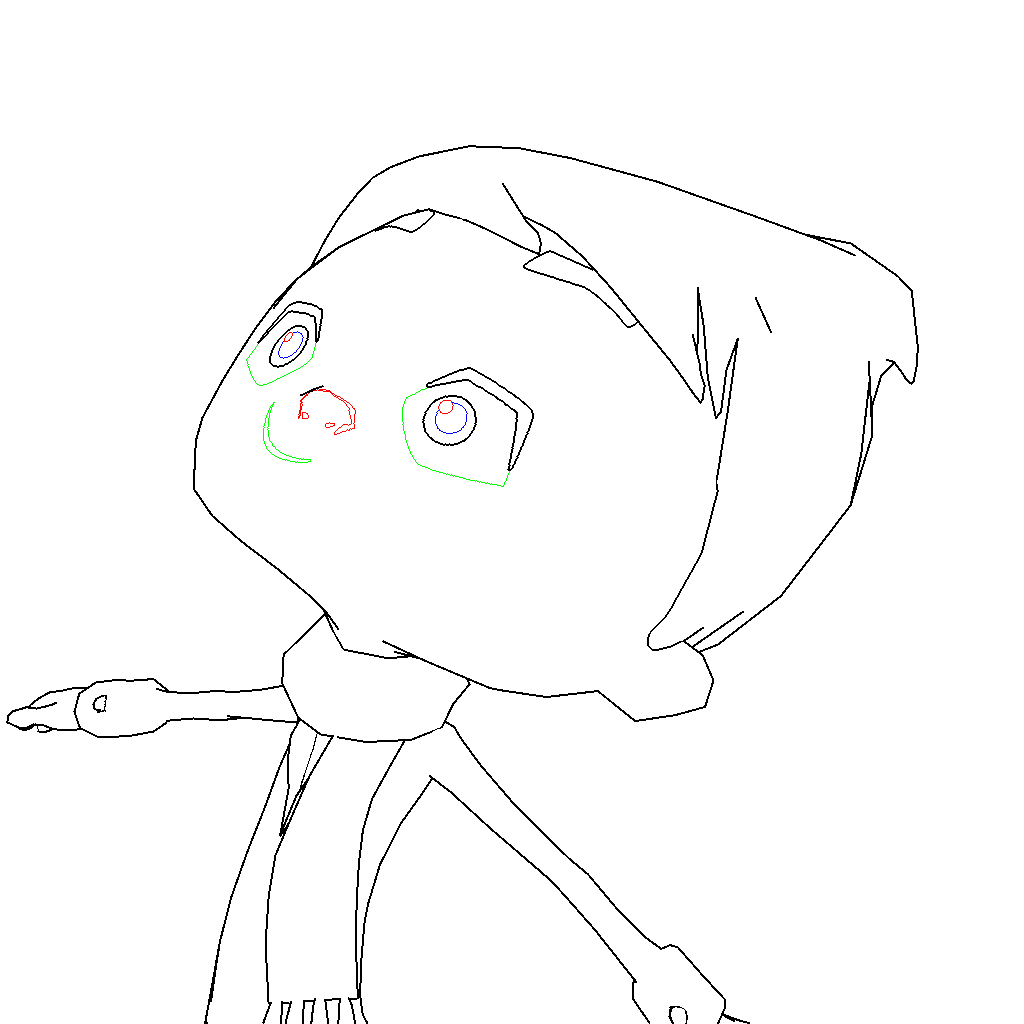} &
        \includegraphics[width=0.132\linewidth]{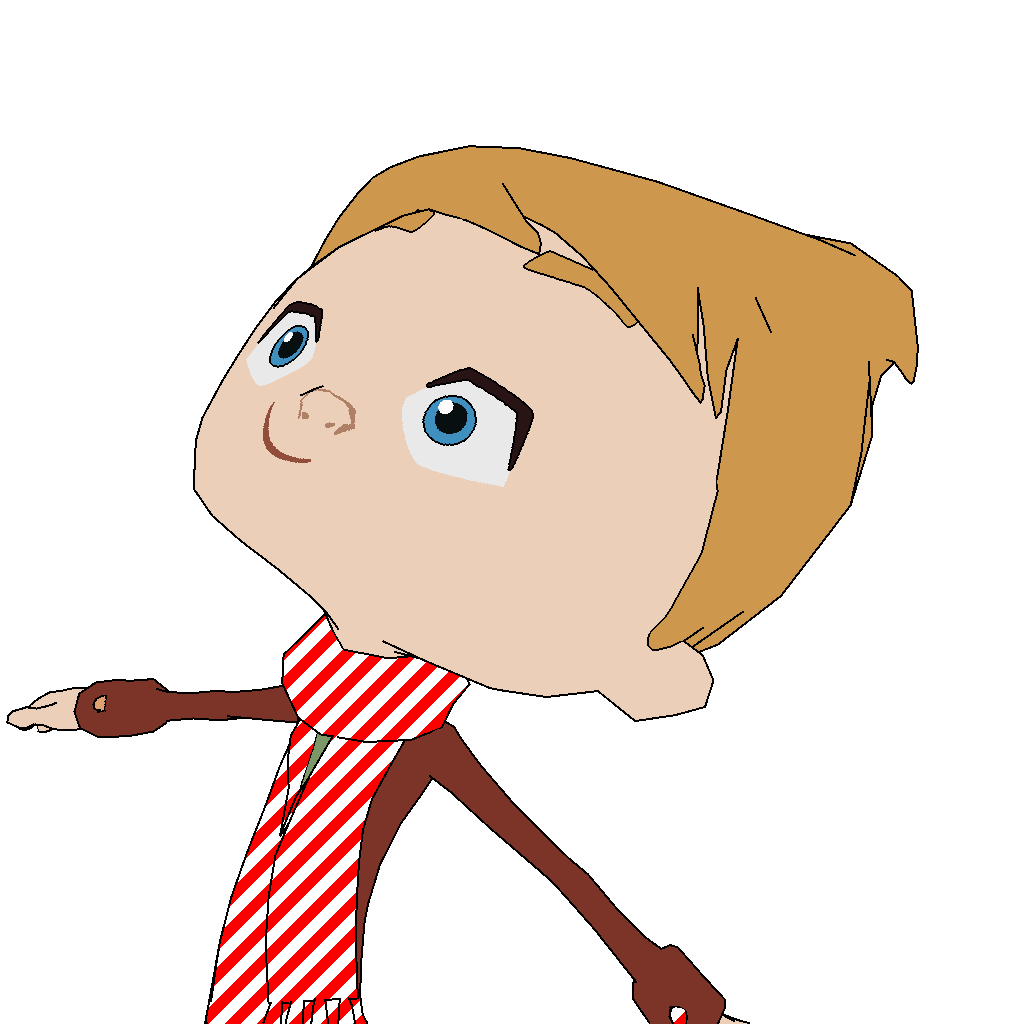} &
        \includegraphics[width=0.132\linewidth]{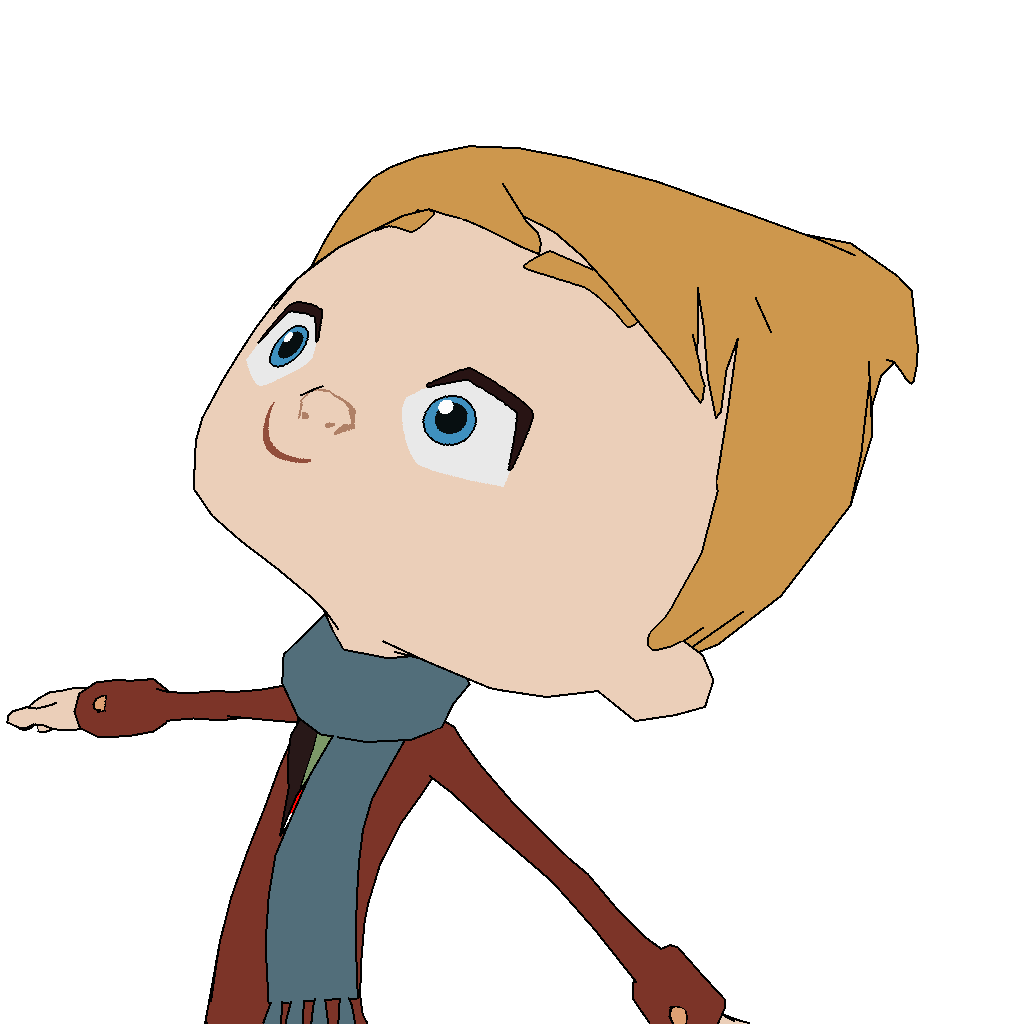} &
        \includegraphics[width=0.132\linewidth]{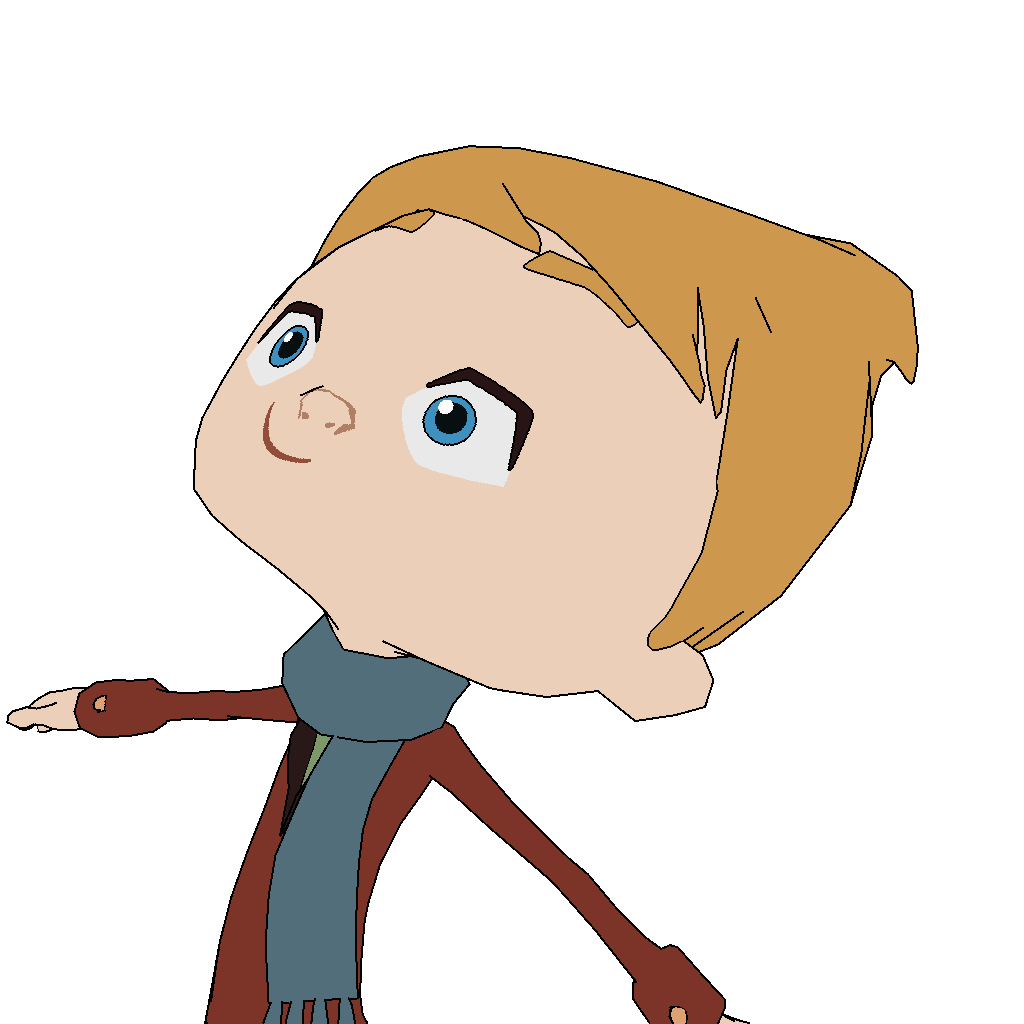} &
        \includegraphics[width=0.132\linewidth]{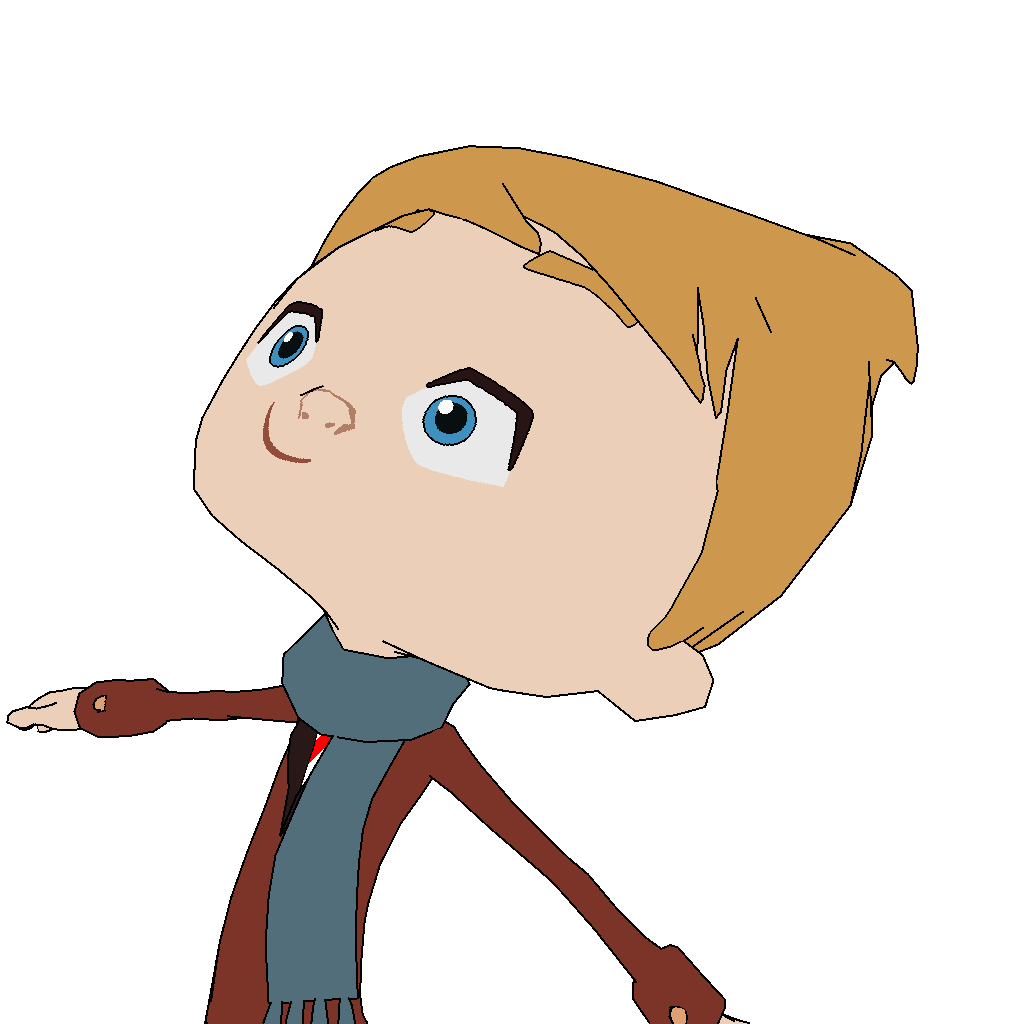} &
        \includegraphics[width=0.132\linewidth]{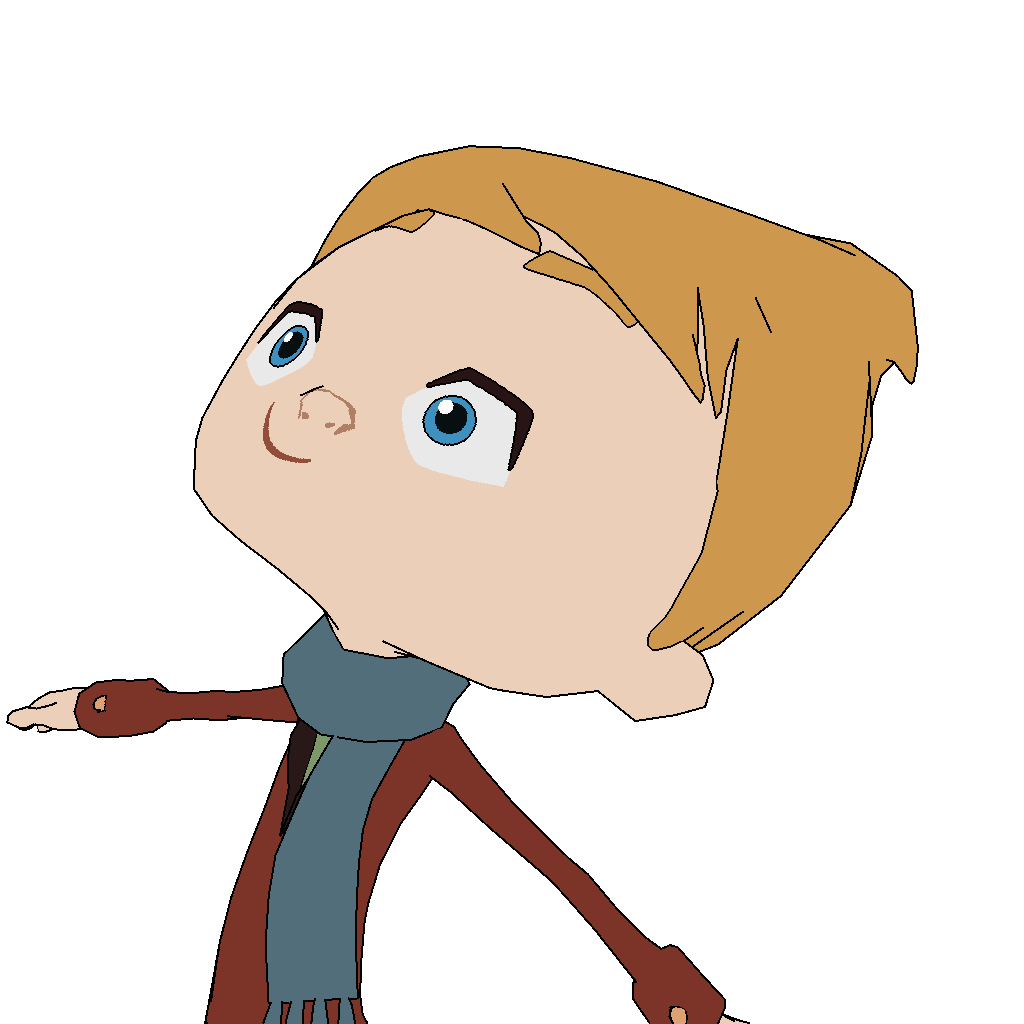} \\
        \includegraphics[width=0.132\linewidth]{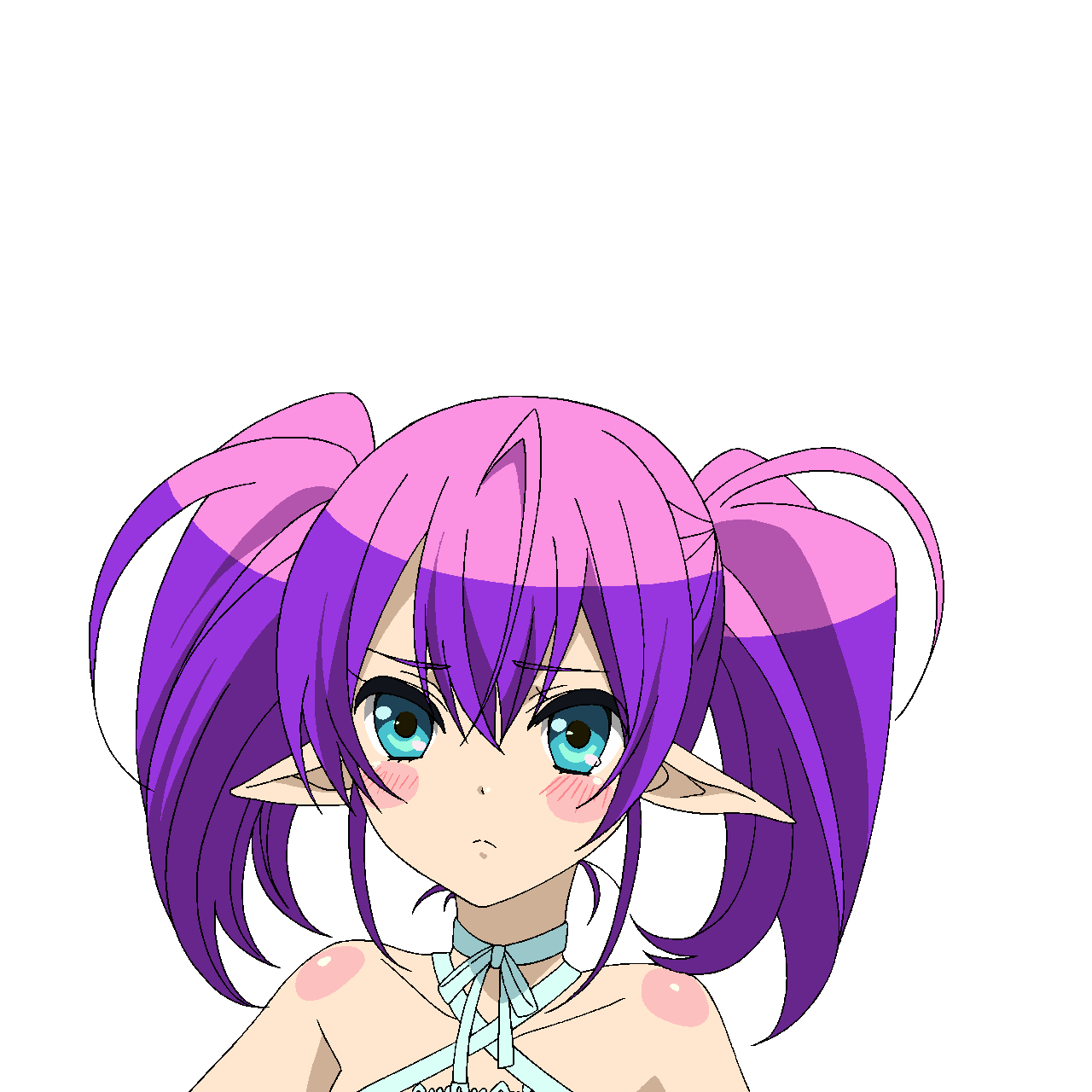} &
        \includegraphics[width=0.132\linewidth]{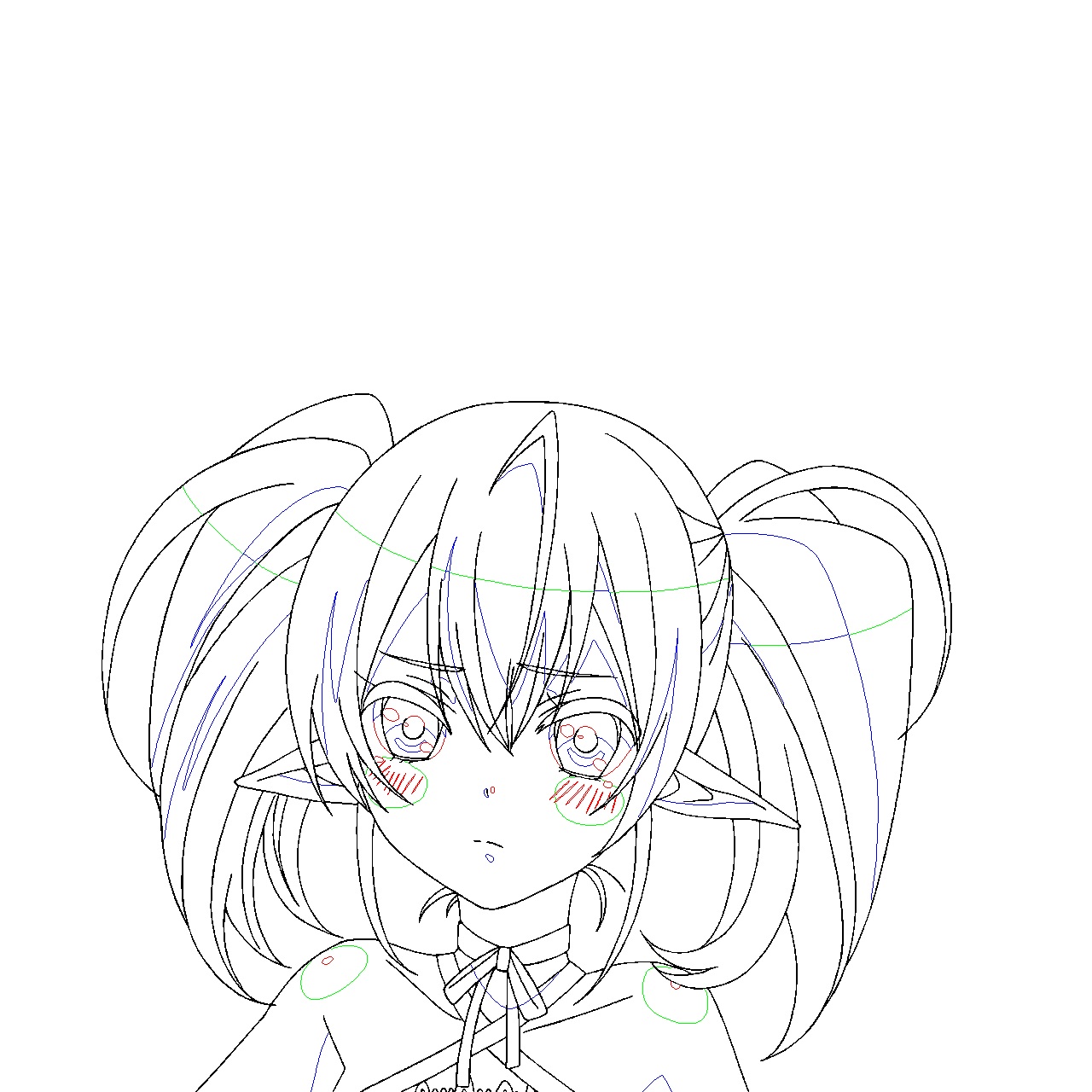} &
        \includegraphics[width=0.132\linewidth]{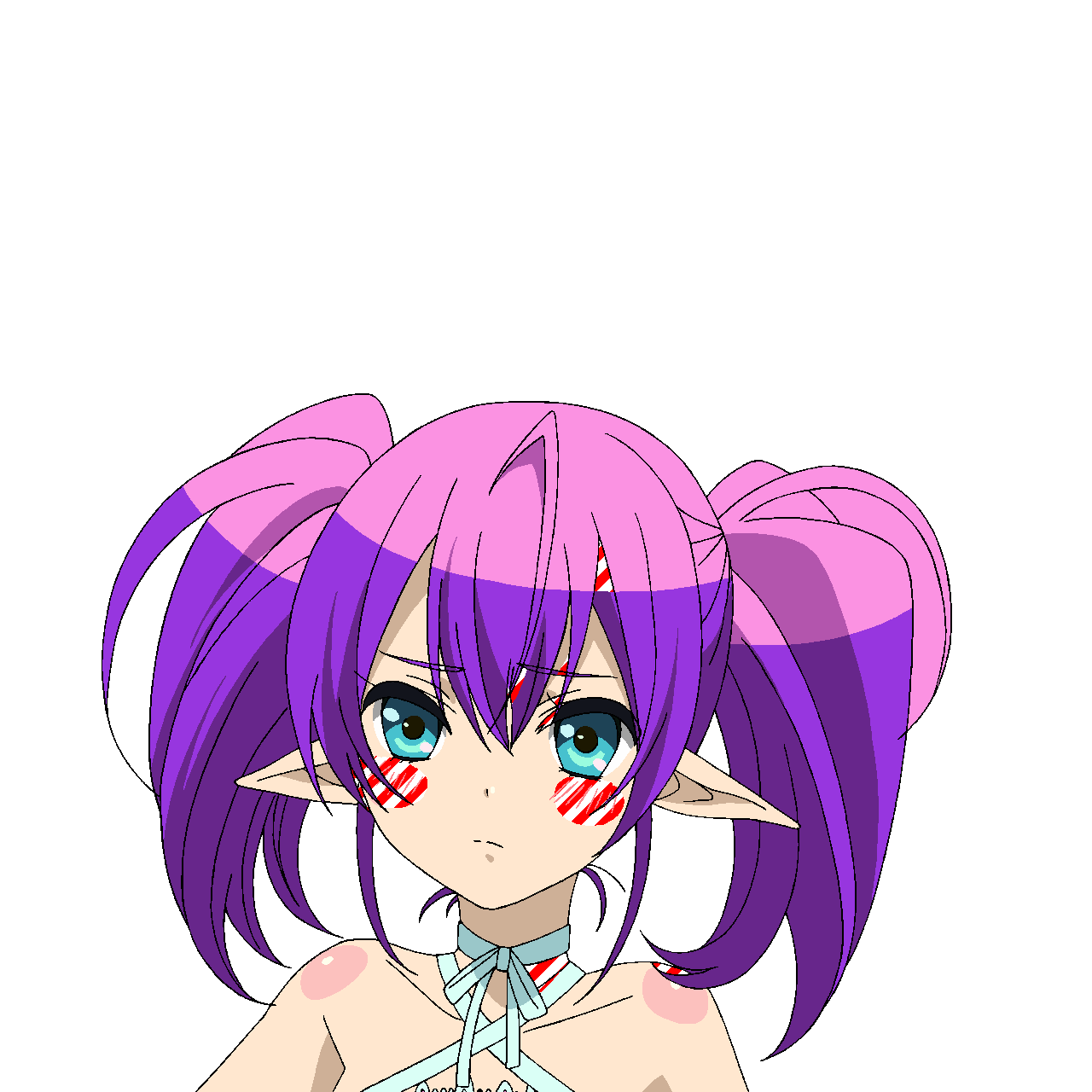} &
        \includegraphics[width=0.132\linewidth]{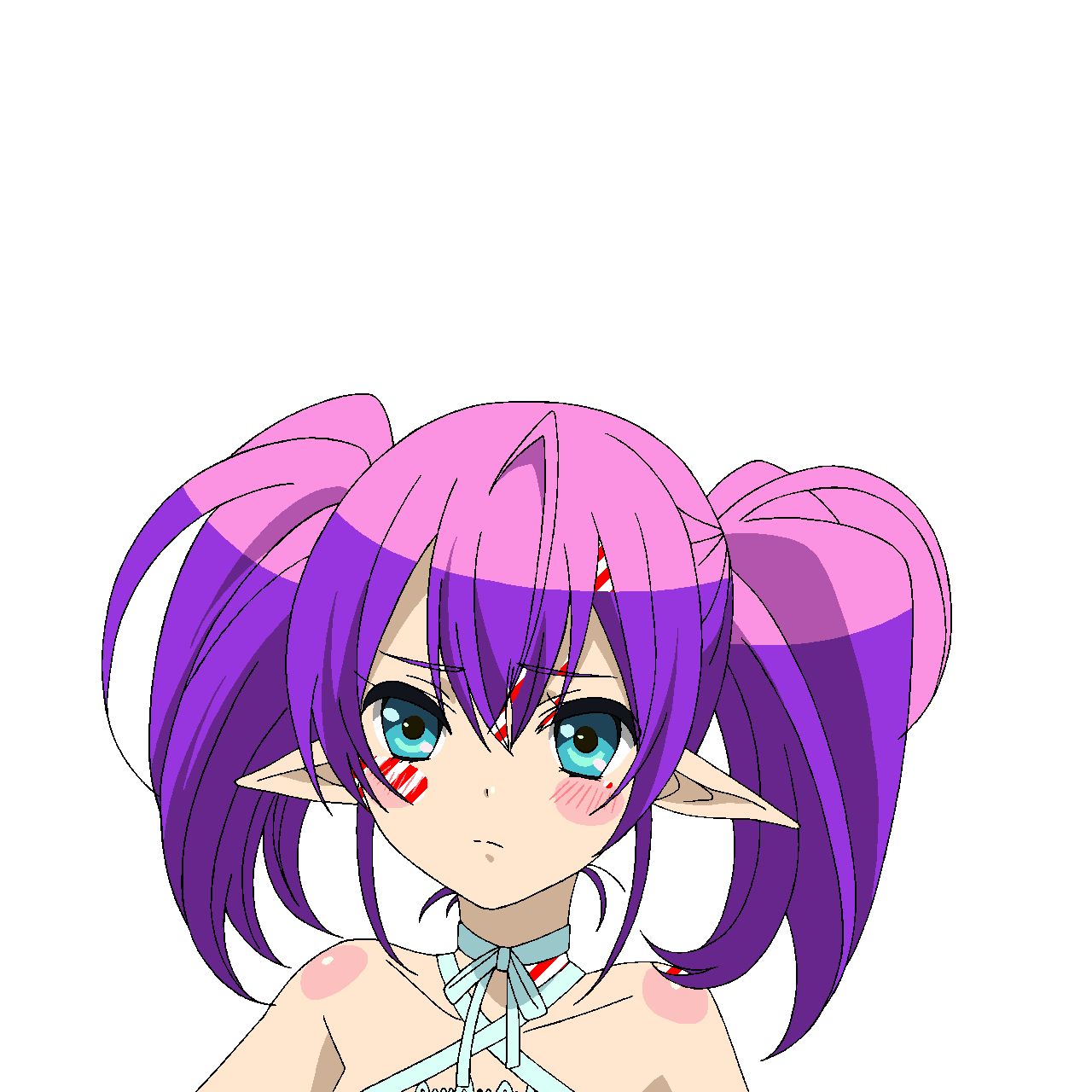} &
        \includegraphics[width=0.132\linewidth]{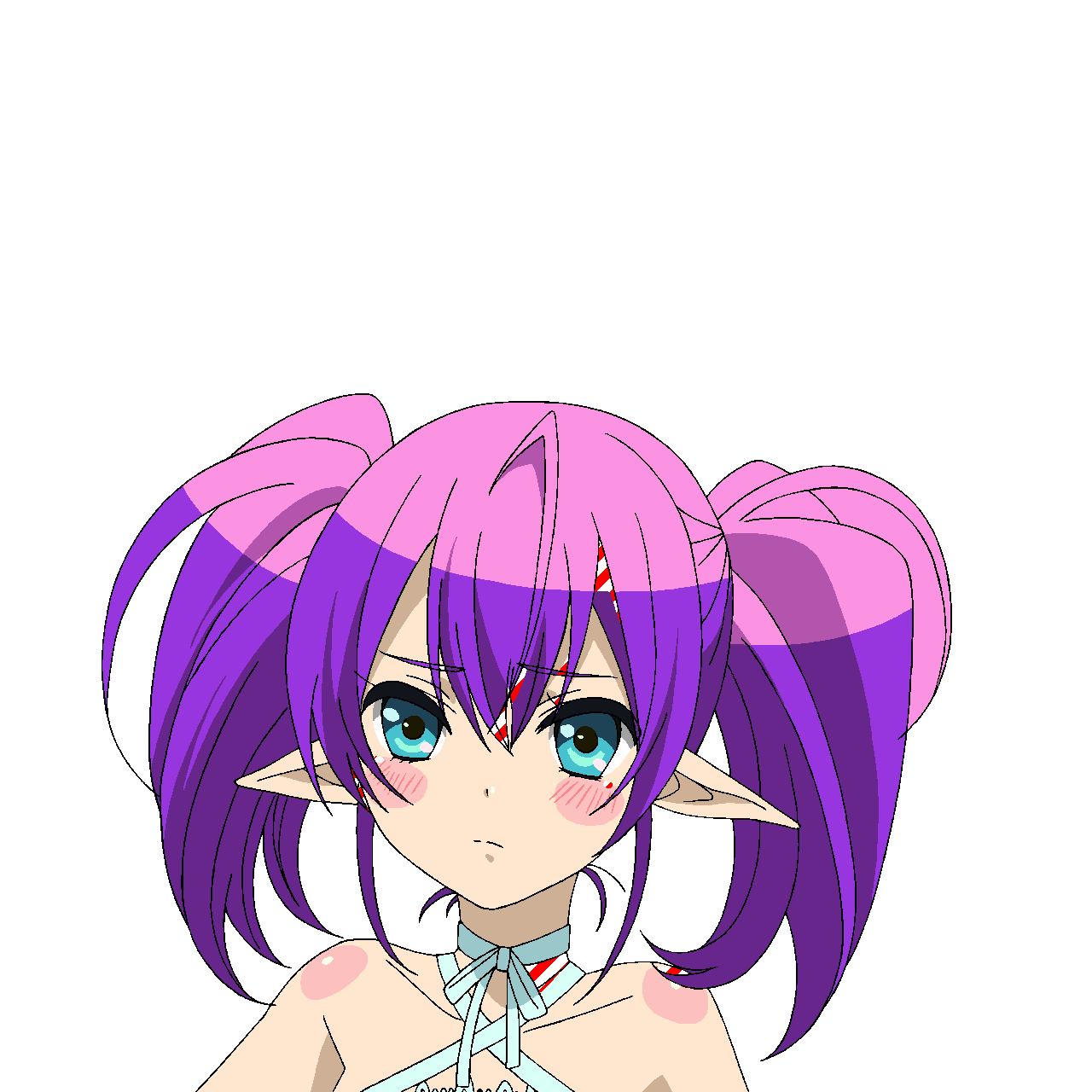} &
        \includegraphics[width=0.132\linewidth]{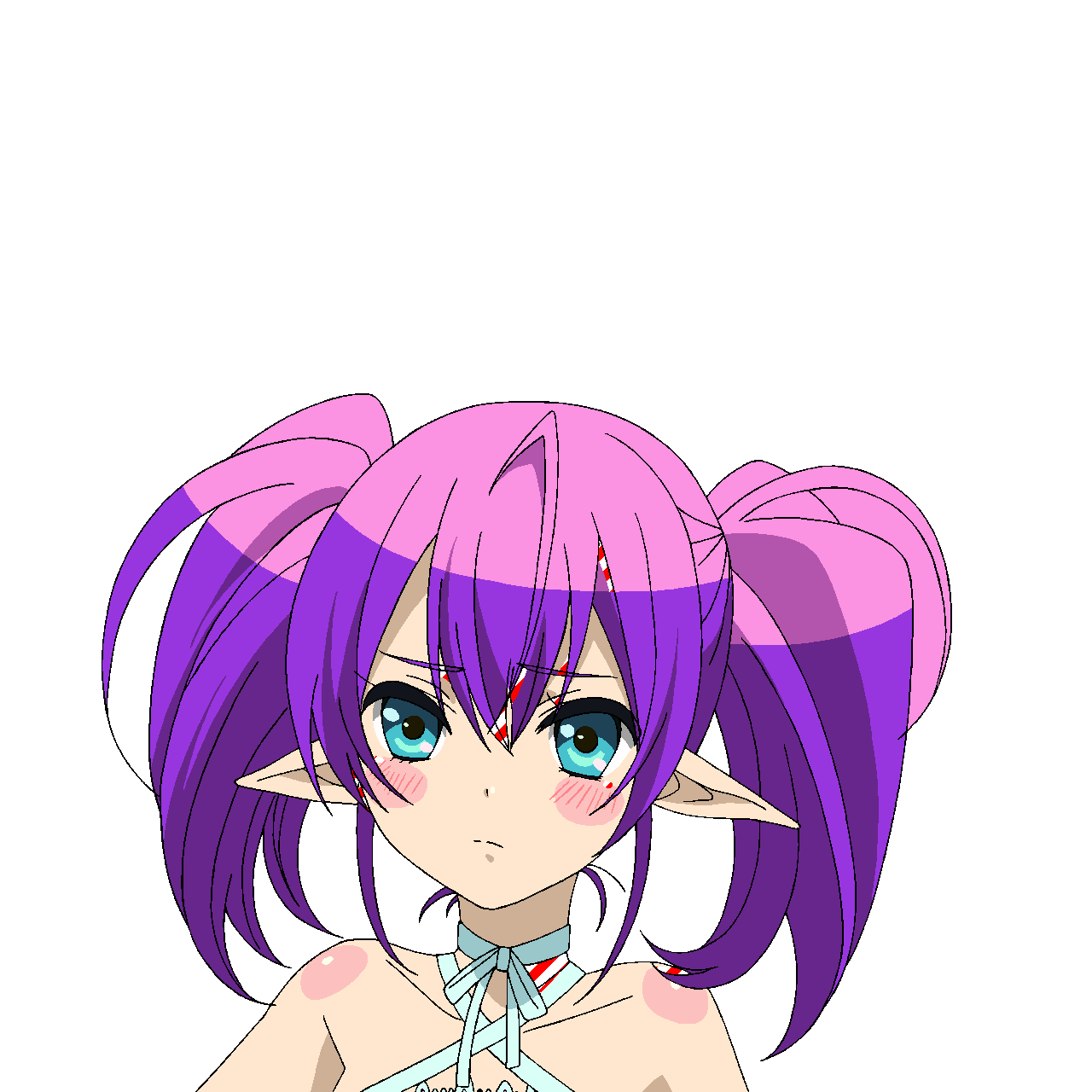} &
        \includegraphics[width=0.132\linewidth]{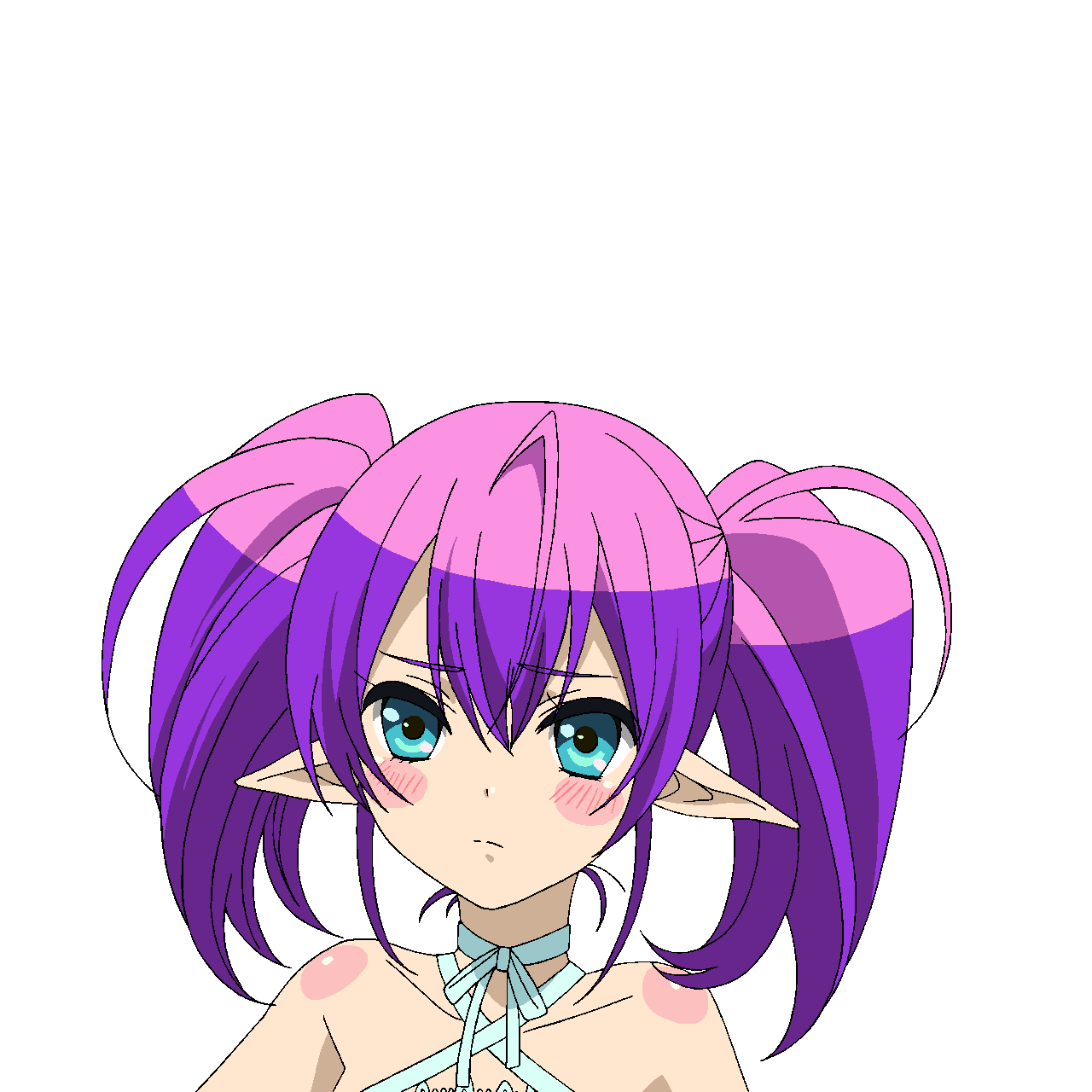}
    \end{tabular}}
    \caption{\textbf{Qualitative hyperparameter comparisons.} Rows 1--2 vary PA settings, and rows 3--4 vary ARE selection budget. \zebraswatch{} marks wrong-colour foreground regions under each setting.}
    \label{fig:supp_hparam_qual}
\end{figure}

The numerical analysis above therefore provides explanations to the provided qualitative examples for extreme and default hyperparameter settings in \cref{fig:supp_hparam_qual}. The results show a trend aligning with quantitative hyperparameter analysis. Overly hard PA under-aggregates useful evidence, while broader PA moves closer to indiscriminate colour mixing; very small selection budgets reduce reference support, whereas larger budgets saturate around the fixed default. These qualitative behaviours are consistent with the quantitative hyperparameter surfaces in \cref{fig:ablation_surfaces}.

\section{Additional Temporal Stability Analysis}
\label{sec:supp_longclip_stability}

To further validate robustness throughout the video, we compare \textit{Base} and \textsc{PeCA} using aggregate curves over relative clip position under the one-shot design-sheet key-frame colourisation setting for PBC-3D \cite{dai2024learning} dataset. We report five standard per-frame quality metrics, together with an additional temporal stability metric defined below. 
\begin{figure*}[h!]
    \centering
    \begin{subfigure}[t]{0.31\textwidth}
        \centering
        \includegraphics[width=\linewidth]{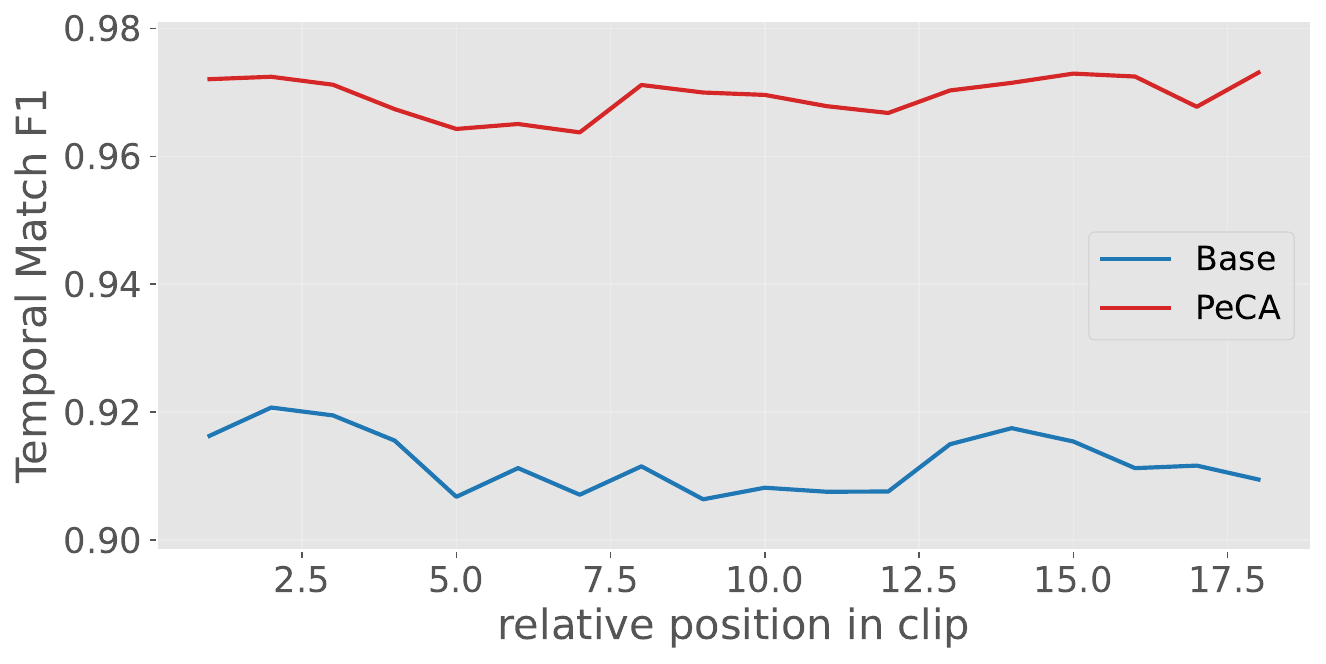}
        \caption{TM-F1 ($\uparrow$)}
        \label{fig:supp_long_tmf1}
    \end{subfigure}
    \hfill
    \begin{subfigure}[t]{0.31\textwidth}
        \centering
        \includegraphics[width=\linewidth]{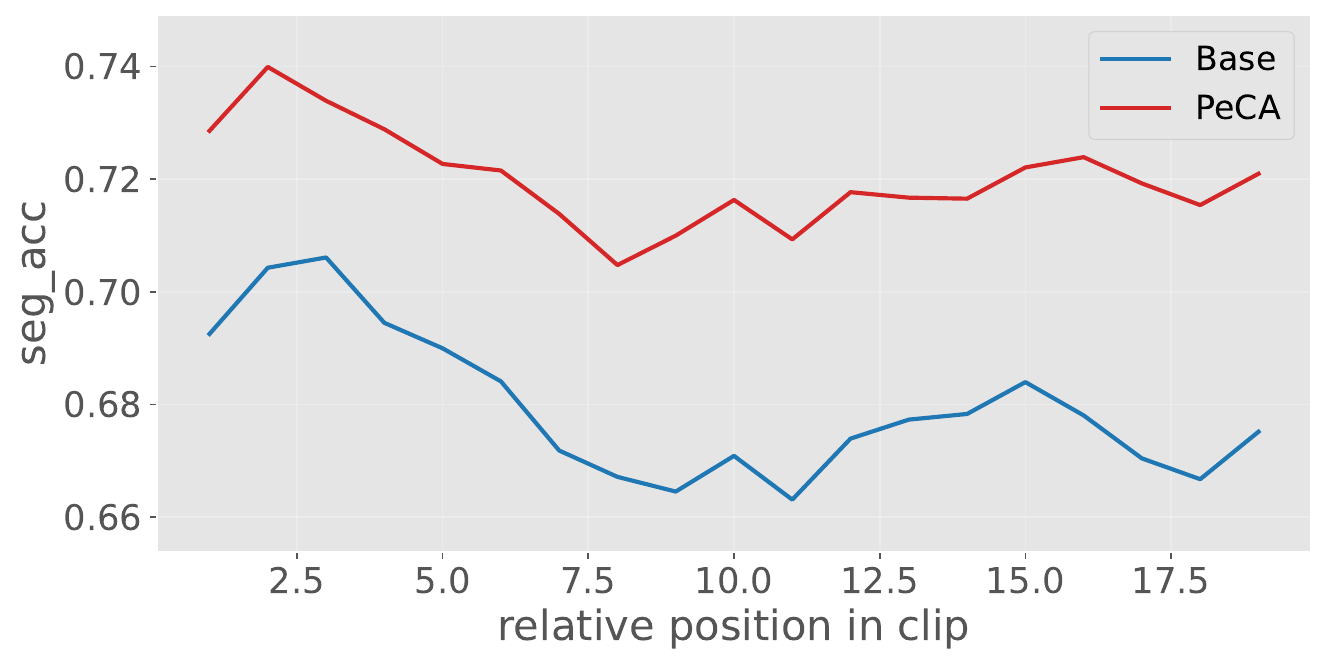}
        \caption{Acc}
        \label{fig:supp_long_segacc}
    \end{subfigure}
    \hfill
    \begin{subfigure}[t]{0.31\textwidth}
        \centering
        \includegraphics[width=\linewidth]{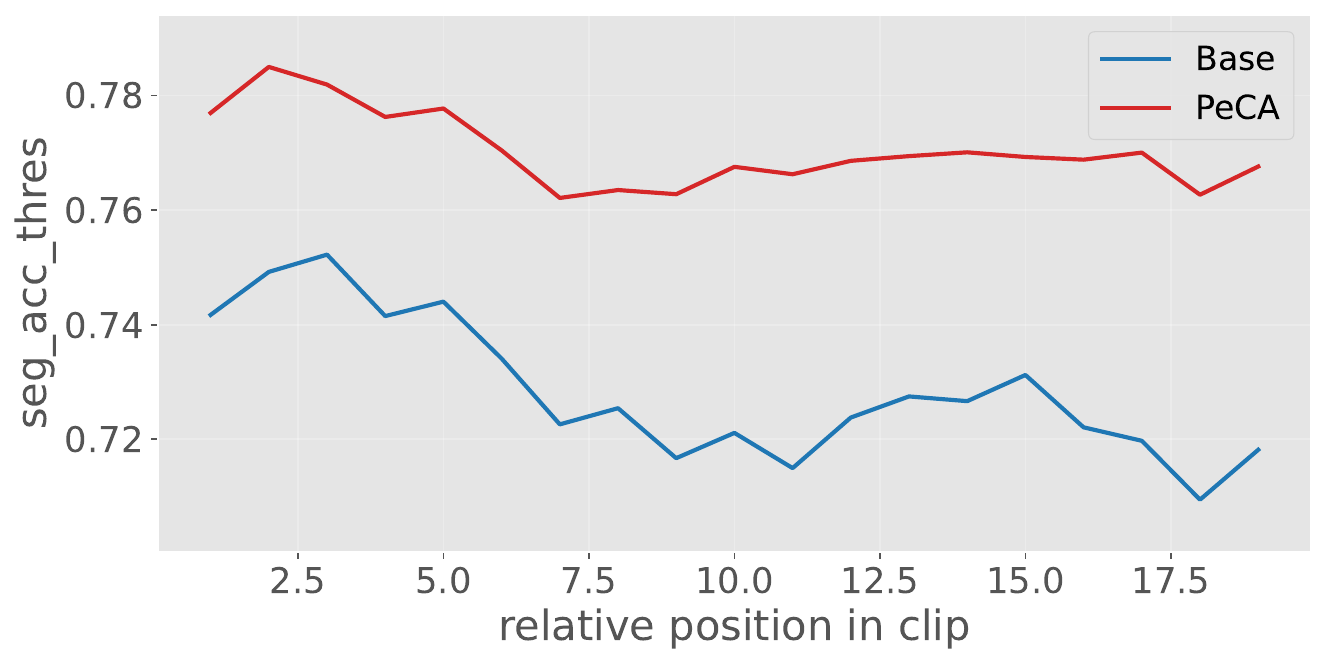}
        \caption{Acc-Thresh}
        \label{fig:supp_long_segacc_thresh}
    \end{subfigure}

    \begin{subfigure}[t]{0.31\textwidth}
        \centering
        \includegraphics[width=\linewidth]{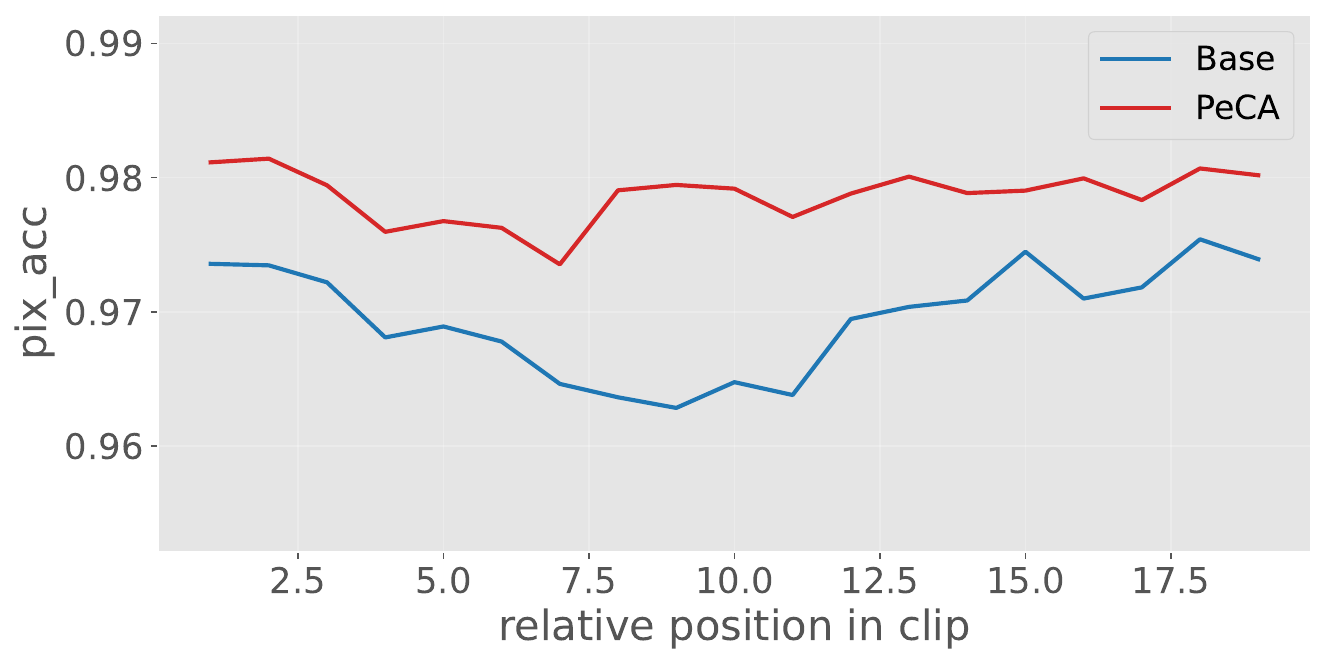}
        \caption{Pix-Acc}
        \label{fig:supp_long_pixacc}
    \end{subfigure}
    \hfill
    \begin{subfigure}[t]{0.31\textwidth}
        \centering
        \includegraphics[width=\linewidth]{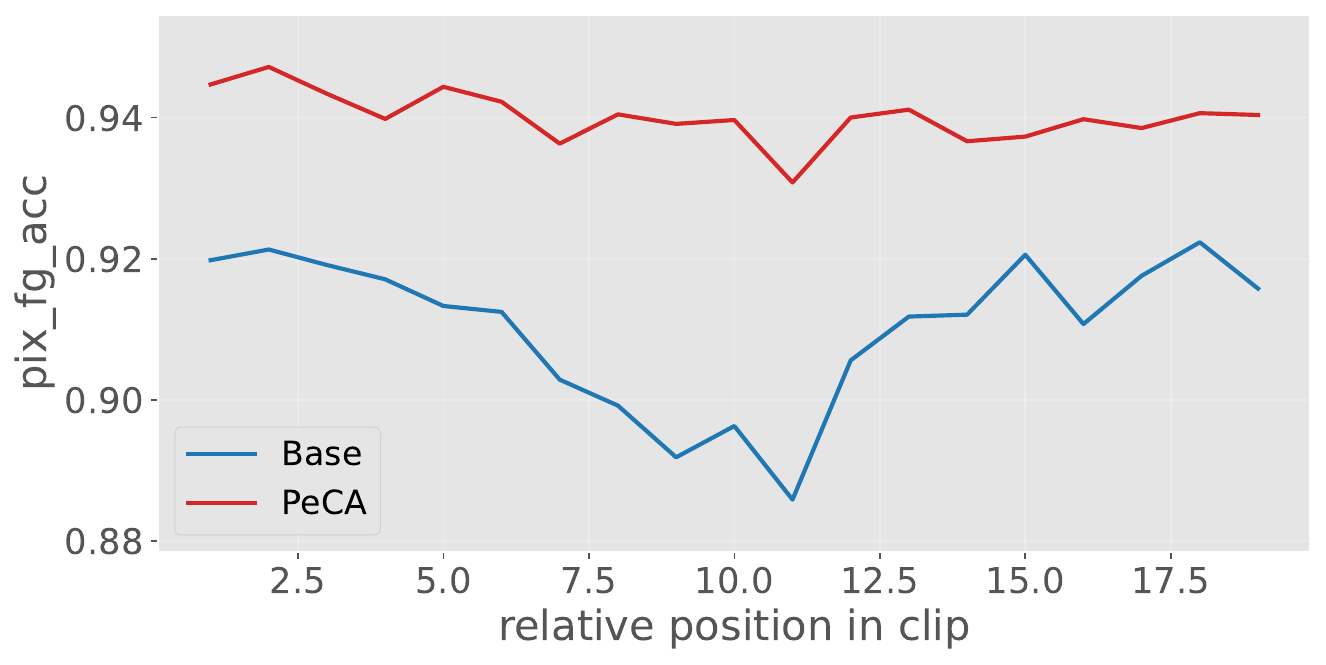}
        \caption{Pix-F-Acc}
        \label{fig:supp_long_pixfg}
    \end{subfigure}
    \hfill
    \begin{subfigure}[t]{0.31\textwidth}
        \centering
        \includegraphics[width=\linewidth]{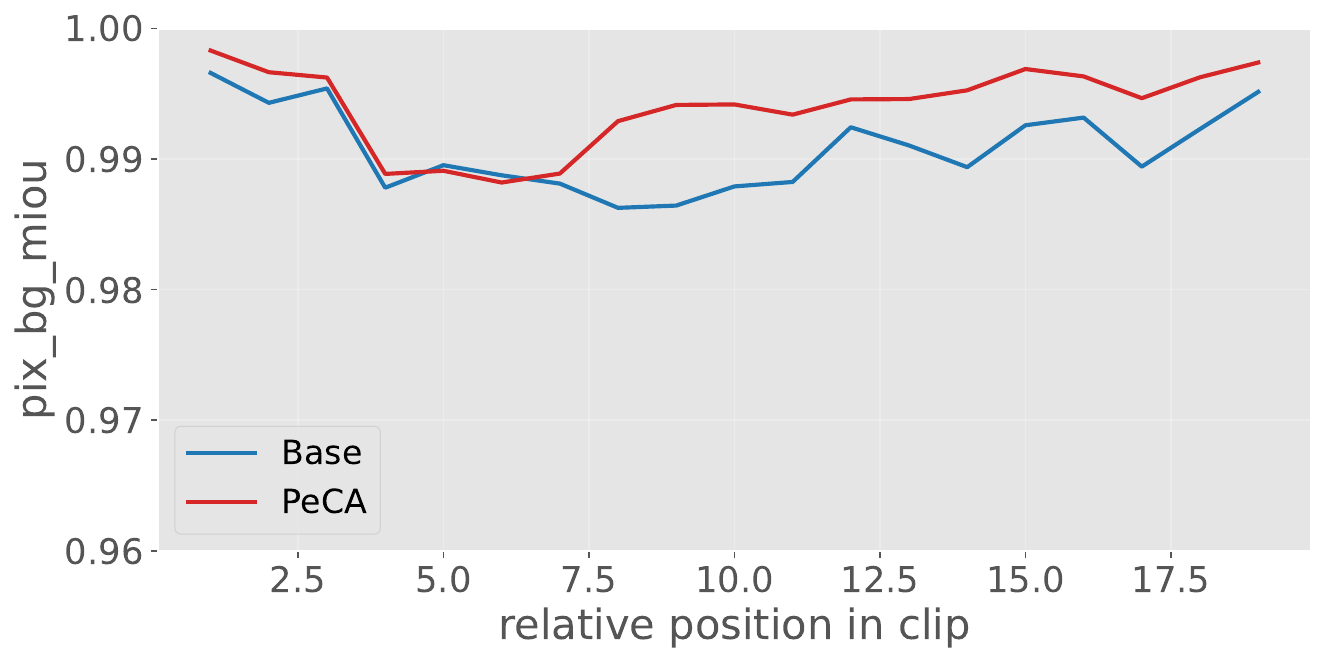}
        \caption{Pix-B-MIoU}
        \label{fig:supp_long_pixbg}
    \end{subfigure}
    \caption{\textbf{Metric curves over relative video clip position (\textit{Base} vs.\ \textsc{PeCA}).} \textsc{PeCA} improves temporal consistency (TM-F1) and consistently improves segment/pixel quality over time, mitigating long-horizon error accumulation.}
    \label{fig:supp_longclip_curves}
\end{figure*}
\paragraph{\textbf{Temporal stability metric definition.}}
Since exact ground-truth segment trajectories between adjacent frames are unavailable, we evaluate temporal stability on surrogate temporal correspondences, following the same strategy used in CT.
Concretely, for each target segment \(i\) at frame \(t\), we first define its nearest temporal link to frame \(t{-}1\) by feature matching using DACoN~1.1 features~\cite{nagata2025dacon}:
\begin{equation}
j^\ast_t(i)=\arg\max_{j}\ \left\langle \bar{\mathbf f}_{t,i}, \bar{\mathbf f}_{t-1,j}\right\rangle .
\label{eq:supp_tm_link}
\end{equation}

Not all nearest-neighbour links are reliable, so we further retain only cycle-consistent links as {stable}:
\begin{equation}
\mathrm{stable}(t,i)=
\mathds{1}\!\left[
i=\arg\max_{i'}\ \left\langle \bar{\mathbf f}_{t-1,j^\ast_t(i)}, \bar{\mathbf f}_{t,i'}\right\rangle
\right].
\label{eq:supp_tm_stable}
\end{equation}

This filtering removes spurious matches and restricts evaluation to reliably trackable regions. On each stable link, we then evaluate whether the {ground-truth} label is temporally consistent, and whether the {prediction} preserves the same consistency pattern:
\begin{equation}
g_{t,i}=\mathds{1}\!\left[y_{t,i}=y_{t-1,j^\ast_t(i)}\right],\qquad
\hat g_{t,i}=\mathds{1}\!\left[\hat y_{t,i}=\hat y_{t-1,j^\ast_t(i)}\right].
\label{eq:supp_tm_consistency}
\end{equation}

Intuitively, \(g_{t,i}=1\) means the true colour should stay the same along the link, while \(\hat g_{t,i}=1\) means the method predicts no colour change. We then compute per-step precision and recall over stable links:
\begin{equation}
P_t=\frac{\sum_i \mathds{1}[\mathrm{stable}(t,i)]\,\hat g_{t,i}g_{t,i}}
{\sum_i \mathds{1}[\mathrm{stable}(t,i)]\,\hat g_{t,i}},
\qquad
R_t=\frac{\sum_i \mathds{1}[\mathrm{stable}(t,i)]\,\hat g_{t,i}g_{t,i}}
{\sum_i \mathds{1}[\mathrm{stable}(t,i)]\,g_{t,i}} .
\label{eq:supp_tm_pr}
\end{equation}

Finally, the temporal stability metric is defined as the per-step F1 score:
\begin{equation}
\mathrm{TM\text{-}F1}_t=\frac{2P_tR_t}{P_t+R_t},\qquad t=2,\dots,T.
\label{eq:supp_tm_f1}
\end{equation}

\begin{figure}[h!]
    \centering
    \includegraphics[width=0.9\linewidth]{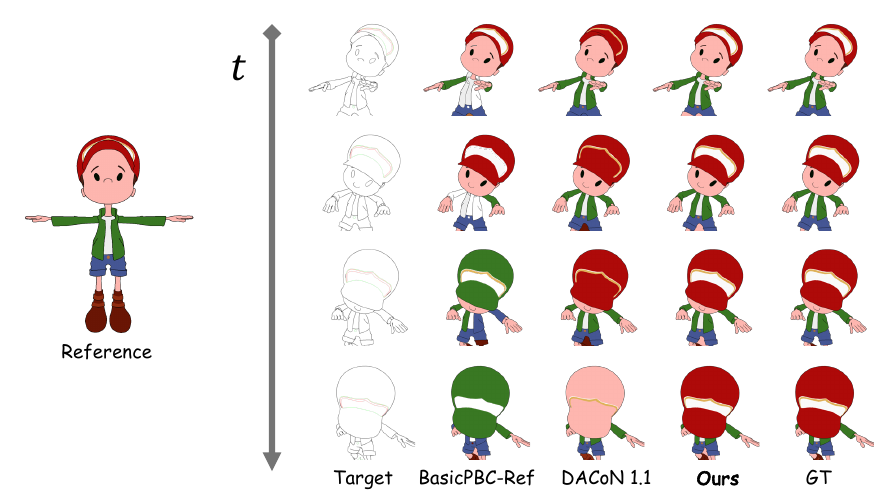}
        \caption{\textbf{Visualisation of temporal stability.}
        Compared with previous methods (BasicPBC-Ref~\cite{dai2024paint}, \textit{Base} DACoN~1.1~\cite{nagata2025dacon}),
        our method shows significantly fewer ``colour flickers'' over time.
        More qualitative results are provided in the \projectvideolink.}
    \label{fig:temporal_stability}
\end{figure}

This completes the calculation of the surrogate temporal stability metric $\mathrm{TM\text{-}F1}$. Higher TM-F1 indicates better temporal consistency, \ie fewer ``colour flickers'' in the output video. As shown in \cref{fig:supp_longclip_curves}, \textsc{PeCA} remains consistently above the base inference across nearly the entire clip for all metrics. In particular, TM-F1 improves from around \(91\%\) (\textit{Base}) to around \(97\%\) (\textsc{PeCA}), indicating substantially better temporal stability throughout the sequence. Overall, these curves suggest that \textsc{PeCA} not only improves average colourisation quality, but also mitigates error accumulation over time. Additional qualitative visualisations of this stability are available in \cref{fig:temporal_stability}.

\section{Extension to Natural Video Region Label Propagation}
\label{sec:vipseg_extension}

\begin{figure}[h!]
    \centering
    \includegraphics[width=1\linewidth]{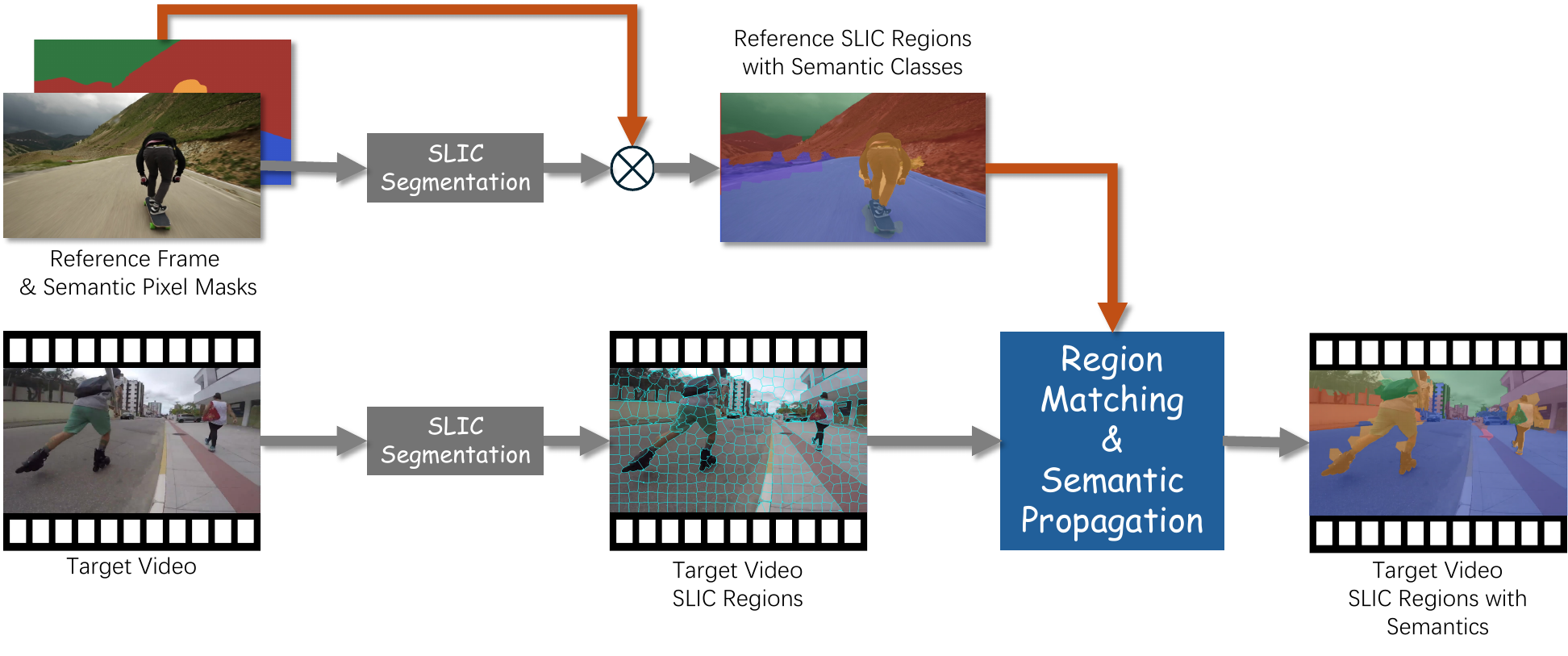}
    \caption{\textbf{Task formulation for reference-guided region label propagation for natural videos.}
    Given an external reference frame with panoptic semantic masks, we compute SLIC superpixels and assign each reference superpixel a semantic class according to the original panoptic annotation. For the target video, we compute SLIC superpixels for each frame. Region matching then propagates semantic labels from reference to target SLIC regions, producing per-region semantic predictions for evaluation.}
    \label{fig:natural_pipeline}
\end{figure}

\begin{figure}[h!]
    \centering
    \includegraphics[width=1\linewidth]{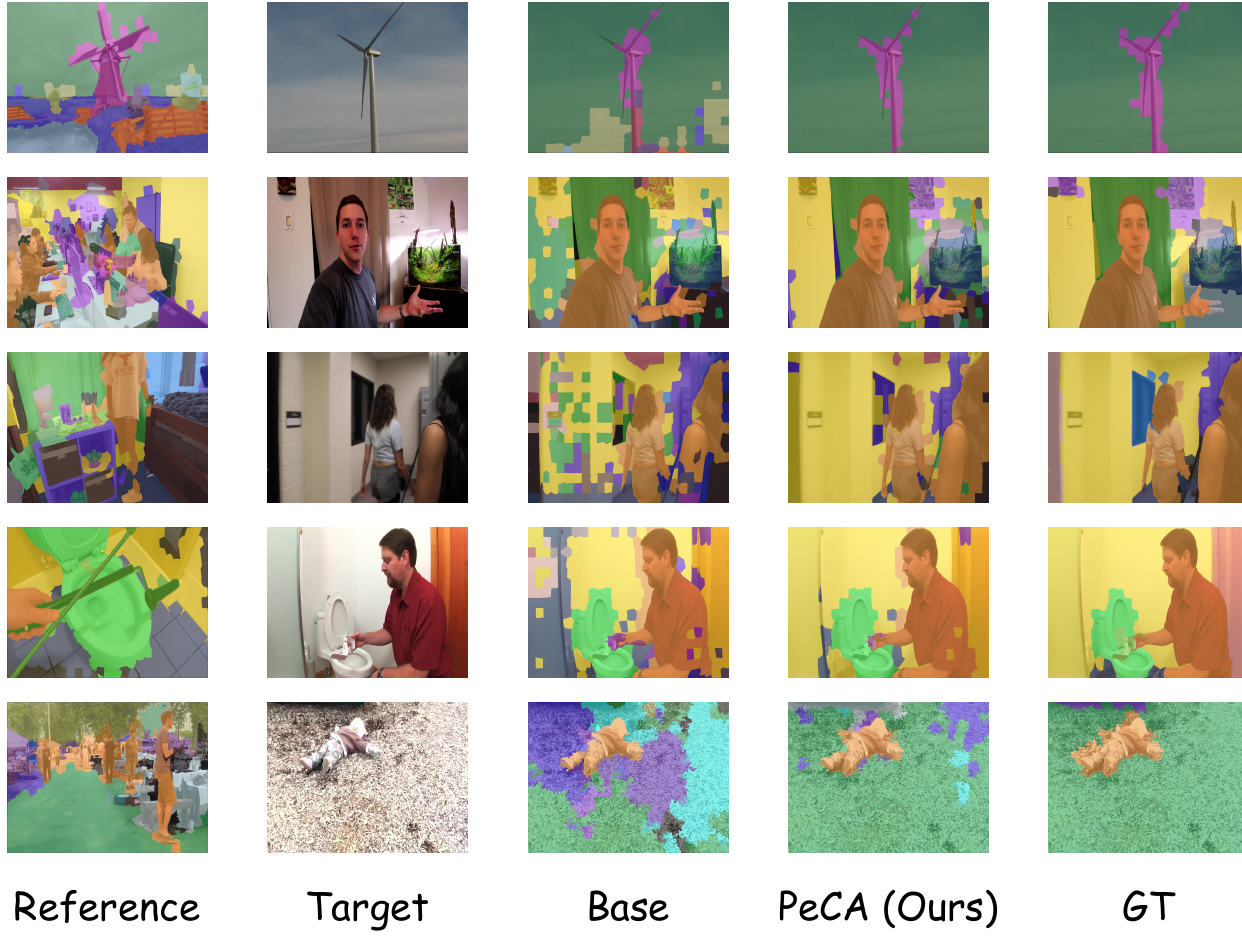}
    \caption{\textbf{Qualitative results on VIPSeg superpixel region label propagation.}
    From left to right: first external reference frame with label map, target RGB frame (input), \textit{Base}, \textsc{PeCA}, and ground truth. Compared with direct matching, \textsc{PeCA} produces more spatially coherent predictions with fewer fragmented labels and better object coverage. Ground truth denotes the superpixel semantic labels induced from the VIPSeg \cite{miao2022large} annotations.}
    \label{fig:VIPSEG_QUAL}
\end{figure}

To test whether \textsc{PeCA} generalises beyond palette-based colour assignment, we evaluate it on a reference-guided {Region Label Propagation} task built on the panoptic video segmentation dataset VIPSeg~\cite{miao2022large}. Given a target video and a small set of external reference frame(s), the goal is to assign each target superpixel a semantic label from its correspondences to reference superpixels. Ground-truth superpixel labels are induced from VIPSeg panoptic annotations via maximum overlap, and we evaluate predictions using Segment-wise accuracy (\texttt{Seg-Acc}), as well as pixel-level accuracy and Mean IoU (\texttt{Pix-Acc, Pix-MIoU}). 

Specifically, we use the VIPSeg validation split (343 videos, 8,255 frames) as targets. For each target frame, we over-segment the RGB image into SLIC superpixels~\cite{slic}. Each target superpixel is assigned a semantic category by maximum overlap with the panoptic mask, which serves as the ground-truth label for evaluation.

For this diagnostic experiment, we use the ground-truth panoptic labels to identify the semantic classes present in each target video, and then greedily select a small set of external reference frames from the VIPSeg training split whose union covers these classes. We apply the same SLIC over-segmentation and maximum-overlap label assignment to the selected reference frames, yielding reference superpixels with semantic labels.

With these reference--target pairs constructed, we compare two inference pipelines as in \cref{sec:experiments}. \textit{Base} uses backbone features and direct matching (\cref{eq:hardmatch}) to perform nearest-neighbour superpixel matching and hard label transfer. \textsc{PeCA} enables the full inference pipeline with the same default hyperparameters as in the colourisation experiments, aggregating matched semantic labels to obtain the final prediction. We report results with two generic pretrained backbones, DINOv2 ViT-L/14~\cite{oquab2023dinov2} and SAM2.1-Large~\cite{ravi2024sam}. Metrics are computed per frame and averaged over all evaluation frames following \cref{sec:metrics}.

\begin{table}[h!]
\centering
\caption{\textbf{VIPSeg Region Label Propagation results.}
All numbers are frame-wise averages. \textsc{PeCA} consistently improves over direct hard matching across both backbones and all metrics.}
\label{tab:vipseg_slic_results}
\resizebox{0.8\linewidth}{!}{
\begin{tabular}{l|l|ccc}
\toprule
\textbf{Backbone} & \textbf{Pipeline} & \textbf{Seg-Acc (\%)} & \textbf{Pix-Acc (\%)} & \textbf{Pix-MIoU (\%)} \\
\midrule
\multirow{2}{*}{SAM2.1-Large}
& \textit{Base}                 & 33.35 & 33.05 & 6.78 \\
& \textsc{PeCA} (ours) & \textbf{38.95} & \textbf{38.79} & \textbf{10.85} \\
\midrule
\multirow{2}{*}{DINOv2 ViT-L/14}
& \textit{Base}                 & 44.12 & 44.03 & 12.68 \\
& \textsc{PeCA} (ours) & \textbf{52.47} & \textbf{52.38} & \textbf{19.23} \\
\bottomrule
\end{tabular}
}
\end{table}

As shown in \cref{tab:vipseg_slic_results}, \textsc{PeCA} yields consistent gains across both backbones and all three metrics, with qualitative examples in \cref{fig:VIPSEG_QUAL}. Although this task is outside our main scope of paint-bucket colourisation, the improvements suggest that \textsc{PeCA} captures a more general form of reference-guided region matching that transfers to natural videos.

\section{More Qualitative Results}
\label{sec:more_qual}
\begin{figure*}[h!]
    \centering
    \setlength{\tabcolsep}{1pt}
    \renewcommand{\arraystretch}{1.0}

    \begin{subfigure}[t]{\textwidth}
        \centering
        \resizebox{\textwidth}{!}{%
        \begin{tabular}{c c c c c c c}
            \tiny \textbf{Reference} &
            \tiny \textbf{Target} &
            \tiny \textbf{MangaNinja~\cite{liu2025manganinja}} &
            \tiny \textbf{BasicPBC-Ref~\cite{dai2024paint}} &
            \tiny \textbf{DACoN~\cite{nagata2025dacon}} &
            \tiny \textbf{Ours} &
            \tiny \textbf{Groundtruth} \\[-2pt]

            \includegraphics[width=0.135\textwidth]{figures/qual/bocchi_ref0.png} &
            \includegraphics[width=0.135\textwidth]{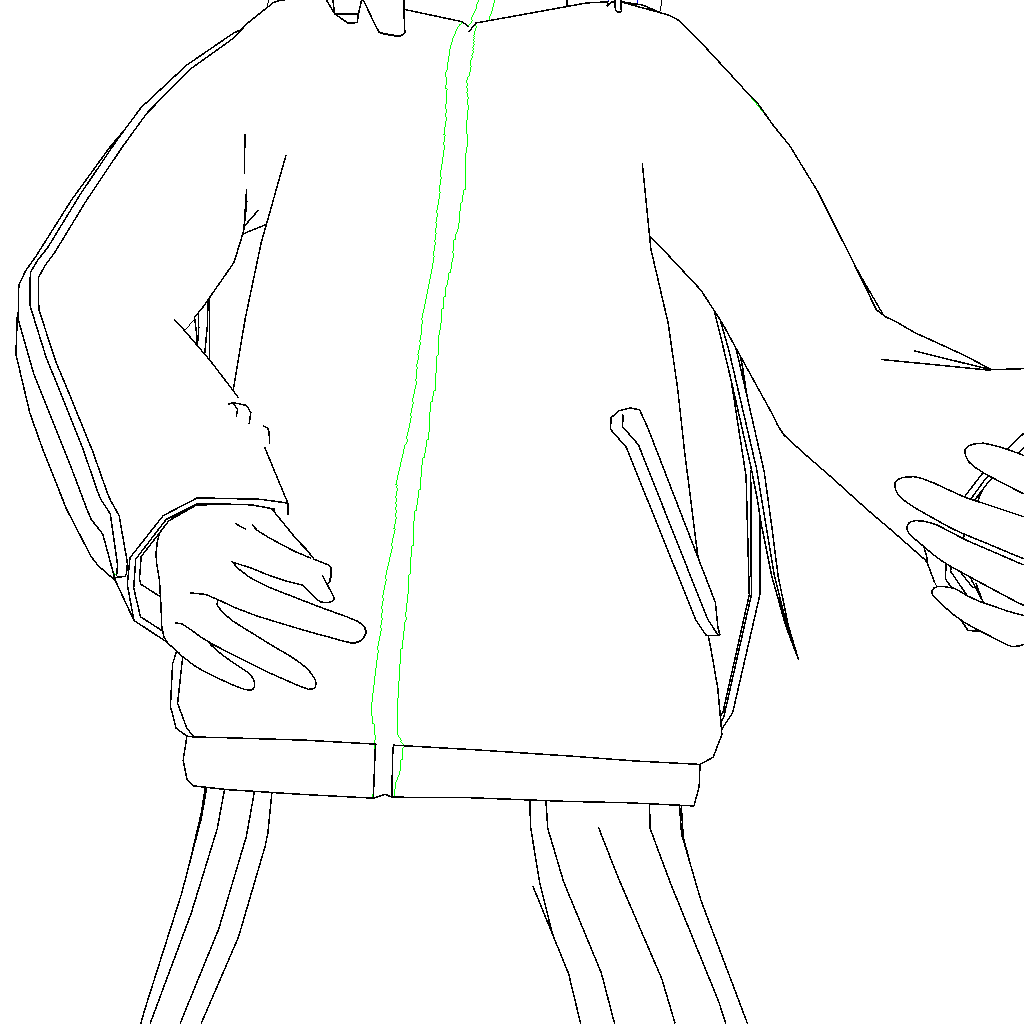} &
            \includegraphics[width=0.135\textwidth]{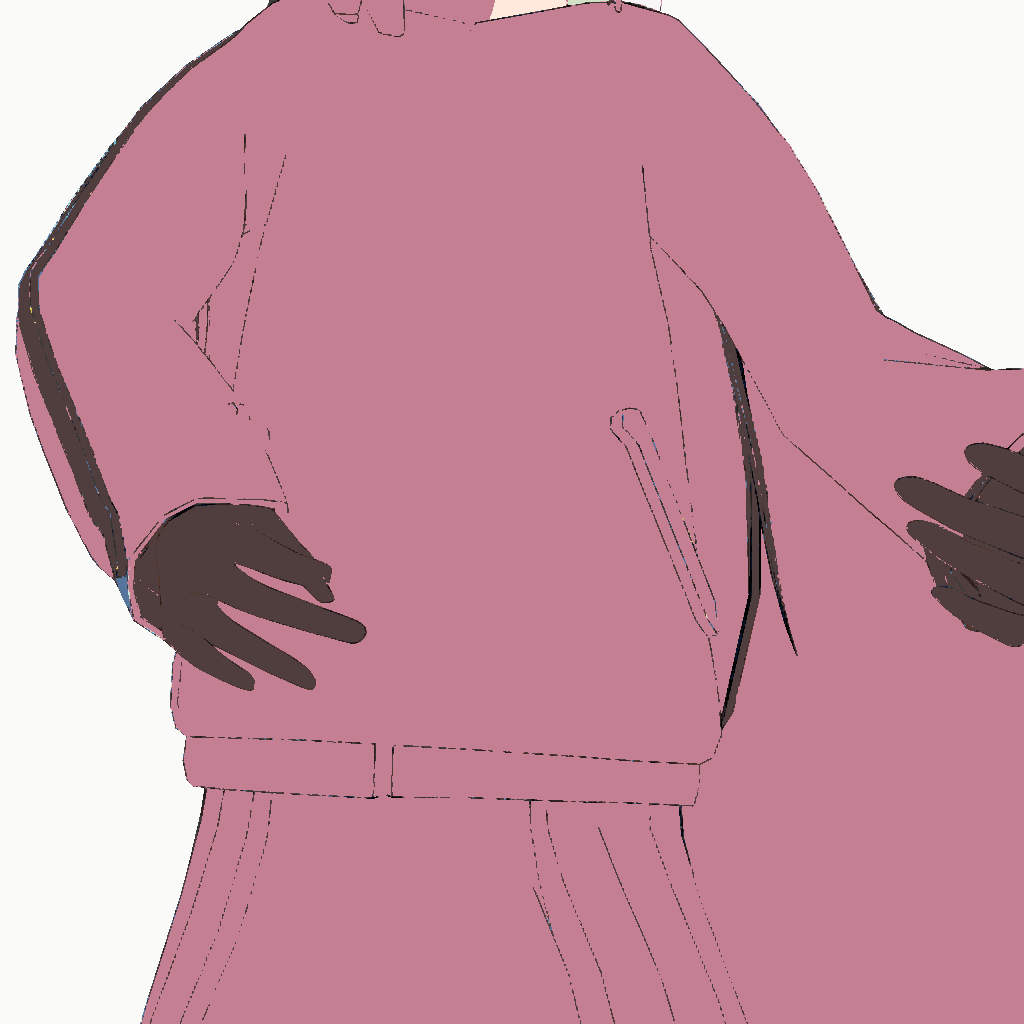} &
            \includegraphics[width=0.135\textwidth]{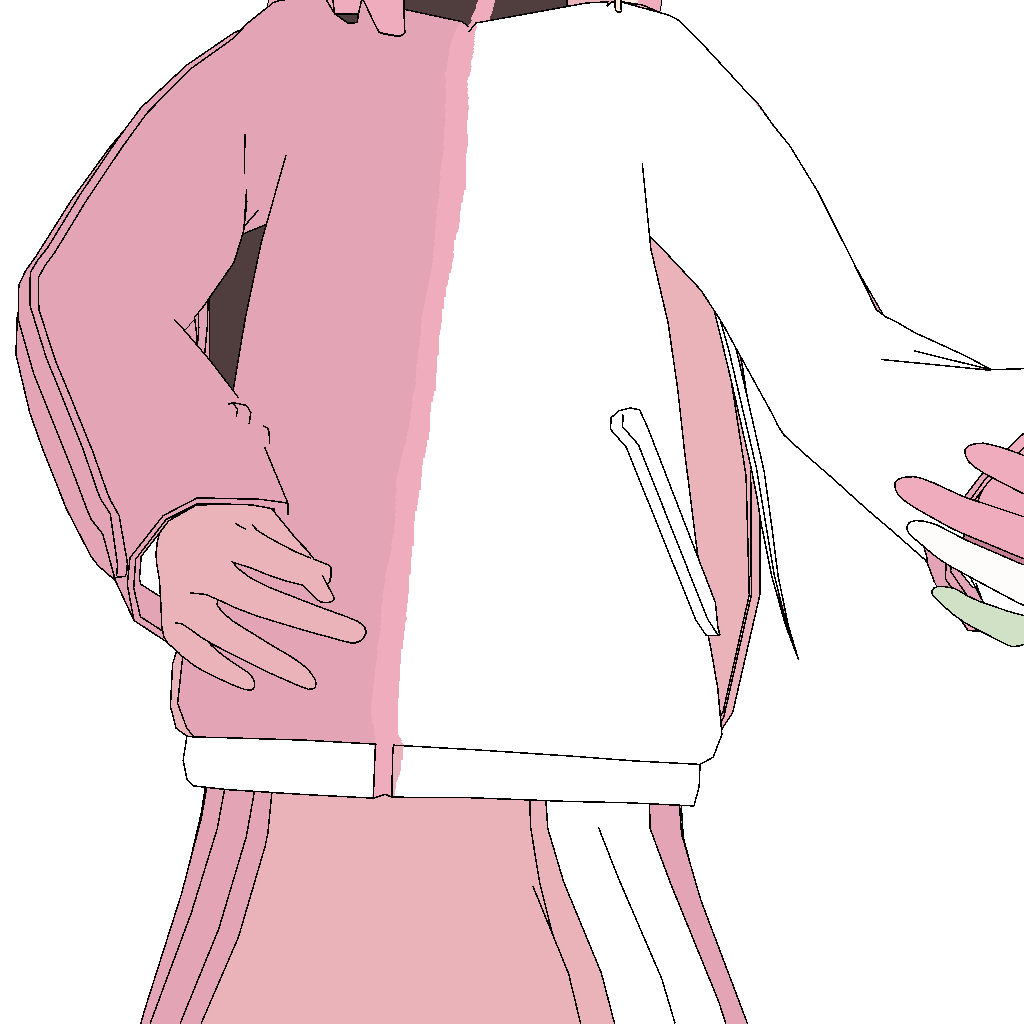} &
            \includegraphics[width=0.135\textwidth]{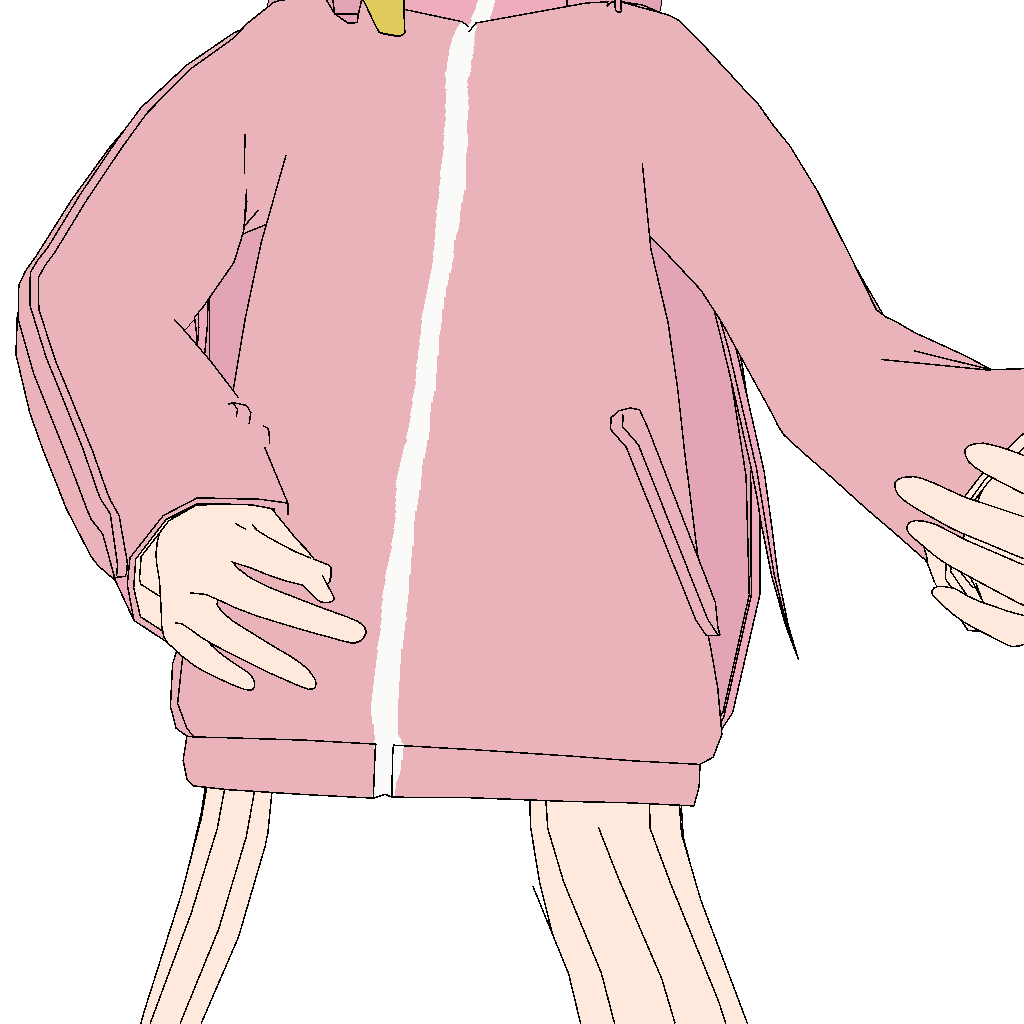} &
            \includegraphics[width=0.135\textwidth]{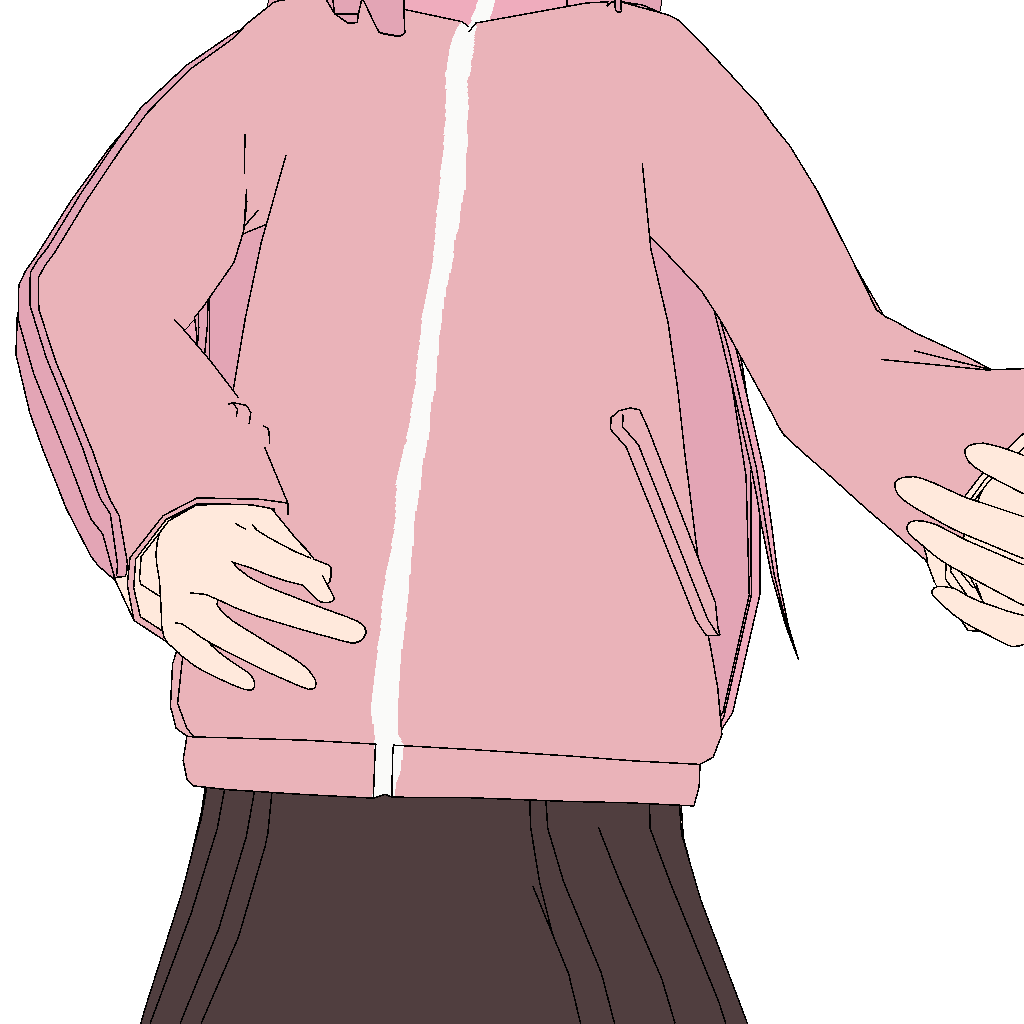} &
            \includegraphics[width=0.135\textwidth]{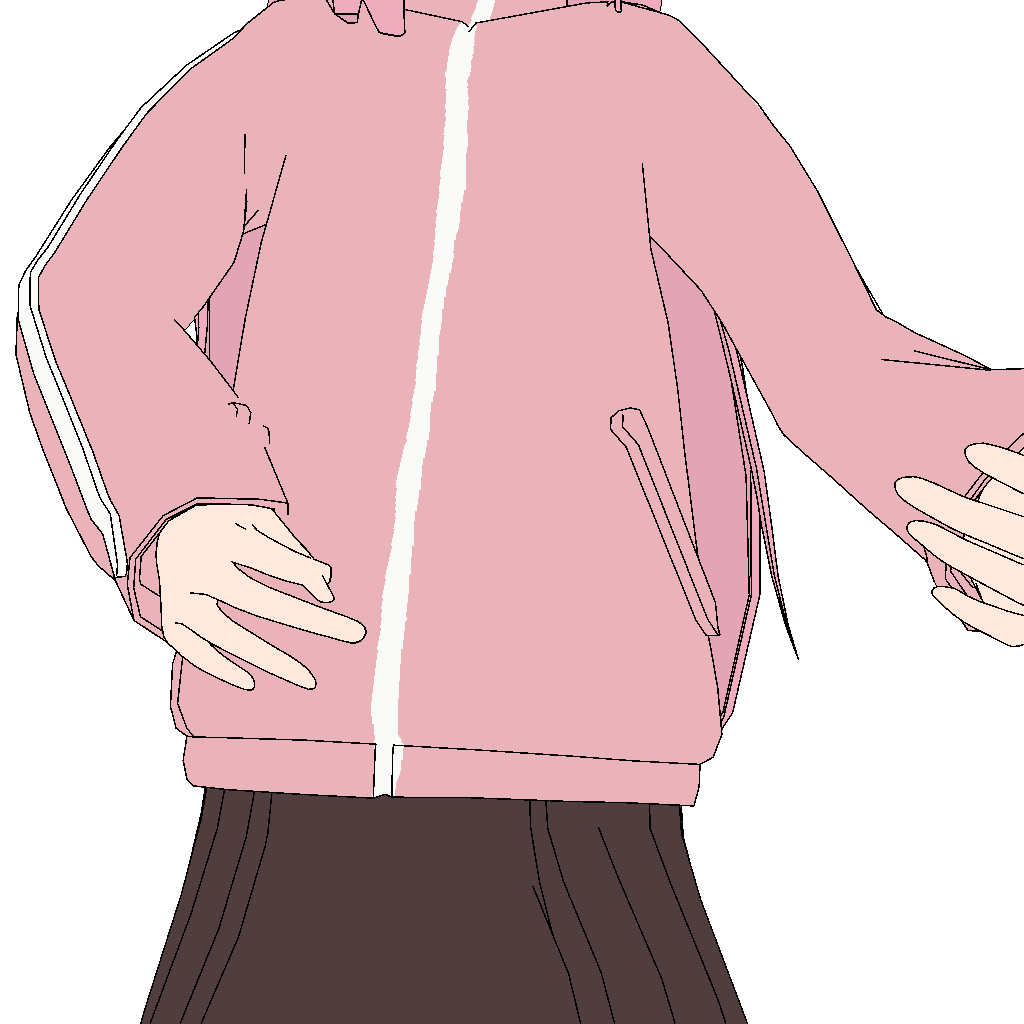} \\

            \includegraphics[width=0.135\textwidth]{figures/qual/0000michelle.png} &
            \includegraphics[width=0.135\textwidth]{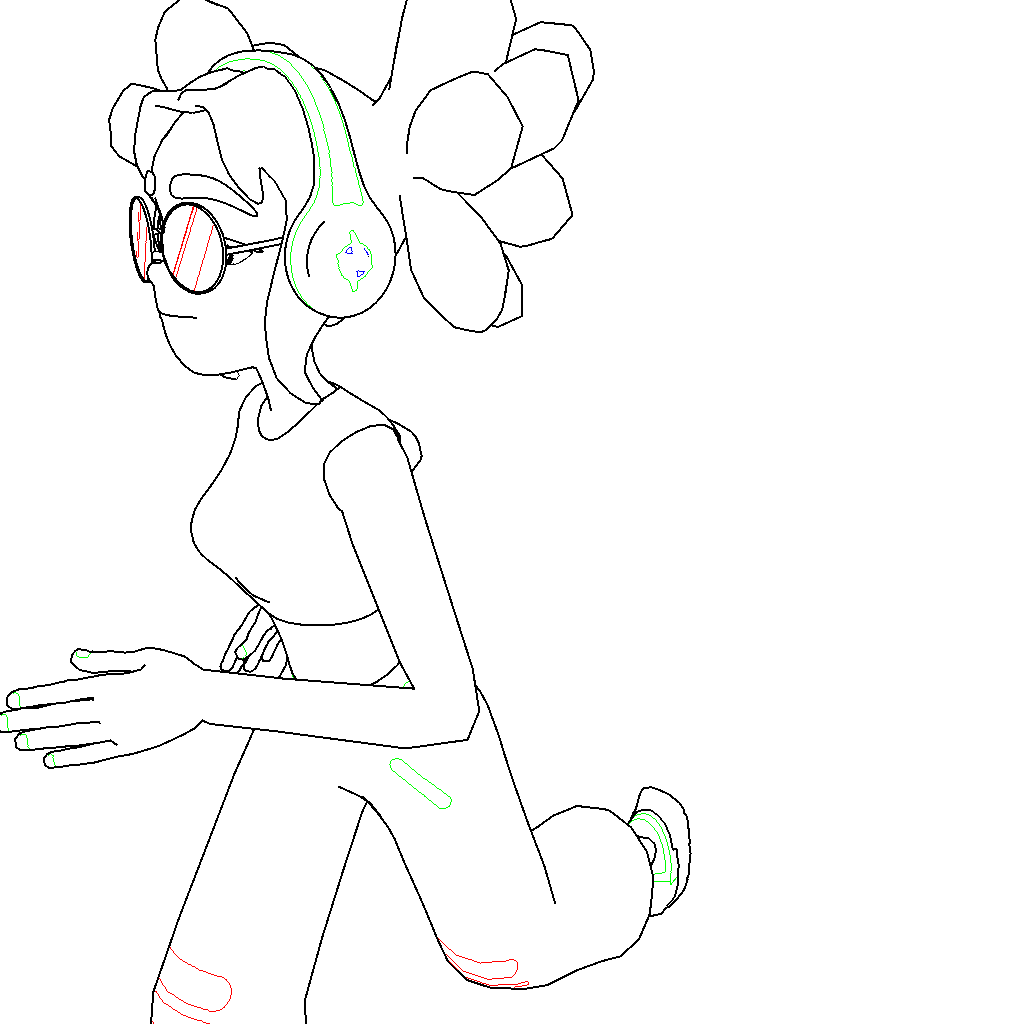} &
            \includegraphics[width=0.135\textwidth]{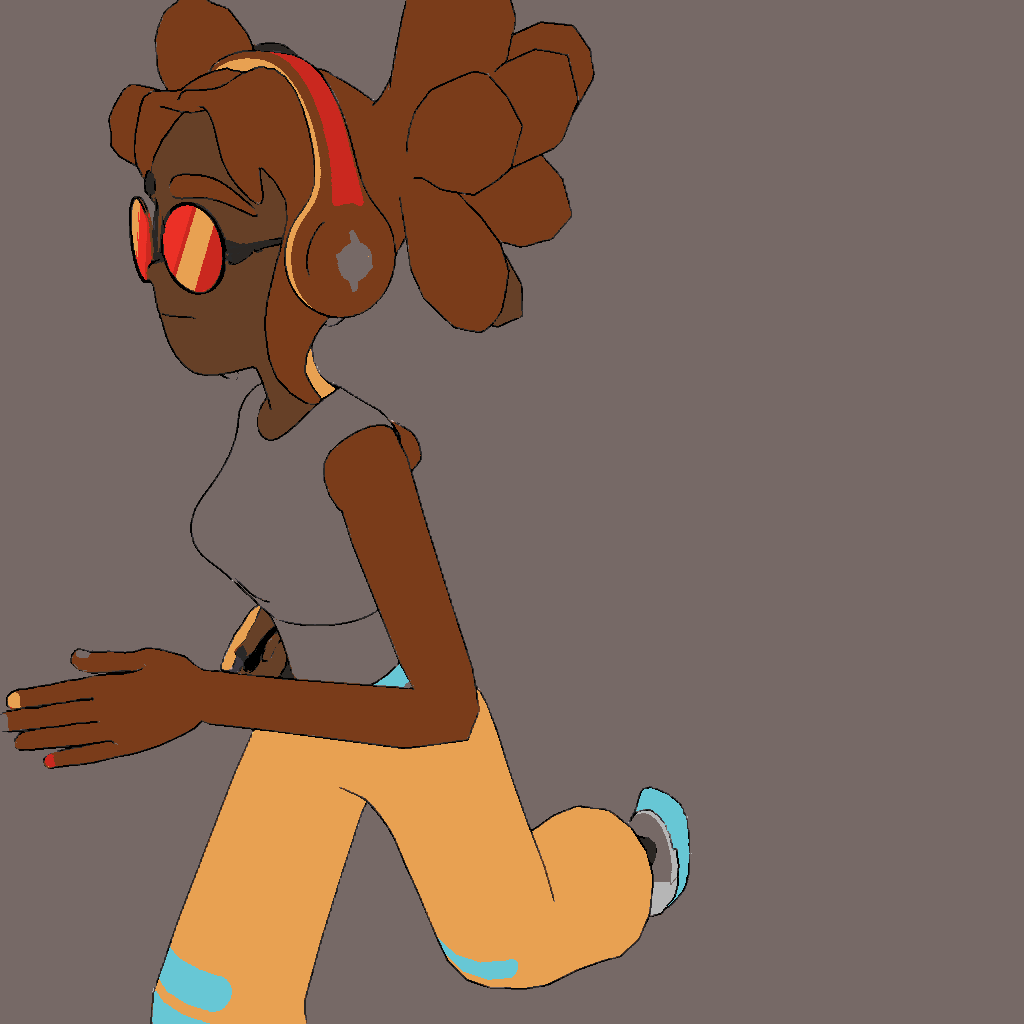} &
            \includegraphics[width=0.135\textwidth]{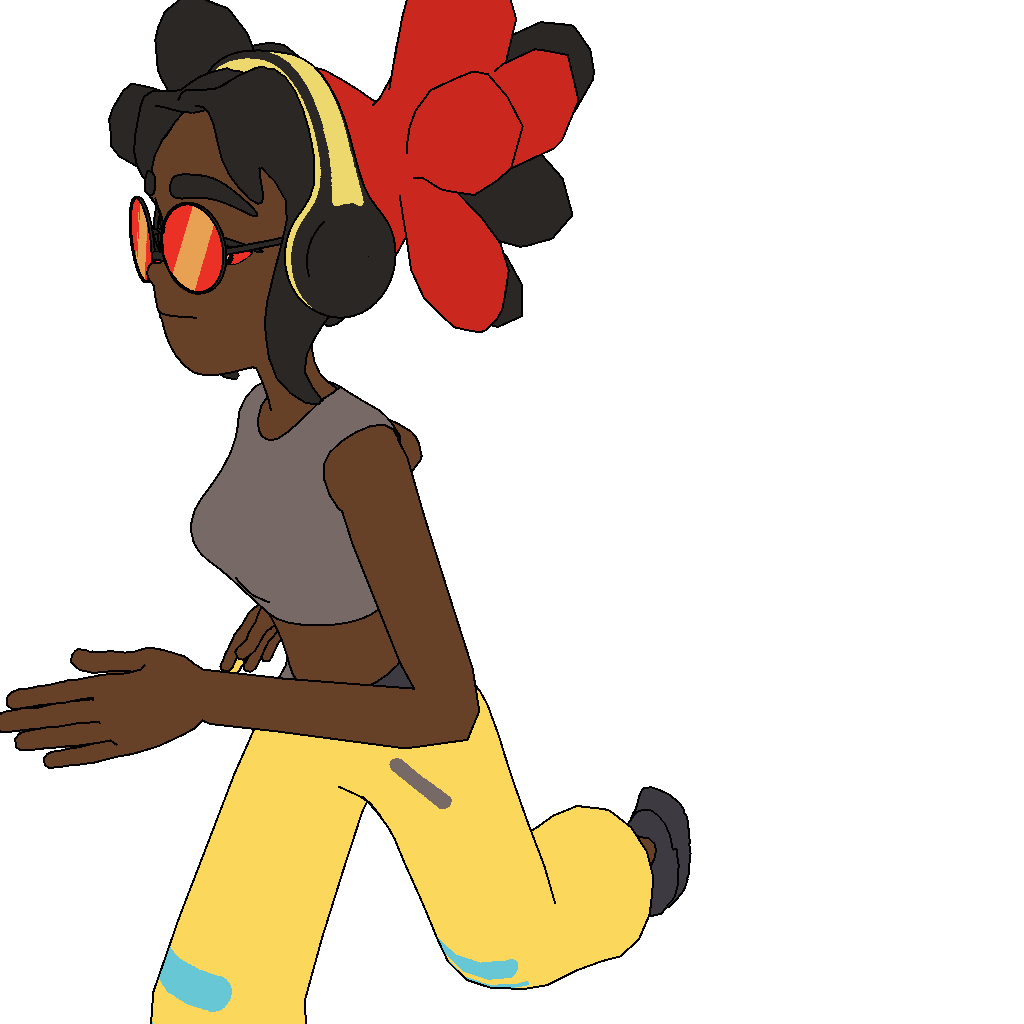} &
            \includegraphics[width=0.135\textwidth]{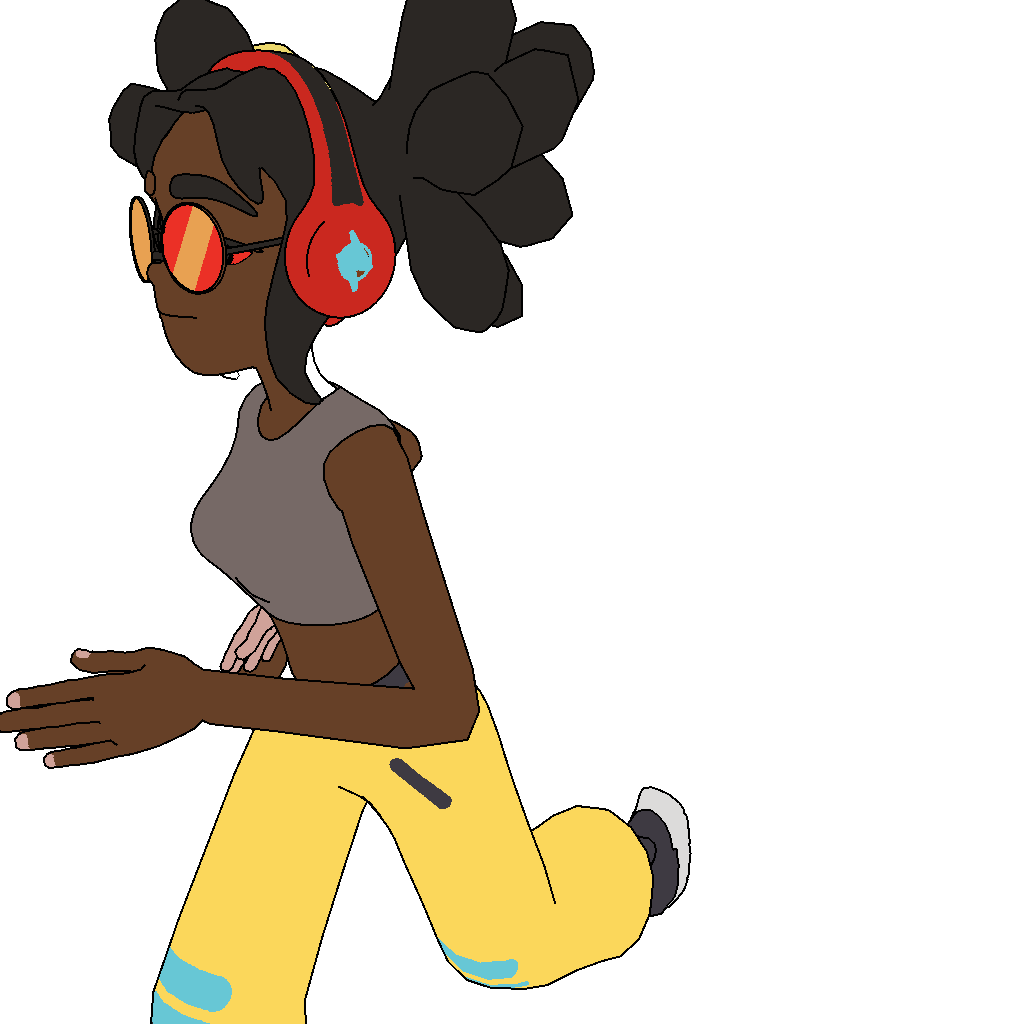} &
            \includegraphics[width=0.135\textwidth]{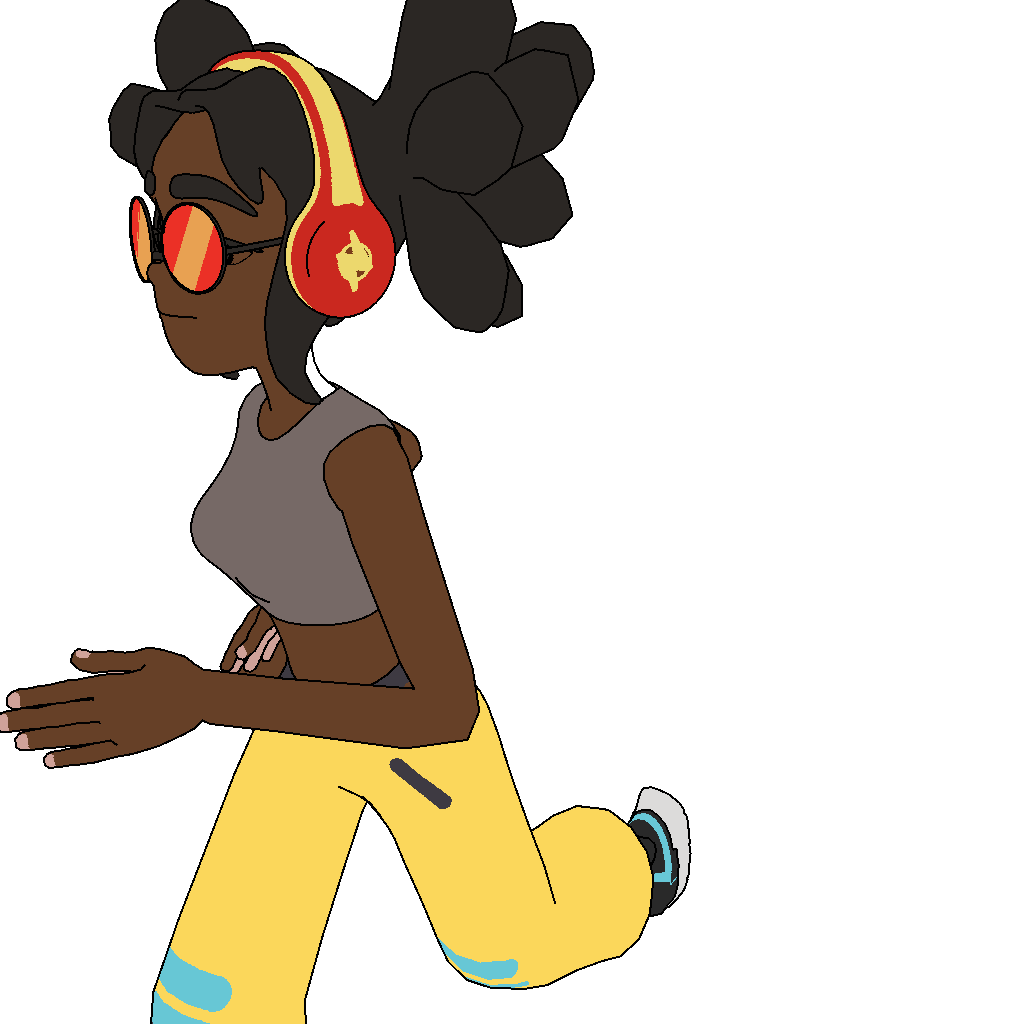} &
            \includegraphics[width=0.135\textwidth]{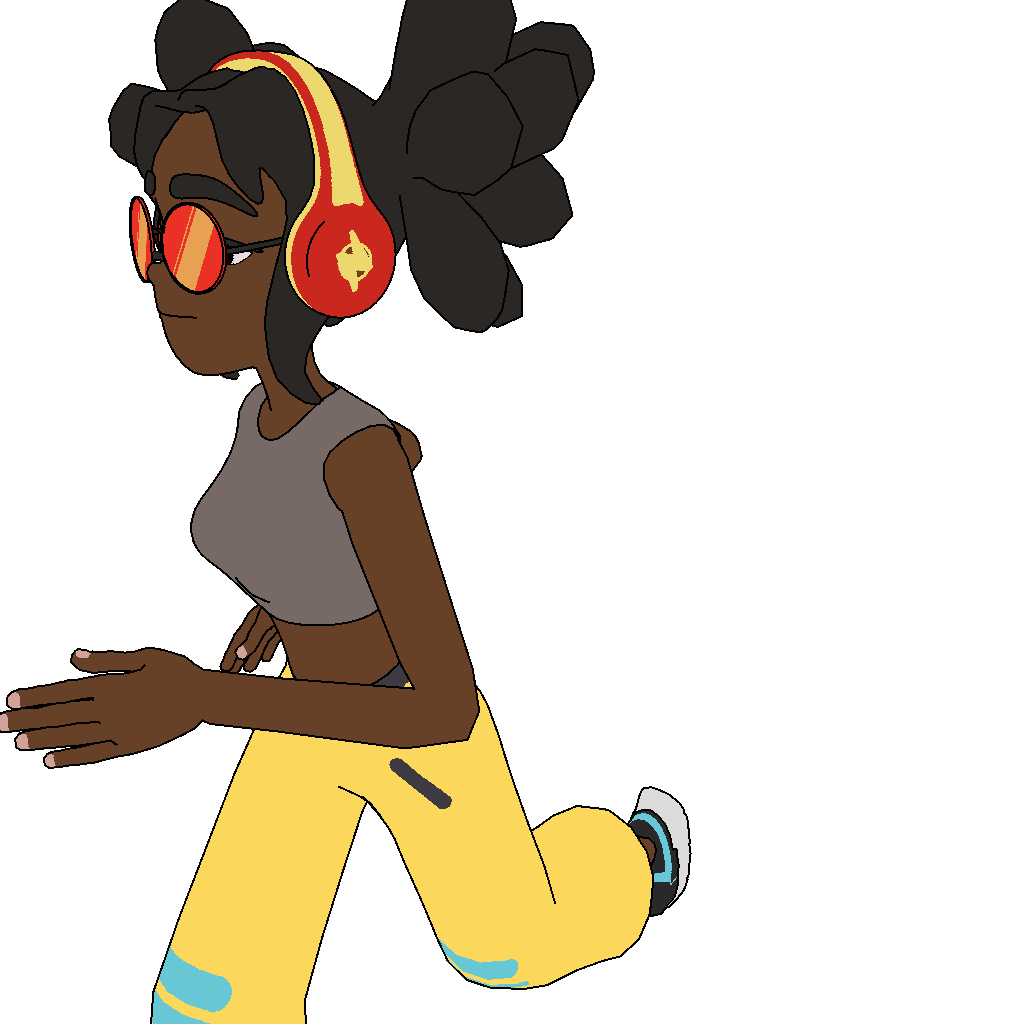} \\

            \includegraphics[width=0.135\textwidth]{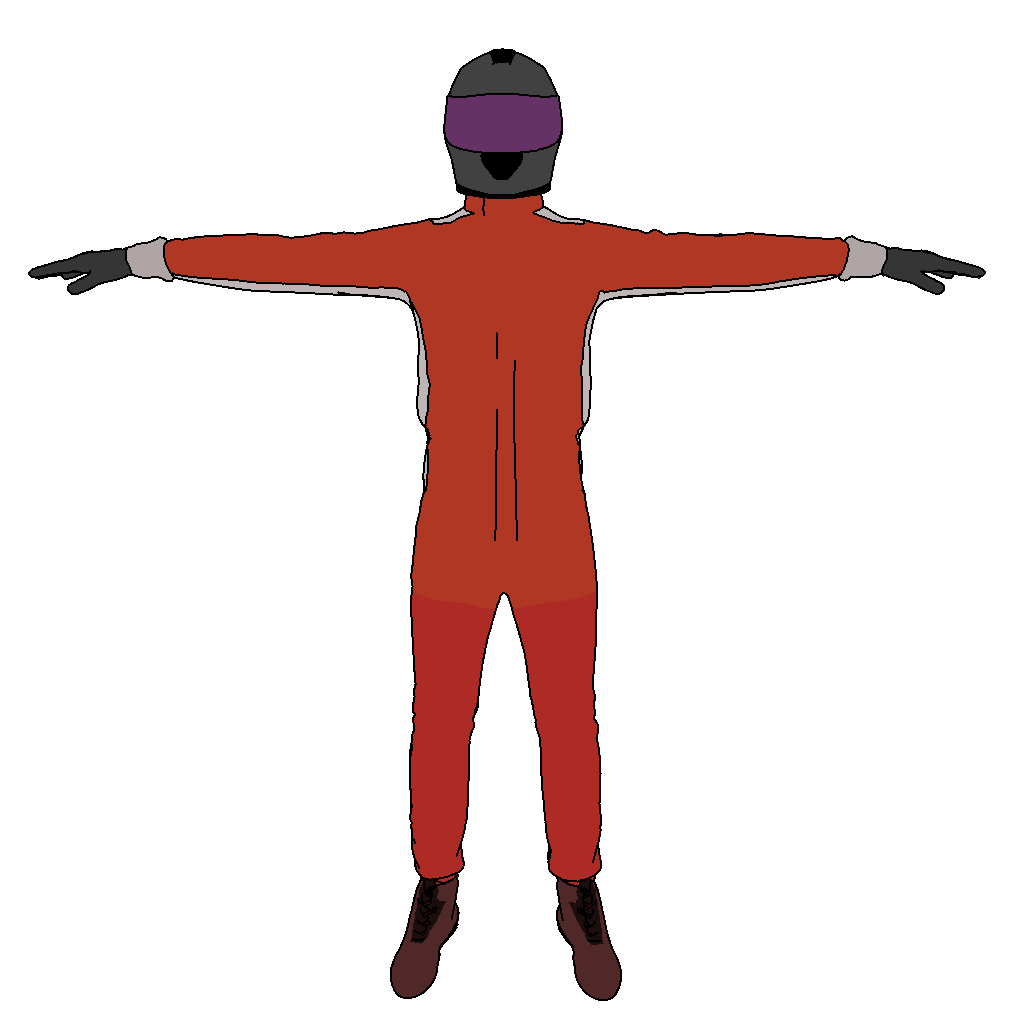} &
            \includegraphics[width=0.135\textwidth]{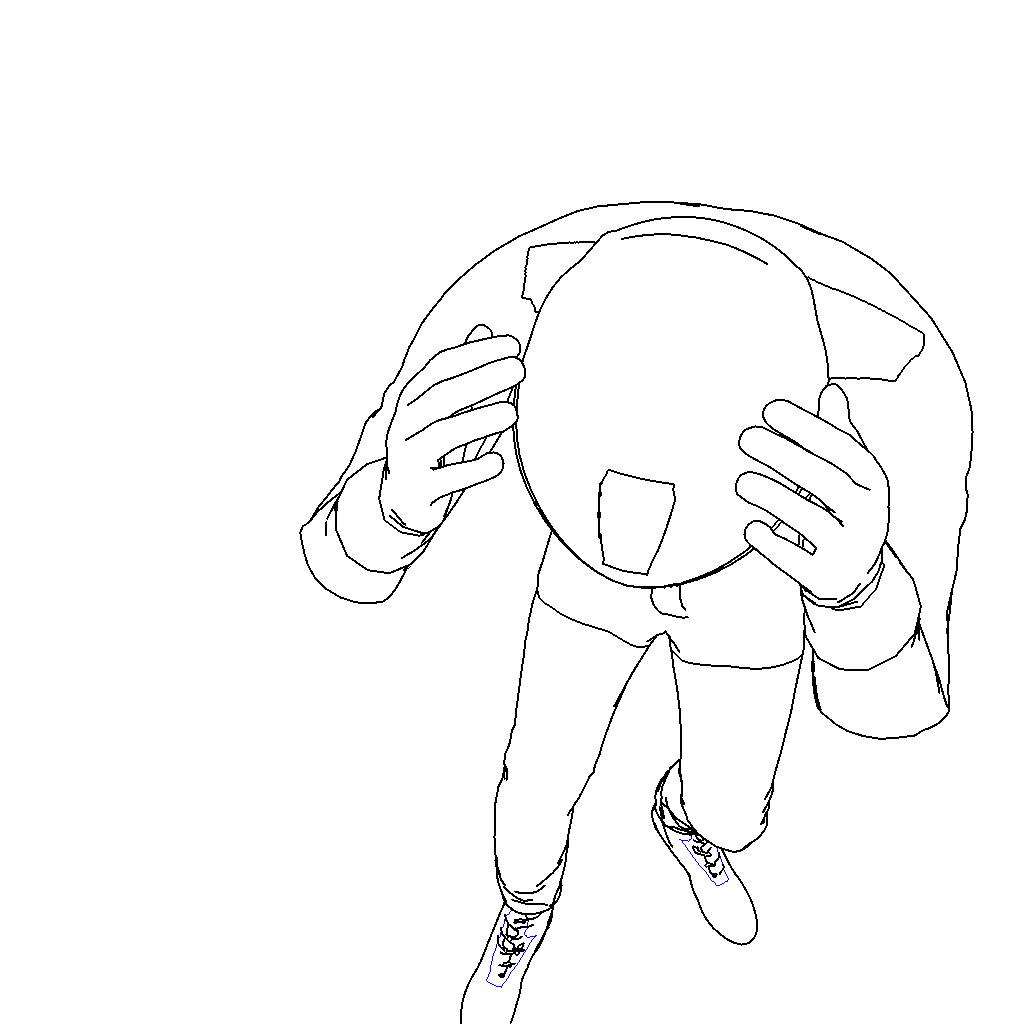} &
            \includegraphics[width=0.135\textwidth]{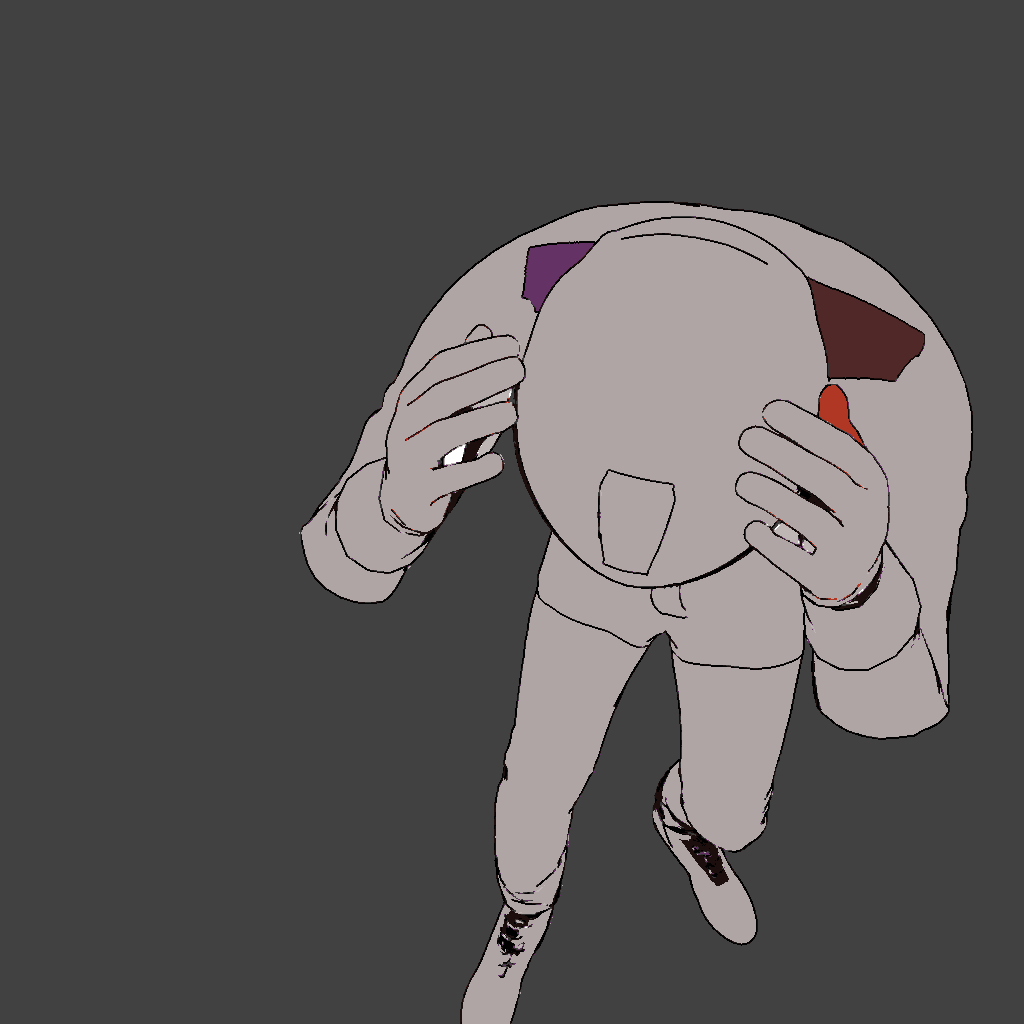} &
            \includegraphics[width=0.135\textwidth]{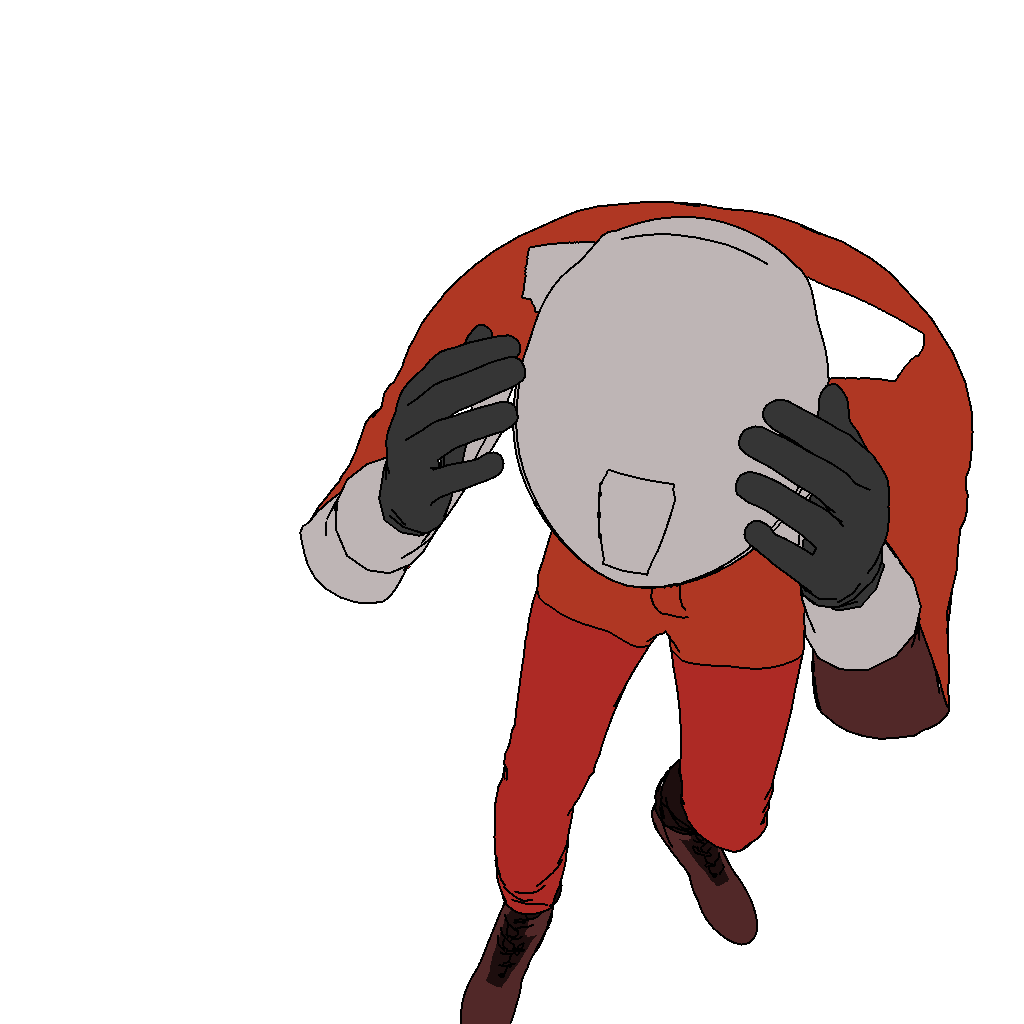} &
            \includegraphics[width=0.135\textwidth]{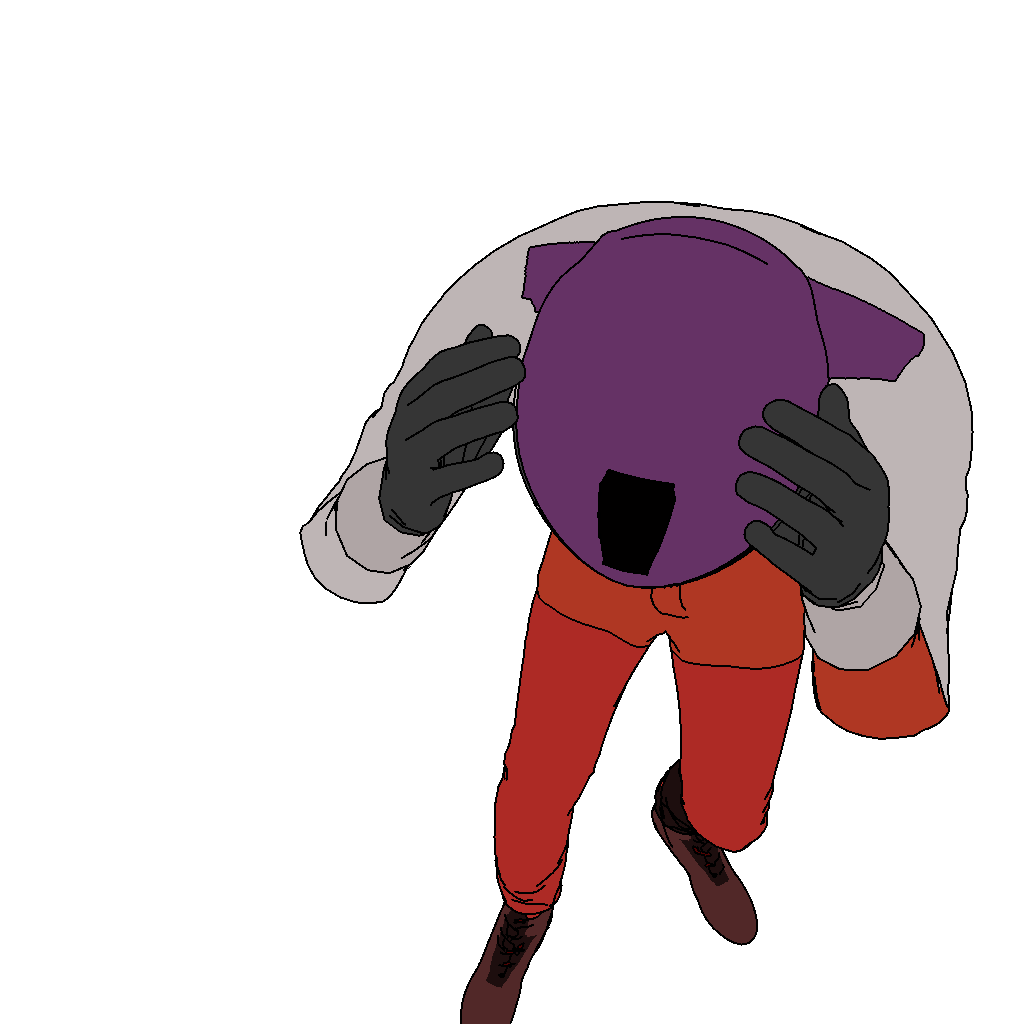} &
            \includegraphics[width=0.135\textwidth]{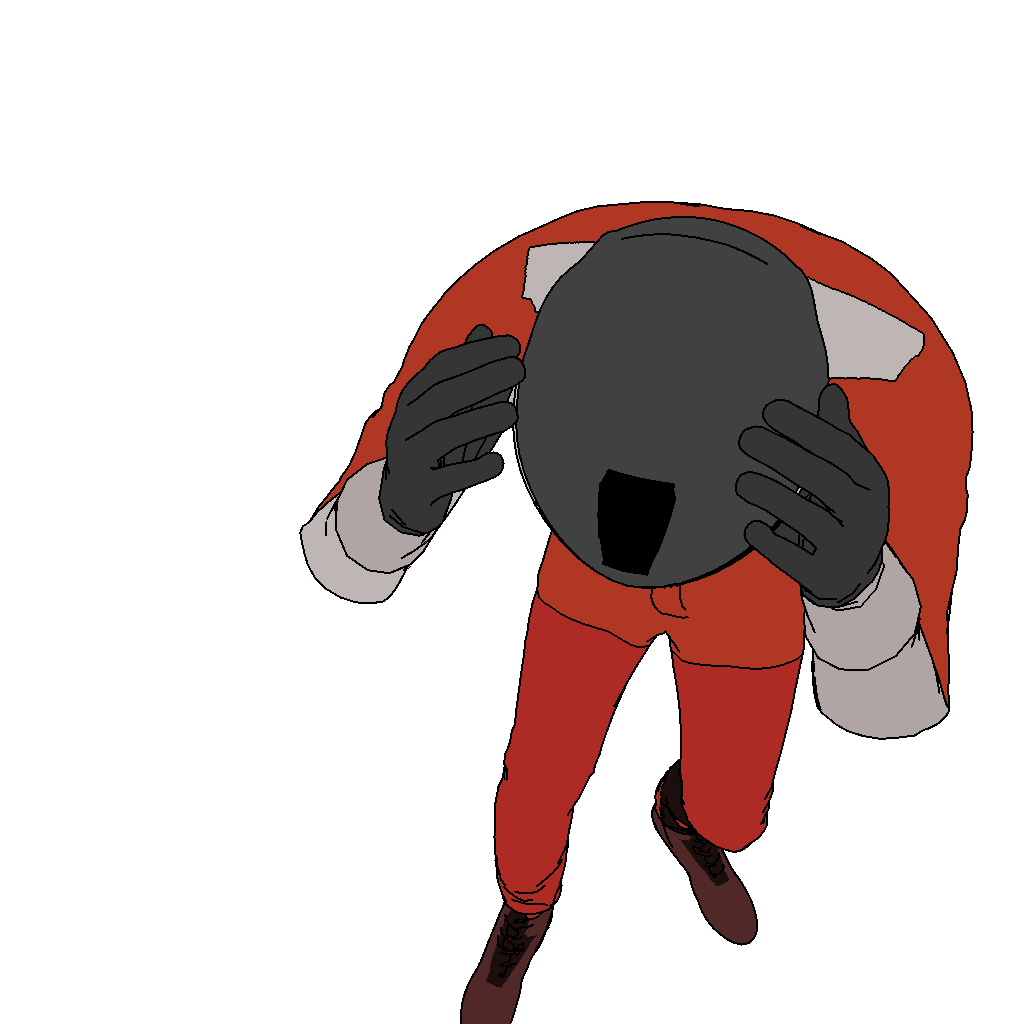} &
            \includegraphics[width=0.135\textwidth]{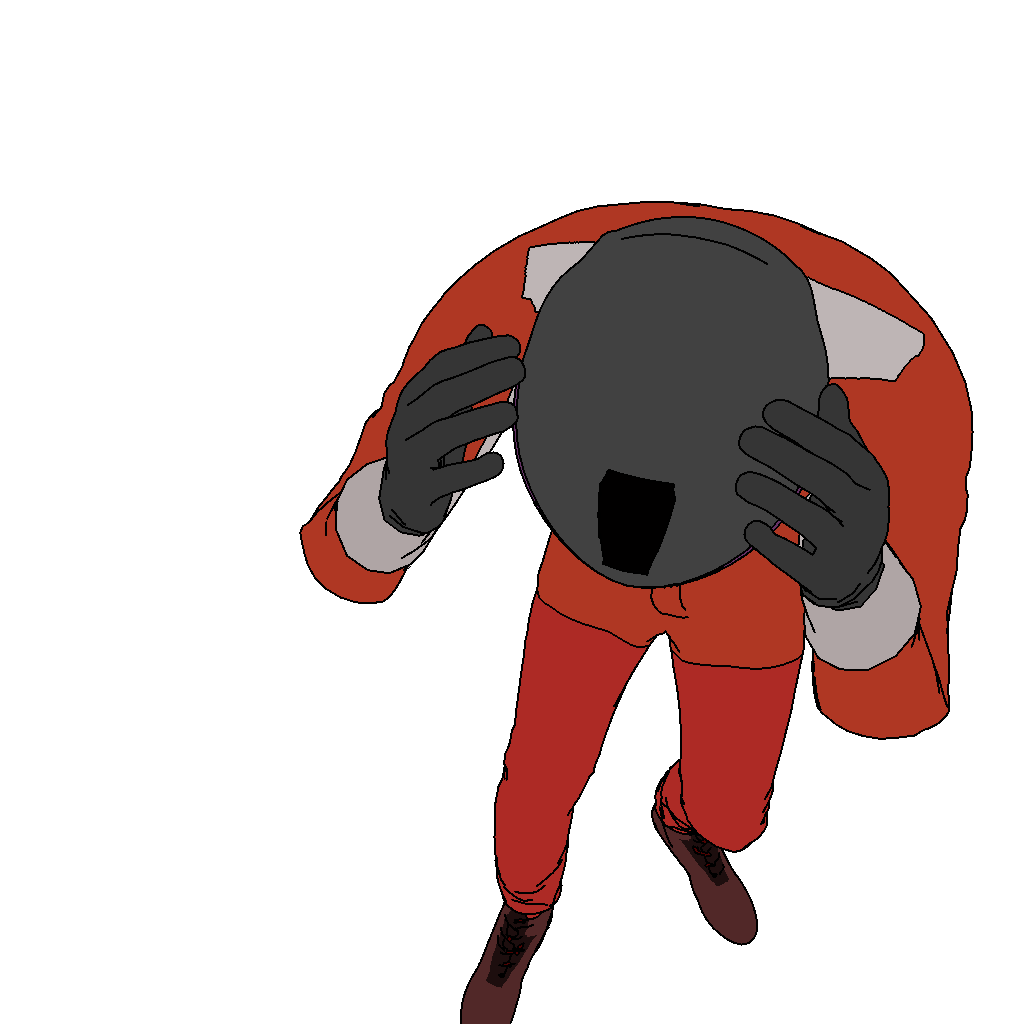} \\

            \includegraphics[width=0.135\textwidth]{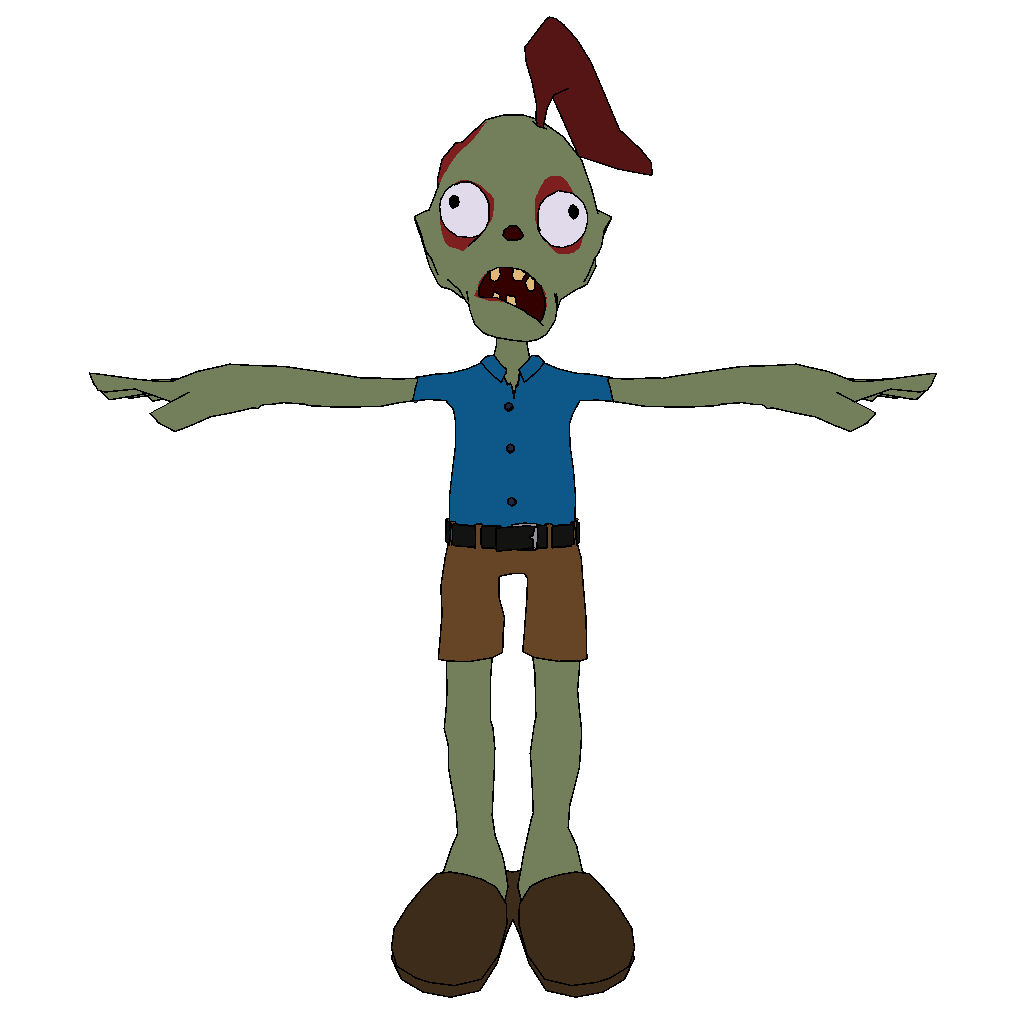} &
            \includegraphics[width=0.135\textwidth]{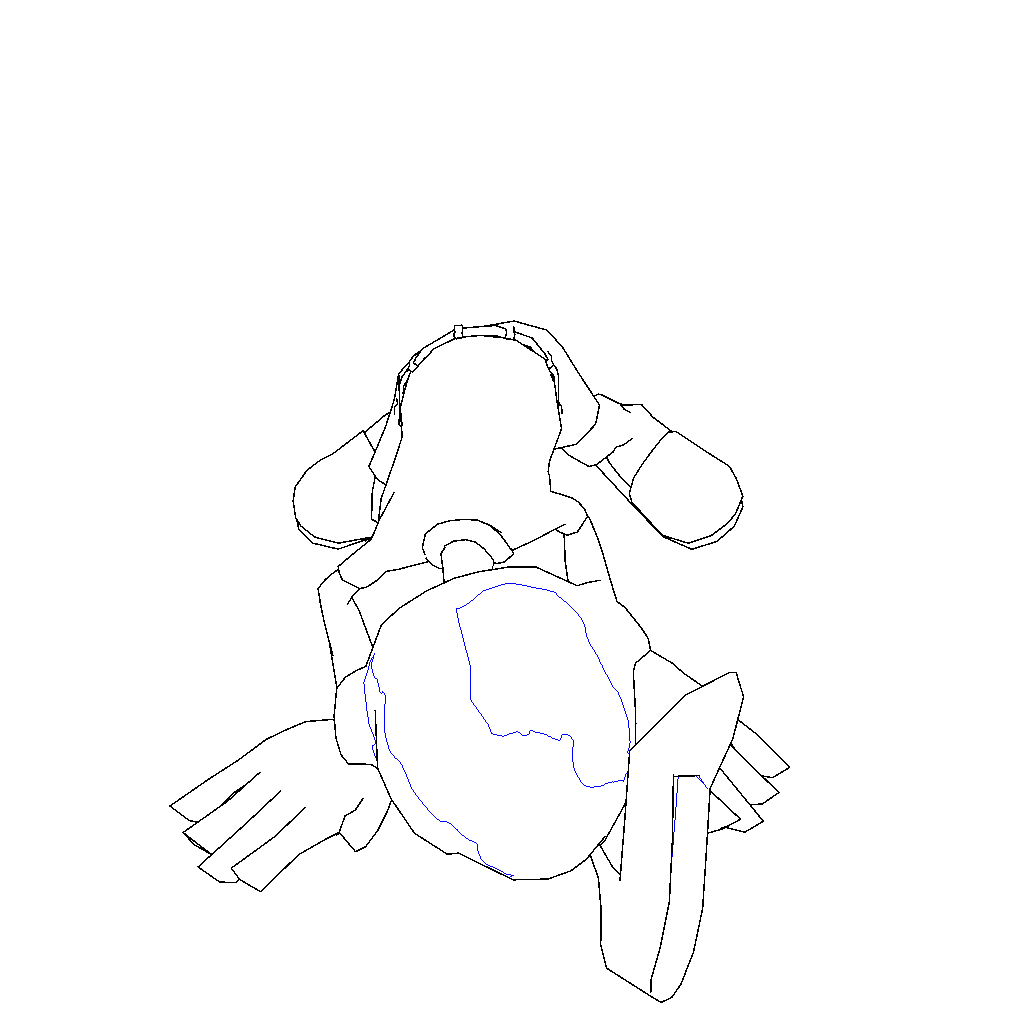} &
            \includegraphics[width=0.135\textwidth]{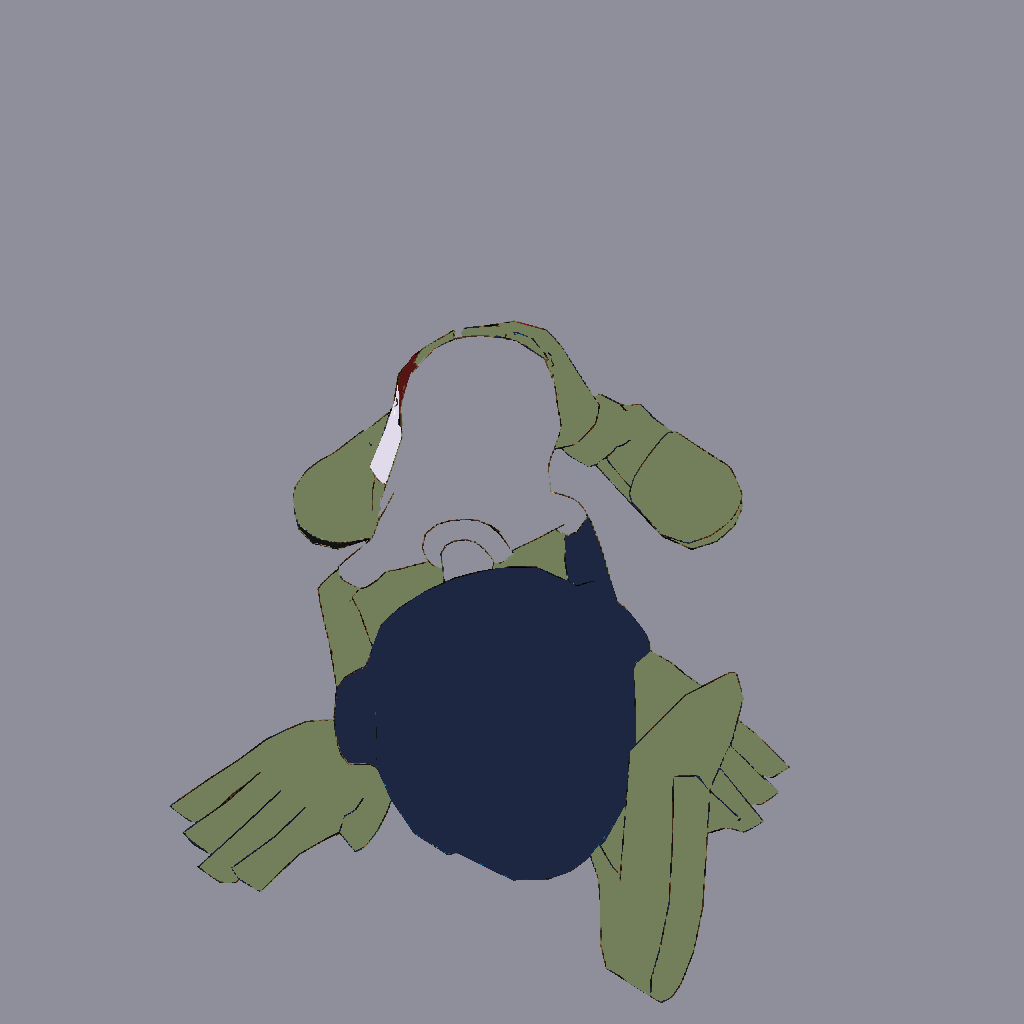} &
            \includegraphics[width=0.135\textwidth]{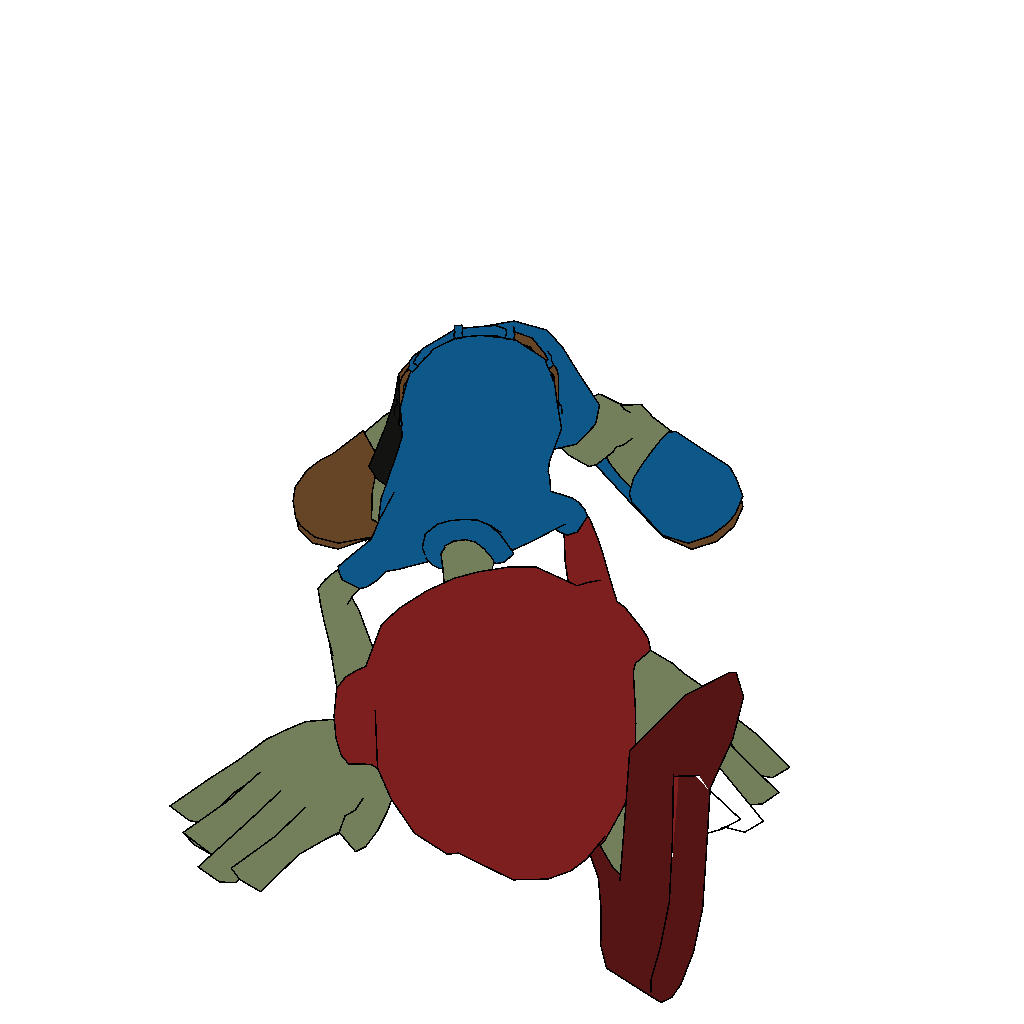} &
            \includegraphics[width=0.135\textwidth]{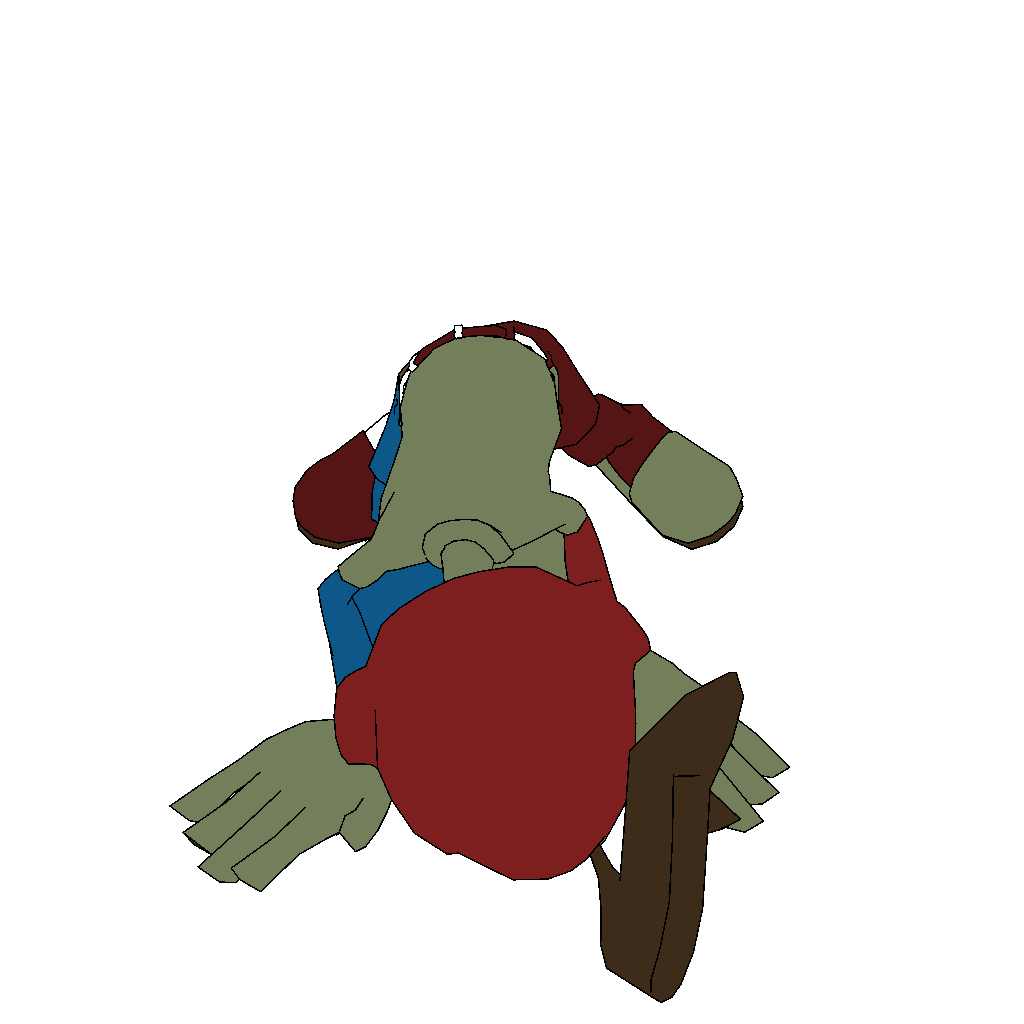} &
            \includegraphics[width=0.135\textwidth]{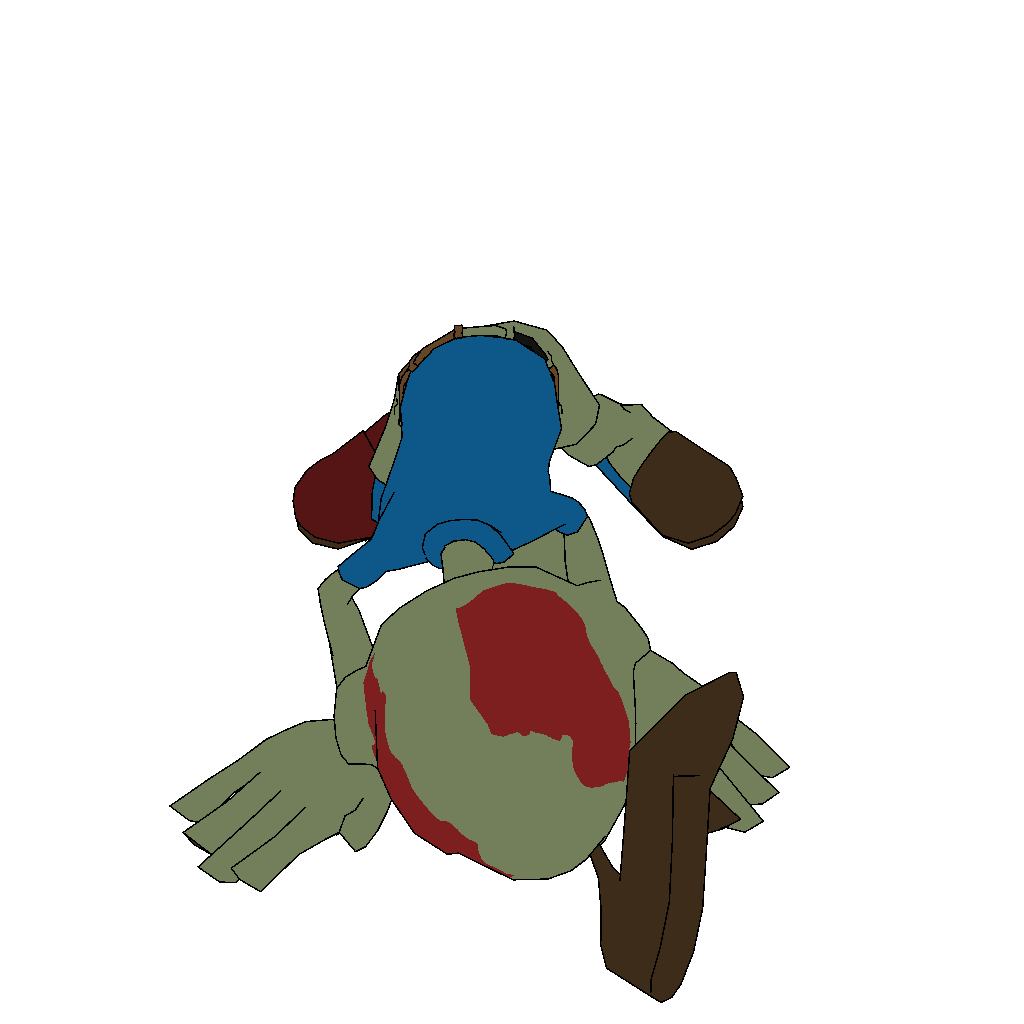} &
            \includegraphics[width=0.135\textwidth]{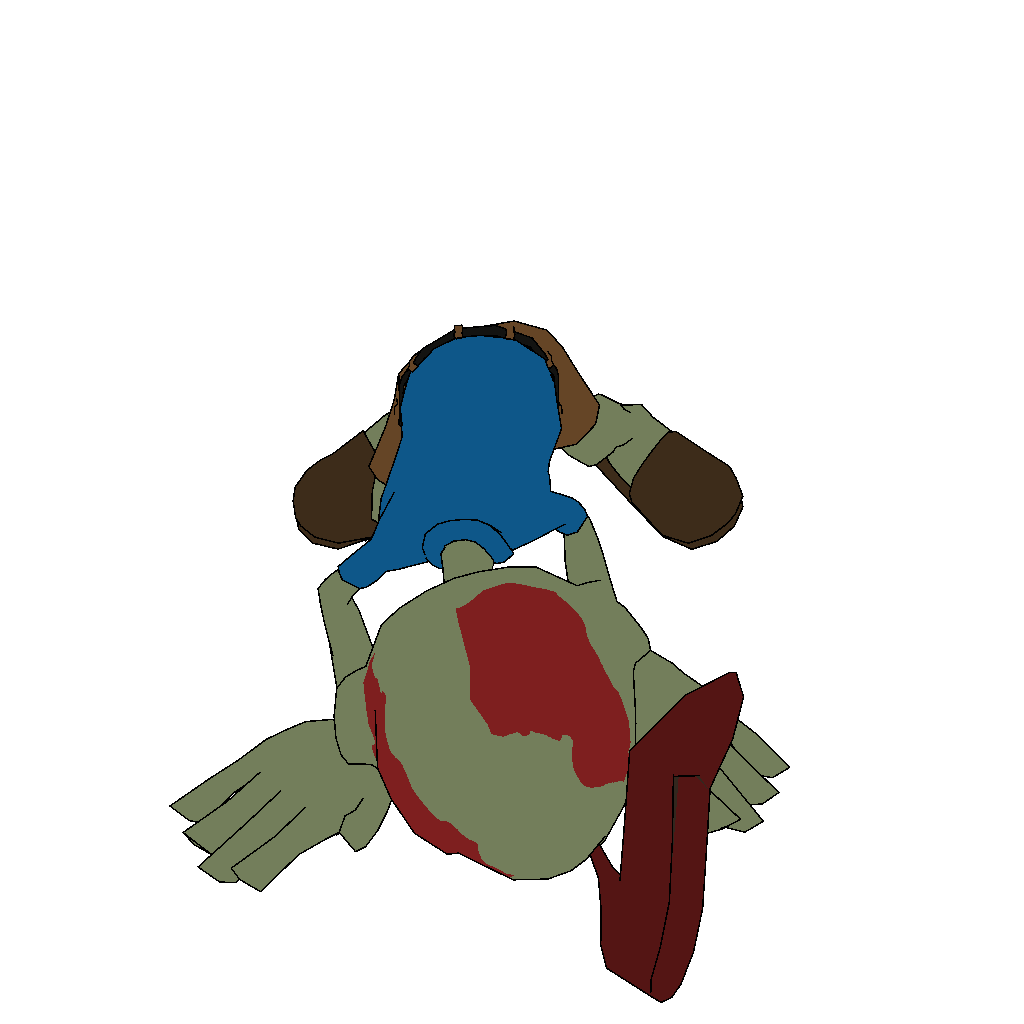} \\
        \end{tabular}%
        }

        \caption{Key-frame (Design-sheet) Reference Colourisation Results}
        \label{fig:morequal_keyframe}
    \end{subfigure}

    \begin{subfigure}[t]{\textwidth}
        \centering
        \resizebox{\textwidth}{!}{%
        \begin{tabular}{c c c c c c c}
            \tiny \textbf{Reference} &
            \tiny \textbf{Target} &
            \tiny \textbf{MangaNinja~\cite{liu2025manganinja}} &
            \tiny \textbf{BasicPBC~\cite{dai2024learning}} &
            \tiny \textbf{DACoN~\cite{nagata2025dacon}} &
            \tiny \textbf{Ours} &
            \tiny \textbf{Groundtruth} \\[-2pt]

            \includegraphics[width=0.135\textwidth]{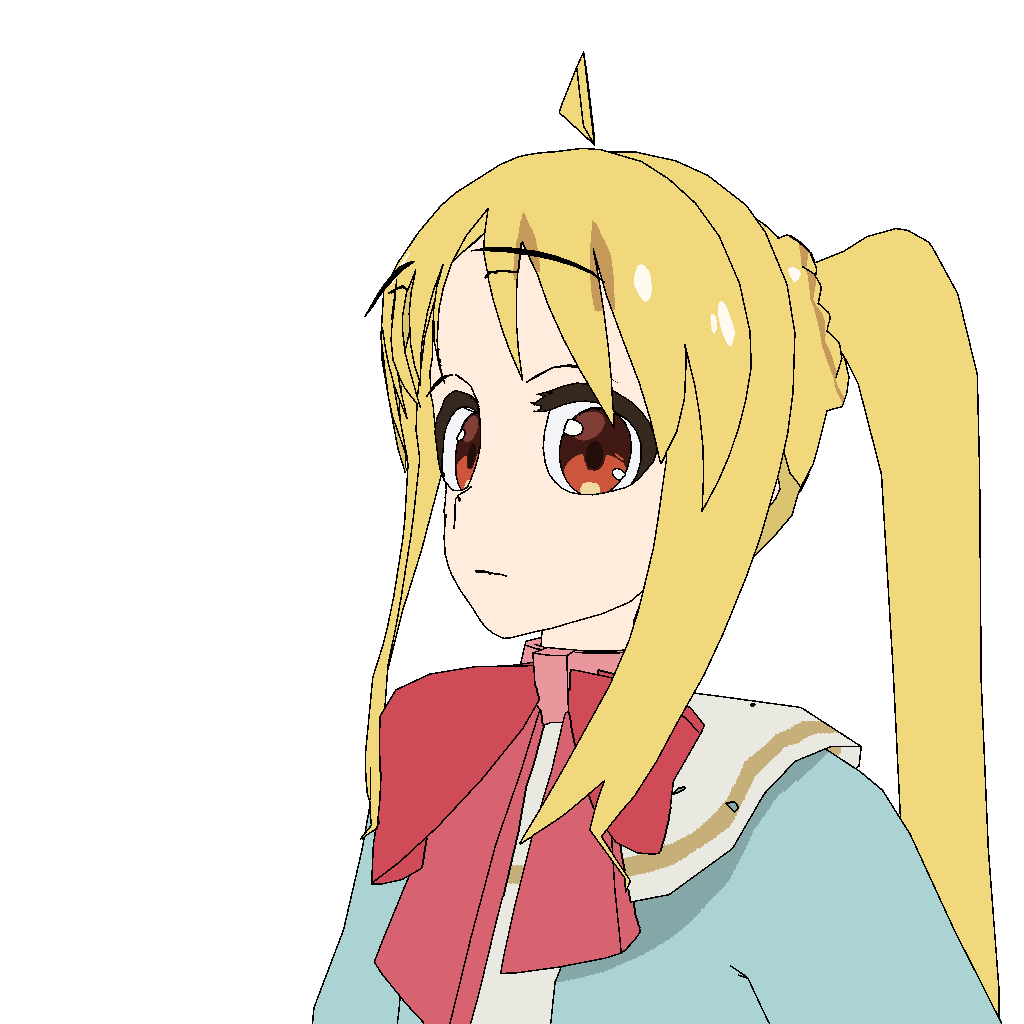} &
            \includegraphics[width=0.135\textwidth]{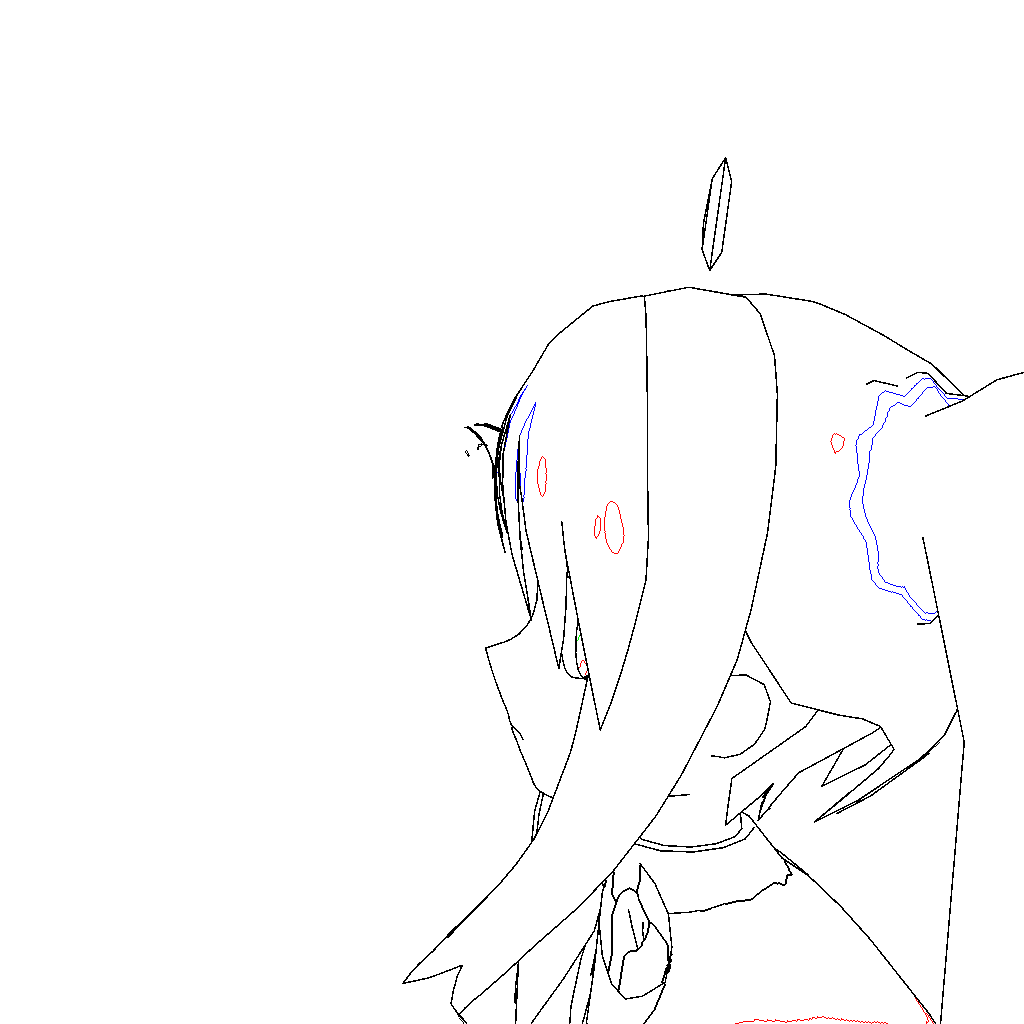} &
            \includegraphics[width=0.135\textwidth]{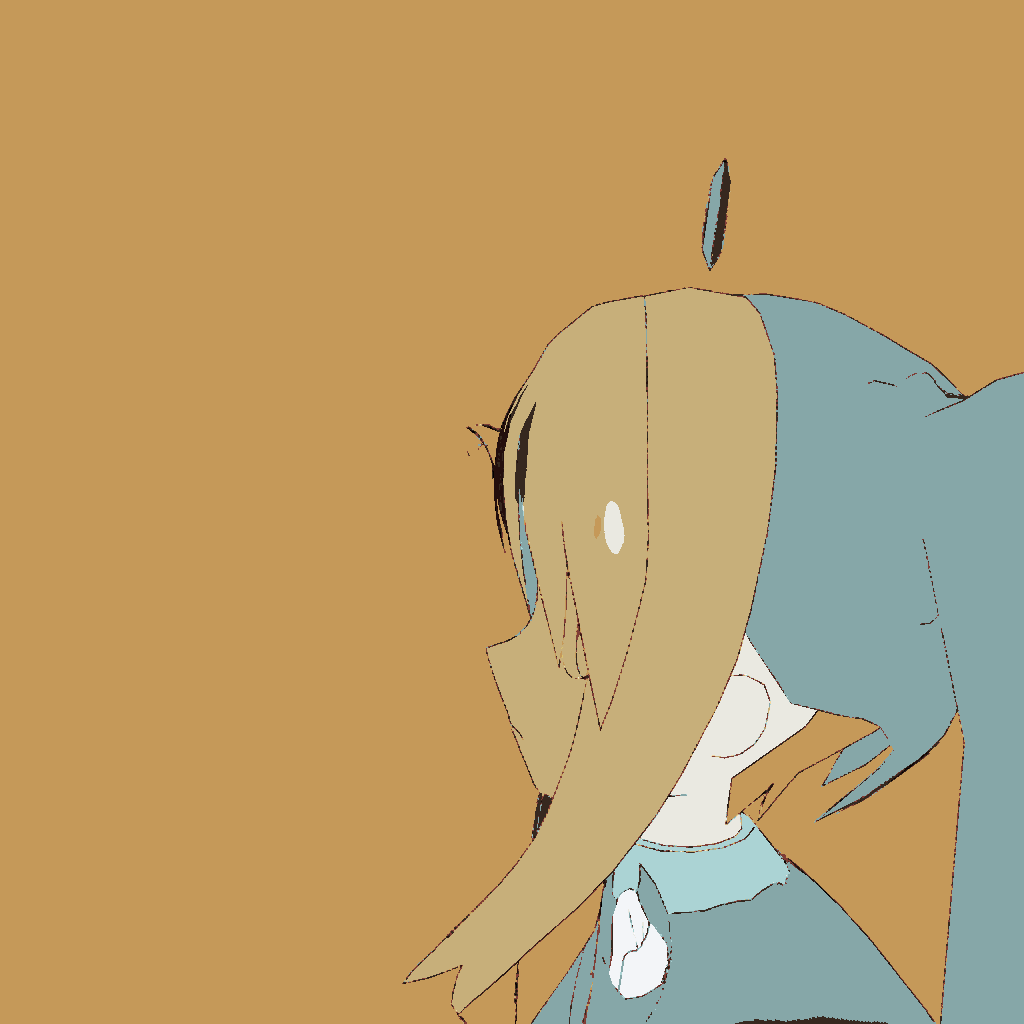} &
            \includegraphics[width=0.135\textwidth]{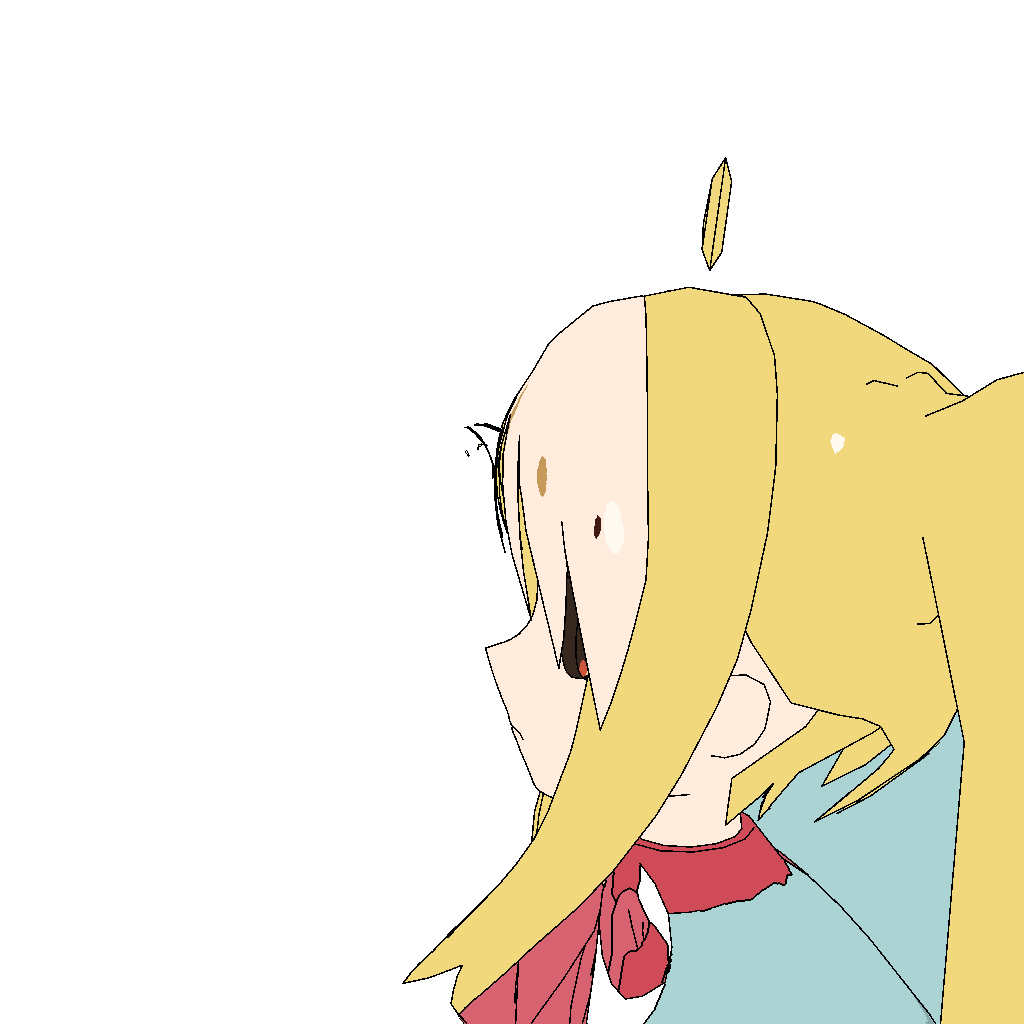} &
            \includegraphics[width=0.135\textwidth]{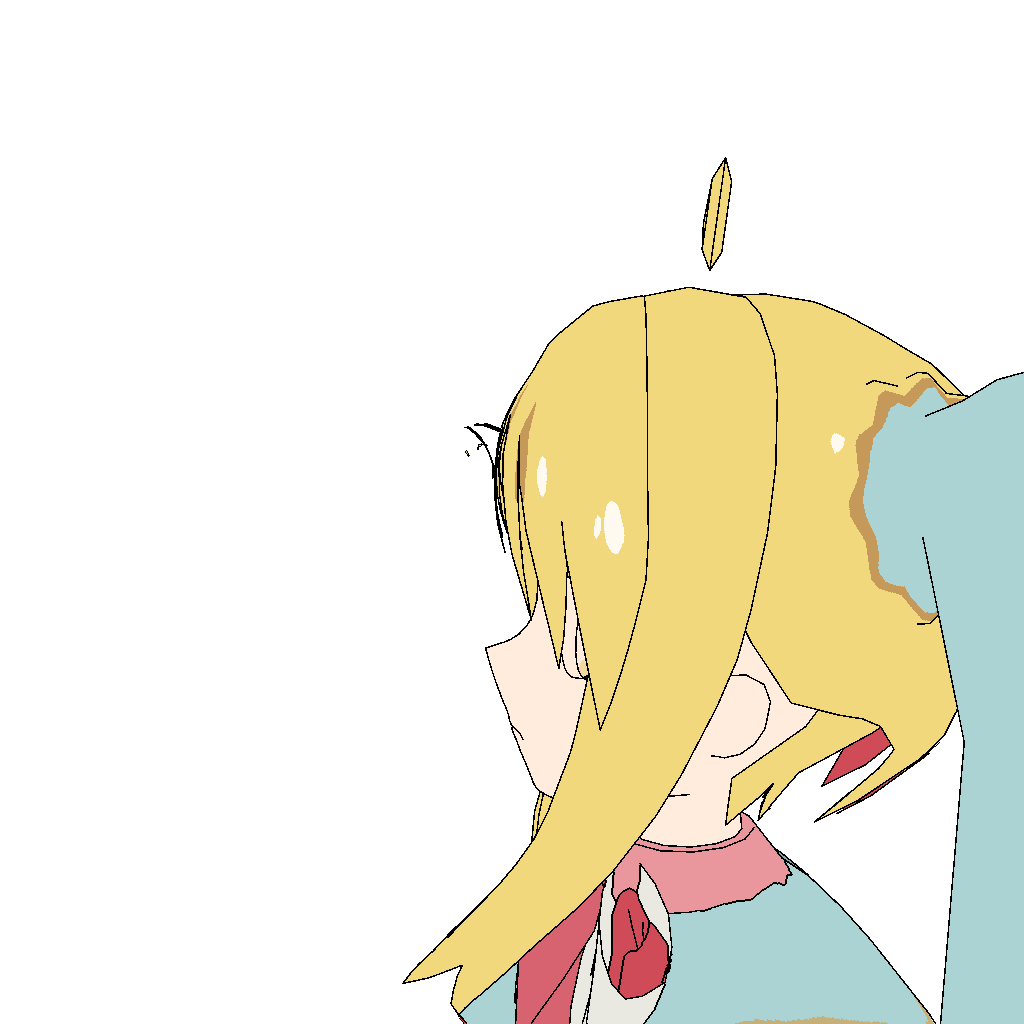} &
            \includegraphics[width=0.135\textwidth]{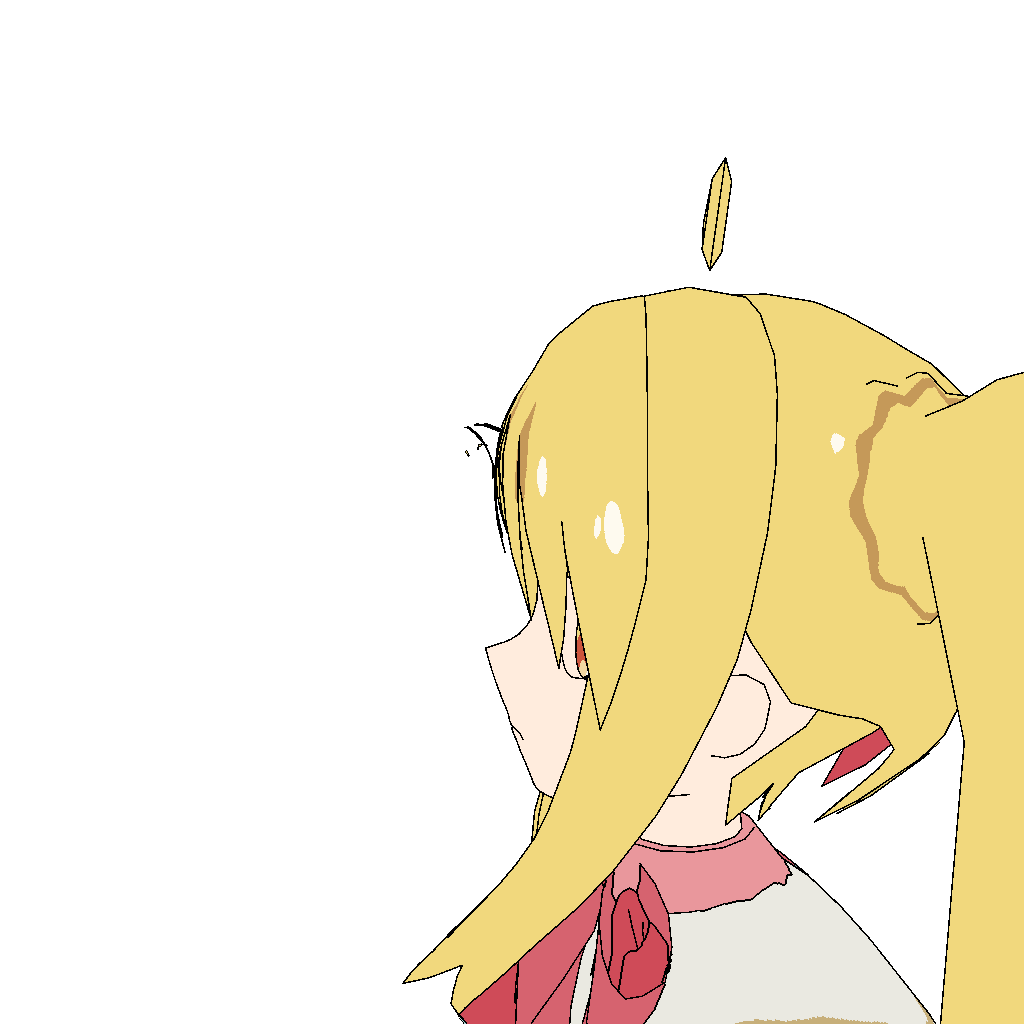} &
            \includegraphics[width=0.135\textwidth]{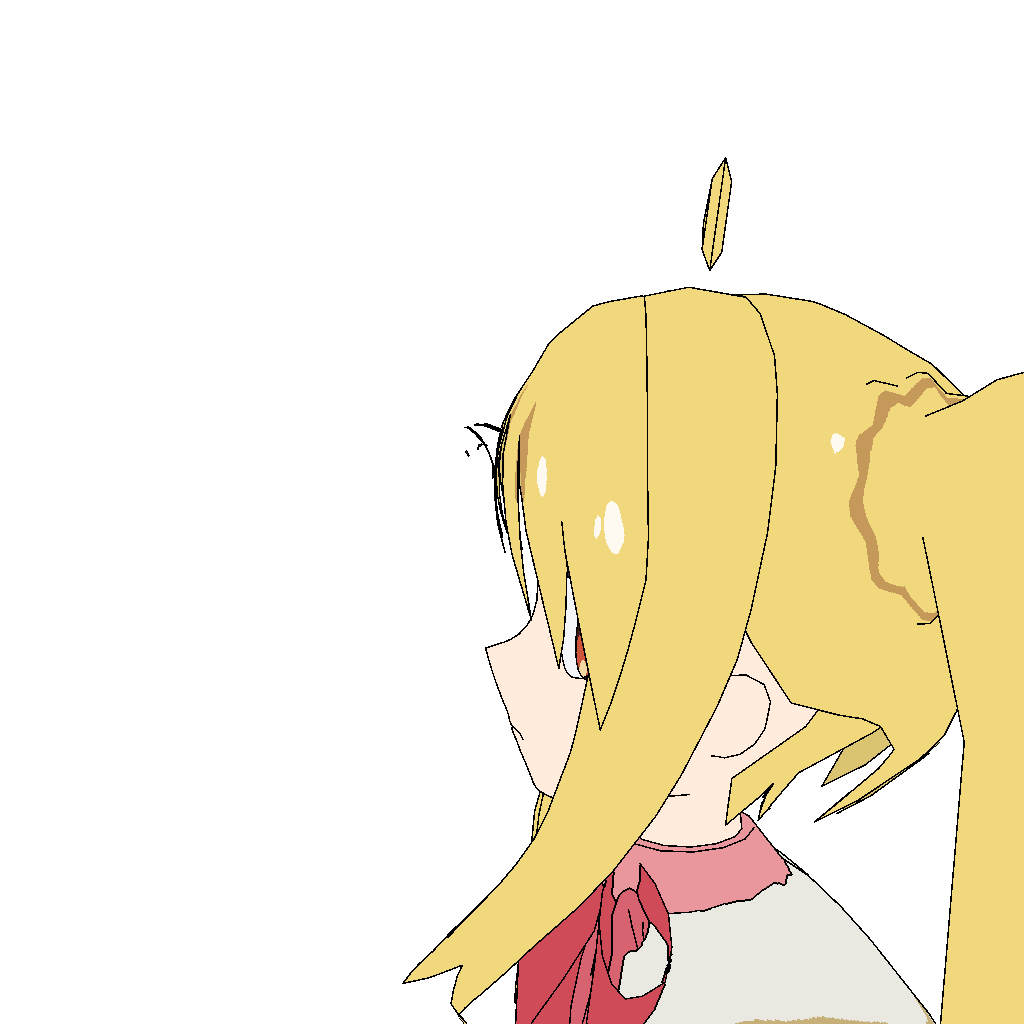} \\

            \includegraphics[width=0.135\textwidth]{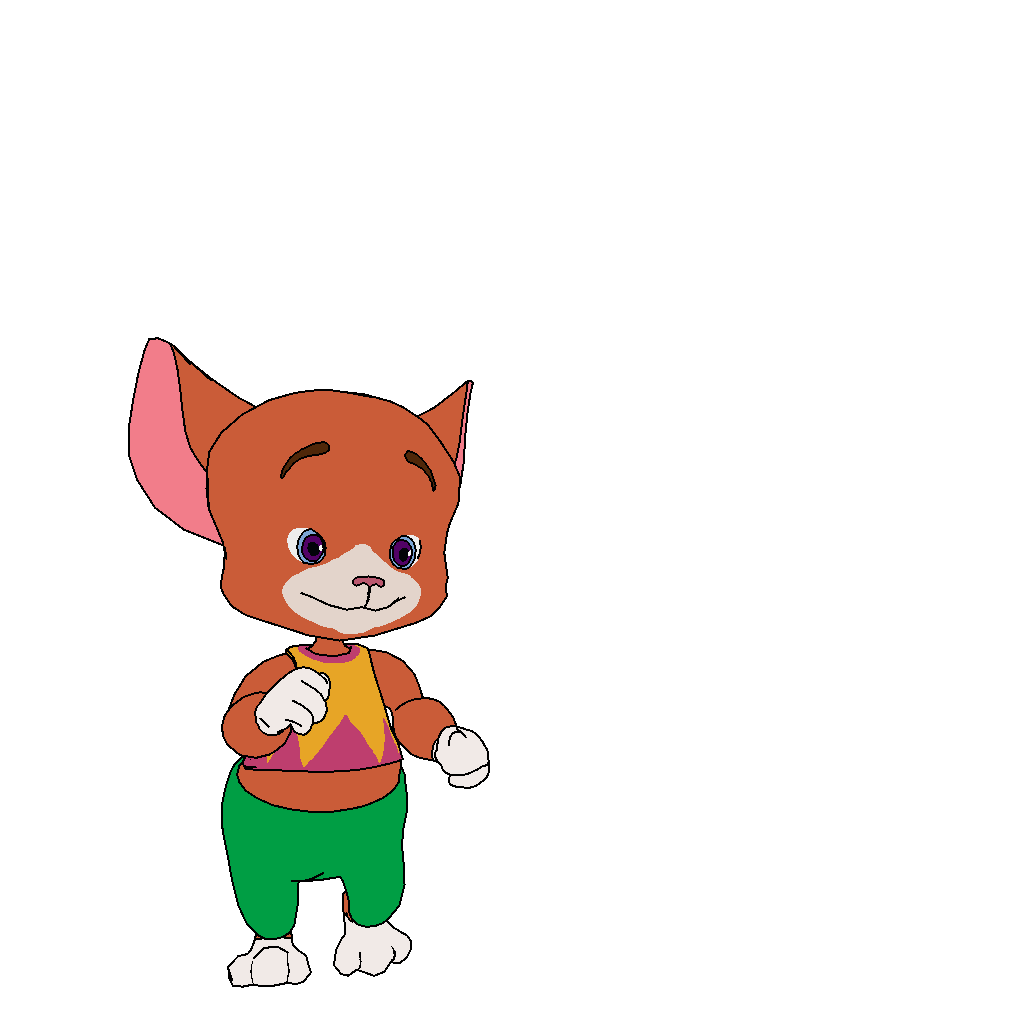} &
            \includegraphics[width=0.135\textwidth]{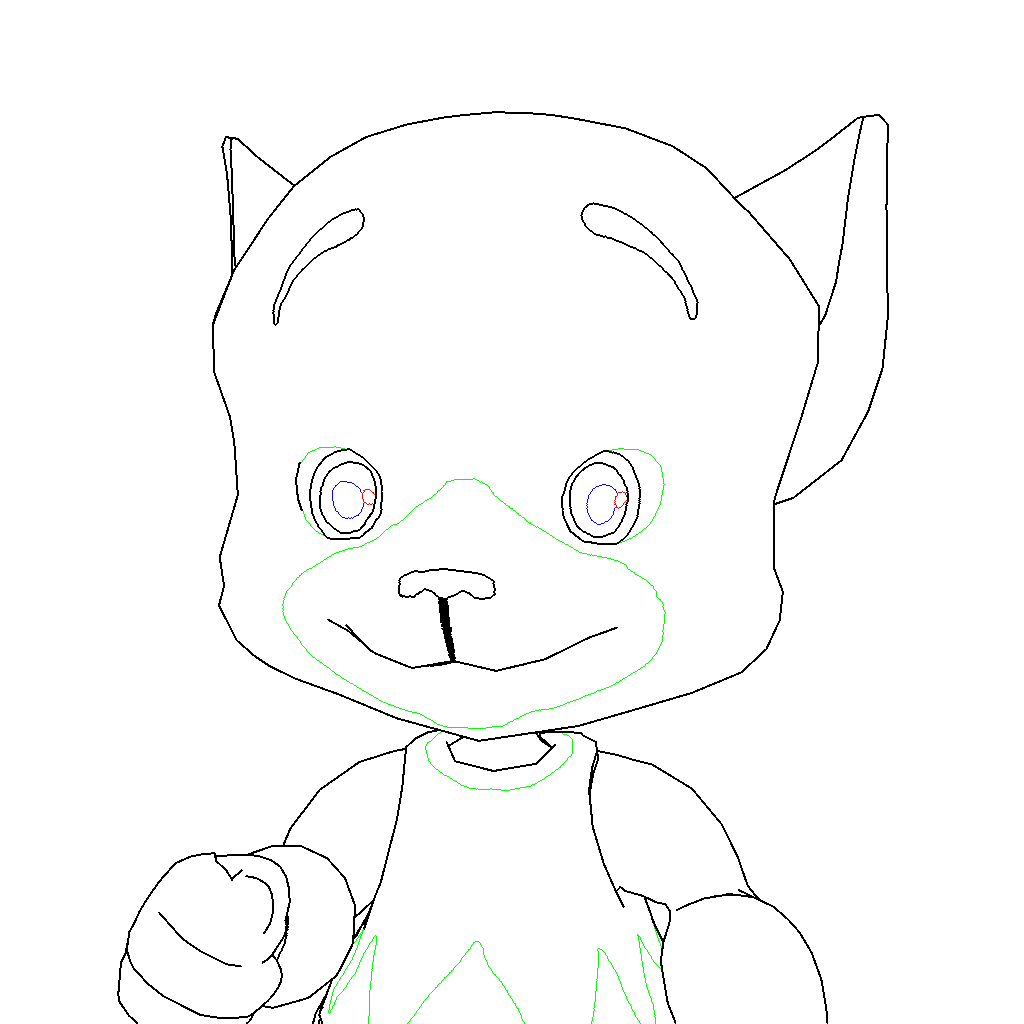} &
            \includegraphics[width=0.135\textwidth]{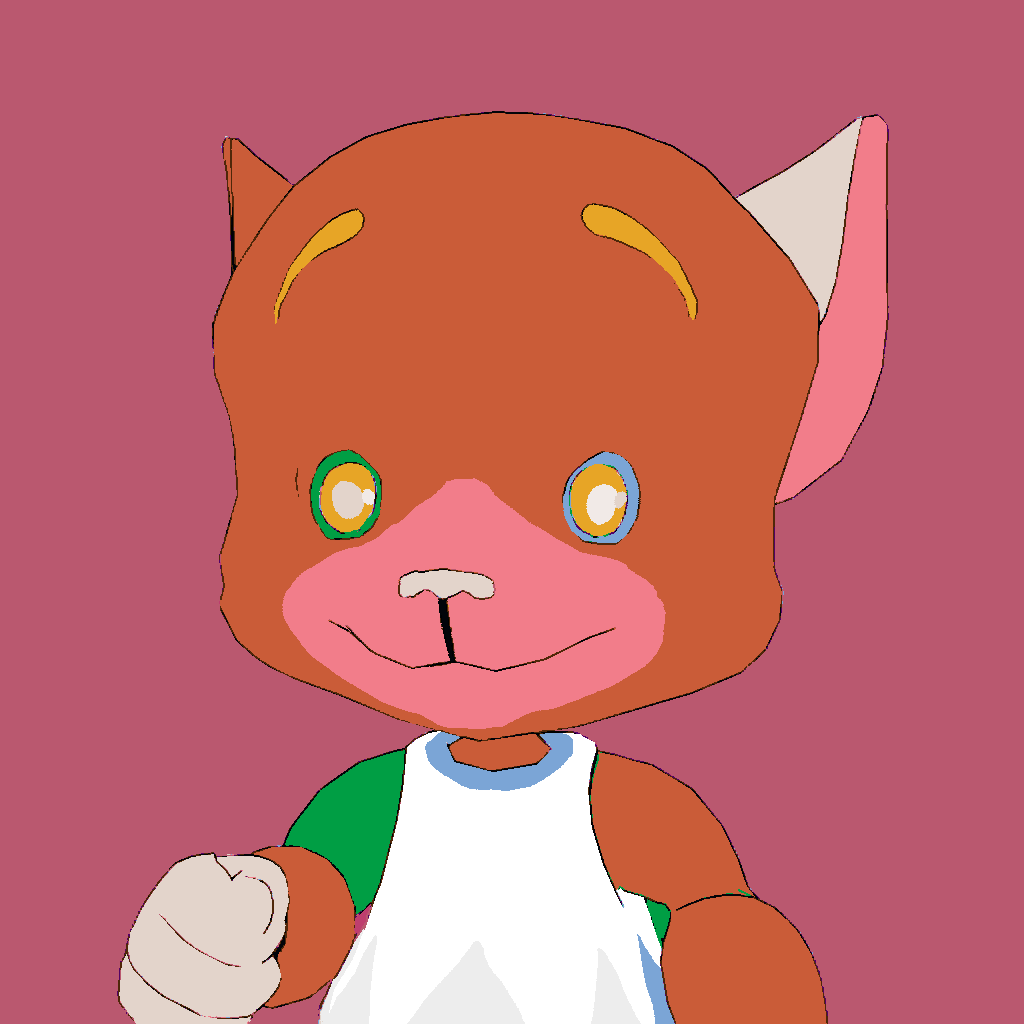} &
            \includegraphics[width=0.135\textwidth]{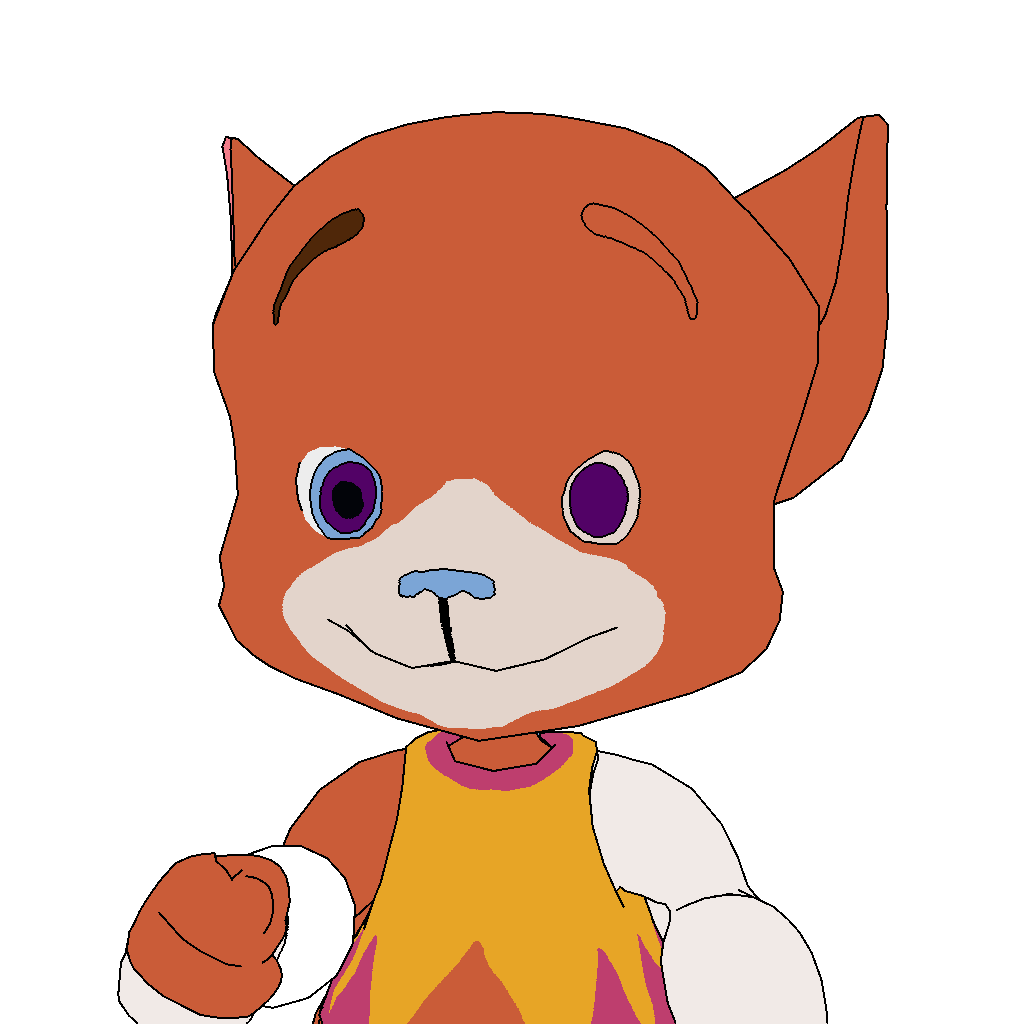} &
            \includegraphics[width=0.135\textwidth]{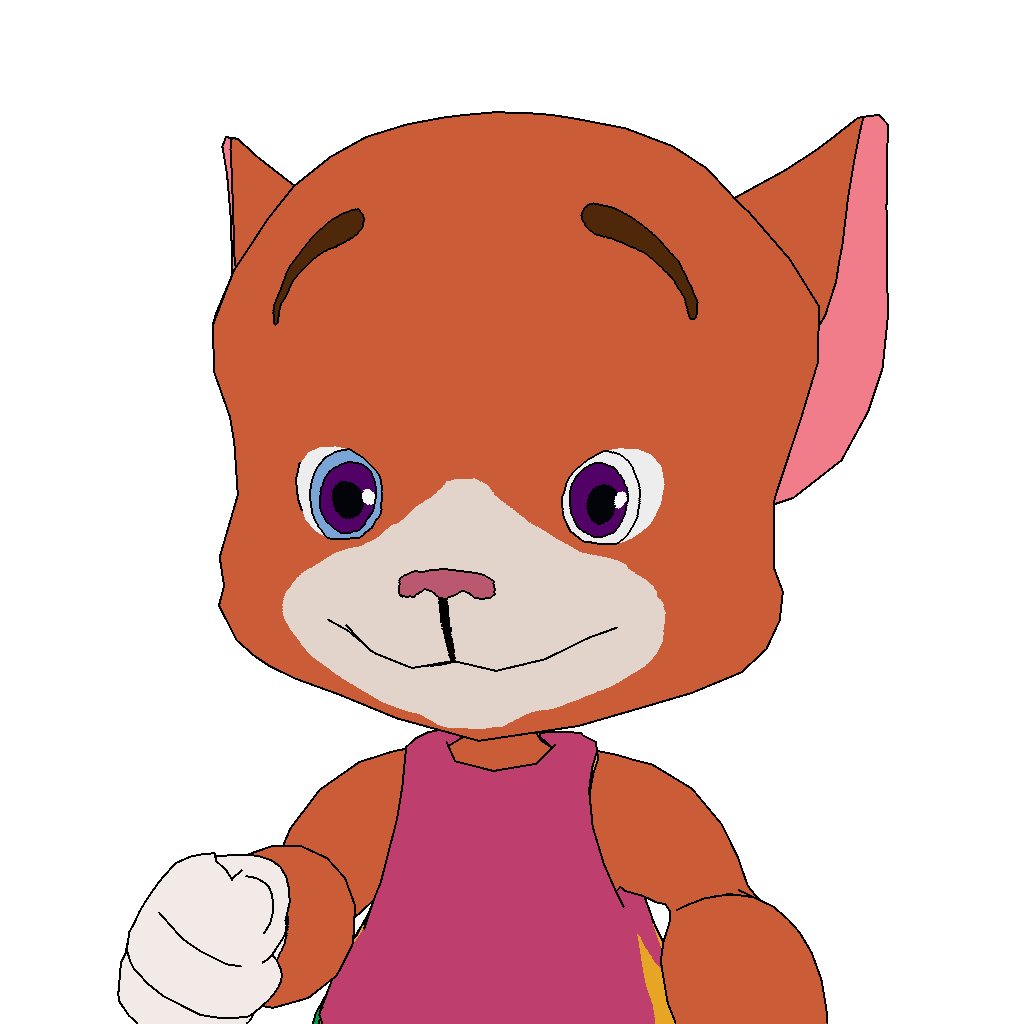} &
            \includegraphics[width=0.135\textwidth]{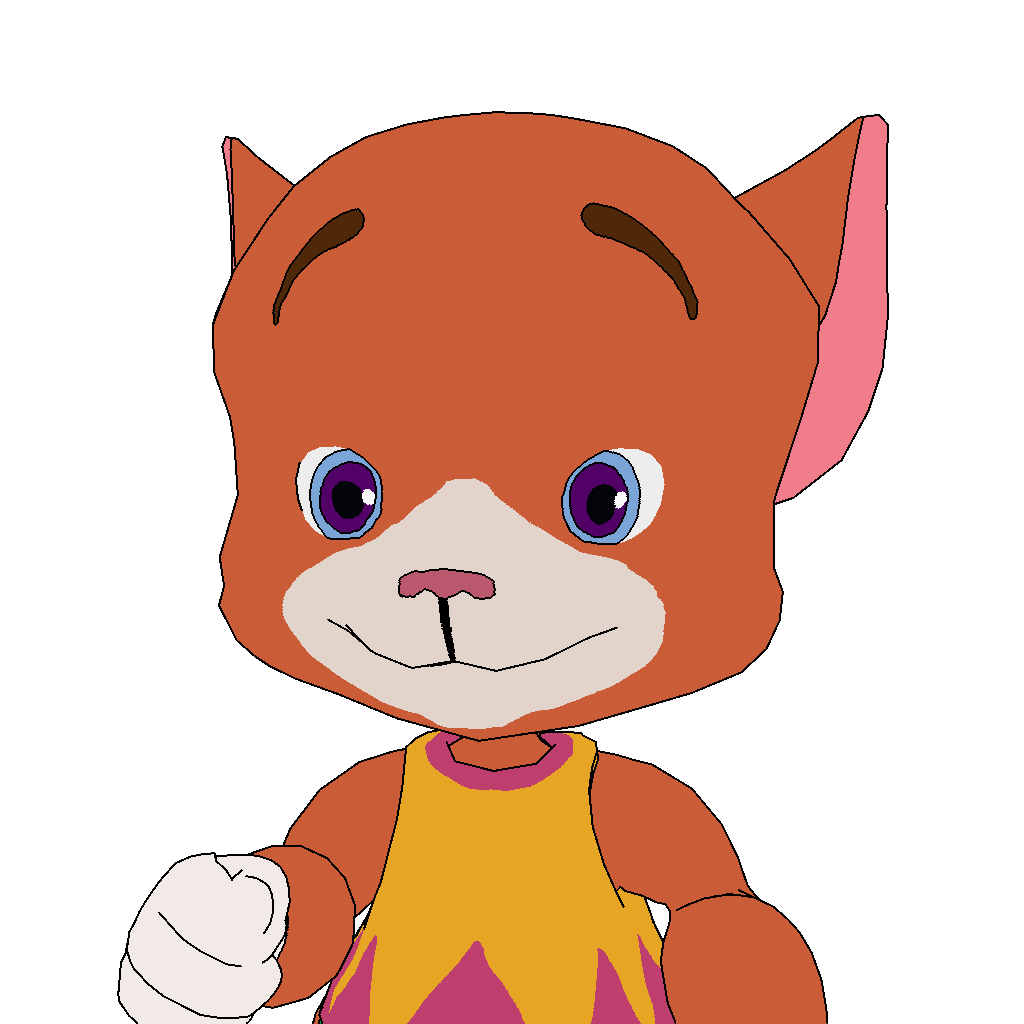} &
            \includegraphics[width=0.135\textwidth]{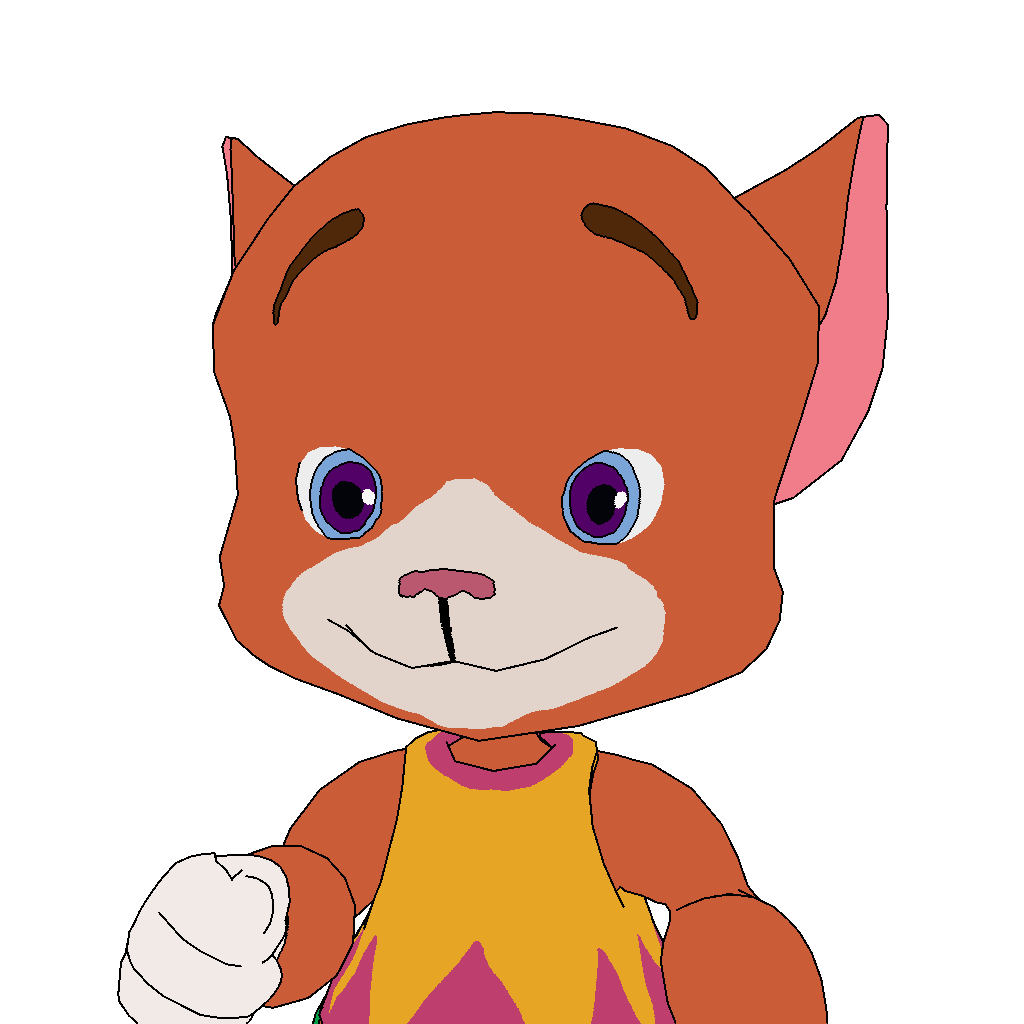} \\

            \includegraphics[width=0.135\textwidth]{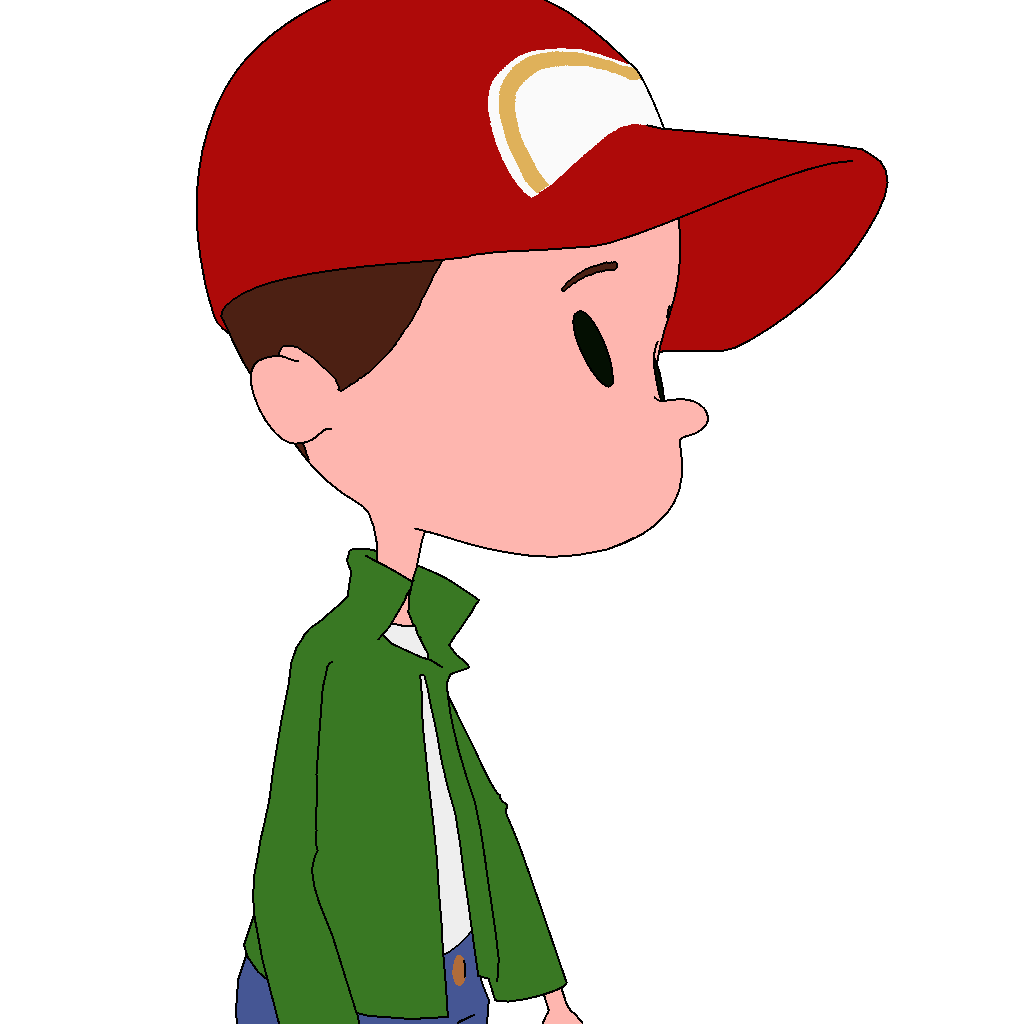} &
            \includegraphics[width=0.135\textwidth]{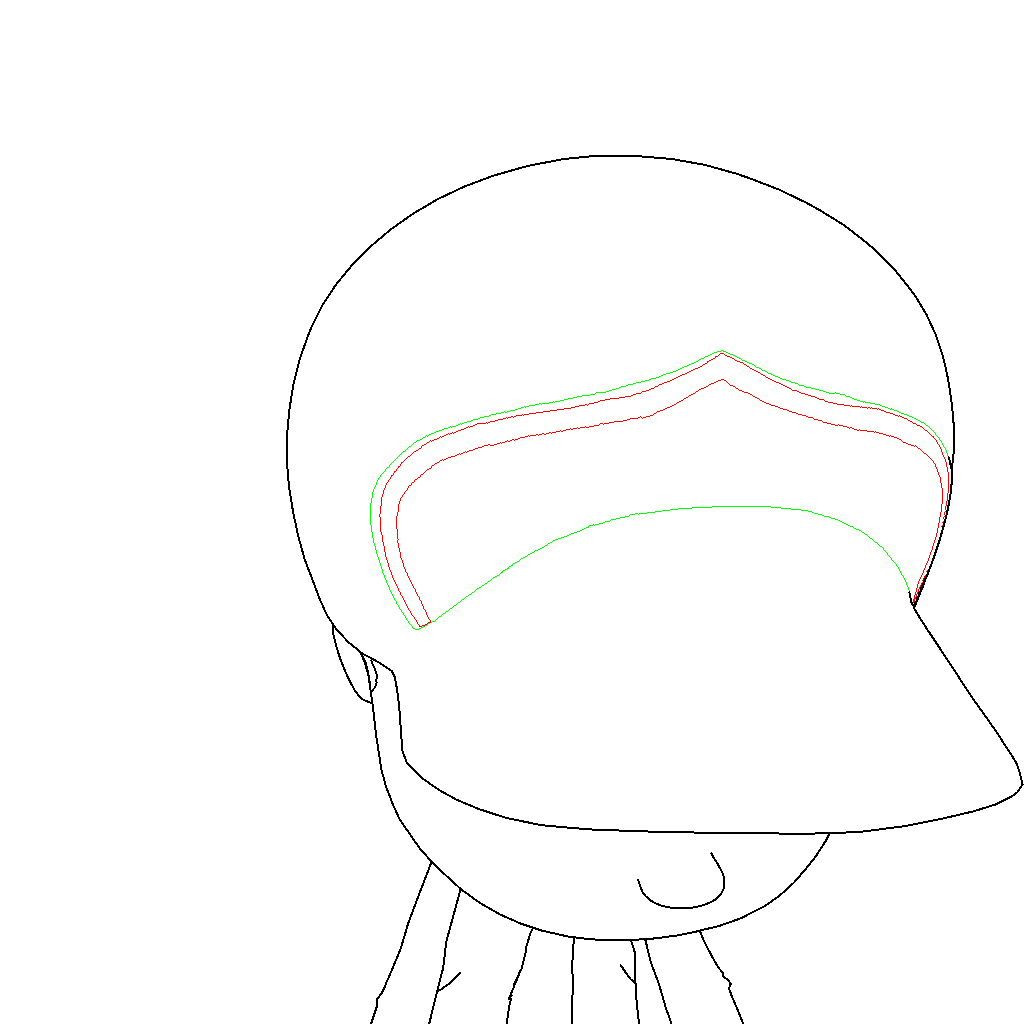} &
            \includegraphics[width=0.135\textwidth]{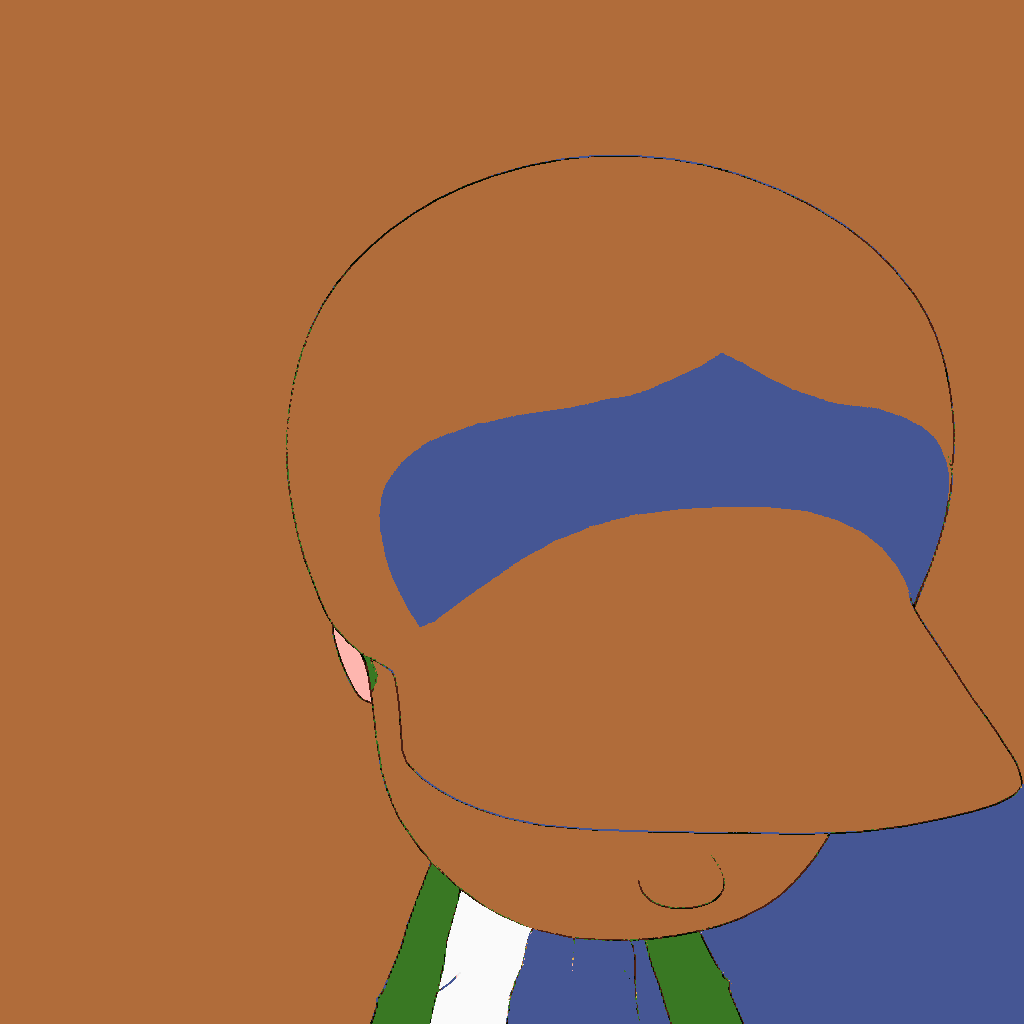} &
            \includegraphics[width=0.135\textwidth]{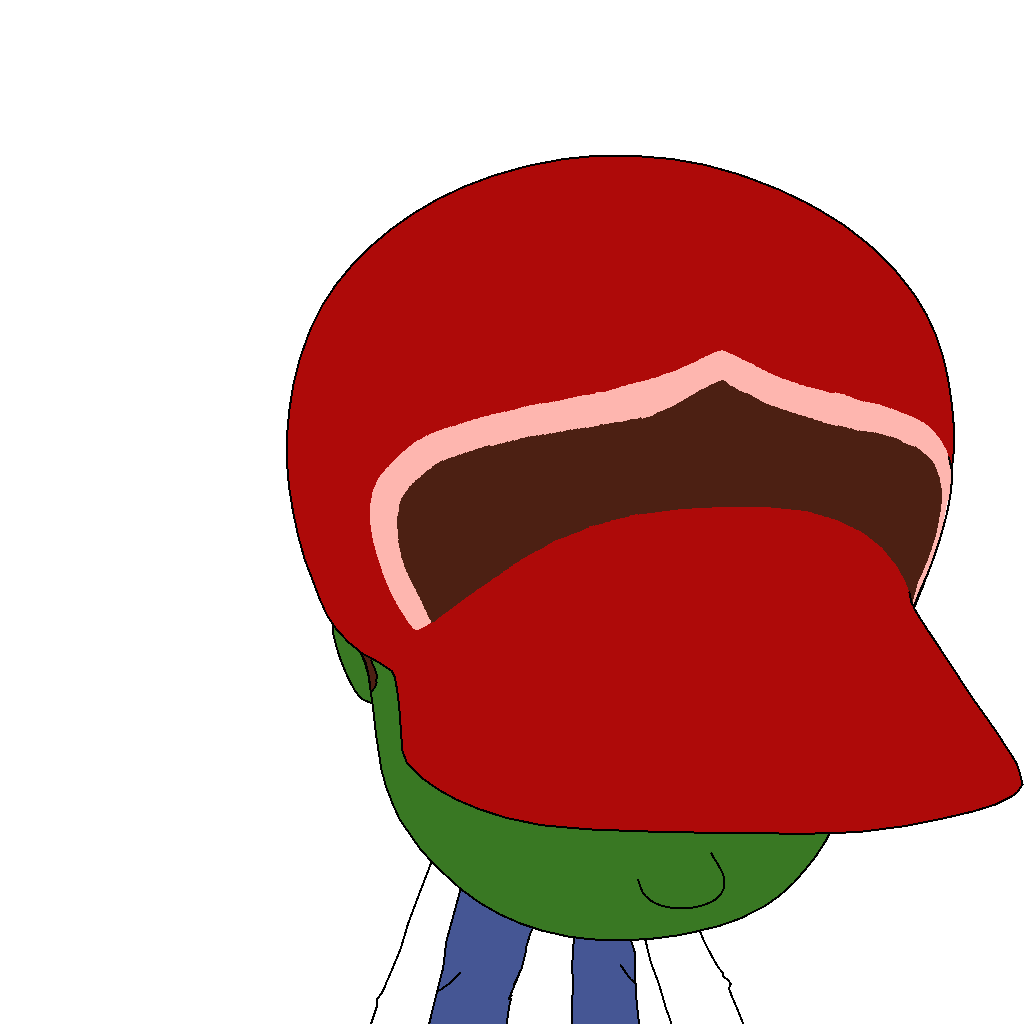} &
            \includegraphics[width=0.135\textwidth]{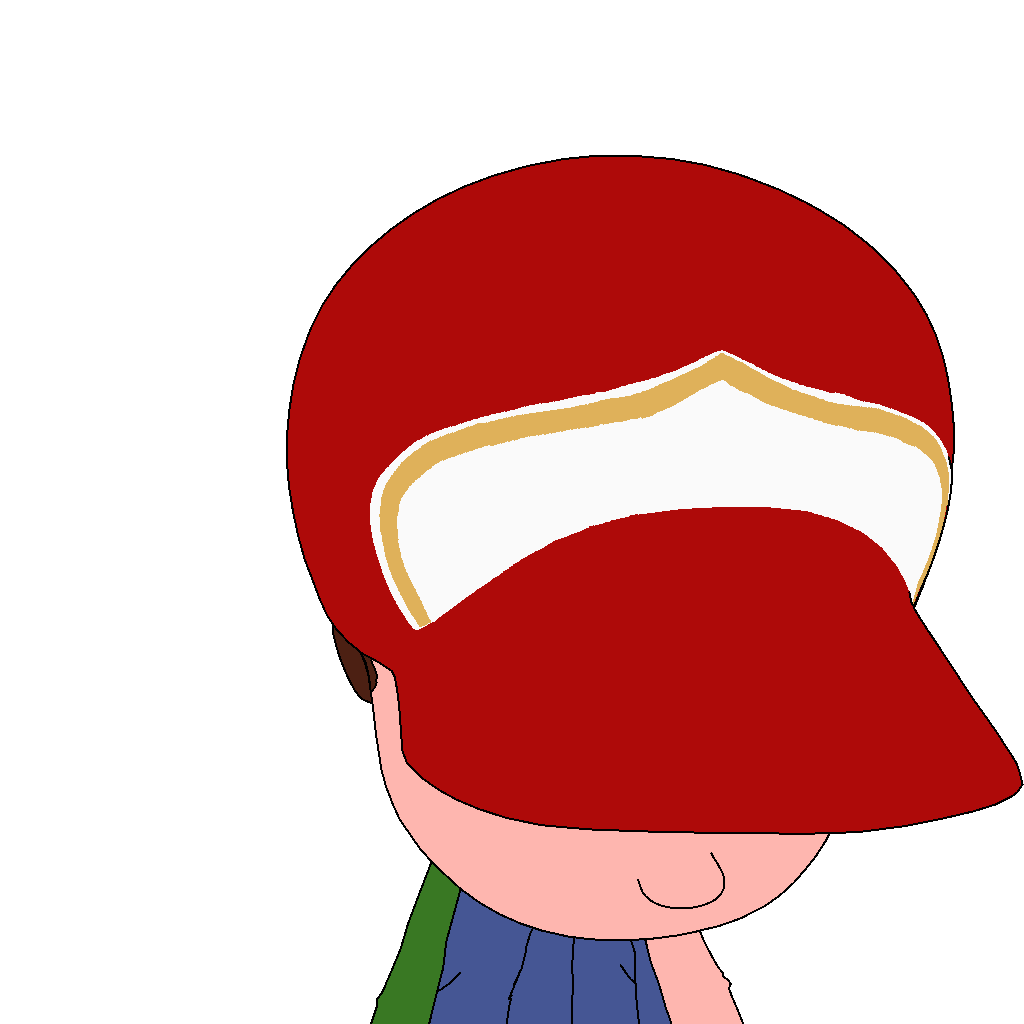} &
            \includegraphics[width=0.135\textwidth]{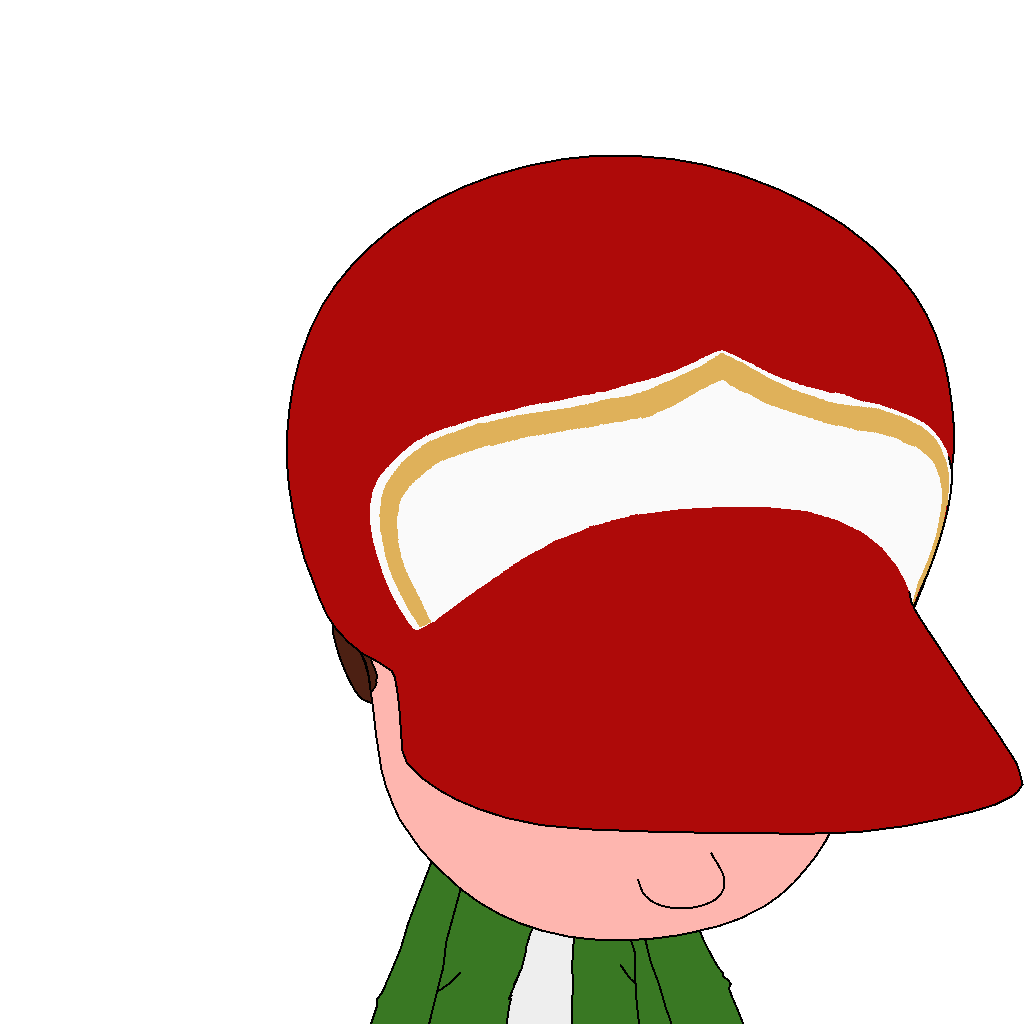} &
            \includegraphics[width=0.135\textwidth]{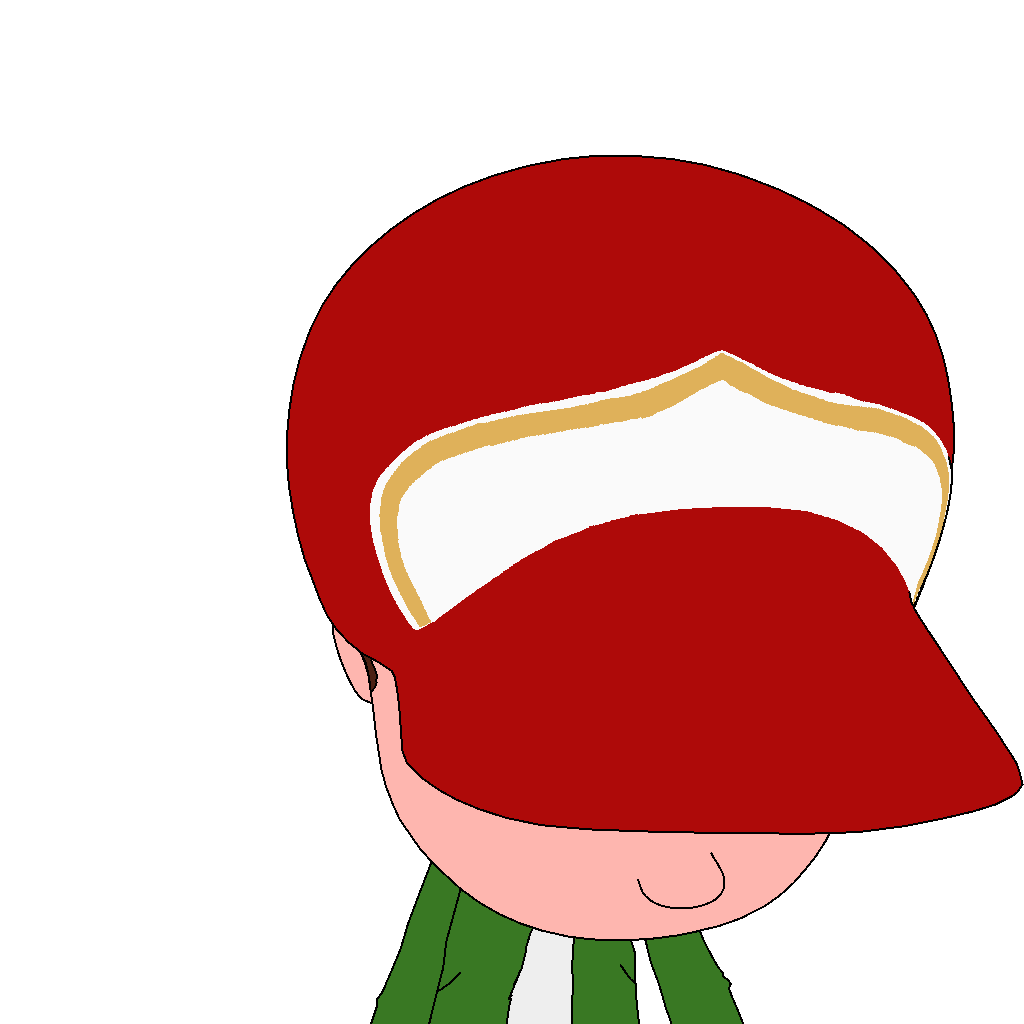} \\

            \includegraphics[width=0.135\textwidth]{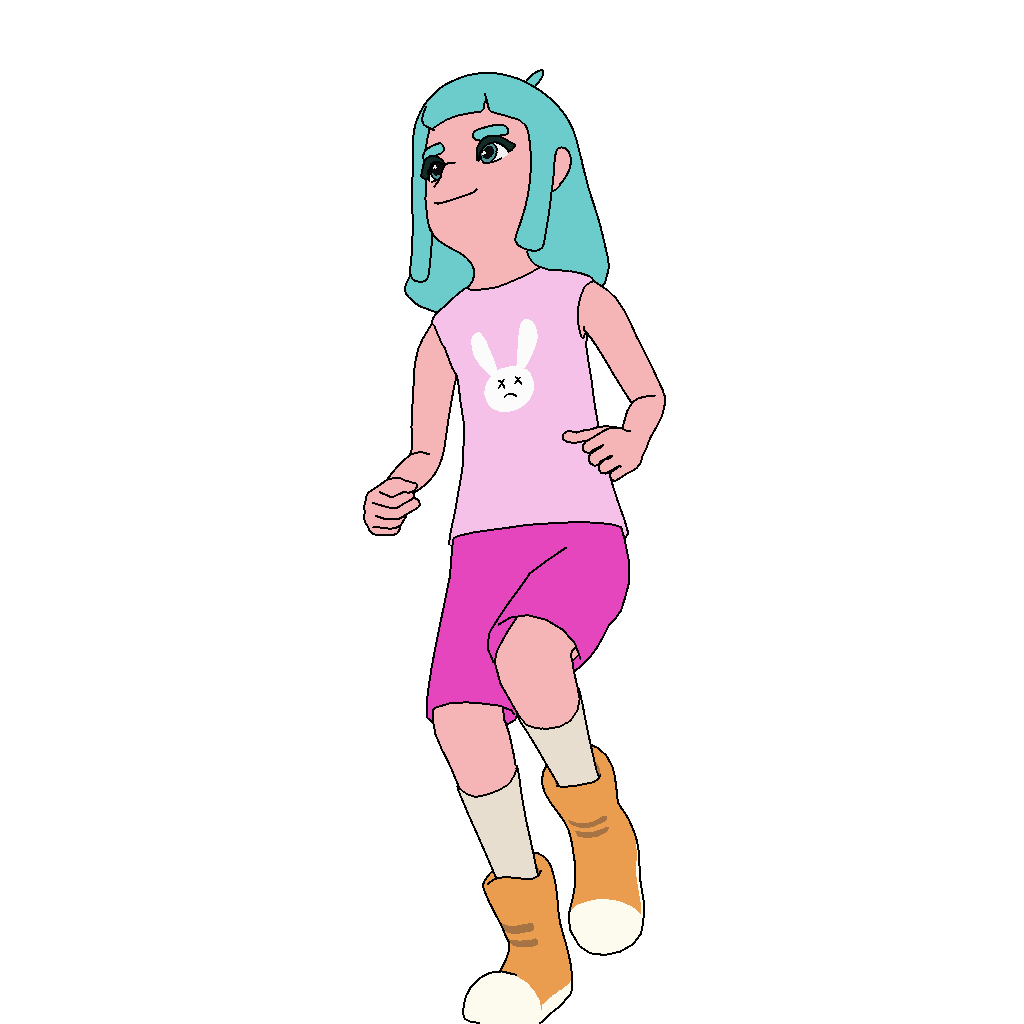} &
            \includegraphics[width=0.135\textwidth]{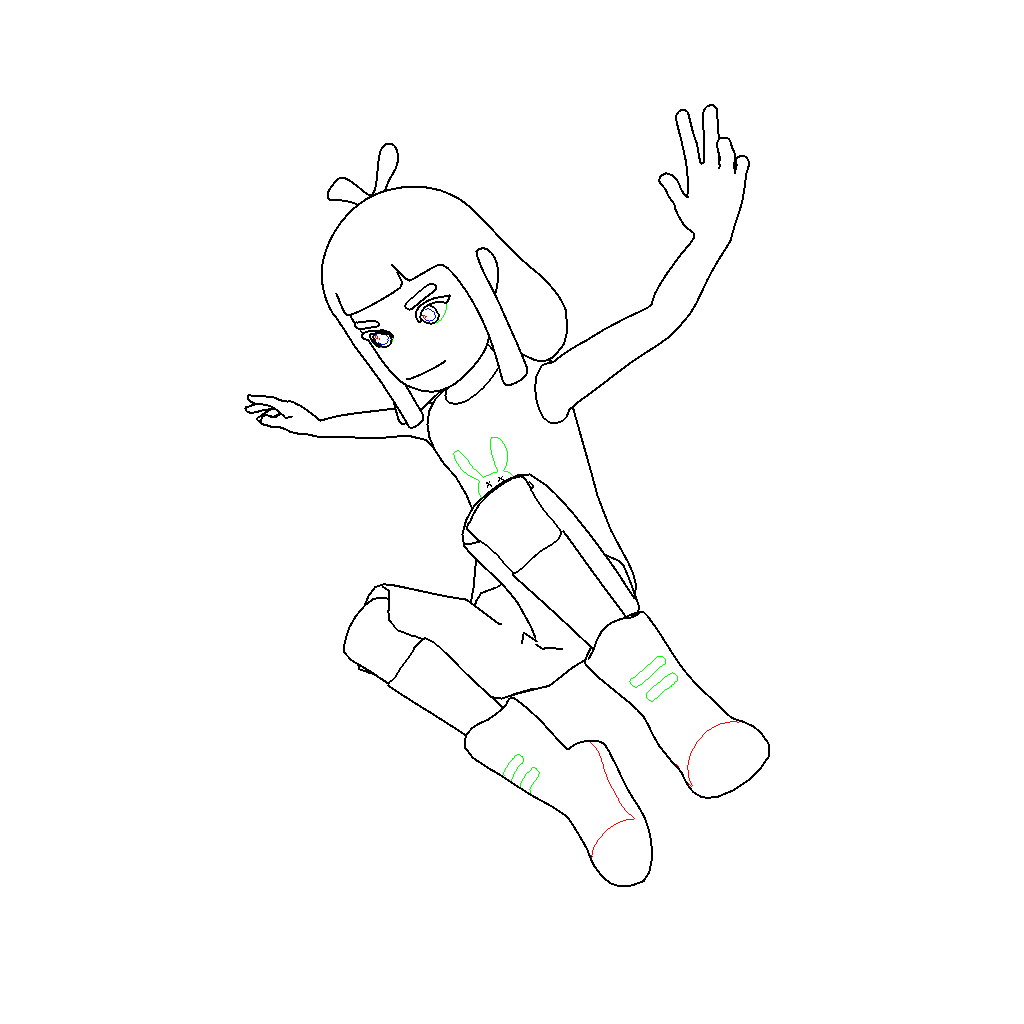} &
            \includegraphics[width=0.135\textwidth]{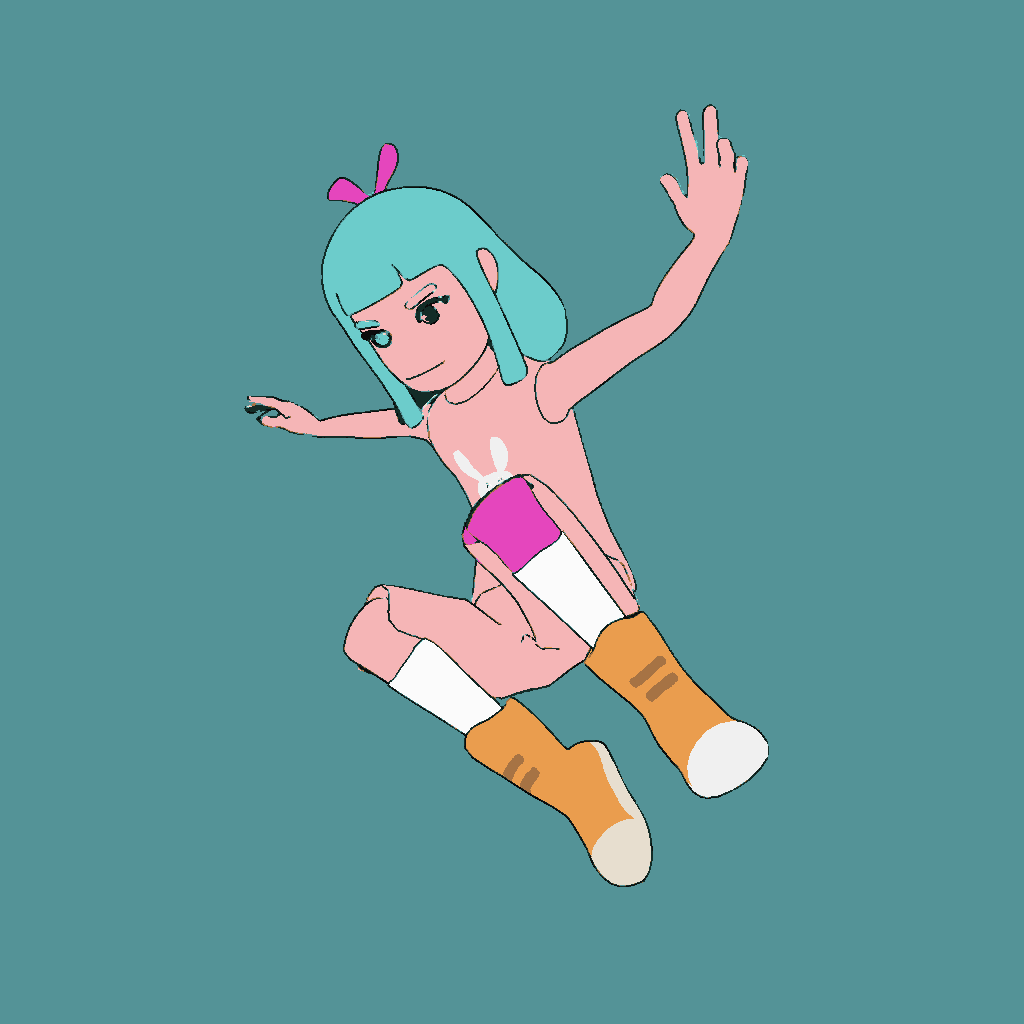} &
            \includegraphics[width=0.135\textwidth]{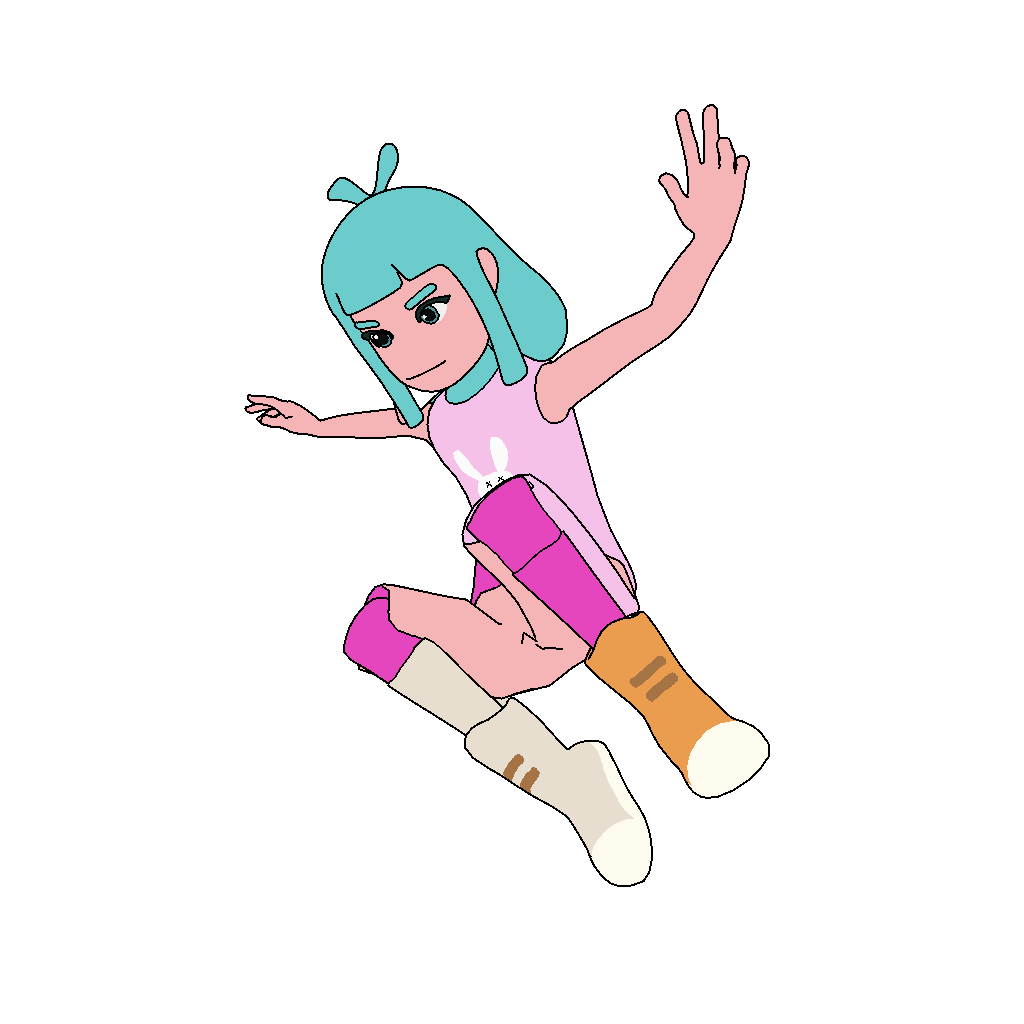} &
            \includegraphics[width=0.135\textwidth]{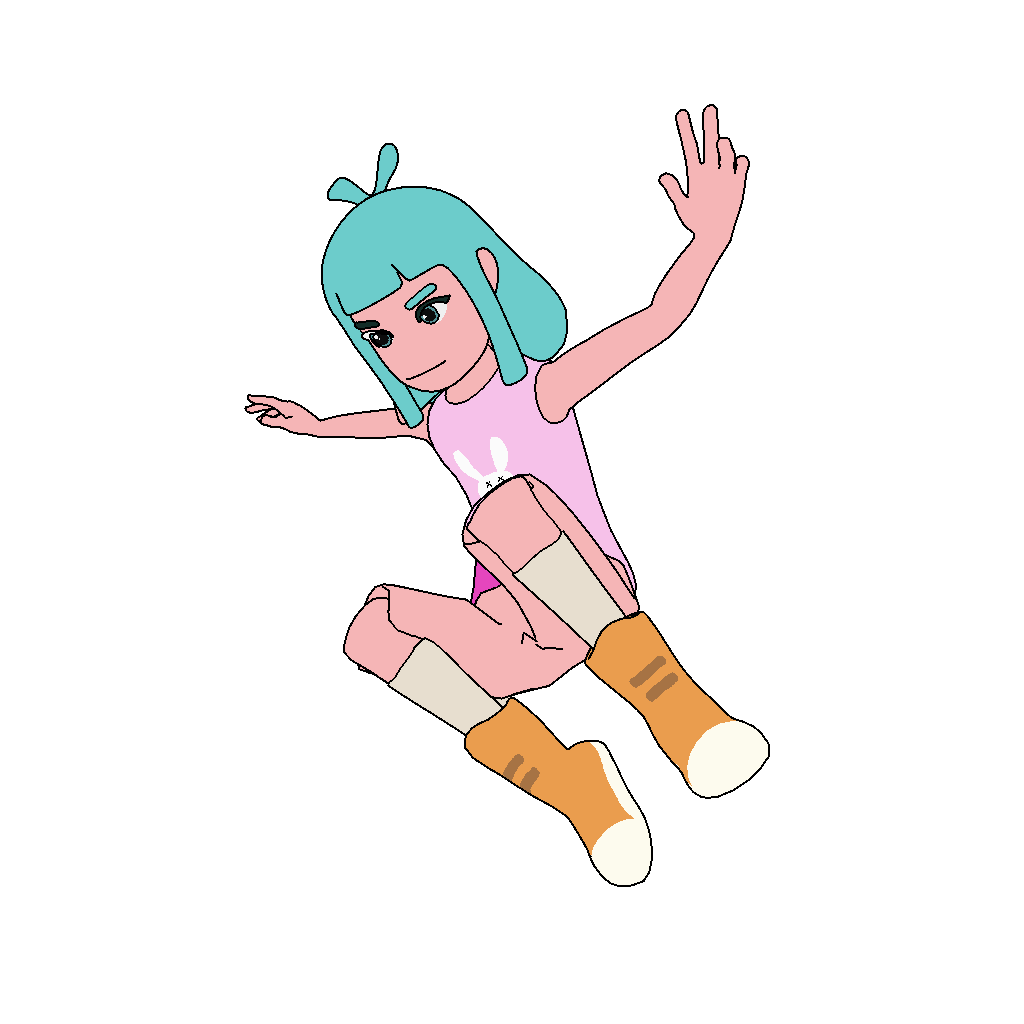} &
            \includegraphics[width=0.135\textwidth]{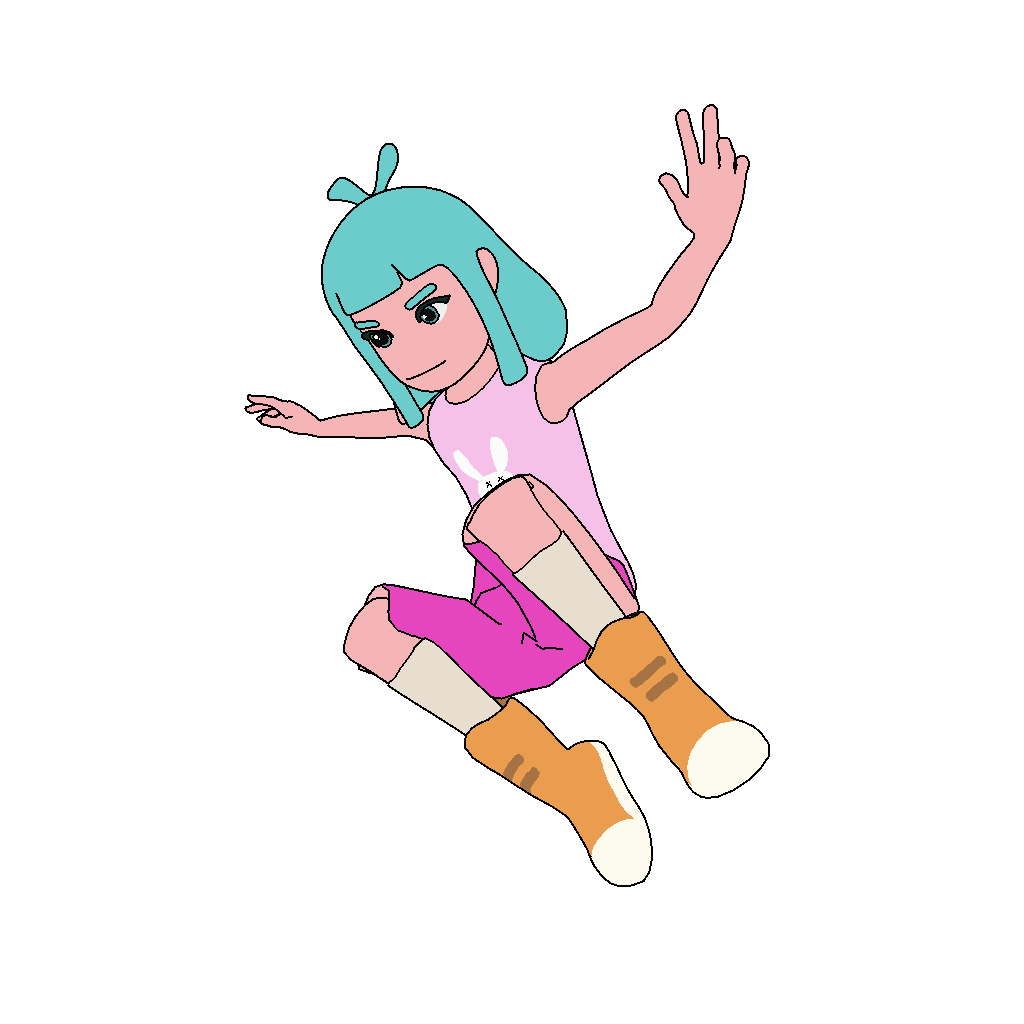} &
            \includegraphics[width=0.135\textwidth]{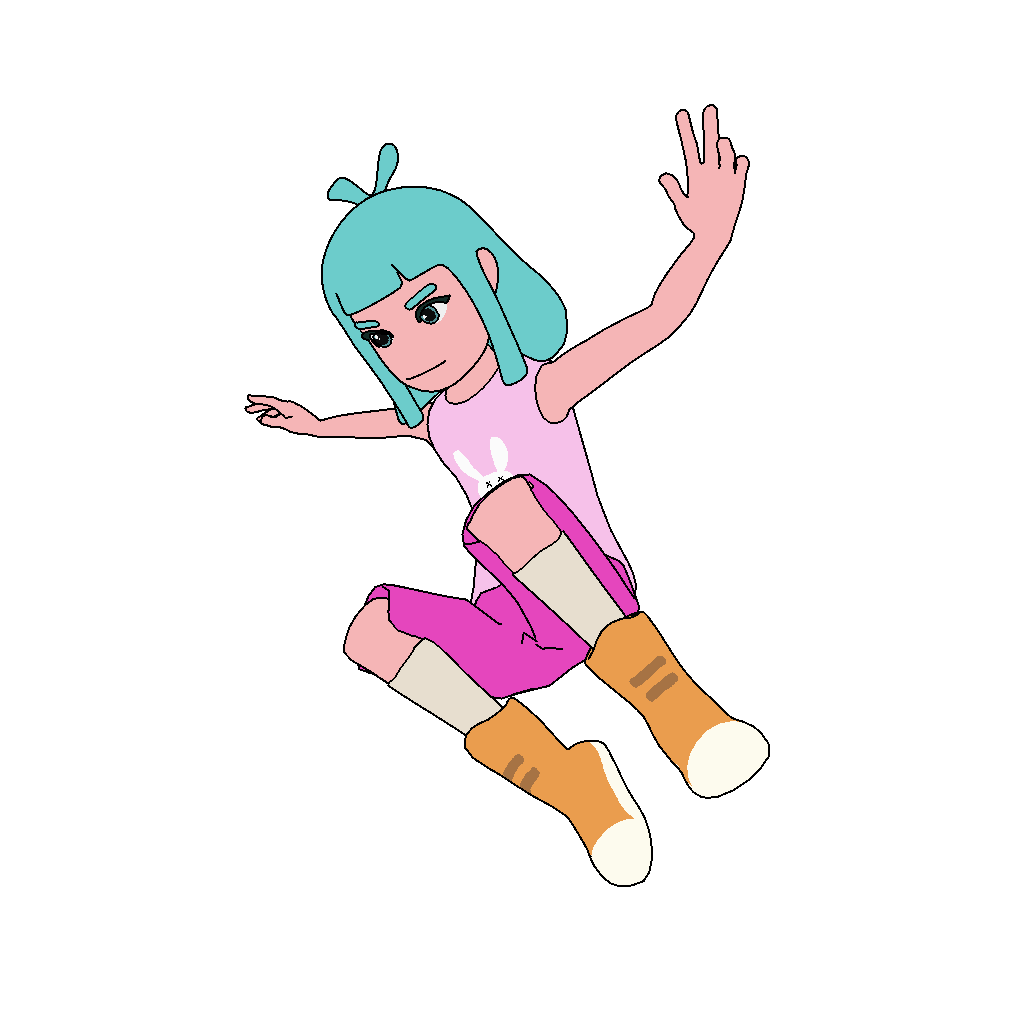} \\           
        \end{tabular}%
        }
        \caption{First-frame Reference Colourisation Results}
        \label{fig:more_qual_continuous}
    \end{subfigure}

    \caption{\textbf{Additional qualitative comparison under different references.}
    We show the one-shot reference and target line sketch, followed by the results from
    \textbf{MangaNinja}~\cite{liu2025manganinja},
    \textbf{BasicPBC(-Ref)}~\cite{dai2024learning,dai2024paint},
    \textbf{DACoN~1.1}~\cite{nagata2025dacon},
    our \textbf{\textsc{PeCA}} on DACoN~1.1, and colour ground-truth (right).}
    \label{fig:more_qual_merged}
\end{figure*}

We provide further qualitative results on various settings in separate figures below. Specifically, we visualise a more comprehensive set of test samples under different methods and references in \cref{fig:more_qual_merged}. Note that for the pixel-generative method \cite{liu2025manganinja}, we follow the post-processing steps in previous work \cite{nagata2025dacon} to convert the raw results to palette-preserving paint-bucket colourisations. In general, our method shows superior performance under challenging test cases.

\section{Limitations}
\label{sec:limitations}
Like previous paint-bucket colourisation methods \cite{dai2024learning,dai2024paint,Feng_2025_ICCV,nagata2025dacon}, \textsc{PeCA} expects line sketches with sufficiently enclosed regions, so that simple flood-fill segmentation~\cite{10.1145/800249.807456} can be applied. This assumption is consistent with standard animation production workflows~\cite{nakanishi2013modeling, tang2025generative} and is therefore largely inherited from the paint-bucket formulation rather than introduced by our method. When applied to raw drafts or amateur sketches with broken strokes and leakage, region extraction may fail and subsequently affect matching and colour assignment. 

\begin{figure}[t]
    \centering
    \includegraphics[width=1\linewidth]{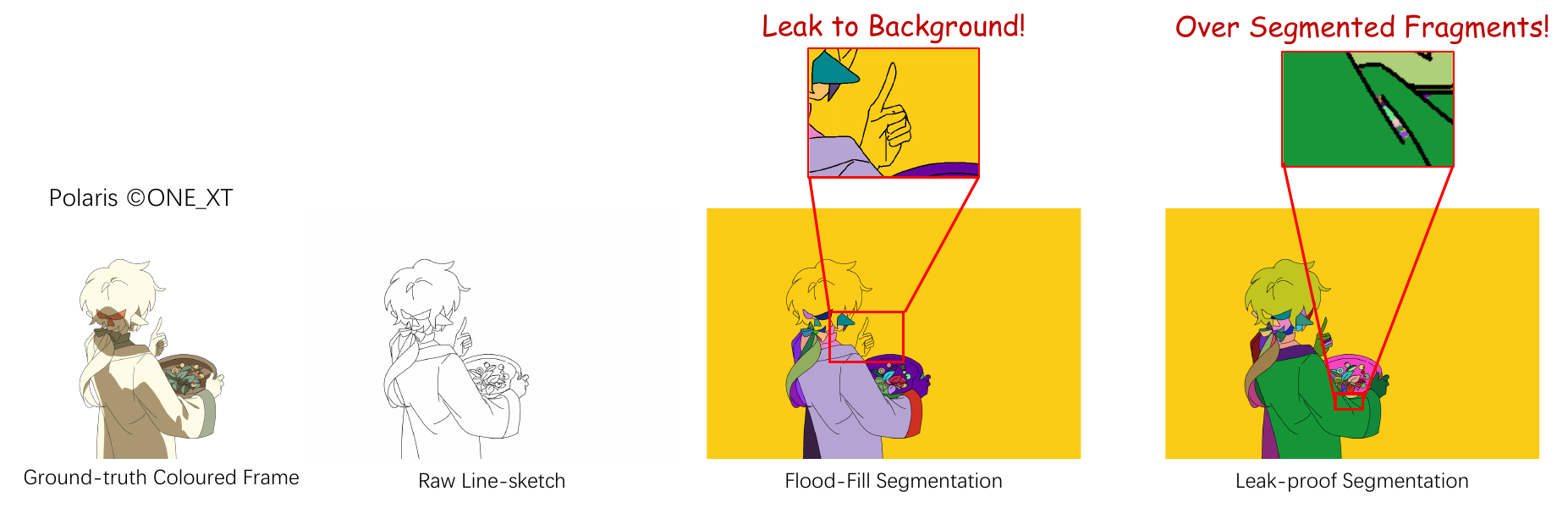}
    \caption{\textbf{Segmentation failure modes on amateur-level line sketches from \cite{Anita2024}.}
    From left to right: coloured reference, raw sketch, standard flood-fill \cite{10.1145/800249.807456} segmentation, and leakage-robust segmentation. Standard flood-fill may leak into the background when strokes are not fully closed, while leakage-robust segmentation reduces leakage, but often over-segments the drawing into fragments. These results highlight the gaps between amateur-level and production-ready line sketches that paint-bucket colourisation task formulations \cite{dai2024learning, dai2024paint} and industry-level animation workflows \cite{nakanishi2013modeling} assume.}
    \label{fig:seg_fail}
\end{figure}

Possible ways to relax this assumption include draft-line gap closing~\cite{sasaki2017joint, xu2022deep} and leakage-robust region segmentation~\cite{10.1145/3681758.3698003, DanbooRegion2020}. The latter preserves the original drawing and therefore has been preferred as additional lines may break the original structure of the target animation. But it often produces more fragmented regions, which can make matching less stable and increase manual colourisation workload (with many more fragments to colour) as shown in \cref{fig:seg_fail}. Bridging the gap between raw or amateur-level sketches and production-ready line art is therefore a promising and largely orthogonal direction for future work \cite{guajardo2024generative}.

\begin{figure}[t]
    \centering
    \includegraphics[width=0.75\linewidth]{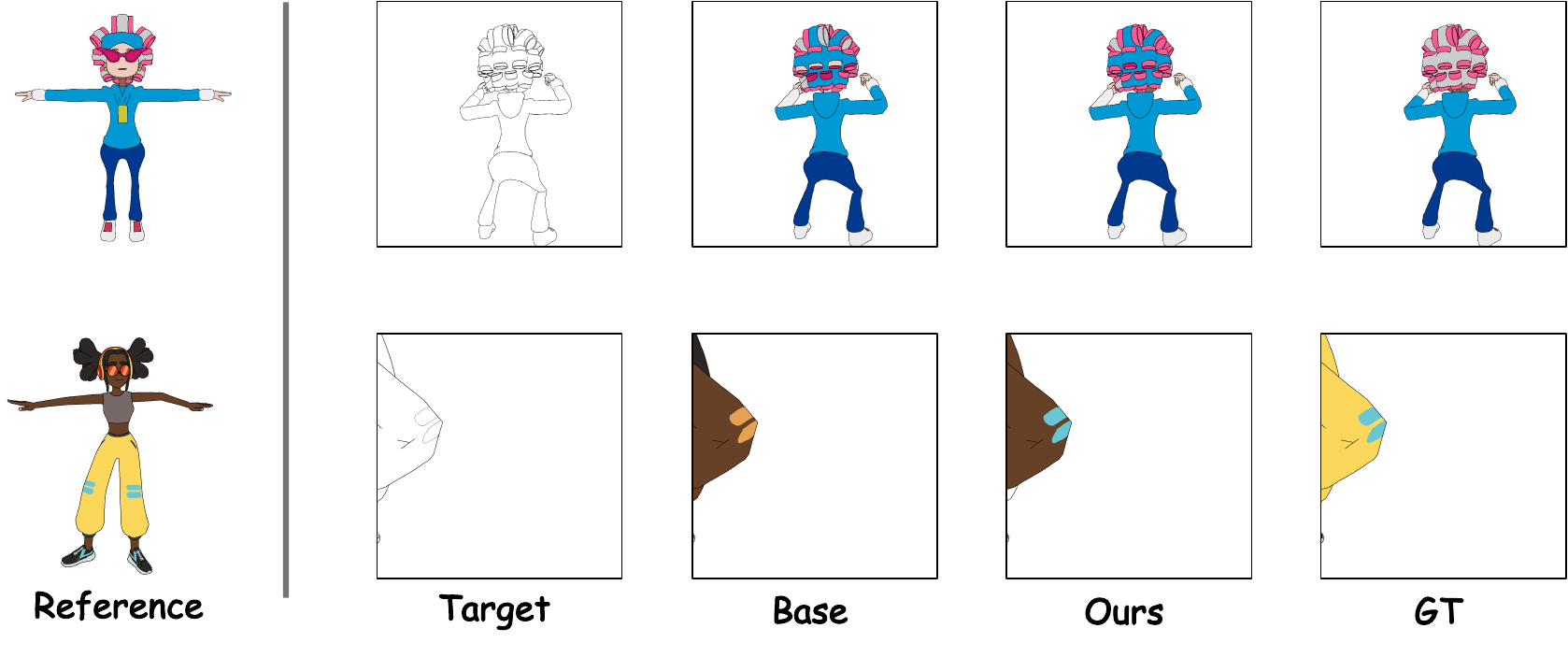}
    \caption{\textbf{Failure cases caused by incomplete reference coverage.}
    From left to right: reference, target line sketch, \textit{Base} and \textsc{PeCA} predictions, and ground truth. Top: missing reference coverage for the target pose leads to incorrect colour assignment on the helmet back regions. Bottom: a very small visible part in the target frame lacks sufficient colour evidence in the reference, resulting in local colour confusion.}
    \label{fig:failure2}
\end{figure}

Another limitation comes from incomplete reference coverage. Like other reference-guided colourisation methods~\cite{dai2024learning,dai2024paint,Feng_2025_ICCV,nagata2025dacon}, \textsc{PeCA} can only propagate colours that are represented in the available references. Missing views, colours, or part appearances may lead to systematic colour confusion as shown in \cref{fig:failure2}. Accordingly, the geometric transformations in \textsc{PeCA} should be understood as lightweight in-plane spatial support, which further reduces moderate pose or layout gaps, but do not synthesise out-of-plane 3D rotations or colours for surfaces never observed in the reference pool.

A possible practical direction is a more interactive reference-selection workflow that allows dynamic reference growth (as an online setting) in paint-bucket colourisation process: instead of simply increasing the number of manually coloured keyframes (which shifts the workload back to the user and thus reduces the benefit of automation), artists may choose to colour a small subset of the most informative keyframes or design-sheet views for downstream automatic colourisation. Because \textsc{PeCA} selects and reweights from the supplied reference pool, it can also benefit from expanded pools provided by artist interaction \cite{nakanishi2013modeling}, external knowledge base/prompt retrieval \cite{zhang2026omnicolorunifiedframeworkmultimodal, liu2026pixels}, or generated novel-view references \cite{peng2024charactergen, liu2023zero} without changing the paint-bucket output interface. An efficient selection strategy based on layout complexity or feature coverage, as we already attempted with \textsc{PeCA}, may be a valuable direction to further accelerate automation. To sum up, we view this gap as a promising orthogonal direction for future work.

\bibliographystyle{splncs04}
\bibliography{main}

@String(ICCV  = {Int. Conf. Comput. Vis.})

@String(ECCV  = {Eur. Conf. Comput. Vis.})

@String(AAAI  = {AAAI})

@String(ICIP  = {IEEE Int. Conf. Image Process.})

@String(TOG   = {ACM Trans. Graph.})

@String(ICCV  = {ICCV})

@String(ECCV  = {ECCV})

@String(ICIP  = {ICIP})

@String(TOG   = {ACM TOG})

@article{sakuga42m2024,
    title   = {Sakuga-42M Dataset: Scaling Up Cartoon Research},
    author  = {Zhenglin Pan, Yu Zhu, Yuxuan Mu},
    journal = {arXiv preprint arXiv:2405.07425},
    year    = {2024}
}

@article{Peebles2022DiT,
  title={Scalable Diffusion Models with Transformers},
  author={William Peebles and Saining Xie},
  year={2022},
  journal={arXiv preprint arXiv:2212.09748},
}

@incollection{guajardo2024generative,
  title={Generative ai for 2d character animation},
  author={Guajardo, Jaime and Bursalioglu, Ozgun and Goldman, Dan B},
  booktitle={ACM SIGGRAPH 2024 Posters},
  pages={1--2},
  year={2024}
}

@incollection{maejima2021anime,
  title={Anime character colorization using few-shot learning},
  author={Maejima, Akinobu and Kubo, Hiroyuki and Shinagawa, Seitaro and Funatomi, Takuya and Yotsukura, Tatsuo and Nakamura, Satoshi and Mukaigawa, Yasuhiro},
  booktitle={SIGGRAPH Asia 2021 Technical Communications},
  pages={1--4},
  year={2021}
}

@article{tang2025generative,
  title={Generative {AI} for cel-animation: A survey},
  author={Tang, Yunlong and Guo, Junjia and Liu, Pinxin and Wang, Zhiyuan and Hua, Hang and Zhong, Jia-Xing and Xiao, Yunzhong and Huang, Chao and Song, Luchuan and Liang, Susan and others},
  journal={arXiv preprint arXiv:2501.06250},
  year={2025}
}

@inproceedings{dai2024learning,
  title={Learning inclusion matching for animation paint bucket colorization},
  author={Dai, Yuekun and Zhou, Shangchen and Li, Qinyue and Li, Chongyi and Loy, Chen Change},
  booktitle={Proceedings of the IEEE/CVF conference on Computer Vision and Pattern Recognition},
  pages={25544--25553},
  year={2024}
}

@article{nakanishi2013modeling,
  title={Modeling the Process of Animation Production.},
  author={Nakanishi, Hisato and Shichijo, Naohiro and Sugi, Masao and Ogata, Taiki and Hara, Tatsunori and Ota, Jun},
  journal={Int. J. Autom. Technol.},
  volume={7},
  number={4},
  pages={439--450},
  year={2013}
}

@article{dai2024paint,
  title={Paint Bucket Colorization Using Anime Character Color Design Sheets},
  author={Dai, Yuekun and Li, Qinyue and Zhou, Shangchen and Luo, Yihang and Li, Chongyi and Loy, Chen Change},
  journal={arXiv preprint arXiv:2410.19424},
  year={2024}
}

@inproceedings{nagata2025dacon,
  title={{DACoN}: DINO for Anime Paint Bucket Colorization with Any Number of Reference Images},
  author={Nagata, Kazuma and Kaneko, Naoshi},
  booktitle={Proceedings of the IEEE/CVF International Conference on Computer Vision},
  pages={17899--17908},
  year={2025}
}

@InProceedings{Feng_2025_ICCV,
    author    = {Feng, Xiaoyi and Huang, Tao and Wang, Peng and Huang, Zizhou and Haihang, Zhang and Zou, Yuntao and Li, Dagang and Zou, Kaifeng},
    title     = {A Unified Framework for Industrial Cel-Animation Colorization with Temporal-Structural Awareness},
    booktitle = {Proceedings of the IEEE/CVF International Conference on Computer Vision},
    month     = {October},
    year      = {2025},
    pages     = {19301-19310}
}

@inproceedings{meng2025anidoc,
  title={{AniDoc}: Animation creation made easier},
  author={Meng, Yihao and Ouyang, Hao and Wang, Hanlin and Wang, Qiuyu and Wang, Wen and Cheng, Ka Leong and Liu, Zhiheng and Shen, Yujun and Qu, Huamin},
  booktitle={Proceedings of the Computer Vision and Pattern Recognition Conference},
  pages={18187--18197},
  year={2025}
}

@article{zhuang2024colorflow,
  title={{ColorFlow}: Retrieval-Augmented Image Sequence Colorization},
  author={Zhuang, Junhao and Ju, Xuan and Zhang, Zhaoyang and Liu, Yong and Zhang, Shiyi and Yuan, Chun and Shan, Ying},
  journal={arXiv preprint arXiv:2412.11815},
  year={2024}
}

@article{huang2024lvcd,
  title={{LVCD}: reference-based lineart video colorization with diffusion models},
  author={Huang, Zhitong and Zhang, Mohan and Liao, Jing},
  journal={ACM Transactions on Graphics (TOG)},
  volume={43},
  number={6},
  pages={1--11},
  year={2024},
  publisher={ACM New York, NY, USA}
}

@inproceedings{liu2025manganinja,
  title={Manganinja: Line art colorization with precise reference following},
  author={Liu, Zhiheng and Cheng, Ka Leong and Chen, Xi and Xiao, Jie and Ouyang, Hao and Zhu, Kai and Liu, Yu and Shen, Yujun and Chen, Qifeng and Luo, Ping},
  booktitle={Proceedings of the Computer Vision and Pattern Recognition Conference},
  pages={5666--5677},
  year={2025}
}

@article{xing2024tooncrafter,
  title={Tooncrafter: Generative cartoon interpolation},
  author={Xing, Jinbo and Liu, Hanyuan and Xia, Menghan and Zhang, Yong and Wang, Xintao and Shan, Ying and Wong, Tien-Tsin},
  journal={ACM Transactions on Graphics (TOG)},
  volume={43},
  number={6},
  pages={1--11},
  year={2024},
  publisher={ACM New York, NY, USA}
}

@inproceedings{radford2021learning,
  title={Learning transferable visual models from natural language supervision},
  author={Radford, Alec and Kim, Jong Wook and Hallacy, Chris and Ramesh, Aditya and Goh, Gabriel and Agarwal, Sandhini and Sastry, Girish and Askell, Amanda and Mishkin, Pamela and Clark, Jack and others},
  booktitle={International Conference on Machine Learning},
  pages={8748--8763},
  year={2021},
  organization={PMLR}
}

@article{oquab2023dinov2,
  title={{DINOv2}: Learning robust visual features without supervision},
  author={Oquab, Maxime and Darcet, Timoth{\'e}e and Moutakanni, Th{\'e}o and Vo, Huy and Szafraniec, Marc and Khalidov, Vasil and Fernandez, Pierre and Haziza, Daniel and Massa, Francisco and El-Nouby, Alaaeldin and others},
  journal={arXiv preprint arXiv:2304.07193},
  year={2023}
}

@inproceedings{kirillov2023segment,
  title={Segment anything},
  author={Kirillov, Alexander and Mintun, Eric and Ravi, Nikhila and Mao, Hanzi and Rolland, Chloe and Gustafson, Laura and Xiao, Tete and Whitehead, Spencer and Berg, Alexander C and Lo, Wan-Yen and others},
  booktitle={Proceedings of the IEEE/CVF International Conference on Computer Vision},
  pages={4015--4026},
  year={2023}
}

@article{ravi2024sam,
  title={{SAM} 2: Segment anything in images and videos},
  author={Ravi, Nikhila and Gabeur, Valentin and Hu, Yuan-Ting and Hu, Ronghang and Ryali, Chaitanya and Ma, Tengyu and Khedr, Haitham and R{\"a}dle, Roman and Rolland, Chloe and Gustafson, Laura and others},
  journal={arXiv preprint arXiv:2408.00714},
  year={2024}
}

@article{zhang2021tip,
  title={Tip-adapter: Training-free clip-adapter for better vision-language modeling},
  author={Zhang, Renrui and Fang, Rongyao and Zhang, Wei and Gao, Peng and Li, Kunchang and Dai, Jifeng and Qiao, Yu and Li, Hongsheng},
  journal={arXiv preprint arXiv:2111.03930},
  year={2021}
}

@inproceedings{shu2022tpt,
  author    = {Manli, Shu and Weili, Nie and De-An, Huang and Zhiding, Yu and Tom, Goldstein and Anima, Anandkumar and Chaowei, Xiao},
  title     = {Test-Time Prompt Tuning for Zero-shot Generalization in Vision-Language Models},
  booktitle = {Advances in Neural Information Processing Systems},
  year      = {2022},
}

@inproceedings{wang2021tent,
  title={Tent: Fully Test-Time Adaptation by Entropy Minimization},
  author={Wang, Dequan and Shelhamer, Evan and Liu, Shaoteng and Olshausen, Bruno and Darrell, Trevor},
  booktitle={International Conference on Learning Representations},
  year={2021},
  url={https://openreview.net/forum?id=uXl3bZLkr3c}
}

@article{schuurmans2018efficient,
  title={Efficient semantic image segmentation with superpixel pooling},
  author={Schuurmans, Mathijs and Berman, Maxim and Blaschko, Matthew B},
  journal={arXiv preprint arXiv:1806.02705},
  year={2018}
}

@article{krause2014submodular,
  title={Submodular function maximization.},
  author={Krause, Andreas and Golovin, Daniel},
  journal={Tractability},
  volume={3},
  number={71-104},
  pages={3},
  year={2014}
}

@book{wittgenstein1977remarks,
  title={Remarks on Colour},
  author={Wittgenstein, L.},
  url={https://books.google.co.uk/books?id=xQhbwgEACAAJ},
  year={1977},
  publisher={University of California Press}
}

@incollection{ramassamy2018pre,
  title={Pre-and post-processes for automatic colorization using a fully convolutional network},
  author={Ramassamy, Sophie and Kubo, Hiroyuki and Funatomi, Takuya and Ishii, Daichi and Maejima, Akinobu and Nakamura, Satoshi and Mukaigawa, Yasuhiro},
  booktitle={SIGGRAPH Asia 2018 Posters},
  pages={1--2},
  year={2018}
}

@incollection{maejima2019graph,
  title={Graph matching based anime colorization with multiple references},
  author={Maejima, Akinobu and Kubo, Hiroyuki and Funatomi, Takuya and Yotsukura, Tatsuo and Nakamura, Satoshi and Mukaigawa, Yasuhiro},
  booktitle={ACM SIGGRAPH 2019 Posters},
  pages={1--2},
  year={2019}
}

@inproceedings{casey2021animation,
  title={The animation transformer: Visual correspondence via segment matching},
  author={Casey, Evan and P{\'e}rez, V{\'\i}ctor and Li, Zhuoru},
  booktitle={Proceedings of the IEEE/CVF International Conference on Computer Vision},
  pages={11323--11332},
  year={2021}
}

@inproceedings{shlapentokh2024region,
  title={Region-based representations revisited},
  author={Shlapentokh-Rothman, Michal and Blume, Ansel and Xiao, Yao and Wu, Yuqun and TV, Sethuraman and Tao, Heyi and Lee, Jae Yong and Torres, Wilfredo and Wang, Yu-Xiong and Hoiem, Derek},
  booktitle={Proceedings of the IEEE/CVF conference on Computer Vision and Pattern Recognition},
  pages={17107--17116},
  year={2024}
}

@inproceedings{kim2023semantic,
  title={Semantic-aware superpixel for weakly supervised semantic segmentation},
  author={Kim, Sangtae and Park, Daeyoung and Shim, Byonghyo},
  booktitle={Proceedings of the AAAI Conference on Artificial Intelligence},
  volume={37},
  number={1},
  pages={1142--1150},
  year={2023}
}

@article{tang2023emergent,
  title={Emergent correspondence from image diffusion},
  author={Tang, Luming and Jia, Menglin and Wang, Qianqian and Phoo, Cheng Perng and Hariharan, Bharath},
  journal={Advances in Neural Information Processing Systems},
  volume={36},
  pages={1363--1389},
  year={2023}
}

@misc{Anita2024,
    title = {Anita Dataset},
    howpublished = {{https://zhenglinpan.github.io/AnitaDataset\_homepage/}},
    note = {Accessed: 2024-06-24}
}

@inproceedings{zhang2025animecolor,
  title={AnimeColor: Reference-based Animation Colorization with Diffusion Transformers},
  author={Zhang, Yuhong and Wang, Liyao and Wang, Han and Wu, Danni and Lin, Zuzeng and Wang, Feng and Song, Li},
  booktitle={Proceedings of the 33rd ACM International Conference on Multimedia},
  pages={6682--6690},
  year={2025}
}

@inproceedings{zhao2023improving,
  title={Improving video colorization by test-time tuning},
  author={Zhao, Yaping and Zheng, Haitian and Luo, Jiebo and Lam, Edmund Y},
  booktitle={2023 IEEE International Conference on Image Processing (ICIP)},
  pages={166--170},
  year={2023},
  organization={IEEE}
}

@inproceedings{RTTLC,
author = {Li, Jinjing and Liang, Qirong and Li, Qipei and Gang, Ruipeng and Fang, Ji and Lin, Chichen and Feng, Shuang and Liu, Xiaofeng},
year = {2023},
month = {06},
pages = {1722-1730},
title = {{RTTLC}: Video Colorization with Restored Transformer and Test-time Local Converter},
doi = {10.1109/CVPRW59228.2023.00173}
}

@misc{simeoni2025dinov3,
      title={{DINOv3}}, 
      author={Oriane Siméoni and Huy V. Vo and Maximilian Seitzer and Federico Baldassarre and Maxime Oquab and Cijo Jose and Vasil Khalidov and Marc Szafraniec and Seungeun Yi and Michaël Ramamonjisoa and Francisco Massa and Daniel Haziza and Luca Wehrstedt and Jianyuan Wang and Timothée Darcet and Théo Moutakanni and Leonel Sentana and Claire Roberts and Andrea Vedaldi and Jamie Tolan and John Brandt and Camille Couprie and Julien Mairal and Hervé Jégou and Patrick Labatut and Piotr Bojanowski},
      year={2025},
      eprint={2508.10104},
      archivePrefix={arXiv},
      primaryClass={cs.CV},
      url={https://arxiv.org/abs/2508.10104}, 
}

@article{tschannen2025siglip,
  title={{SIGLIP} 2: Multilingual vision-language encoders with improved semantic understanding, localization, and dense features},
  author={Tschannen, Michael and Gritsenko, Alexey and Wang, Xiao and Naeem, Muhammad Ferjad and Alabdulmohsin, Ibrahim and Parthasarathy, Nikhil and Evans, Talfan and Beyer, Lucas and Xia, Ye and Mustafa, Basil and others},
  journal={arXiv preprint arXiv:2502.14786},
  year={2025}
}

@InProceedings{Rombach_2022_CVPR,
    author    = {Rombach, Robin and Blattmann, Andreas and Lorenz, Dominik and Esser, Patrick and Ommer, Bj\"orn},
    title     = {High-Resolution Image Synthesis With Latent Diffusion Models},
    booktitle = {Proceedings of the IEEE/CVF conference on Computer Vision and Pattern Recognition},
    month     = {June},
    year      = {2022},
    pages     = {10684-10695}
}

@inproceedings{liu2023zero,
  title={Zero-1-to-3: Zero-shot One Image to 3D Object},
  author={Liu, Ruoshi and Wu, Rundi and Van Hoorick, Basile and Tokmakov, Pavel and Zakharov, Sergey and Vondrick, Carl},
  booktitle={2023 IEEE/CVF International Conference on Computer Vision},
  pages={9264--9275},
  year={2023},
  organization={IEEE}
}

@inproceedings{carreira2017quo,
  title={Quo vadis, action recognition? a new model and the kinetics dataset},
  author={Carreira, Joao and Zisserman, Andrew},
  booktitle={proceedings of the IEEE Conference on Computer Vision and Pattern Recognition},
  pages={6299--6308},
  year={2017}
}

@inproceedings{karaev2024cotracker,
  title={{CoTracker}: It is better to track together},
  author={Karaev, Nikita and Rocco, Ignacio and Graham, Benjamin and Neverova, Natalia and Vedaldi, Andrea and Rupprecht, Christian},
  booktitle={European Conference on Computer Vision},
  pages={18--35},
  year={2024},
  organization={Springer}
}

@inproceedings{wang2019learning,
  title={Learning correspondence from the cycle-consistency of time},
  author={Wang, Xiaolong and Jabri, Allan and Efros, Alexei A},
  booktitle={Proceedings of the IEEE/CVF conference on Computer Vision and Pattern Recognition},
  pages={2566--2576},
  year={2019}
}

@inproceedings{cao2021line,
  title={Line art colorization based on explicit region segmentation},
  author={Cao, Ruizhi and Mo, Haoran and Gao, Chengying},
  booktitle={Computer Graphics Forum},
  volume={40},
  number={7},
  pages={1--10},
  year={2021},
  organization={Wiley Online Library}
}

@article{shi2022reference,
  title={Reference-based deep line art video colorization},
  author={Shi, Min and Zhang, Jia-Qi and Chen, Shu-Yu and Gao, Lin and Lai, Yu-Kun and Zhang, Fang-Lue},
  journal={IEEE Transactions on Visualization and Computer Graphics},
  volume={29},
  number={6},
  pages={2965--2979},
  year={2022},
  publisher={IEEE}
}

@article{cao2024animediffusion,
  title={{AnimeDiffusion}: Anime diffusion colorization},
  author={Cao, Yu and Meng, Xiangqiao and Mok, PY and Lee, Tong-Yee and Liu, Xueting and Li, Ping},
  journal={IEEE Transactions on Visualization and Computer Graphics},
  volume={30},
  number={10},
  pages={6956--6969},
  year={2024},
  publisher={IEEE}
}

@inproceedings{zhaommicl,
  title={{MMICL}: Empowering Vision-language Model with Multi-Modal In-Context Learning},
  author={Zhao, Haozhe and Cai, Zefan and Si, Shuzheng and Ma, Xiaojian and An, Kaikai and Chen, Liang and Liu, Zixuan and Wang, Sheng and Han, Wenjuan and Chang, Baobao},
  booktitle={The Twelfth International Conference on Learning Representations}
}

@article{qiao2025v,
  title={V-thinker: Interactive thinking with images},
  author={Qiao, Runqi and Tan, Qiuna and Yang, Minghan and Dong, Guanting and Yang, Peiqing and Lang, Shiqiang and Wan, Enhui and Wang, Xiaowan and Xu, Yida and Yang, Lan and others},
  journal={arXiv preprint arXiv:2511.04460},
  year={2025}
}

@article{radovanovic2010hubs,
  title={Hubs in space: Popular nearest neighbors in high-dimensional data},
  author={Radovanovic, Milos and Nanopoulos, Alexandros and Ivanovic, Mirjana},
  journal={Journal of machine learning research},
  volume={11},
  number={sept},
  pages={2487--2531},
  year={2010}
}

@inproceedings{shanmugam2021better,
  title={Better aggregation in test-time augmentation},
  author={Shanmugam, Divya and Blalock, Davis and Balakrishnan, Guha and Guttag, John},
  booktitle={Proceedings of the IEEE/CVF International Conference on Computer Vision},
  pages={1214--1223},
  year={2021}
}

@article{shorten2019surveytru,
  title={A survey on image data augmentation for deep learning},
  author={Shorten, Connor and Khoshgoftaar, Taghi M},
  journal={Journal of big data},
  volume={6},
  number={1},
  pages={1--48},
  year={2019},
  publisher={Springer}
}

@article{sadihin2026timecolor,
  title={{TimeColor}: Flexible Reference Colorization via Temporal Concatenation},
  author={Sadihin, Bryan Constantine and Meng, Yihao and Wang, Michael Hua and Chen, Matteo Jiahao and Su, Hang},
  journal={arXiv preprint arXiv:2601.00296},
  year={2026}
}

@inproceedings{10.1145/800249.807456,
author = {Smith, Alvy Ray},
title = {Tint fill},
year = {1979},
isbn = {0897910044},
publisher = {Association for Computing Machinery},
address = {New York, NY, USA},
url = {https://doi.org/10.1145/800249.807456},
doi = {10.1145/800249.807456},
booktitle = {Proceedings of the 6th Annual Conference on Computer Graphics and Interactive Techniques},
pages = {276--283},
numpages = {8},
location = {Chicago, Illinois, USA},
series = {SIGGRAPH '79}
}

@inproceedings{miao2022large,

  title={Large-scale Video Panoptic Segmentation in the Wild: A Benchmark},

  author={Miao, Jiaxu and Wang, Xiaohan and  Wu, Yu and Li, Wei and Zhang, Xu and Wei, Yunchao and Yang, Yi},

  booktitle={Proceedings of the {IEEE} Conference on Computer Vision and Pattern Recognition},

  year={2022}

}

@article{slic,
author = {Achanta, Radhakrishna and Shaji, Appu and Smith, Kevin and Lucchi, Aurélien and Fua, Pascal and Süsstrunk, Sabine},
year = {2010},
month = {06},
pages = {},
title = {{SLIC} superpixels},
journal = {Technical report, EPFL}
}

@inproceedings{sasaki2017joint,
  title={Joint gap detection and inpainting of line drawings},
  author={Sasaki, Kazuma and Iizuka, Satoshi and Simo-Serra, Edgar and Ishikawa, Hiroshi},
  booktitle={Proceedings of the IEEE conference on computer vision and pattern recognition},
  pages={5725--5733},
  year={2017}
}

@article{xu2022deep,
  title={Deep learning for free-hand sketch: A survey},
  author={Xu, Peng and Hospedales, Timothy M and Yin, Qiyue and Song, Yi-Zhe and Xiang, Tao and Wang, Liang},
  journal={IEEE transactions on pattern analysis and machine intelligence},
  volume={45},
  number={1},
  pages={285--312},
  year={2022},
  publisher={IEEE}
}

@inproceedings{teed2020raft,
  title={Raft: Recurrent all-pairs field transforms for optical flow},
  author={Teed, Zachary and Deng, Jia},
  booktitle={European conference on computer vision},
  pages={402--419},
  year={2020},
  organization={Springer}
}

@InProceedings{DanbooRegion2020,
author={Lvmin Zhang and Yi JI and Chunping Liu}, 
booktitle={European Conference on Computer Vision (ECCV)}, 
title={DanbooRegion: An Illustration Region Dataset}, 
year={2020}, 
}

@inproceedings{10.1145/3681758.3698003,
author = {Allen, Benjamin and Maejima, Akinobu and Anjyo, Ken},
title = {Fast Leak-Resistant Segmentation for Anime Line Art},
year = {2024},
isbn = {9798400711404},
publisher = {Association for Computing Machinery},
address = {New York, NY, USA},
url = {https://doi.org/10.1145/3681758.3698003},
doi = {10.1145/3681758.3698003},
booktitle = {SIGGRAPH Asia 2024 Technical Communications},
articleno = {4},
numpages = {4},
location = {
},
series = {SA '24}
}

@article{10.1023/B:VISI.0000029664.99615.94,
author = {Lowe, David G.},
title = {Distinctive Image Features from Scale-Invariant Keypoints},
year = {2004},
issue_date = {November 2004},
publisher = {Kluwer Academic Publishers},
address = {USA},
volume = {60},
number = {2},
issn = {0920-5691},
url = {https://doi.org/10.1023/B:VISI.0000029664.99615.94},
doi = {10.1023/B:VISI.0000029664.99615.94},
journal = {Int. J. Comput. Vision},
month = nov,
pages = {91--110},
numpages = {20}
}

@misc{blogNanoBanana,
	author = {Google},
	title = {{N}ano {B}anana 2: {C}ombining {P}ro capabilities with lightning-fast speed --- blog.google},
	howpublished = {\url{https://blog.google/innovation-and-ai/technology/ai/nano-banana-2/}},
	year = {},
	note = {[Accessed 21-06-2026]},
}

@inproceedings{zhuang2025cobra,
  title={Cobra: Efficient line art colorization with broader references},
  author={Zhuang, Junhao and Li, Lingen and Ju, Xuan and Zhang, Zhaoyang and Yuan, Chun and Shan, Ying},
  booktitle={Proceedings of the Special Interest Group on Computer Graphics and Interactive Techniques Conference Conference Papers},
  pages={1--11},
  year={2025}
}

@misc{zhang2025followyourcolormultiinstancesketchcolorization,
      title={Follow-Your-Color: Multi-Instance Sketch Colorization}, 
      author={Yinhan Zhang and Yue Ma and Bingyuan Wang and Qifeng Chen and Zeyu Wang},
      year={2025},
      eprint={2503.16948},
      archivePrefix={arXiv},
      primaryClass={cs.CV},
      url={https://arxiv.org/abs/2503.16948}, 
}

@article{li2025tooncomposer,
  title={ToonComposer: Streamlining Cartoon Production with Generative Post-Keyframing},
  author={Li, Lingen and Wang, Guangzhi and Zhang, Zhaoyang and Li, Yaowei and Li, Xiaoyu and Dou, Qi and Gu, Jinwei and Xue, Tianfan and Shan, Ying},
  journal={arXiv preprint arXiv:2508.10881},
  year={2025}
}

@article{peng2024charactergen,
  title={Charactergen: Efficient 3d character generation from single images with multi-view pose canonicalization},
  author={Peng, Hao-Yang and Zhang, Jia-Peng and Guo, Meng-Hao and Cao, Yan-Pei and Hu, Shi-Min},
  journal={ACM Transactions on Graphics (TOG)},
  volume={43},
  number={4},
  pages={1--13},
  year={2024},
  publisher={ACM New York, NY, USA}
}

@InProceedings{Yang_2025_ICCV,
    author    = {Yang, Yuxue and Fan, Lue and Lin, Zuzeng and Wang, Feng and Zhang, Zhaoxiang},
    title     = {LayerAnimate: Layer-level Control for Animation},
    booktitle = {Proceedings of the IEEE/CVF International Conference on Computer Vision (ICCV)},
    month     = {October},
    year      = {2025},
    pages     = {10865-10874}
}

@misc{zhang2026omnicolorunifiedframeworkmultimodal,
      title={OmniColor: A Unified Framework for Multi-modal Lineart Colorization}, 
      author={Xulu Zhang and Haoqian Du and Xiaoyong Wei and Qing Li},
      year={2026},
      eprint={2603.27531},
      archivePrefix={arXiv},
      primaryClass={cs.CV},
      url={https://arxiv.org/abs/2603.27531}, 
}

@inproceedings{liu2026pixels,
  title={From Pixels to Personas: Tracking the Evolution of Anime Characters},
  author={Liu, Rongze and Pei, Jiaxin and Zhu, Jian},
  booktitle={Proceedings of the International AAAI Conference on Web and Social Media},
  volume={20},
  number={1},
  pages={1488--1504},
  year={2026}
}

\end{document}